%% file: Thesis.tex
\documentclass{mimosis}

\usepackage{metalogo}
\usepackage{enumitem}
\usepackage{algorithmic}
\usepackage{algorithm}
\usepackage{tabularx}
\usepackage[utf8]{inputenc}
\usepackage[T1]{fontenc}
\usepackage[figuresright]{rotating}
\usepackage{microtype}
\usepackage{breakcites}

\newcolumntype{Y}{>{\centering\arraybackslash}X}

\newcommand{\txtsub}[1]{$_{\text{#1}}$}

\usepackage{etoolbox}

\usepackage[binary-units=true]{siunitx}
\DeclareSIUnit\px{px}

\usepackage[%
  colorlinks = true,
  citecolor  = RoyalBlue,
  linkcolor  = RoyalBlue,
  urlcolor   = RoyalBlue,
  unicode,
  ]{hyperref}

\usepackage{bookmark}

\usepackage[%
  autocite     = plain,
  backend      = biber,
  doi          = true,
  url          = true,
  giveninits   = true,
  hyperref     = true,
  maxbibnames  = 99,
  maxcitenames = 99,
  sortcites    = true,
  style        = numeric,
  ]{biblatex}

\input{bibliography-mimosis}
\ifxetexorluatex
\else
  \usepackage[lf]{ebgaramond}
  \usepackage[oldstyle,scale=0.7]{sourcecodepro}
\fi

\makeindex
\makeglossaries

\title{Imbalanced data preprocessing techniques utilizing local data characteristics}
\author{Michał Koziarski}

\begin{document}

\frontmatter
  \include{Sources/Title}
  \include{Sources/Abstract}
  \include{Sources/Abstract_PL}

  \tableofcontents

\mainmatter

  \include{Sources/Introduction}
  \include{Sources/Preliminaries}
\include{Sources/Binary}
  \include{Sources/Multiclass}
  \include{Sources/Applications}
  \include{Sources/Summary}



  \printbibliography

\end{document}

%% file: bibliography-mimosis.tex
\AtEveryBibitem{\clearfield{month}}
\AtEveryCitekey{\clearfield{month}}

\AtBeginBibliography{%
  \renewcommand*{\finalnamedelim}{%
    \ifthenelse{\value{listcount} > 2}{%
      \addcomma
      \addspace
      \bibstring{and}%
    }{%
      \addspace
      \bibstring{and}%
    }
  }
}

\renewbibmacro{in:}{%
  \ifentrytype{article}
  {%
  }%
  {%
    \printtext{\bibstring{in}\intitlepunct}%
  }%
}

\renewbibmacro*{issue+date}{%
  \setunit{\addcomma\space}
    \iffieldundef{issue}
      {\usebibmacro{date}}
      {\printfield{issue}%
       \setunit*{\addspace}%
       \usebibmacro{date}}%
  \newunit}

\renewbibmacro*{volume+number+eid}{%
  \printfield{volume}%
  \setunit*{\addcolon}%
  \printfield{number}%
  \setunit{\addcomma\space}%
  \printfield{eid}%
}

\renewbibmacro*{publisher+location+date}{%
  \printlist{publisher}%
  \setunit*{\addcomma\space}%
  \printlist{location}%
  \setunit*{\addcomma\space}%
  \usebibmacro{date}%
  \newunit%
}

\renewbibmacro*{organization+location+date}{%
  \printlist{location}%
  \setunit*{\addcomma\space}%
  \printlist{organization}%
  \setunit*{\addcomma\space}%
  \usebibmacro{date}%
  \newunit%
}

\DeclareFieldFormat{labelnumberwidth}{#1\adddot}

\DeclareFieldFormat{doi}{%
  \mkbibacro{DOI}\addcolon\addnbspace
    \ifhyperref
      {\href{http://dx.doi.org/#1}{\nolinkurl{#1}}}
      {\nolinkurl{#1}}
}

\DeclareLanguageMapping{english}{english-mimosis}

%% file: Sources/Title.tex
\begin{titlepage}
  \begin{figure}
  \centering
    \includegraphics[width=.4\textwidth]{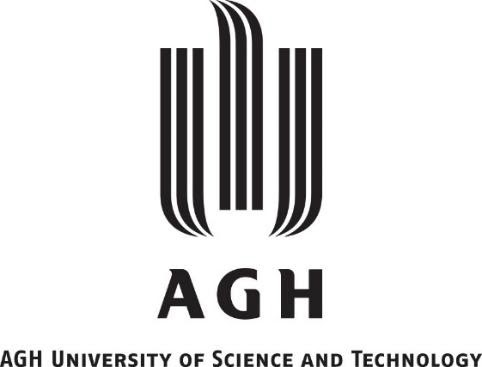}
  \end{figure}
  
  \vspace*{0.1cm}

  \begin{center}
    \begin{large}
        \textbf{FIELD OF SCIENCE: TECHNICAL SCIENCE}
    \end{large}\\[0.1cm]
    SCIENTIFIC DISCIPLINE: COMPUTER SCIENCE
  \end{center}
  
  \vspace*{1cm}
  
  \begin{center}
    \begin{huge}
        \textbf{DOCTORAL THESIS}
    \end{huge}\\[0.1cm]
  \end{center}
  
  \vspace*{0.5cm}
  
  \makeatletter
  \begin{center}
    \begin{LARGE}
      \@title
    \end{LARGE}\\[0.1cm]
  
    \vspace*{2.5cm}

    Author: \@author, M.Sc.
    
    \vspace*{0.5cm}
  
    First supervisor: Professor Bogusław Cyganek, Ph.D., D.Sc. \\
    Assisting supervisor: Bartosz Krawczyk, Ph.D. \\
    
    \vspace*{0.5cm}
    
    Completed in: AGH University of Science and Technology, \\
    Faculty of Computer Science, Electronics and Telecommunications
    
    \vfill
  \end{center}
  
  \begin{center}
      Kraków, 2021
  \end{center}
 
  \makeatother
\end{titlepage}

\newpage
\null
\thispagestyle{empty}
\newpage

%% file: Sources/Abstract.tex
\begin{center}
  \textsc{Abstract}
\end{center}
\noindent Data imbalance, that is the disproportion between the number of training observations coming from different classes, remains one of the most significant challenges affecting contemporary machine learning. The negative impact of data imbalance on traditional classification algorithms can be reduced by the data preprocessing techniques, methods that manipulate the training data to artificially reduce the degree of imbalance. However, the existing data preprocessing techniques, in particular SMOTE and its derivatives, which constitute the most prevalent paradigm of imbalanced data preprocessing, tend to be susceptible to various data difficulty factors. This is in part due to the fact that the original SMOTE algorithm does not utilize the information about majority class observations. The focus of this thesis is development of novel data resampling strategies natively utilizing the information about the distribution of both minority and majority class. The thesis summarizes the content of 12 research papers focused on the proposed binary data resampling strategies, their translation to the multi-class setting, and the practical application to the problem of histopathological data classification.

\newpage
\null
\thispagestyle{empty}
\newpage

%% file: Sources/Abstract_PL.tex
\begin{center}
  \textsc{Streszczenie}
\end{center}

\noindent Niezbalansowanie danych, czyli dysproporcja pomiędzy liczbą obserwacji treningowych należących do różnych klas, pozostaje jednym z najbardziej istotnych wyzwań współczesnego uczenia maszynowego. Negatywny wpływ niezbalansowania danych na tradycyjne algorytmy klasyfikacji może być zredukowany poprzez zastosowanie metod wstępnego przetwarzania danych, czyli algorytmów modyfikujących zbiór danych treningowych w celu sztucznego zredukowania stopnia niezbalansowania. Istniejące metody wstępnego przetwarzania danych, w szczególności SMOTE oraz jego pochodne, są jednak podatne na obecność skomplikowanych dystrybucji danych. Wynika to po części z faktu, że SMOTE nie uwzględnia informacji na temat pozycji obiektów z klasy większościowej. Celem poniższej rozprawy jest opracowanie nowych metod wstępnego przetwarzania danych, natywnie uwzględniających informację na temat dystrybucji zarówno klasy większościowej jak i mniejszościowej. Rozprawa zawiera podsumowanie zawartości 12 prac naukowych skupiających się na opracowaniu nowych metod wstępnego przetwarzania danych w problemie binarnym, ich przeniesienie do problemu wieloklasowego, oraz praktyczne zastosowanie w zadaniu klasyfikacji obrazów histopatologicznych.

%% file: Sources/Introduction.tex
\chapter{Introduction}

\begin{center}
  \begin{minipage}{0.5\textwidth}
    \begin{small}
      In which the content and the layout of the thesis are described, and the main contributions are outlined.
    \end{small}
  \end{minipage}
  \vspace{0.5cm}
\end{center}

\noindent The aim of this thesis is to describe the resampling algorithms designed to reduce the negative impact of data imbalance on the performance of traditional classification algorithms, as well as their applications to the histopathological image recognition task.

\section{Content of the thesis}

The thesis is composed of the following twelve research papers:

\begin{enumerate}[label={[\Roman*]}]
\item \fullcite{koziarski2017ccr} \\
(journal paper; Impact Factor: 0.967)
\item \fullcite{koziarski2017hais} \\
(conference paper)
\item \fullcite{koziarski2019neuro} \\
(journal paper; Impact Factor: 4.438)
\item \fullcite{koziarski2020radial} \\
(journal paper; Impact Factor: 7.196)
\item \fullcite{rbccr} \\
(preprint; journal paper; Impact Factor: 2.672)
\item \fullcite{koziarski2020csmoute} \\
(preprint; conference paper)
\item \fullcite{pa} \\
(preprint; journal paper; Impact Factor: 7.196)
\item \fullcite{krawczyk2019radial} \\
(journal paper; Impact Factor: 8.793)
\item \fullcite{koziarski2020combined} \\
(journal paper; Impact Factor: 5.921)
\item \fullcite{koziarski2018convolutional} \\
(conference paper)
\item \fullcite{koziarski2019radial} \\
(conference paper)
\item \fullcite{diagset} \\
(preprint; journal paper; Impact Factor: 4.383)
\end{enumerate}

Seven of the papers ([I] - [VII]) focus on the proposed over- and undersampling algorithms for handling data imbalance in the binary classification setting. Two of the papers ([VIII], [IX]) extend selected binary oversampling algorithms to the multiclass setting with a novel class decomposition strategy. Three of the papers ([X] - [XII]) concentrate on the practical applications of the proposed algorithms in the histopathological image recognition task.

The dependencies between the above papers can be summarized as follows:

\begin{itemize}
    \item {[I]}, [II] and [VI] do not depend on any other paper,
    \item {[III]} is a direct extension of [II],
    \item {[IV]} utilizes the ideas introduced in papers [II] and [III],
    \item {[V]} combines the techniques introduced in papers [I], [II] and [III],
    \item {[VII]} utilizes the ideas introduced in papers [II], [III], [IV] and [VI],
    \item {[VIII]} contains multiclass extension of [II] and [III],
    \item {[IX]} contains multiclass extension of [I],
    \item {[X]} applies methods introduced in [I], [II] and [III],
    \item {[XI]} applies method introduced in [IV],
    \item {[XII]} applies method introduced in [IX].
\end{itemize}


\section{Motivation}

\begin{figure}
\centering
    \begin{subfigure}[t]{0.4\textwidth}
        \centering
        \includegraphics[width=1\textwidth]{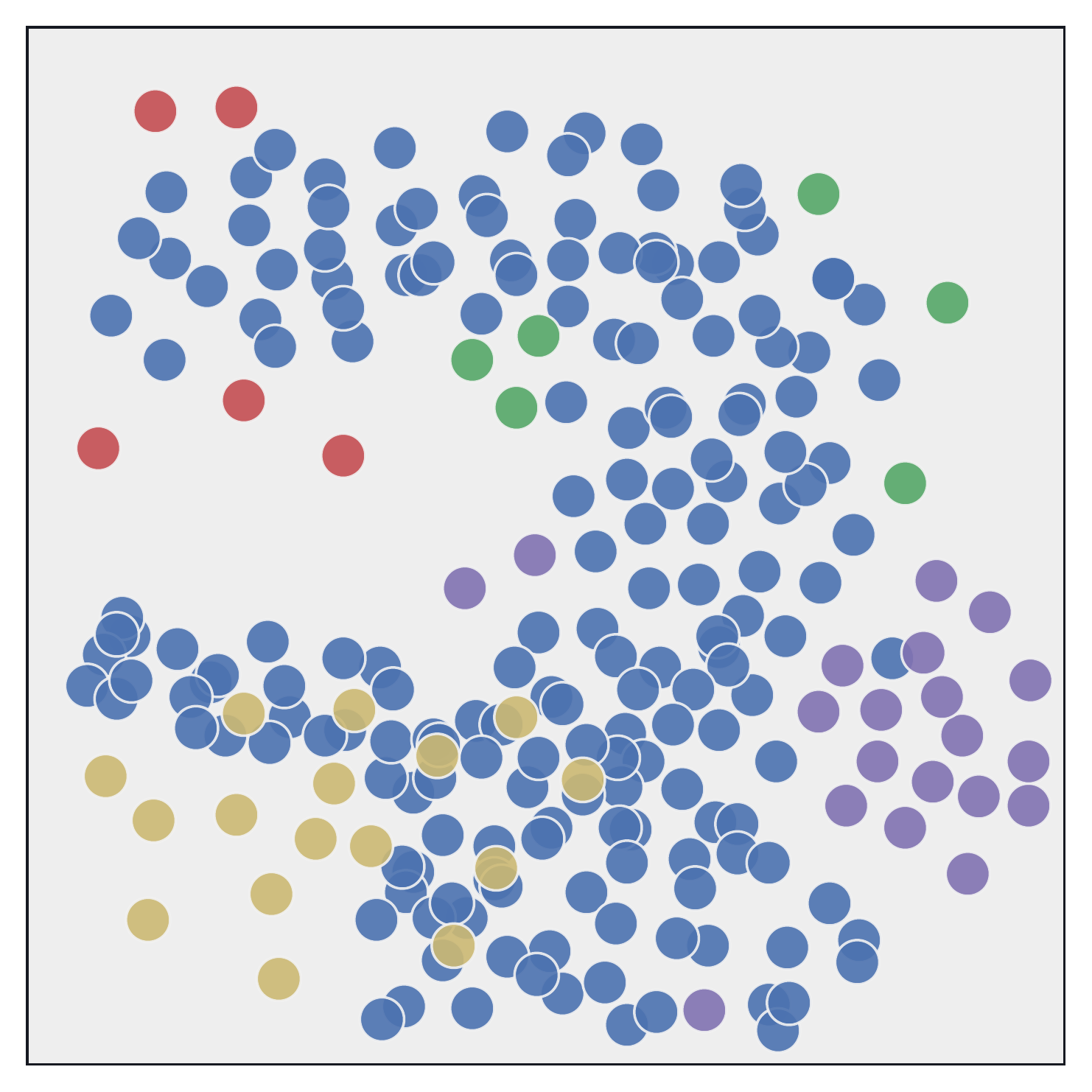}
        \caption{Initial dataset.}
    \end{subfigure}
    ~ 
    \begin{subfigure}[t]{0.4\textwidth}
        \centering
        \includegraphics[width=1\textwidth]{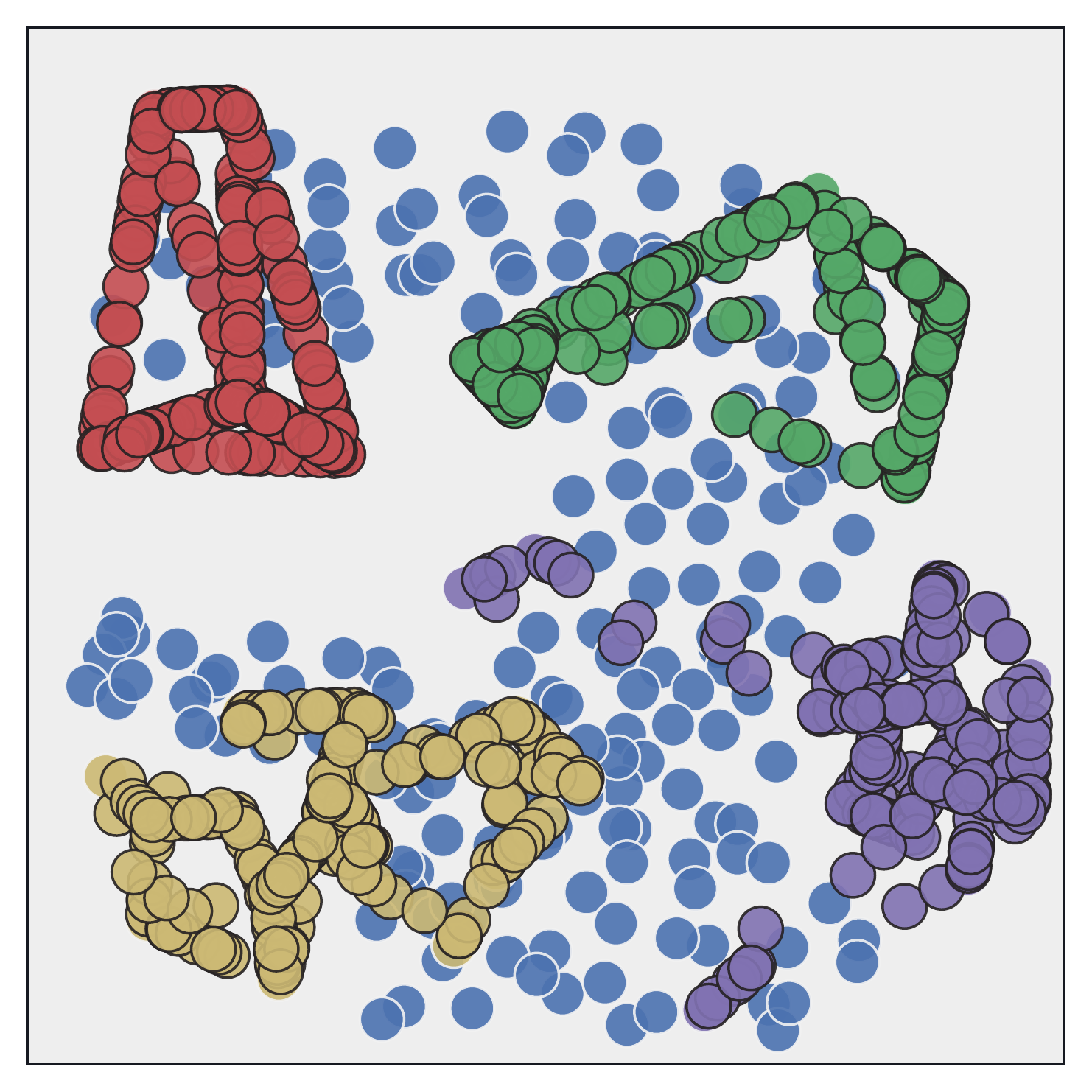}
        \caption{Dataset after balancing with SMOTE.}
    \end{subfigure}
\caption{Limitations of neighborhood-based oversampling strategies in the multiclass setting illustrated with an example of SMOTE. Neighborhood-based strategies are susceptible to disjoint data distributions, presence of noise and outliers.} 
\label{fig:difficult}
\end{figure}

The main motivations behind this thesis are the observed shortcomings of Synthetic Minority Oversampling Technique (SMOTE) \cite{chawla2002smote} and its derivatives, the most prevalent data-level approach for handling data imbalance. Despite their widespreadness, SMOTE-based techniques tend to be susceptible to various data difficulty factors, such as disjoint class distributions, small junctions, presence of noise, insufficient amount of training data, etc. This, to some extent, is caused by the fact that the original SMOTE algorithm does not utilize the information about majority class observations: placement of the synthetic observations generated by SMOTE is based solely on the relative position the neighboring minority class observations. As a result, SMOTE can produce inappropriately placed samples and introduce class overlap, which is illustrated in Figure~\ref{fig:difficult}. While some of the numerous extensions of SMOTE try to address this issue, their scope is limited by the backbone of the underlying algorithm. The claim made in this thesis is that to fully avoid the described pitfalls caused by data difficulty factors, development of novel data resampling strategies, natively utilizing the information about distribution of both minority and majority classes, is necessary.

The negative effect of said difficulty factors is even greater in the multiclass setting, in which the relationships between the classes are less clearly defined. For instance, even the typical division into majority and minority class, which occurs for the binary datasets, is more nuanced in the multiclass setting, as single class can at the same time constitute majority when compared with some of the classes, minority when compared to the others, and have comparable size to the rest. Similarly, the presence and the effect of data difficulty factors can vary depending on the chosen class pair. As a result, similar to the data difficulty factors, translation to the multiclass setting is another consideration that needs to be addressed to achieve a satisfactory performance of the developed methods for handling data imbalance. In conjunction, this thesis focuses on the development of techniques that a) address specific data difficulty factors that limit the performance of existing resampling strategies, b) leverage information about both the majority and the minority class distribution, including the relation between the two, and c) can be scaled to the multiclass setting in a manner that utilizes the interclass relationships.

\section{Research hypothesis}

Based on the described motivations the following research hypothesis is formulated:

\begin{quote}
    \textit{It is possible to design imbalanced data preprocessing techniques that, due to utilizing local data characteristics, can produce qualitatively better classification algorithms than the existing preprocessing methods.}
\end{quote}

The rest of this thesis aims to confirm the formulated research hypothesis, with the primary goal of designing novel imbalanced data preprocessing techniques utilizing local data characteristics, and the secondary goal of empirically evaluating the quality of the proposed methods.

\section{Summary of the contributions}

The main contributions of this thesis can be summarized as follows:

\begin{itemize}
    \item Proposal of the Combined Cleaning and Resampling (CCR) algorithm \cite{koziarski2017ccr}. CCR utilizes an energy-based approach to modeling the regions suitable for oversampling, which is less affected by small disjuncts and outliers than SMOTE. It combines it with a simultaneous cleaning operation, the aim of which is to reduce the effect of overlapping class distributions on the performance of the learning algorithms.
    
    \item Introduction of the concept of mutual class potential, a real-valued function used to estimate regions of interest during the resampling procedure. Mutual class potential was first used in \cite{koziarski2017hais}, in which the Radial-Based Oversampling (RBO) algorithm was originally introduced. RBO was further examined in \cite{koziarski2019neuro}, in which the computational complexity analysis of the algorithm was conducted, and the experimental analysis was extended, with a particular focus given to the algorithms performance on noisy data. Finally, the concept of mutual class potential was extended to the undersampling procedure in the form of the Radial-Based Undersampling (RBU) algorithm \cite{koziarski2020radial}. RBU addressed one of the most significant drawbacks of RBO, its high computational complexity, at the same time preserving or surpassing its performance.

    \item Utilization of the class potential within the CCR framework in the form of Radial-Based Combined Cleaning and Resampling (RB-CCR) algorithm \cite{rbccr}. RB-CCR uses class potential to refine the energy-based resampling approach of CCR. In particular, RB-CCR exploits the class potential to accurately locate sub-regions of the data-space for synthetic oversampling.

    \item Proposal of the Synthetic Majority Undersampling Technique (SMUTE), an extension of SMOTE instance interpolation to the undersampling setting, as well as the Combined Synthetic Oversampling and Undersampling Technique  (CSMOUTE), which integrates SMOTE oversampling with SMUTE undersampling \cite{koziarski2020csmoute}.
    
    \item Extension of the concept of class potential to the Potential Anchoring (PA) \cite{pa} algorithm, which instead of boosting the specific regions, focused on preserving the underlying class distribution in a unified over- and undersampling framework.

    \item Proposal of a novel class decomposition strategy that utilizes the information coming from all of the classes and its application to two of the proposed binary oversampling algorithms in form of Multiclass Combined Cleaning and Resampling (MC-CCR) \cite{koziarski2020combined} and Multiclass Radial-Based Oversampling (MC-RBO) \cite{krawczyk2019radial}.

    \item Application to the introduced algorithms to the histopathological image recognition problem domain in three different experimental studies, with the first one \cite{koziarski2018convolutional} aimed at assessing the impact of data imbalance on the performance of convolutional neural networks, the second one \cite{koziarski2019radial} evaluating the possibility of utilizing RBU undersampling in the space of high-level features extracted from a convolutional neural network, and the third one \cite{diagset} dealing with the issue of large-scale multi-class histopathological image classification, with MC-CCR used as a method of reducing the impact of data imbalance.
\end{itemize}

\section{Structure of the thesis}

The rest of this thesis is organized as follows. In Chapter~\ref{chapter:preliminaries} basic concepts related to dealing with data imbalance are briefly outlined. In Chapter~\ref{chapter:binary} the binary data resampling strategies developed as a part of this thesis are presented. In Chapter~\ref{chapter:multiclass} the proposed multi-class decomposition strategy and its extension to the selected binary algorithms are described. In both chapters the emphasis is put on introducing the algorithms and summarizing the empirical findings, instead of detailed description of the obtained experimental results, which can be found in the corresponding papers. In contrast, the focus of Chapter~\ref{chapter:applications} is the application of the previously developed methods in a histopathological image classification task, and a presentation of a more in-depth description of the conducted experiments. Finally, in Chapter~\ref{chapter:summary} the thesis is concluded and possible future research directions are presented.


% ...

\section{Acknowledgment}

Papers [IV] - [VII] and [IX] - [XI] were funded by the National Science Centre, Poland, PRELUDIUM grant no. 2017/27/N/ST6/01705, for which the author was a principal investigator. Papers [I] - [III] and [VIII] were funded by the National Science Centre, Poland, OPUS grant no. 2015/19/B/ST6/01597, for which the author was an investigator. Paper [XII] was funded by the European Regional Development Fund in the Intelligent Development 2014-2020 Programme grant no. POIR.01.01.01-00-0861/16-00, supported by the National Center for Research and Development, Poland, and Diagnostyka Consilio, for which the author was an investigator. Papers [I] - [XI] were also supported in part by the PLGrid Infrastructure.

%% file: Sources/Preliminaries.tex
\chapter{Preliminaries}
\label{chapter:preliminaries}

\begin{center}
  \begin{minipage}{0.5\textwidth}
    \begin{small}
      In which basic concepts, essential for the later part of this thesis, are introduced.
    \end{small}
  \end{minipage}
  \vspace{0.5cm}
\end{center}

\section{Imbalanced data classification}

The presence of data imbalance can significantly impact the performance of traditional learning algorithms \cite{Branco:2016}. The disproportion between the number of majority and minority observations influences the process of optimization concerning a zero-one loss function, leading to a bias towards the majority class and accompanying degradation of the predictive capabilities for the minority classes. While the problem of data imbalance is well established in the literature, it was traditionally studied in the context of binary classification problems, with the sole goal of reducing the degree of imbalance. However, recent studies point to the fact that it is not the imbalanced data itself, but rather other data difficulty factors, amplified by the data imbalance, that pose a challenge during the learning process \cite{stefanowski2016dealing,Fernandez:2018}. Such factors include small sample size, presence of disjoint and overlapping data distributions, and presence of outliers and noisy observations.

\section{Approaches to handling data imbalance}

The strategies for dealing with data imbalance can be divided into two categories. First of all, the data-level methods: algorithms that perform data preprocessing with the aim of reducing the imbalance ratio, either by decreasing the number of majority observations (undersampling) or increasing the number of minority observations (oversampling). After applying such preprocessing, the transformed data can be later classified using traditional learning algorithms. The simplest data-level approaches are the unguided data preprocessing techniques, that is random oversampling (ROS) and random undersampling (RUS). However, despite their simplicity and computational efficiency, they tend to be outperformed by the guided resampling algorithms.

By far, the most prevalent data-level approach is SMOTE \cite{chawla2002smote} algorithm. It is a guided oversampling technique, in which synthetic minority observations are being created by interpolation of the existing instances. It is nowadays considered a cornerstone for the majority of the following oversampling methods \cite{Perez-Ortiz:2016,Bellinger:2018}. However, due to the underlying assumption about the homogeneity of the clusters of minority observations, SMOTE can inappropriately alter the class distribution when factors such as disjoint data distributions, noise, and outliers are present. Numerous modifications of the original SMOTE algorithm have been proposed in the literature. The most notable include Borderline SMOTE \cite{Han:2005}, which focuses on the process of synthetic observation generation around the instances close to the decision border; Safe-level SMOTE~\cite{Bunkhumpornpat:2009} and LN-SMOTE~\cite{Maciejewski:2011}, which aim to reduce the risk of introducing synthetic observations inside regions of the majority class; and ADASYN \cite{He:2008}, that prioritizes the difficult instances.

Similar to the case of oversampling, finding the regions of interest, in the case of undersampling indicating which observations are to be discarded, is essential choice in the algorithm design process. Besides the random methods, over the years a number of guided undersampling strategies was proposed. For instance, Anand et al. \cite{anand2010approach} propose sorting the undersampled observations based on the weighted Euclidean distance from the positive samples. Smith et al. \cite{smith2014instance}, in their study of instance level data complexity, advocate for using the instance hardness criterion, with the hardness estimated based on the certainty of the classifiers predictions. Another family of methods that can be distinguished are the cluster-based undersampling algorithms, notably the methods proposed by Yen and Lee \cite{yen2009cluster}, which use clustering to select the most representative subset of data. Finally, as has been originally demonstrated by Liu et al. \cite{liu2008exploratory}, undersampling algorithms are well-suited for forming classifier ensembles, an idea that was further extended in form of evolutionary undersampling \cite{galar2013eusboost} and boosting \cite{lu2017adaptive}.

Finally, the second category of methods for dealing with data imbalance consists of algorithm-level solutions. These techniques alter the traditional learning algorithms to eliminate the shortcomings they display when applied to imbalanced data problems. Notable examples of algorithm-level solutions include: kernel functions \cite{Mathew:2018}, splitting criteria in decision trees \cite{Li:2018}, one-class classification algorithms \cite{cyganek2012one,devi2019learning}, and modifications off the underlying loss function to make it cost-sensitive \cite{Khan:2018}. However, contrary to the data-level approaches, algorithm-level solutions necessitate a choice of a specific classifier. Still, in many cases, they are reported to lead to a better performance than sampling approaches \cite{Fernandez:2018}.

\section{Limitations of the existing methods}
\label{sec:limitations}

Neighborhood-based oversampling strategies, such as SMOTE and its derivatives, are by far the most prevalent approaches to dealing with the imbalance on the data level. However, despite their popularity, this family of methods does not remain without its own shortcomings. In the remainder of this section some of disadvantages of the neighborhood-based oversampling strategies are discussed, putting a particular emphasis on the issue of noisy data.

\begin{figure*}
\centering
\includegraphics[width=0.35\textwidth]{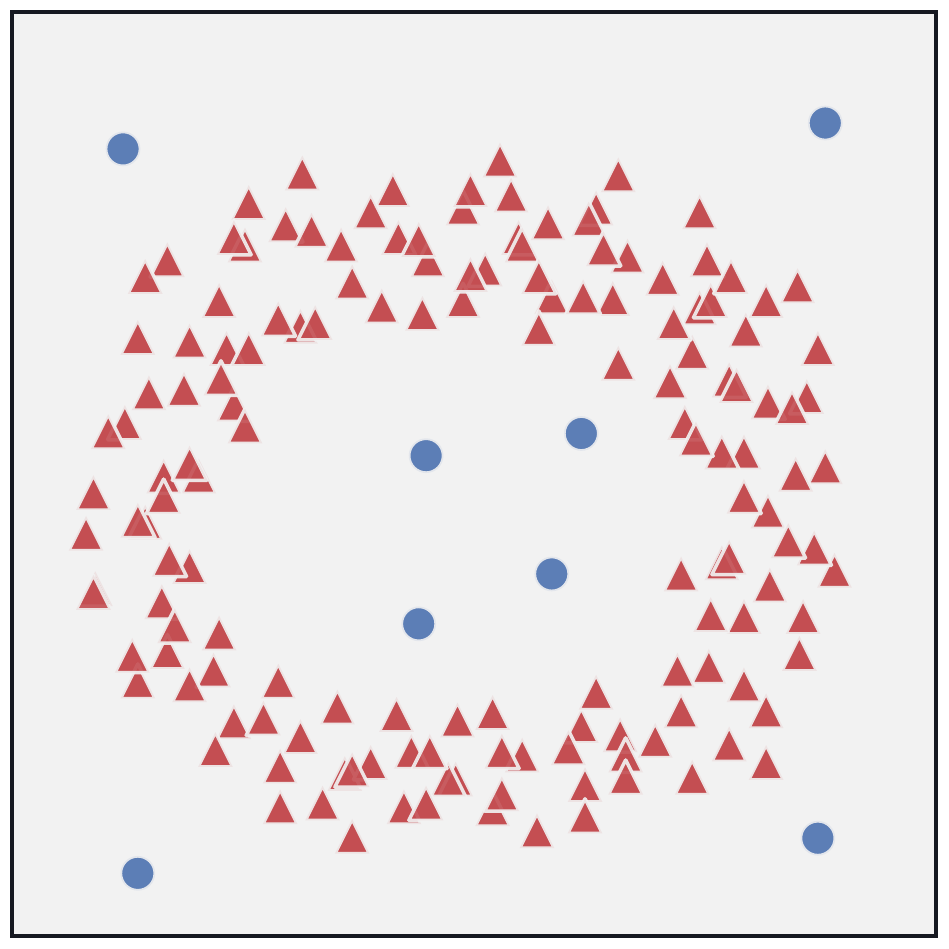}
\includegraphics[width=0.35\textwidth]{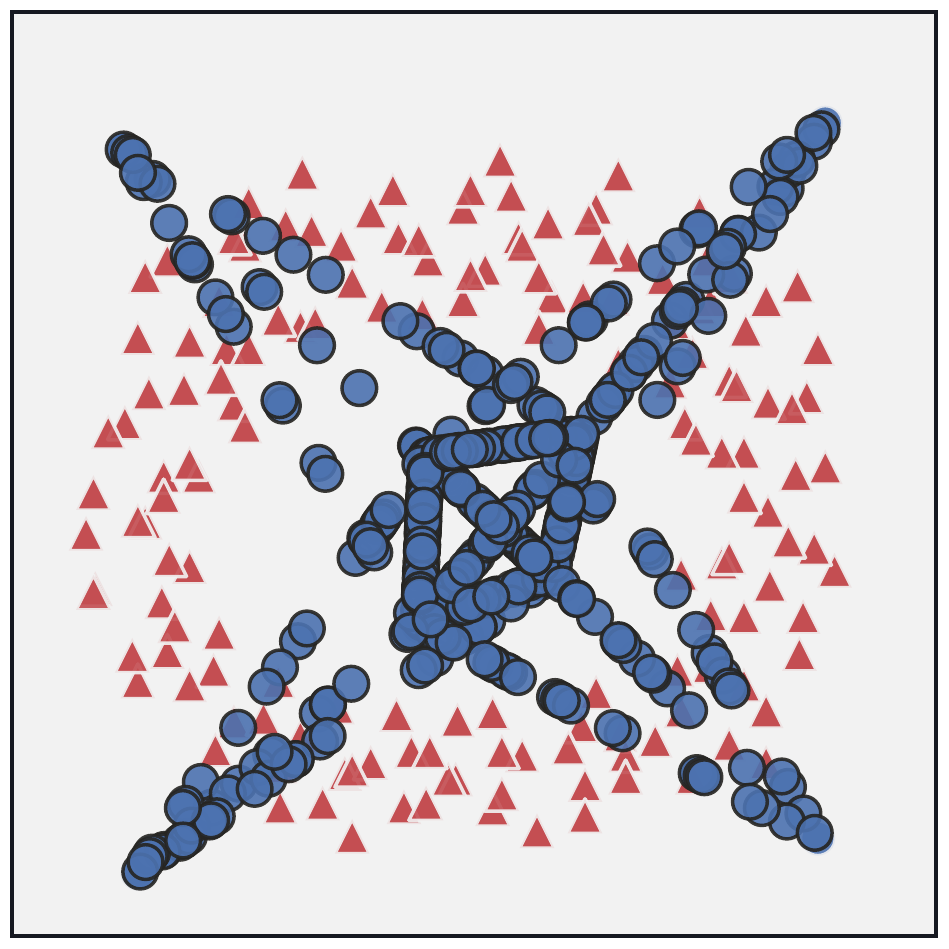}
\caption{An example of a difficult dataset. Neighborhood-based methods are not resilient to small number of minority objects, disjoint data distributions, or presence of the outliers. On the left: original imbalanced dataset. On the right: data after the oversampling with SMOTE. Synthetically generated samples overlap the original majority distribution.}
\label{fig:smote}
\end{figure*}

Conceptually simplest oversampling-based approach to dealing with the imbalanced data is random oversampling (ROS). When applying ROS new objects are being generated by duplicating randomly chosen, existing objects. The drawback of this approach is that it leads to the minority objects being grouped in small areas, in which the original objects were placed. This can present a problem for some of the classifiers, especially those prone to overfitting. SMOTE algorithm and its derivatives were proposed specifically to alleviate this issue. Instead of duplicating existing objects, SMOTE aimed at synthesizing new ones. In the original SMOTE the synthetic objects are being placed on the lines connecting the existing minority objects with its nearest minority neighbors. This lines act as a regions of interest, in which the creation of new objects is being considered. Compared to ROS, this approach leads to a more spread-out clusters of minority objects, leading to a reduced risk of overfitting. However, SMOTE makes an implicit assumption that the produced regions of interest are indeed suitable for the oversampling. It can be argued that oftentimes this is not the case. An example of a dataset for which the regions of interest produced by SMOTE might be inappropriate is presented in Figure~\ref{fig:smote}. Due to the presence of spread-out, minority outliers SMOTE's regions of interest overlap with the original majority class distribution. This behavior is intuitively harmful, as synthetically generated, overlapping objects are not representative of the original data distribution. So while the expansion of the regions of interest compared to ROS is desirable, neighborhood-based approaches might often expand them in unwarranted directions, not taking into the account the majority class distribution. This issue is prevalent in case of noisy data, for which the likelihood of observing disjoint outliers is high, especially if the noise occurs on the label level.

Some recent works suggest using kernel functions \cite{Tang:2015,Perez-Ortiz:2016} or density-based \cite{Gao:2012} solutions to achieve a more informative oversampling and overcome local instance-level difficulties. Kernel-based solutions perform a high-dimensional mapping into an artificial feature space that should lead to a better separation between classes and thus easier oversampling without the risk of increasing class overlapping. Density-based methods try to avoid using neighborhood-based introduction of artificial instances and instead focus on detecting regions of given density for minority class. Despite being a step forward from neighborhood-based methods, these techniques still suffer from a number of drawbacks. Kernel-based methods increase the dimensionality of the artificial feature space, which may further emphasize the issue of limited access to minority class instances (leading to the problem of small sample size and high dimensionality). Additionally, kernel-based approaches do not take into account individual difficulty of instances, thus allowing for noisy or rare objects to strongly influence the mapping process. On the other hand, density-based solutions use only density from the minority class, thus practically ignoring the role of majority class and suffering for similar limitations as neighborhood-based approaches when facing difficult data distributions.

\section{Multi-class imbalanced problems}

While in the binary classification, one can easily define the majority and the minority class, as well as quantify the degree of imbalance between the classes, this relationship becomes less clear in the multi-class setting. One of the earlier proposals for the taxonomy of multi-class problems used either the concept of multi-minority, that is a single majority class accompanied by multiple minority classes, or multi-majority, that is a single minority class accompanied by multiple majority classes \cite{Wang:2012}. However, in practice, the relationship between the classes tends to be more complicated, and a single class can act as a majority towards some, a minority towards others, and have a similar number of observations to the rest of the classes. Such situations are not well-encompassed by the current taxonomies. Since categorizations such as the one proposed by Napierała and Stefanowski \cite{napierala2016types} played an essential role in the development of specialized strategies for dealing with data imbalance in the binary setting, the lack of a comparable alternative for the multi-class setting can be seen as a limiting factor for the further research.

The difficulties associated with the imbalanced data classification are also further pronounced in the multi-class setting, where each additional class increases the complexity of the classification problem. This includes the problem of overlapping data distributions, where multiple classes can simultaneously overlap a particular region, and the presence of noise and outliers, where on one hand a single outlier can affect class boundaries of several classes at once, and on the other can cease to be an outlier where some of the classes are excluded. Finally, any data-level observation generation or removal must be done by a careful analysis of how action on a single class influences different types of observations in remaining classes. All of the above lead to a conclusion that algorithms designed explicitly to handle the issues associated with multi-class imbalance are required to adequately address the problem of data imbalance.

The existing methods for handling multi-class imbalance can be divided into two categories. First of all, the binarization solutions, which decompose a multi-class problem into either $M(M-1)/2$ (one-vs-one, OVO) or $M$ (one-vs-all, OVA) binary sub-problems \cite{Fernandez:2013}. Each sub-problem can then be handled individually using a selected binary algorithm. An obvious benefit of this approach is the possibility of utilization of existing algorithms \cite{Zhang:2018}. However, binarization solutions have several significant drawbacks.

Most importantly, they suffer from the loss of information about class relationships. In essence, we either completely exclude the remaining classes in a single step of OVO decomposition or discard the inner-class relations by merging classes into a single majority in OVA decomposition. Furthermore, especially in the case of OVO decomposition associated computational cost can quickly grow with the number of classes and observations, making the approach ill-suited for dealing with the big data. On the other hand, OVA decomposition can further exacerbate the level of data imbalance, even on the originally balanced datasets: for instance, in the case of three-class dataset with an equal number of observations belonging to each class, applying OVA would group two of the classes together, introducing a 2:1 artificial imbalance. Among the binarization solutions, the recent literature suggests the efficacy of using ensemble methods with OVO decomposition \cite{Zhang:2016}, augmenting it with cost-sensitive learning \cite{Krawczyk:2016ijcnn}, or applying dedicated classifier combination methods \cite{Japkowicz:2015}.

The second category of methods consists of ad-hoc solutions: techniques that treat the multi-class problem natively, proposing dedicated solutions for exploiting the complex relationships between the individual classes. Ad-hoc solutions require either a significant modification to the existing algorithms, or exploring an entirely novel approach to overcoming the data imbalance, both on the data and the algorithm level. However, they tend to significantly outperform binarization solutions, offering a promising direction for further research. Most-popular data-level approaches include extensions of the SMOTE algorithm into a multi-class setting \cite{Fernandez-Navarro:2011,Saez:2016,Zhu:2017}, strategies using feature selection \cite{Cao:2017,Wu:2017}, and alternative methods for instance generation by using Mahalanobis distance \cite{Abdi:2016,Yang:2017}. Algorithm-level solutions include decision tree adaptations \cite{Hoens:2012}, cost-sensitive matrix learning \cite{Bernard:2016}, and ensemble solutions utilizing Bagging \cite{Lango:2018,Collell:2018} and Boosting \cite{Wang:2012,Guo:2016}. It is also worth mentioning the works employing association rule mining techniques for multi-class imbalanced data classification, such as the proposition of  Huaifeng et al. \cite{ Huaifeng:2007} dedicated to discovering an efficient association rule for highly imbalanced data. In \cite{ Nguyen:2019} CARs algorithm has been proposed for multi-class imbalanced data, which employs k-means clustering algorithm and association rule generation for each cluster. 

%% file: Sources/Binary.tex
\chapter{Binary resampling strategies}
\label{chapter:binary}

\begin{center}
  \begin{minipage}{0.5\textwidth}
    \begin{small}
      In which the proposed binary resampling algorithms are described.
    \end{small}
  \end{minipage}
  \vspace{0.5cm}
\end{center}

\noindent Binary imbalanced classification problems are conceptually simpler than their multiclass counterparts due to the fact that the class of interest is clearly defined, and the degree of either boosting (in the case of oversampling) or reduction (in the case of undersampling) can, to some extent, be deduced based on the imbalance ratio of the two classes. As a result, it is often the case that novel resampling algorithms are first developed with the binary problems in mind, and later extended to the multiclass setting. The same approach was adopted in the development of the algorithms described in the rest of this chapter.

In total five different binary over- and undersampling algorithms were proposed as a part of this thesis. In the remainder of this chapter the motivation behind each of them is briefly outlined, and the algorithms themselves are presented. More detailed discussion, as well as the results of the experimental studies empirically confirming the usefulness of each of the proposed algorithms, can be found in the corresponding papers.

\section{CCR: Combined Cleaning and Resampling}

The Combined Cleaning and Resampling (CCR) \cite{koziarski2017ccr} algorithm is an energy-based approach to modeling the regions suitable for oversampling, the aim of which is being less affected by small disjuncts and outliers than SMOTE. The CCR algorithm is based on two observations. Firstly, the fact that class imbalance does not make the classification problem difficult by itself. This might be easily illustrated by an example of a highly imbalanced, but linearly separable dataset. In such a case finding the decision border leading to the perfect accuracy will not be a problem for the most classifiers. It is only when we deal with noisy data, complicated distributions or insufficient number of observations that imbalance further exacerbates the difficulty of the classification task. Secondly, that in most problems achieving better accuracy on the minority class is the most pressing issue. Data imbalance mainly lowers classifiers' performance on the examples from the minority class, leaving precision in large part unaffected. At the same time, misclassification of the minority examples is often more costly in the practical applications such as medical diagnosis or fraud detection. Therefore, while we would like to achieve highest possible accuracy for all the classes, in practice sacrificing some of the precision to improve recall is often desirable.

As the name indicates, the proposed algorithm consists of two distinct steps. Firstly, cleaning the neighborhoods of the minority samples from the majority objects. The aim of this step is to simplify the task of classification of the examples from the minority class. Secondly, selectively generating the synthetic samples, with highest number of synthetic objects created near the least safe observations. In the remainder of this section a thorough description of both the cleaning and sample generating steps is given.

\noindent\textbf{Cleaning the minority neighborhoods.} As a step preceding the oversampling itself, CCR proposes performing a data preprocessing in the form of cleaning the majority observations located in proximity to the minority instances. The aim of such an operation is twofold. First of all, to reduce the problem of class overlap: by designing the regions from which majority observations are being removed, the original dataset is transformed with an intention of simplifying it for further classification. Secondly, to skew the classifiers' predictions towards the minority class: since in the case of the imbalanced data such regions, bordering two-class distributions or consisting of overlapping instances, tend to produce predictions biased towards the majority class. By performing clean-up, this trend is either reduced or reversed.

Two key components of such cleaning operation are a mechanism of the designation of regions from which the majority observations are to be removed, and a removal procedure itself. The former, especially when dealing with data affected by label noise, should be able to adapt to the surroundings of any given minority observation, and adjust its behavior depending on whether the observation resembles a mislabeled instance or a legitimate outlier from an underrepresented region, which is likely to occur in the case of imbalanced data with scarce volume. The later should limit the loss of information that could occur due to the removal of a large number of majority observations.

\begin{figure}
\centering
\includegraphics[width=0.4\linewidth]{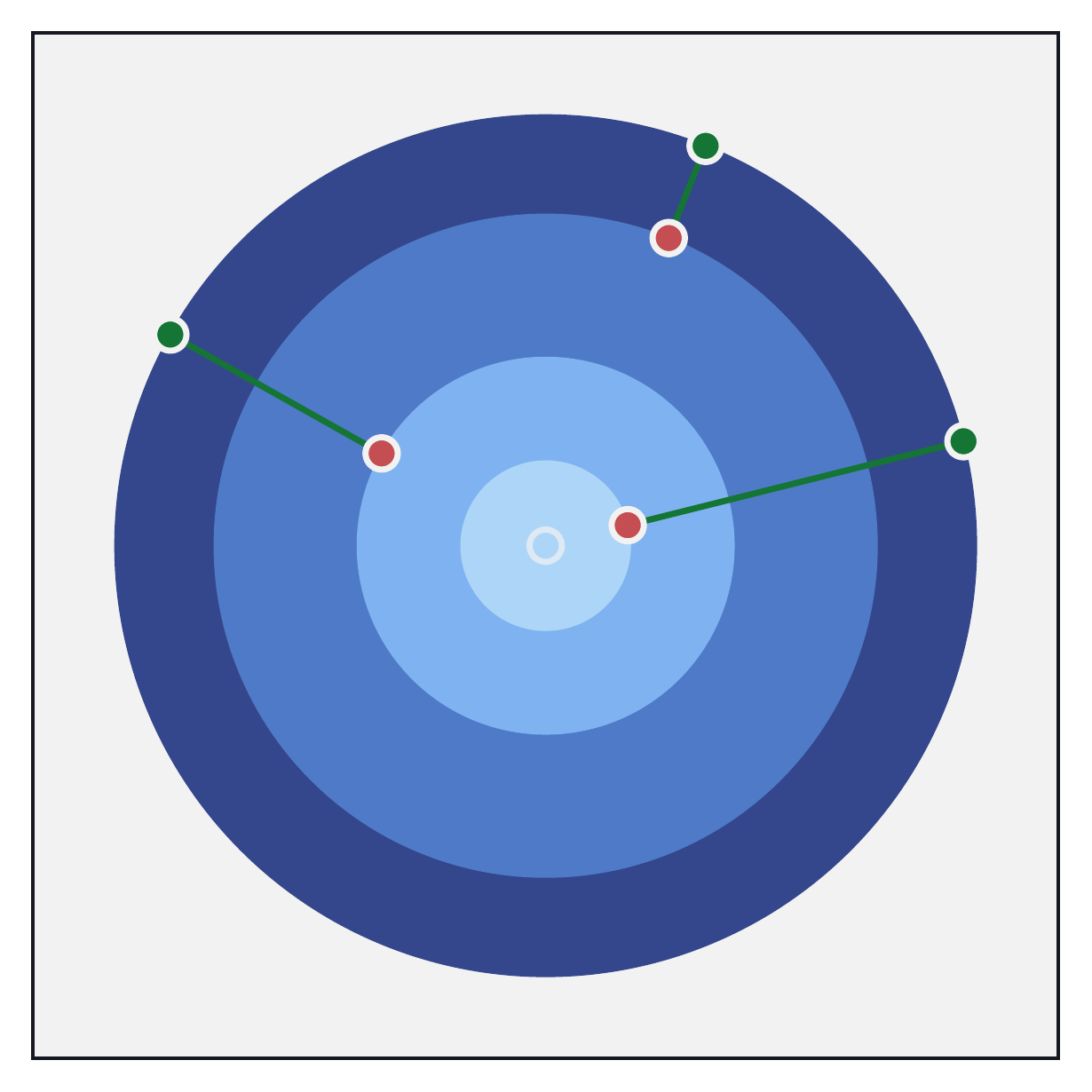}
\caption{An illustration of the sphere creation for an individual minority observation (in the center) surrounded by majority observations (in red). Sphere expends at a normal cost until it reaches a majority observation, at which point the further expansion cost increases (depicted by blue orbits with an increasingly darker color). Finally, after the expansions, the majority observations within the sphere are being pushed outside (in green).}
\label{fig:cleaning}
\end{figure}

To implement such preprocessing in practice, CCR proposes an energy-based approach, in which spherical regions are constructed around every minority observation. Spheres expand using the available energy, a parameter of the algorithm, with the cost increasing for every majority observation encountered during the expansion. More formally, for a given minority observation denoted by $x_i$, current radius of an associated sphere denoted by $r_i$, a function returning the number of majority observations inside a sphere centered around $x_i$ with radius $r$ denoted by $f_n(r)$, a target radius denoted by $r_i'$, and $f_n(r_i') = f_n(r_i) + 1$, the energy change caused by the expansion from $r_i$ to $r_i'$ is defined as
\begin{equation}
    \Delta e = - (r_i' - r_i) \cdot f_n(r_i').
\end{equation}
During the sphere expansion procedure, the radius of a given sphere increases up to the point of completely depleting the energy, with the cost increases after each encountered majority observation. Finally, the majority observations inside the sphere are being pushed out to its outskirts. The whole process is illustrated in Figure~\ref{fig:cleaning}.

The proposed cleaning approach meets both of the outlined criteria. First of all, due to the increased expansion cost after each encountered majority observation, it distinguishes the likely mislabeled instances: minority observations surrounded by a large number of majority observations lead to a creation of smaller spheres and, as a result, more constrained cleaning regions. On the other hand, in case of overlapping class distributions, or other words in the presence of a large number of both minority and majority observations, despite the small size of individual spheres, their large volume still leads to large cleaning regions. Secondly, since the majority observations inside the spheres are being translated instead of being completely removed, the information associated with their original positions is to a large extent preserved, and the distortion of class density in specific regions is limited.

\noindent\textbf{Selectively oversampling the minority class.} After the cleaning stage is concluded, new synthetic minority observations are being generated. To further exploit the spheres created during the cleaning procedure, new synthetic instances are being sampled within the previously designed cleaning regions. This not only prevents the synthetic observations from overlapping the majority class distribution but also constraints the oversampling areas for observations displaying the characteristics of mislabeled instances. 

Moreover, in addition to designating the oversampling regions, CCR proposes employing the size of the calculated spheres in the process of weighting the selection of minority observations used as the oversampling origin. Analogous to the ADASYN \cite{He:2008}, CCR focuses on the difficult observations, with difficulty estimated based on the radius of an associated sphere. More formally, for a given minority observation denoted by $x_i$, the radius of an associated sphere denoted by $r_i$, the vector of all calculated radii denoted by $r$, collection of majority observations denoted by $\mathcal{X}_{maj}$, collection of minority observations denoted by $\mathcal{X}_{min}$, and assuming that the oversampling is performed up to the point of achieving balanced class distribution, the number of synthetic observations to be generated around $x_i$ is defined as
\begin{equation}
    \label{eq:prop}
    g_i = \lfloor\dfrac{r_i^{-1}}{\sum_{k = 1}^{|\mathcal{X}_{min}|}{r_k^{-1}}} \cdot (|\mathcal{X}_{maj}| - |\mathcal{X}_{min}|)\rfloor.
\end{equation}
Just like in the ADASYN, such weighting aims to reduce the bias introduced by the class imbalance and to shift the classification decision boundary toward the difficult examples adaptively. However, compared to the ADASYN, in the proposed method the relative distance of the observations plays an important role: while in the ADASYN outlier observations, located in a close proximity of neither majority nor minority instances, based on their far-away neighbors could be categorized as difficult, that is not the case under the proposed weighting, where the full sphere expansion would occur.

\begin{figure*}
\centering
        \includegraphics[width=0.24\textwidth]{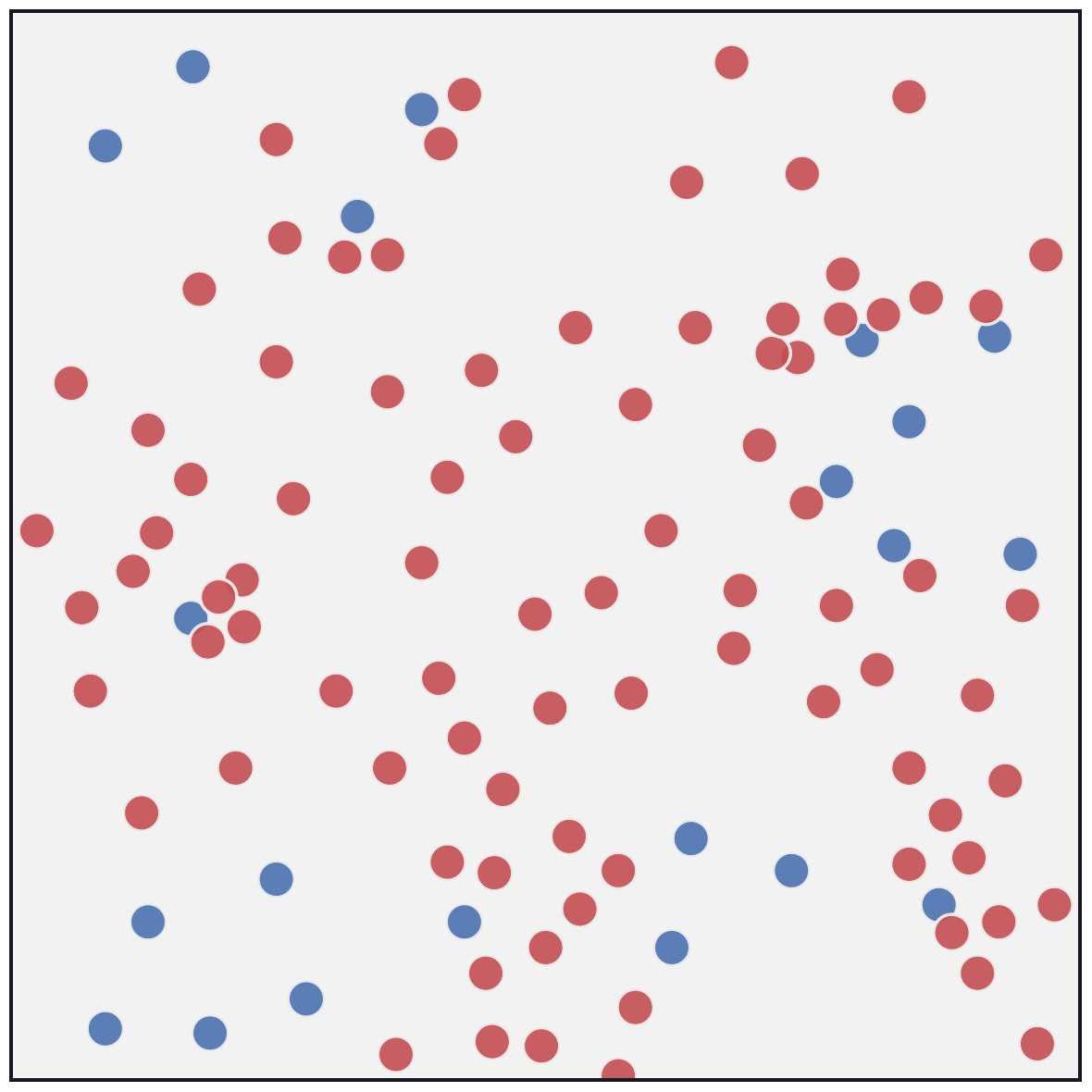}
        \includegraphics[width=0.24\textwidth]{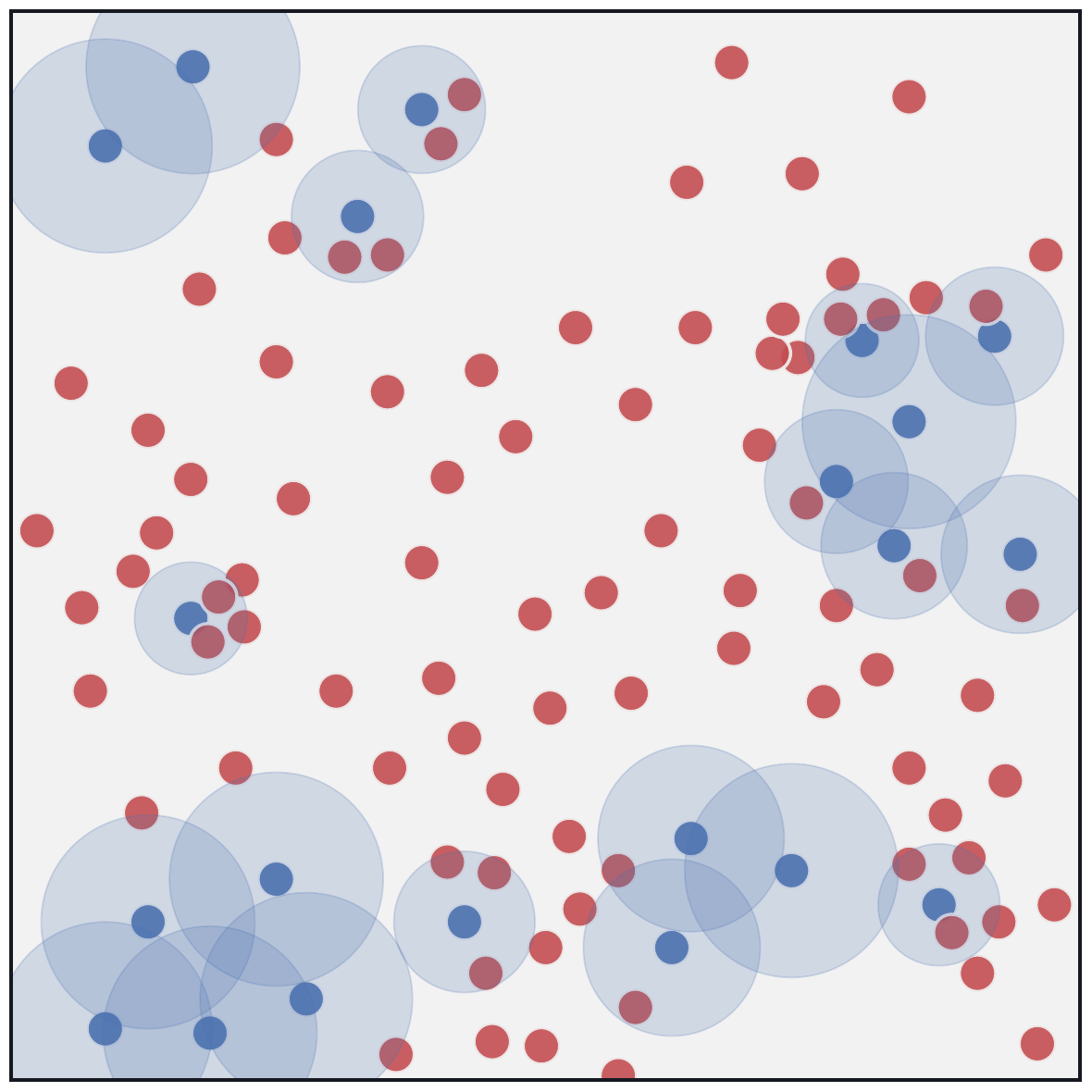}
        \includegraphics[width=0.24\textwidth]{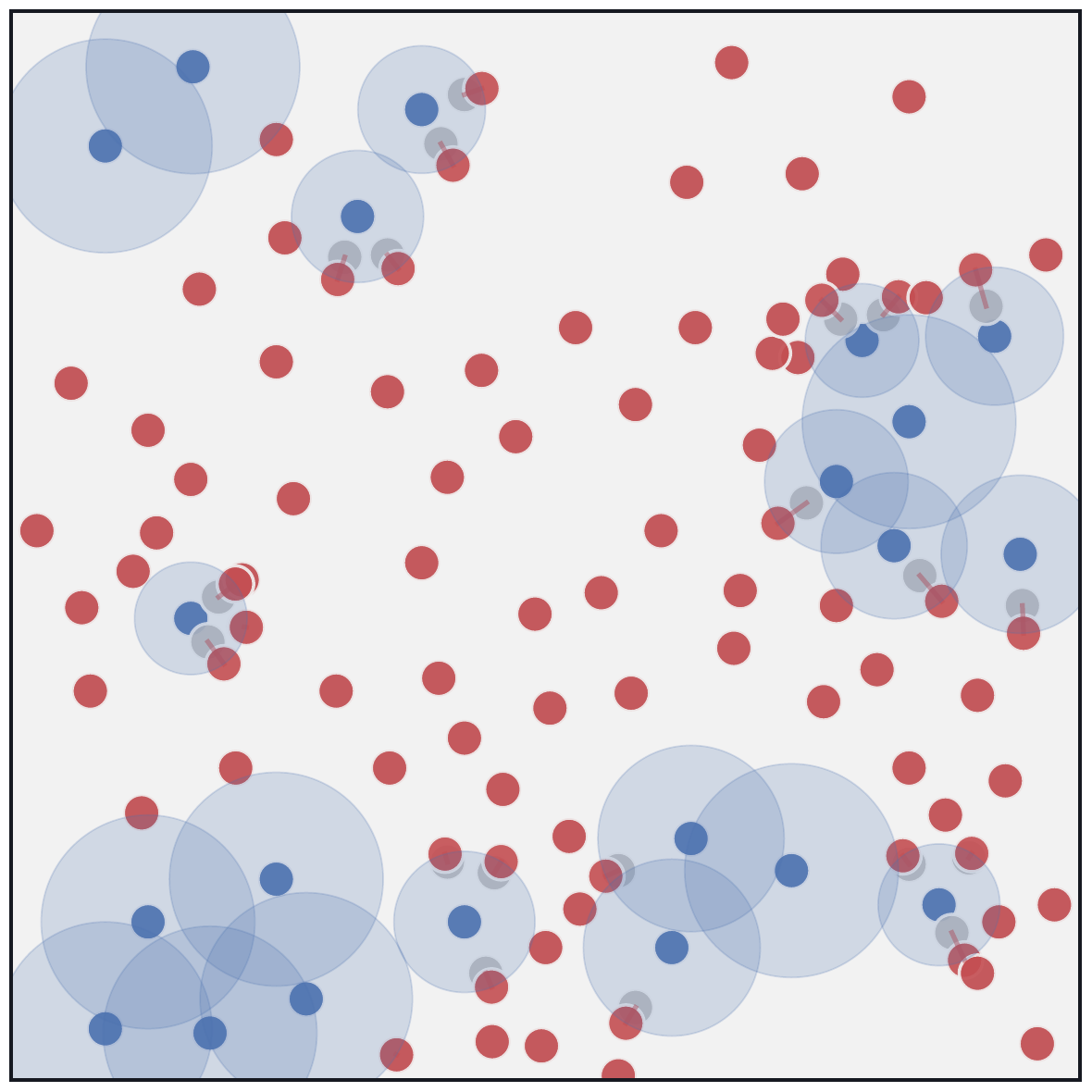}
        \includegraphics[width=0.24\textwidth]{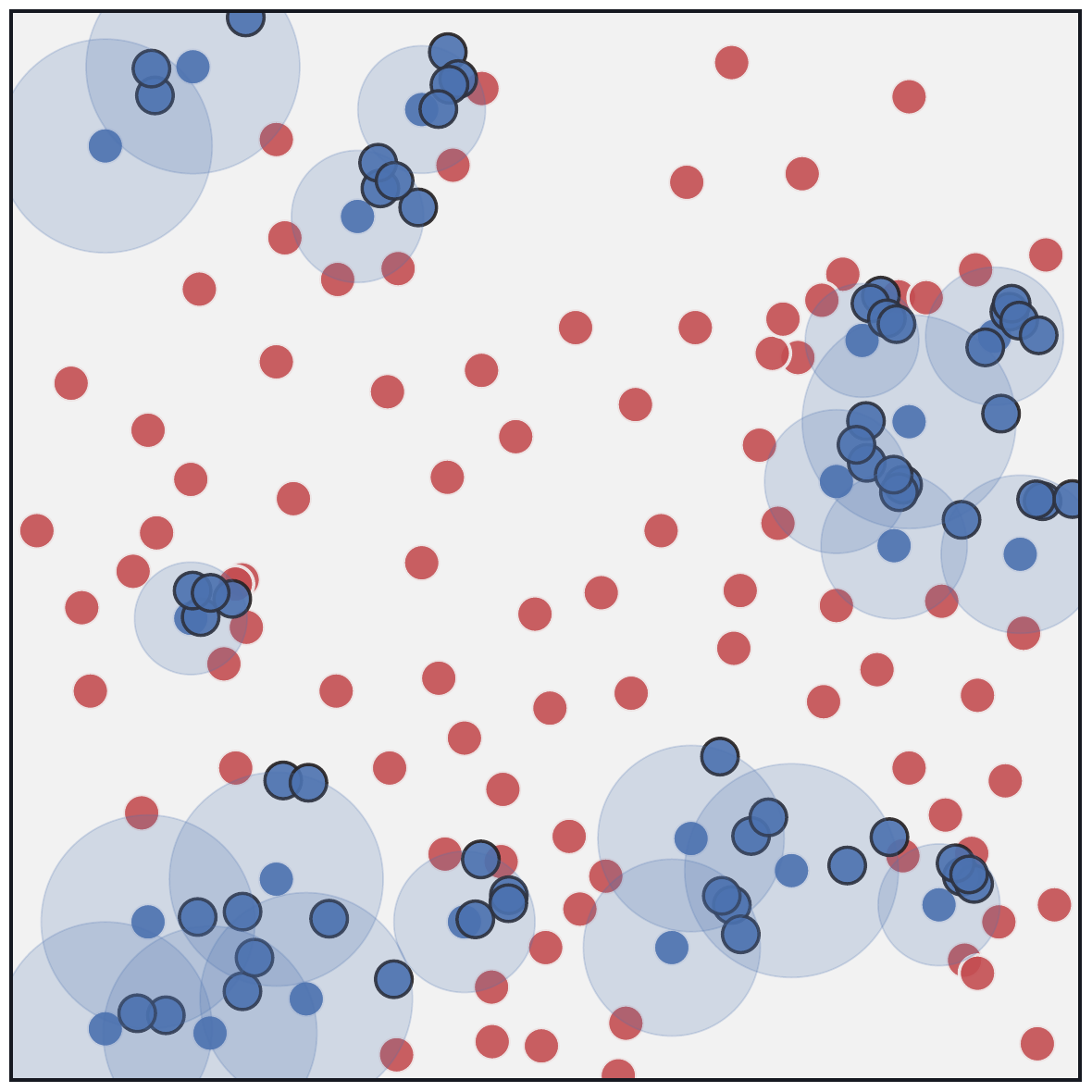}
\caption{An illustration of the algorithms behavior in a binary case. From the left: 1) original dataset, 2) sphere calculation for individual minority objects, with smaller spheres created for the observations surrounded by the majority objects, 3) pushing out the majority objects outside the sphere radius, 4) generating synthetic minority observations inside the spheres, in the number inversely proportional to the sphere radius.}
\label{fig:b-ccr}
\end{figure*}

\noindent\textbf{Combined algorithm.} A complete pseudocode of the proposed method is presented in Algorithm~\ref{algorithm:b-ccr}. Furthermore, the behavior of the algorithm in a binary case is illustrated in Figure~\ref{fig:b-ccr}. All three major stages of the proposed procedure are outlined: forming spheres around the minority observations, clean-up of the majority observations inside the spheres, and adaptive oversampling based on the sphere radii.

\begin{algorithm}[!htb]
	\caption{Combined Cleaning and Resampling}
	\footnotesize
	
	\textbf{Input:} collections of majority observations $\mathcal{X}_{maj}$ and minority observations $\mathcal{X}_{min}$ \\
	\textbf{Parameters:} $energy$ budget for sphere expansion, $p$-norm used for distance calculation \\
    \textbf{Output:} translated majority observations $\mathcal{X}_{maj}'$ and synthetic minority observations $S$
    
	\label{algorithm:b-ccr}
	
	\vspace{-0.5\baselineskip}
	\hrulefill
	\begin{algorithmic}[1]
		\STATE \textbf{function} CCR($\mathcal{X}_{maj}$, $\mathcal{X}_{min}$, $energy$, $p$):
		\STATE $S \gets \emptyset$ \COMMENT{synthetic minority observations}
		\STATE $t \gets $ zero matrix of size $|\mathcal{X}_{maj}| \times m$, with $m$ denoting the number of features \COMMENT{translations of majority observations}
		\STATE $r \gets $ zero vector of size $|\mathcal{X}_{min}|$ \COMMENT{radii of spheres associated with the minority observations}
        \FORALL{minority observations $x_i$ in $\mathcal{X}_{min}$}
        \STATE $e$ $\gets$ $energy$ 
        \STATE $n_r \gets 0$ \COMMENT{number of majority observations inside the sphere generated around $x_i$}
        \FORALL{majority observations $x_j$ in $\mathcal{X}_{maj}$}
        \STATE $d_j \gets \lVert x_i - x_j \rVert_p$
        \ENDFOR
        \STATE sort $\mathcal{X}_{maj}$ with respect to $d$
        \FORALL{majority observations $x_j$ in $\mathcal{X}_{maj}$}
        \STATE $n_r \gets n_r + 1$
        \STATE $\Delta e \gets - (d_j - r_i) \cdot n_r$
        \IF{$e + \Delta e > 0$}
        \STATE $r_i \gets d_j$
        \STATE $e \gets e + \Delta e$
        \ELSE
        \STATE $r_i \gets r_i + \frac{e}{n_r}$
        \STATE \textbf{break}
        \ENDIF
        \ENDFOR
        \FORALL{majority observations $x_j$ in $\mathcal{X}_{maj}$}
        \IF{$d_j < r_i$}
        \STATE $t_j \gets t_j + \dfrac{r_i - d_j}{d_j} \cdot (x_j - x_i)$
        \ENDIF
        \ENDFOR
        \ENDFOR
        \STATE $\mathcal{X}_{maj}' \gets \mathcal{X}_{maj} + t$
        \FORALL{minority observations $x_i$ in $\mathcal{X}_{min}$}
        \STATE $g_i \gets \lfloor\dfrac{r_i^{-1}}{\sum_{k = 1}^{|\mathcal{X}_{min}|}{r_k^{-1}}} \cdot (|\mathcal{X}_{maj}| - |\mathcal{X}_{min}|)\rfloor$ 
        \FOR{1 \textbf{to} $g_i$}
        \STATE $v \gets$ random point inside a zero-centered sphere with radius $r_i$
		\STATE $S \gets S \cup \{x_i + v\}$
        \ENDFOR
        \ENDFOR
		\STATE \textbf{return} $\mathcal{X}_{maj}'$, $S$
	\end{algorithmic}
\end{algorithm}

\noindent\textbf{Main results.} During the experimental comparison described in \cite{koziarski2017ccr} it was demonstrated that CCR tends to outperform the considered reference methods, in particular when combined with the decision trees as a classification algorithm. More specifically, CCR achieved a significantly better recall at the cost of precision of the predictions, which tends to be a favorable trade-off when dedicated imbalance classification metrics, such as AUC and G-mean, are considered. Furthermore, during the described experimental study it was also demonstrated that CCR is heavily reliant on a proper choice of the energy parameter, making the parameter selection a crucial step for achieving the optimal performance.

\section{RBO: Radial-Based Oversampling}

The Radial-Based Oversampling (RBO) \cite{koziarski2017hais,koziarski2019neuro} algorithm is directly motivated by the shortcomings of SMOTE and other neighborhood-based oversampling strategies previously discussed in Section~\ref{sec:limitations}. An essential step in oversampling strategies that synthetically generate new objects is establishing the regions of interest, in which the new objects should be introduced. SMOTE and its derivatives achieve that by connecting the nearby minority objects and generating the new objects alongside those connections. While conceptually very simple, this approach does not take into the account neither the position of existing majority class objects, nor the distance between the connected minority objects. As a result, generated regions of interest can overlap areas with a high density of the majority class objects. Intuitively, this is often not a desired behavior, since the goal of the oversampling is to generate minority objects reflecting the whole data distribution.

\noindent\textbf{Potential estimation with radial basis functions.} To mitigate this issue, RBO proposes an alternative approach for approximating the regions of interest. Instead of generating binary regions between the minority objects, RBO uses a real-valued potential surface, with potential at each point in space representing our preference towards that point belonging to either minority or majority class. To calculate that potential, RBO assigns a Gaussian radial basis function (RBF) to every object in the training dataset, with the polarity dependent on its class. Using the concept of Gaussian RBF, the potential function is defined as follows: given a collection of observations $\mathcal{X}$, parameter $\gamma$ representing the spread of a single RBF, a point in space $x$, and using the L1 norm as a distance measure, the potential function is defined as
\begin{equation}
\Phi(x, \mathcal{X}, \gamma) = \sum_{i=1}^{\mid \mathcal{X} \mid}{e^{-\left(\frac{\lVert \mathcal{X}_i - x \rVert_1}{\gamma}\right)^{2}}}\label{eq:potential}.
\end{equation}
\noindent
The subcollection of observations $\mathcal{X}$ belonging to class $c$ is further denoted by $\mathcal{X}^{(c)}$. Of particular interest is the \textit{class potential}, the potential computed on the collection of observations belonging to a specific class $\mathcal{X}^{(c)}$, and the \textit{mutual class potential}, the difference between the potential of two different classes. While discussing the potentials, in the binary setting the following convention is assumed: high mutual majority class potential indicates the situation, in which the class potential for the majority class is higher than the potential of the minority class, and high mutual minority class potential indicates the opposite.

\begin{figure*}
\centering
\includegraphics[width=0.35\textwidth]{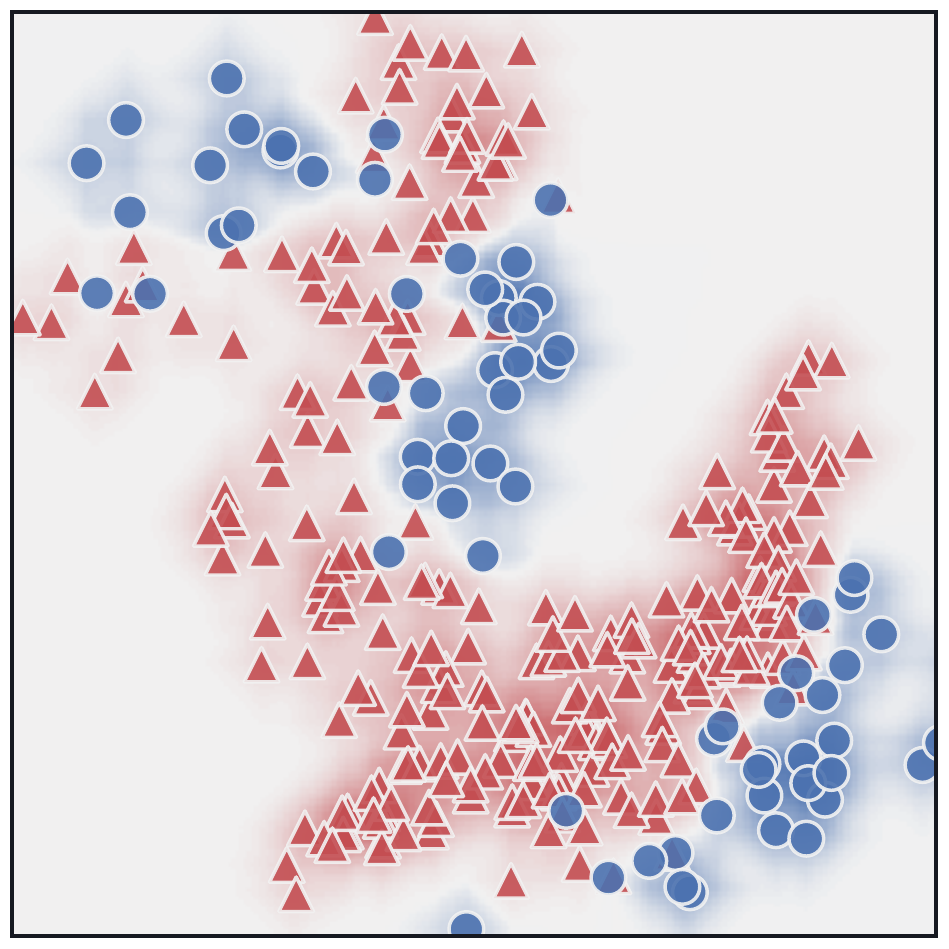}
\includegraphics[width=0.35\textwidth]{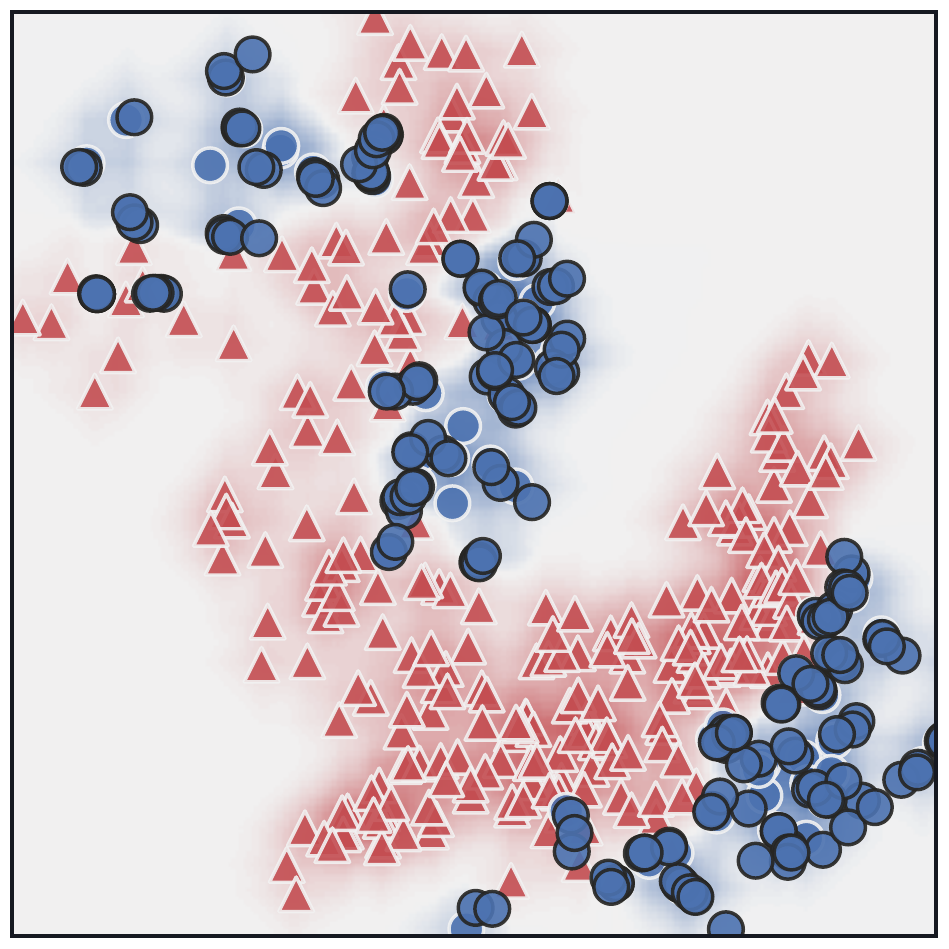}
\caption{On the left: visualized potential function for the original dataset. On the right: the same potential with an addition of the synthetic samples generated by the \textit{Radial-Based Oversampling} algorithm.}
\label{fig:rbo-potential}
\end{figure*}

An example of a potential surface for a two-dimensional dataset is presented in Figure~\ref{fig:rbo-potential}. Minority objects, as well as associated with them negative potentials are represented by a blue color, whereas majority objects and associated with them positive potentials are represented by a red color. Potentials close to zero are indicated by a gray color.

Compared to the regions of interest generated by SMOTE, potential surface has several advantages. First of all, it is resilient to the presence of the outliers, as well as a small number of minority objects combined with disjoint distributions. While in those cases SMOTE is likely to generate synthetic objects over the clusters of a majority objects, using the potential surface would result in a smaller, constrained regions of a negative potential. Secondly, compared to the regions of interest used by SMOTE, potential surface contains more information that can be used to guide more sophisticated placement strategies. For instance, depending on the task at hand and the preference towards either the precision or the recall, synthetic minority objects might be placed either at the regions with a potential close to zero, or those with a very high potential. The former might be interpreted as an uncertain area, and placing new objects in it should move the decision border in favor of the minority class, whereas the latter represents a higher certainty, and should lead to a more conservative decision border.

\noindent\textbf{Oversampling procedure.} Based on a potential surface, various strategies of imbalanced data resampling can be designed. In Radial-Based Oversampling an algorithm that creates artificial objects as a product of an iterative optimization is proposed. RBO optimizes the absolute value of the potential, which corresponds to placing the synthetic minority objects in regions of high uncertainty, close to the predicted decision border. Two alternative optimization criteria were considered, namely the maximization and the minimization of potentials. The final choice is motivated by the intuitive soundness of objects generated in the local minima, especially when compared to the considered alternatives, as well as the visual examination of the datasets oversampled with the different optimization criteria. In all the cases, a random minority objects are chosen as a starting points for the optimization. The maximization of the potential, if not constrained by the distance, would lead to oversampling over the clusters of majority points, the exact behavior that was described as disadvantageous in SMOTE. On the other hand, the minimization of the potential would often be immediately stuck in a local minima, since the existing minority objects are often associated with an especially low potential.

The optimization procedure itself is based on a simple hill climbing algorithm. Starting by generating a new object in a position associated with a randomly chosen minority object, at every iteration the object is translated in a random direction, and the translation is preserved only if the absolute value of the potential decreases. The procedure lasts for a number of iterations given as a parameter of the algorithm, with a small probability of an early stopping, also given as a parameter. The addition of the early stopping possibility is motivated by the idea of covering the considered region of interest more evenly, instead of placing them only on the decision borders. Importantly, during the oversampling, in the proposed version of the algorithm the newly generated, synthetic minority instances are not being considered during the potential calculation. It was observed that synthetic instances could cause a significant drift in the original potential, a behavior deemed potentially dangerous, especially in the case of an extreme imbalance.

Complete pseudocode of the described oversampling strategy is presented in Algorithm~\ref{algorithm:rbo}. In the proposed variant, the oversampling was performed up to the point of balancing the class distributions. An example of an originally imbalanced dataset after applying the RBO was presented in Figure~\ref{fig:rbo-potential}. As can be seen, in the case of a singular outliers, new objects are being generated in a close proximity, and the overlapping of the majority clusters is limited. On the other hand, in the remaining cases synthetically generated objects are being spread across the regions of interest, with a preference towards the difficult areas. It can be argued that in case of a singular outliers, instead of limiting the distance within which the synthetic objects can be generated, the outliers could be ignored completely and regarded as a noise. However, especially in the case of extreme imbalance, such behavior could lead to omitting important instances. Therefore, instead of making an assumption about the nature of the outliers on the level of the oversampling algorithm, it is instead recommended to use data cleaning prior to oversampling when the presence of noise is suspected. 

\begin{algorithm}[!htb]
	\caption{Radial-Based Oversampling}
	\textbf{Input:} collections of majority observations $\mathcal{X}_{maj}$ and minority observations $\mathcal{X}_{min}$ \\
	\textbf{Parameters:} spread of radial basis function $\gamma$, optimization $step$, number of $iterations$ per synthetic observation, $k$ nearest neighbors used for potential approximation \\
    \textbf{Output:} collection of synthetic minority observations $S$
	\label{algorithm:rbo}	
	\vspace{-0.5\baselineskip}
	
	\hrulefill
	\begin{algorithmic}[1]
		\STATE \textbf{function} RBO($\mathcal{X}_{maj}$, $\mathcal{X}_{min}$, $\gamma$, $step$, $iterations$, $k$):
		\STATE $S \gets \emptyset$
		\STATE find $k$ nearest neighbors from $\mathcal{X}_{maj} \cup \mathcal{X}_{min}$ for every observation in $\mathcal{X}_{min}$
		\WHILE{$\vert \mathcal{X}_{min} \vert + \vert S \vert < \vert \mathcal{X}_{maj} \vert$}
		\STATE $x^0 \gets $ randomly chosen observation from $\mathcal{X}_{min}$
		\STATE $\mathcal{X}^0_{maj}$, $\mathcal{X}^0_{min} \gets$ collections of majority and minority observations belonging to $k$ nearest neighbors of $x^0$
		\STATE $\Phi^0 \gets \Phi(x^0, \mathcal{X}^0_{maj}, \gamma) - \Phi(x^0, \mathcal{X}^0_{min}, \gamma)$
		\FOR{$i \gets 1$ \textbf{to} $iterations$}
		\STATE $direction \gets $ randomly chosen standard basis vector in a $m$-dimensional Euclidean space, with $m$ being the number of features
		\STATE $sign \gets $ randomly chosen value from $\{-1, 1\}$
		\STATE $x^\prime \gets x^0 + direction \times sign \times step$
		\STATE $\Phi^\prime \gets \Phi(x^\prime, \mathcal{X}^0_{maj}, \gamma) - \Phi(x^\prime, \mathcal{X}^0_{min}, \gamma)$
		\IF{$|\Phi^\prime| < |\Phi^0|$}
		\STATE $x^0 \gets x^\prime$
		\STATE $\Phi^0 \gets \Phi(x^\prime, \mathcal{X}^0_{maj}, \gamma) - \Phi(x^\prime, \mathcal{X}^0_{min}, \gamma)$
		\ENDIF
		\ENDFOR
		\STATE $S \gets S \cup \{x^0\}$
		\ENDWHILE
		\STATE \textbf{return} $S$
	\end{algorithmic}
\end{algorithm}

\begin{figure*}
\centering
    \begin{subfigure}[t]{0.25\textwidth}
        \centering
        \includegraphics[width=1\textwidth]{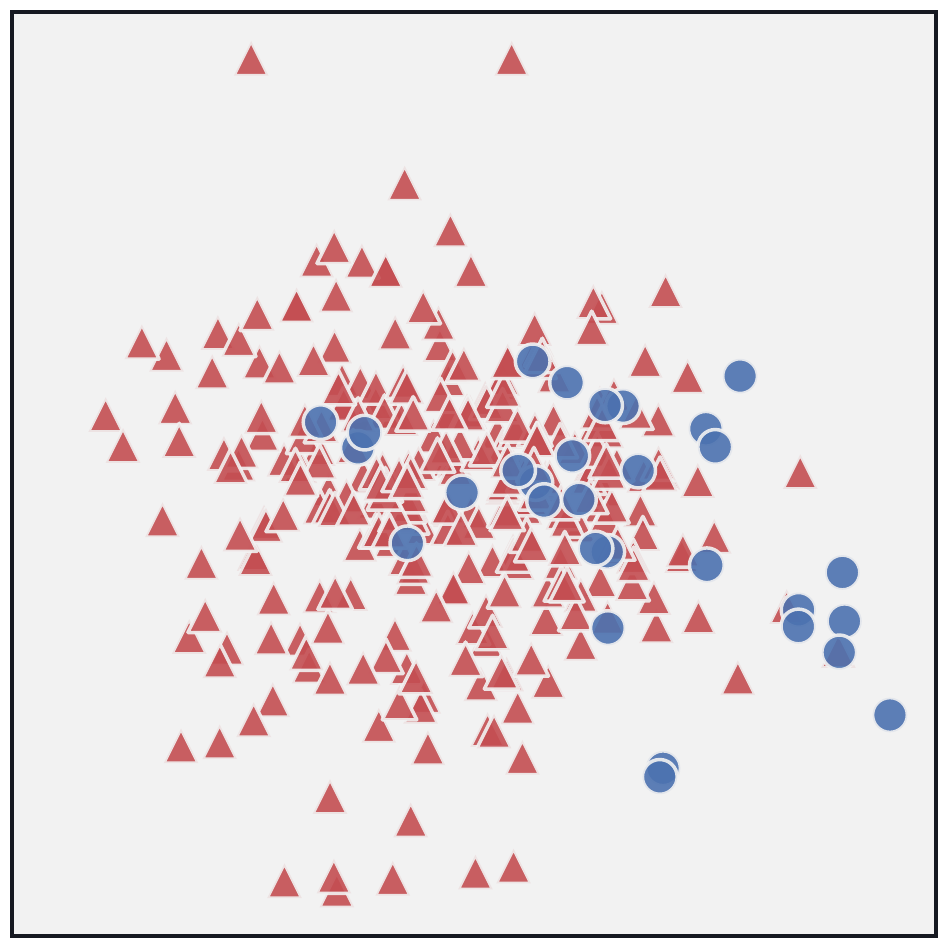}
        \caption{Initial dataset.}
    \end{subfigure}%
    ~ 
    \begin{subfigure}[t]{0.25\textwidth}
        \centering
        \includegraphics[width=1\textwidth]{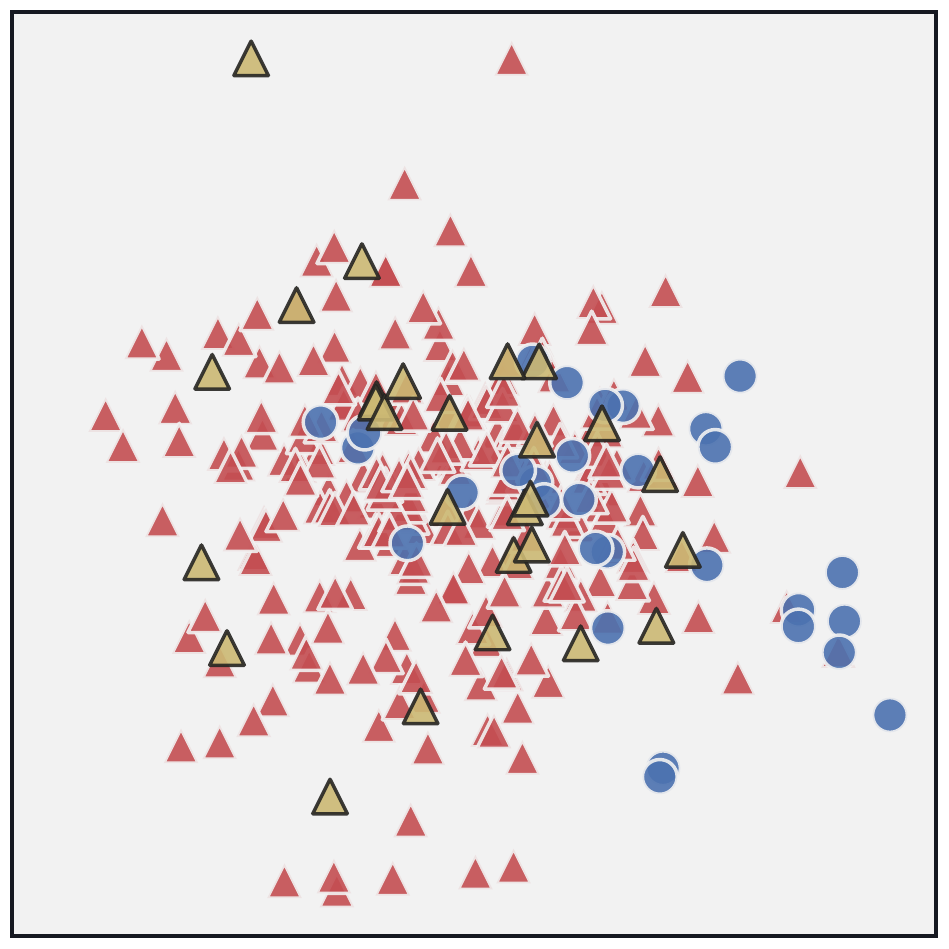}
        \caption{Class label noise introduced.}
    \end{subfigure}
     ~ 
    \begin{subfigure}[t]{0.25\textwidth}
        \centering
        \includegraphics[width=1\textwidth]{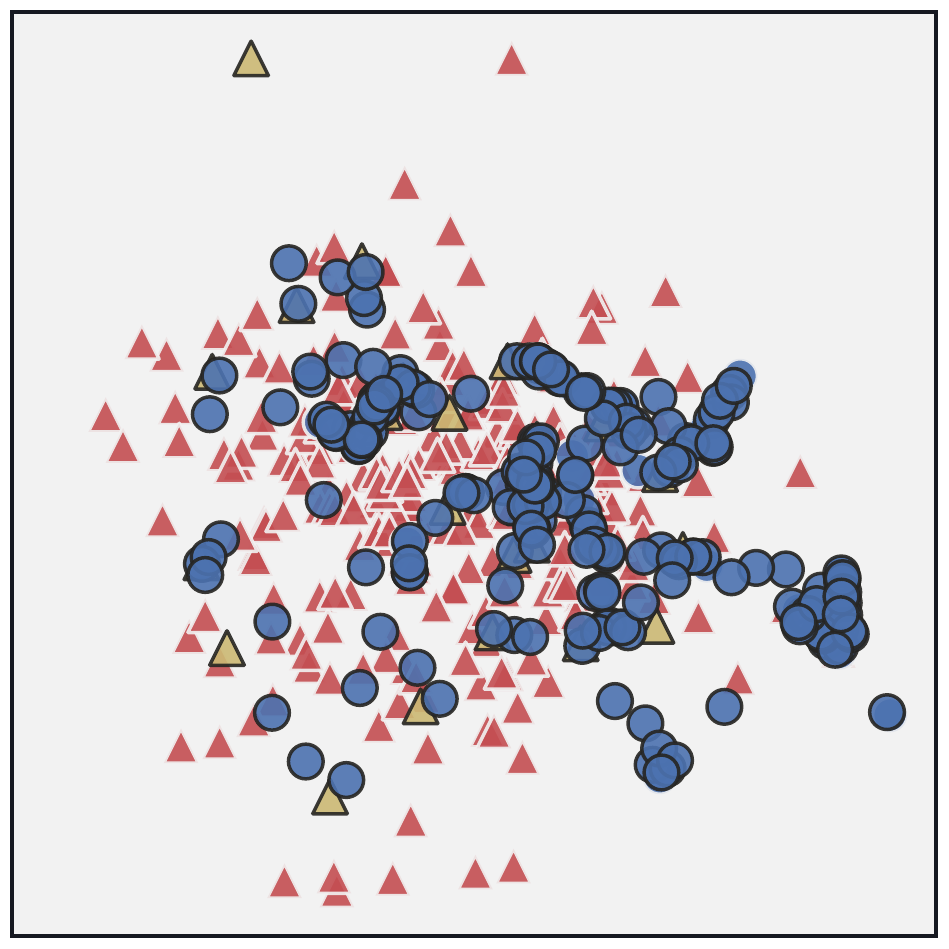}
        \caption{Effects of SMOTE.}
    \end{subfigure}
     ~ 
    \begin{subfigure}[t]{0.25\textwidth}
        \centering
        \includegraphics[width=1\textwidth]{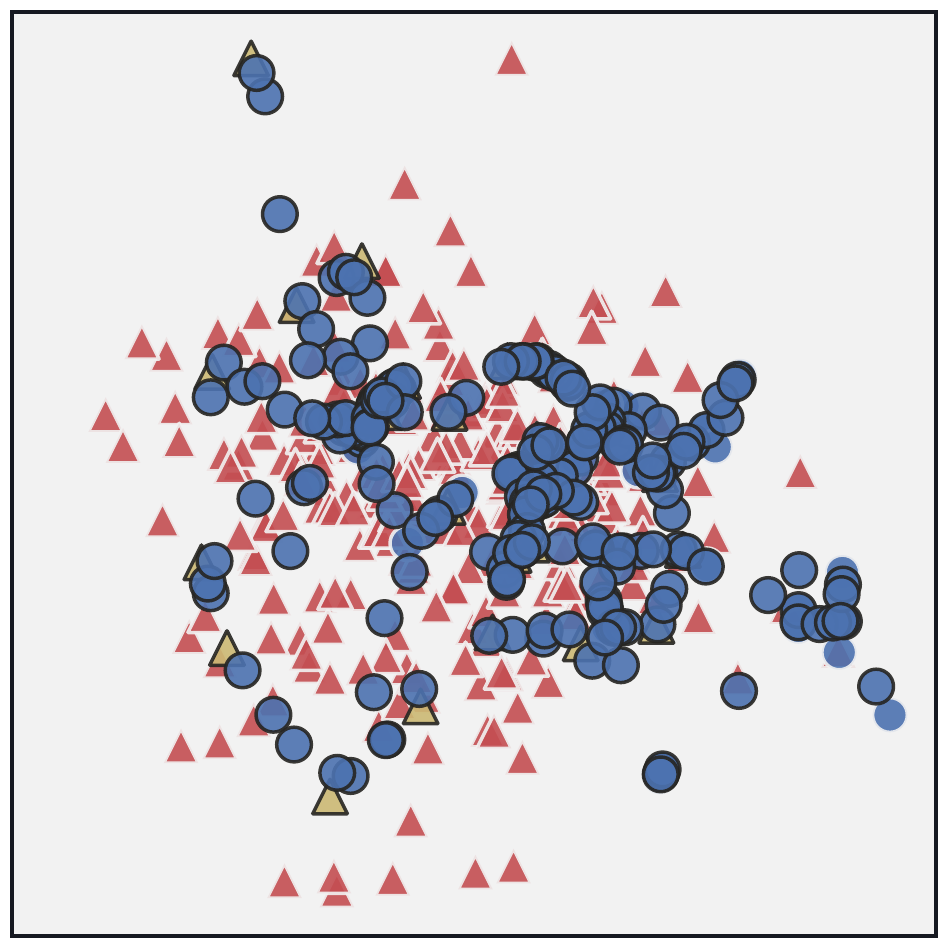}
        \caption{Effects of ADASYN.}
    \end{subfigure}
     ~ 
    \begin{subfigure}[t]{0.25\textwidth}
        \centering
        \includegraphics[width=1\textwidth]{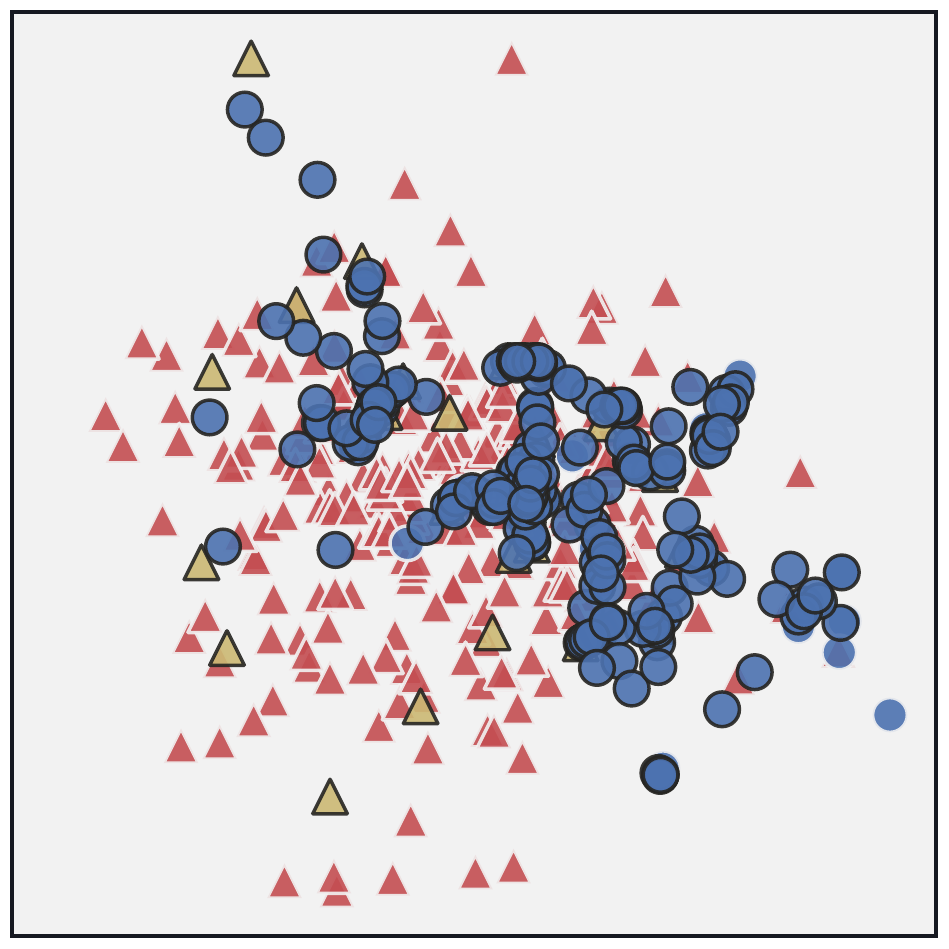}
        \caption{Effects of Borderline-SMOTE.}
    \end{subfigure}
     ~ 
    \begin{subfigure}[t]{0.25\textwidth}
        \centering
        \includegraphics[width=1\textwidth]{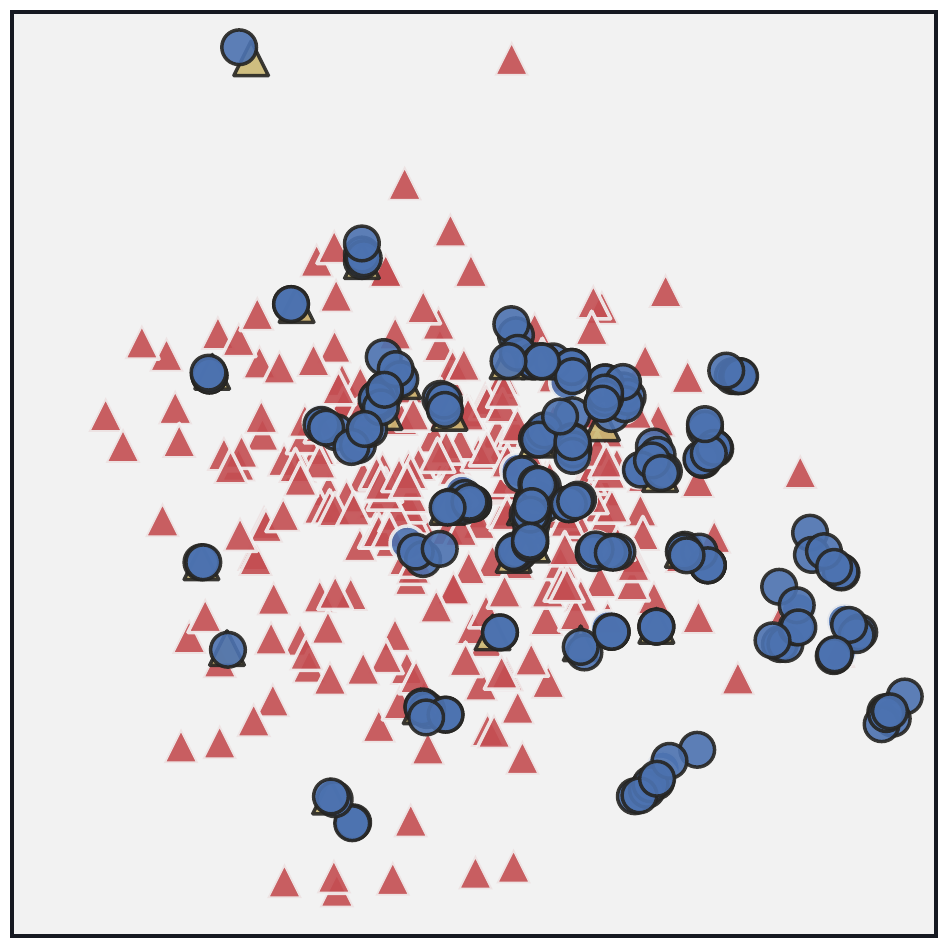}
        \caption{Effects of RBO.}
    \end{subfigure}
\caption{An example of influence of class label noise on imbalanced data oversampling. SMOTE and ADASYN are strongly affected, as they have no mechanisms to reduce the influence of noisy neighbors on introduced artificial objects. Borderline SMOTE is able to detect obvious outliers, but is still prone to generating synthetic minority objects over clusters of majority objects. The proposed RBO approach is less affected due to its potential-based selection of regions to oversample.}
\label{fig:overnoise}
\end{figure*}

Characteristics of the proposed RBO algorithm can also be favorable when dealing with noisy data. Most of existing oversampling schemes suffer from a degraded performance when the noise level is being increased. This phenomenon can be explained using SMOTE as an exemplary algorithm. Label noise will lead to mixed objects in the neighborhood, while feature noise will lead to increased overlapping between classes. As SMOTE creates new artificial objects based on a selected neighbors, both types of noise will affect it significantly. Label noise will lead to either too small neighborhood that will not allow for sufficient empowering of the minority class presence in this region (when majority objects in overlapping areas are mislabeled as minority ones), or to overextended neighborhood that will introduce scattered artificial objects (when local minority objects are mislabeled as majority ones). Feature noise will shift the location of objects in the training set, thus leading to incorrect placement of new objects (as they are placed along the lines joining two selected minority objects). These limitations are shared by most of oversampling techniques from SMOTE family. While recently SMOTE-IPF \cite{Saez:2015} implementation was proposed in order to tackle noisy data, it must be noted that it combines standard SMOTE with computationally costly filtering. Therefore, it seems more interesting to investigate oversampling methodologies that have inbuilt mechanism for handling the presence of noise. RBO falls into this category, as by using potentials it is able to identify which regions should be subject to oversampling. Therefore, shifts in the training set will not affect RBO procedure as much as reference methods. Additionally, RBO generates new objects and positions them in an less random manner than SMOTE, eliminating the problem of artificial objects being spread over the feature space. An illustrative example of differences between SMOTE and RBO on noisy data is given in Figure~\ref{fig:overnoise}. 

\noindent\textbf{Main results.} RBO was first introduced in \cite{koziarski2017hais}, in which the original study demonstrating its performance was conducted. It was further extended in \cite{koziarski2019neuro} in which, in addition to extending the scope of the initial study with additional benchmark datasets and reference methods, the impact of noise on the algorithms performance was considered. Observed results indicated the suitability of RBO for handling data affected by label noise. Finally, it is worth noting that based on the observed results, RBO was much less affected by the choice of its main parameter, $\gamma$, than CCR was affected by the proper choice of energy.

\section{RBU: Radial-Based Undersampling}

The Radial-Based Undersampling (RBU) \cite{koziarski2020radial} algorithm is an extension of the concept of class potential to the undersampling procedure. One of the most significant drawbacks of RBO is its high computational complexity, which makes its usage prohibitive when dealing with large datasets due to the need for recalculation of class potential at every iteration step. RBU addressed this issue, introducing a potential-based ranking procedure for determining the order of removal of majority observations.

\noindent\textbf{Using mutual class potential during undersampling.} While originally proposed to provide the regions of interest in the process of oversampling, mutual class potential can also easily be used to guide the process of undersampling the majority class. Recall that, using the assumed convention, high value of mutual class potential in a given point in space would indicate that in its proximity there is a higher concentration of majority than minority observations. It is therefore possible to rank the existing majority observations based on their mutual class potential. RBU proposes using such ranking mechanism to determine the order of undersampling. Specifically, it makes the assumption that the majority observations with highest corresponding mutual class potential provide the least amount of information and are more redundant than the observations with lower potential. As a result, the undersampling is performed in the order of decreasing potential, with its value updated for the remaining observations after each undersampled object.

\begin{figure*}
\centering
\includegraphics[width=0.195\textwidth]{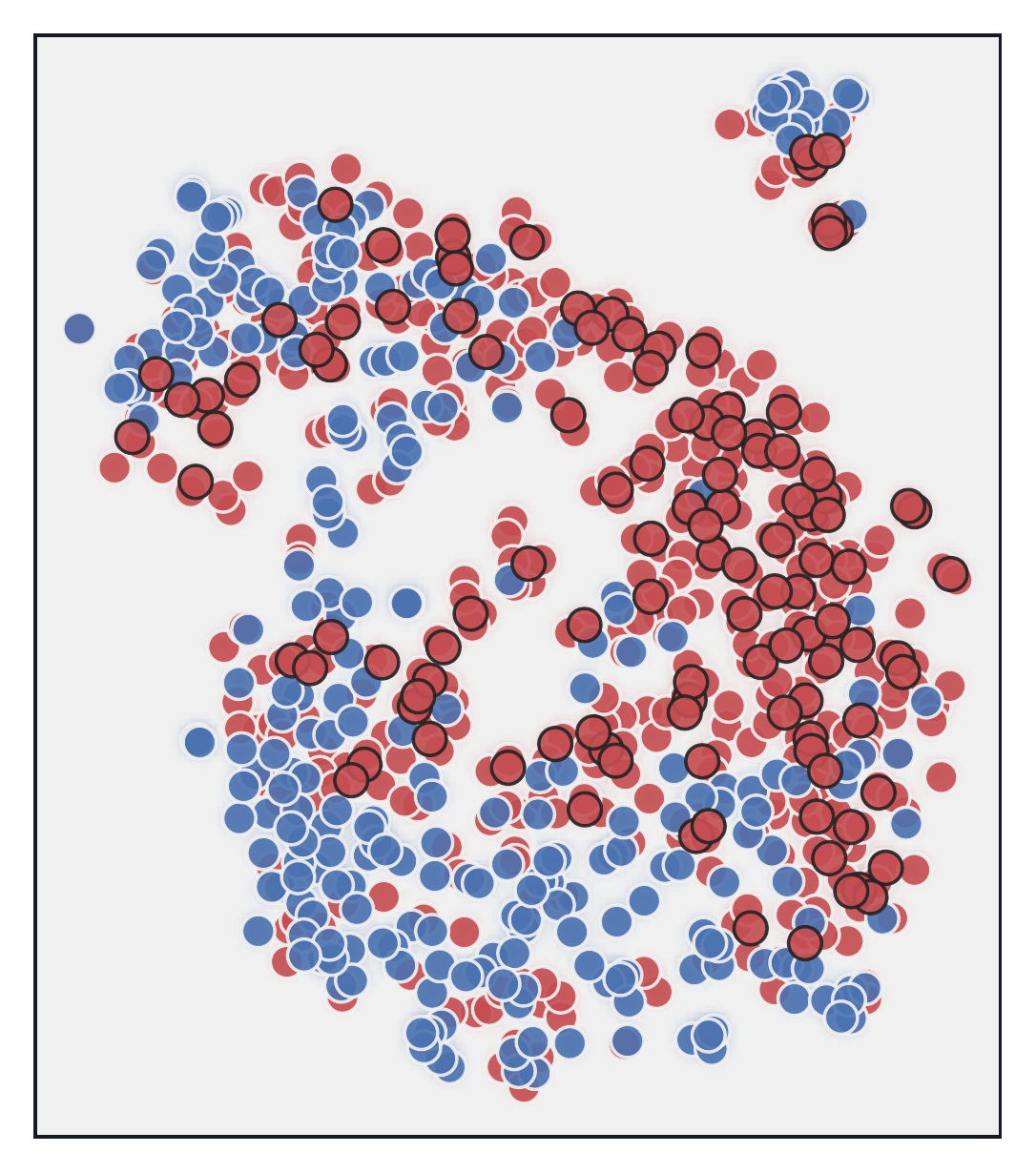}
\includegraphics[width=0.195\textwidth]{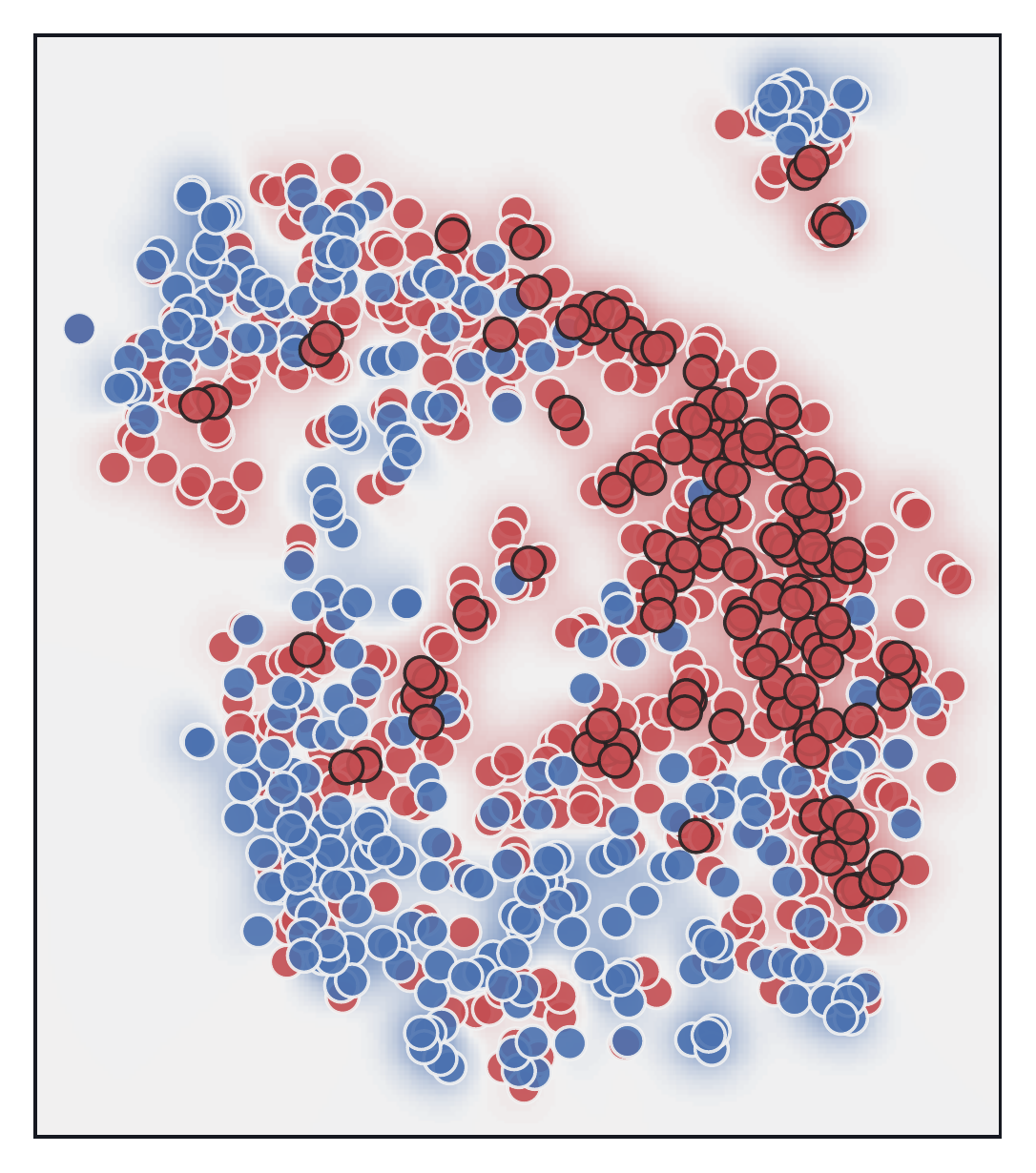}
\includegraphics[width=0.195\textwidth]{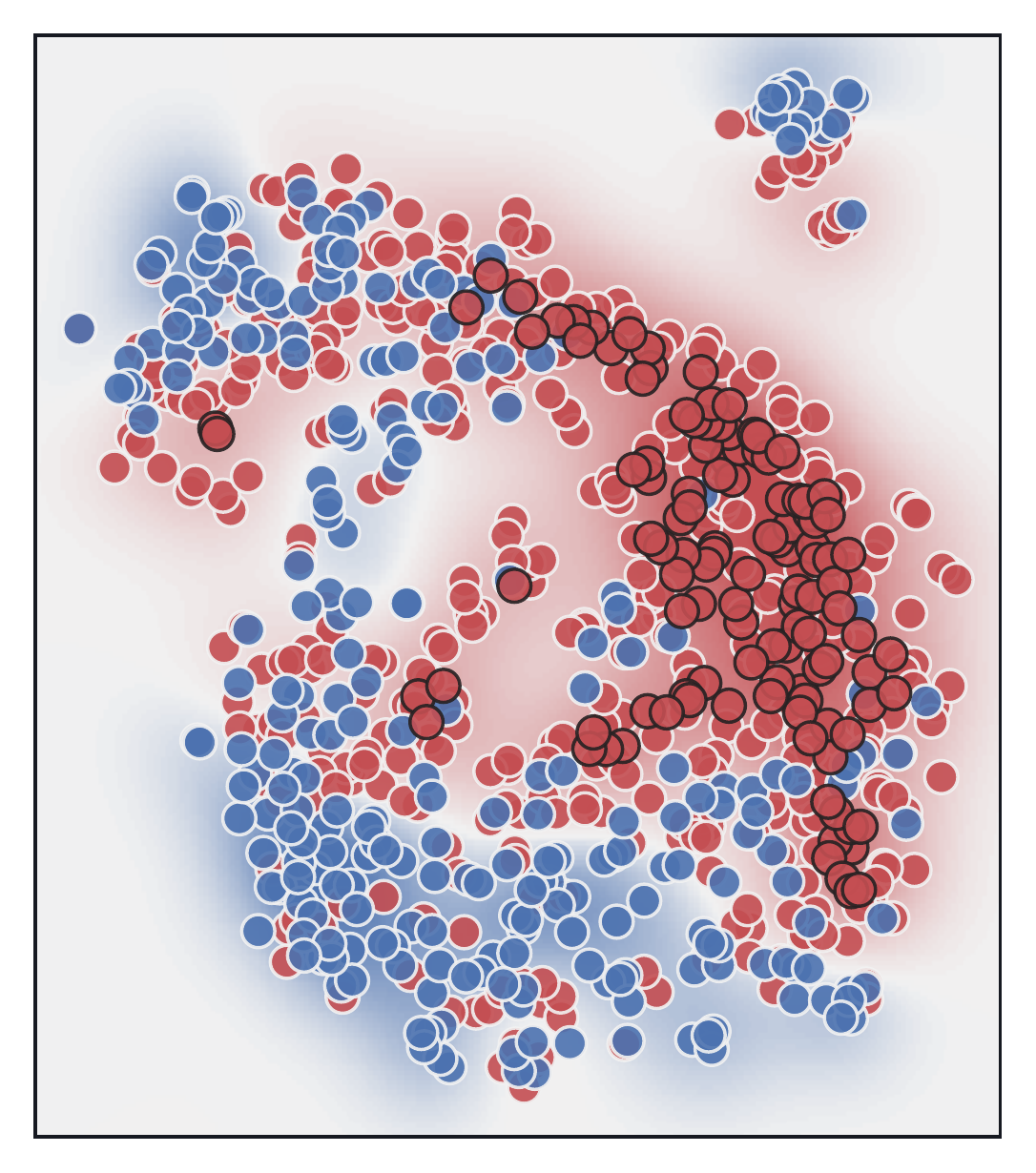}
\includegraphics[width=0.195\textwidth]{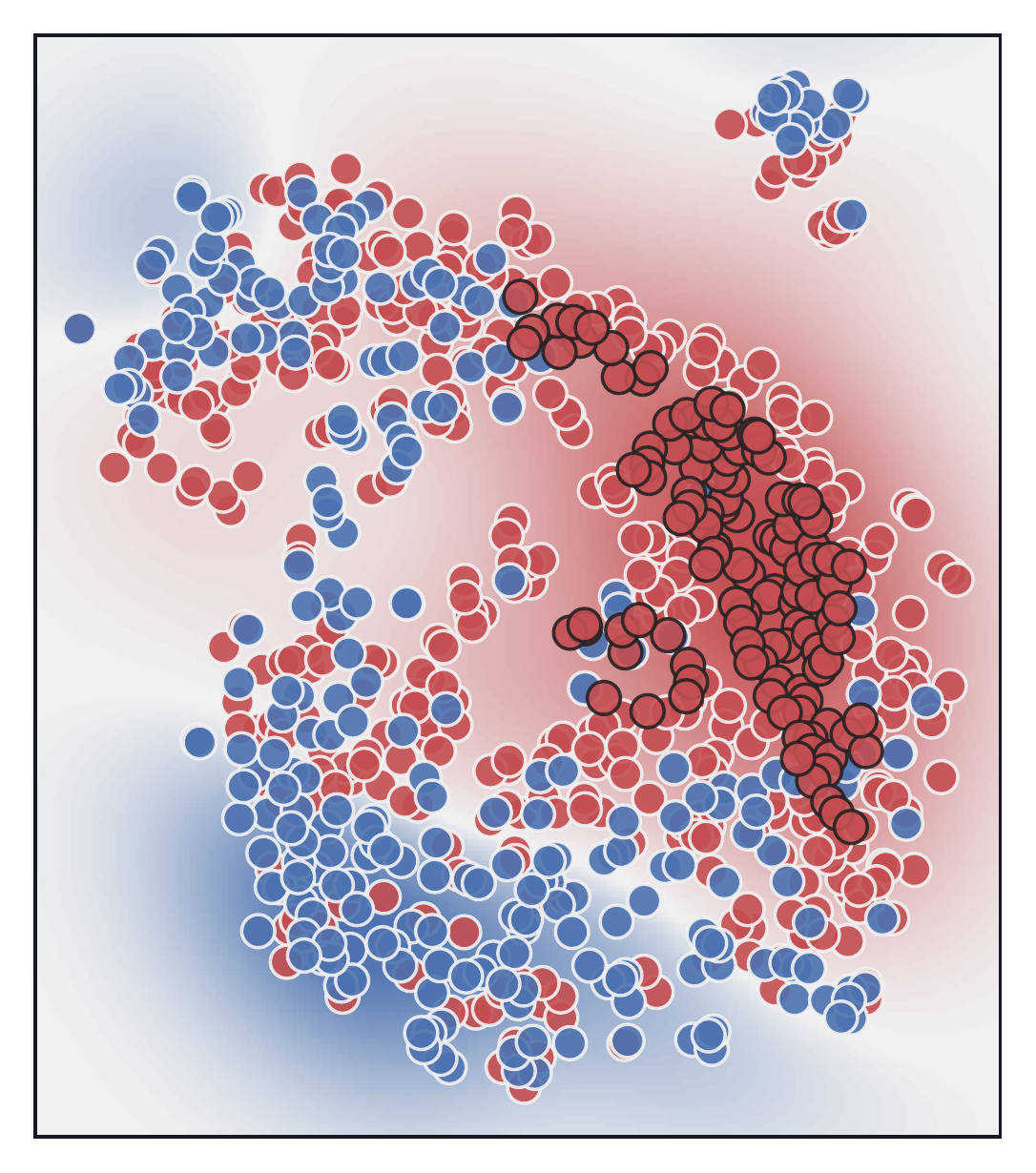}
\includegraphics[width=0.195\textwidth]{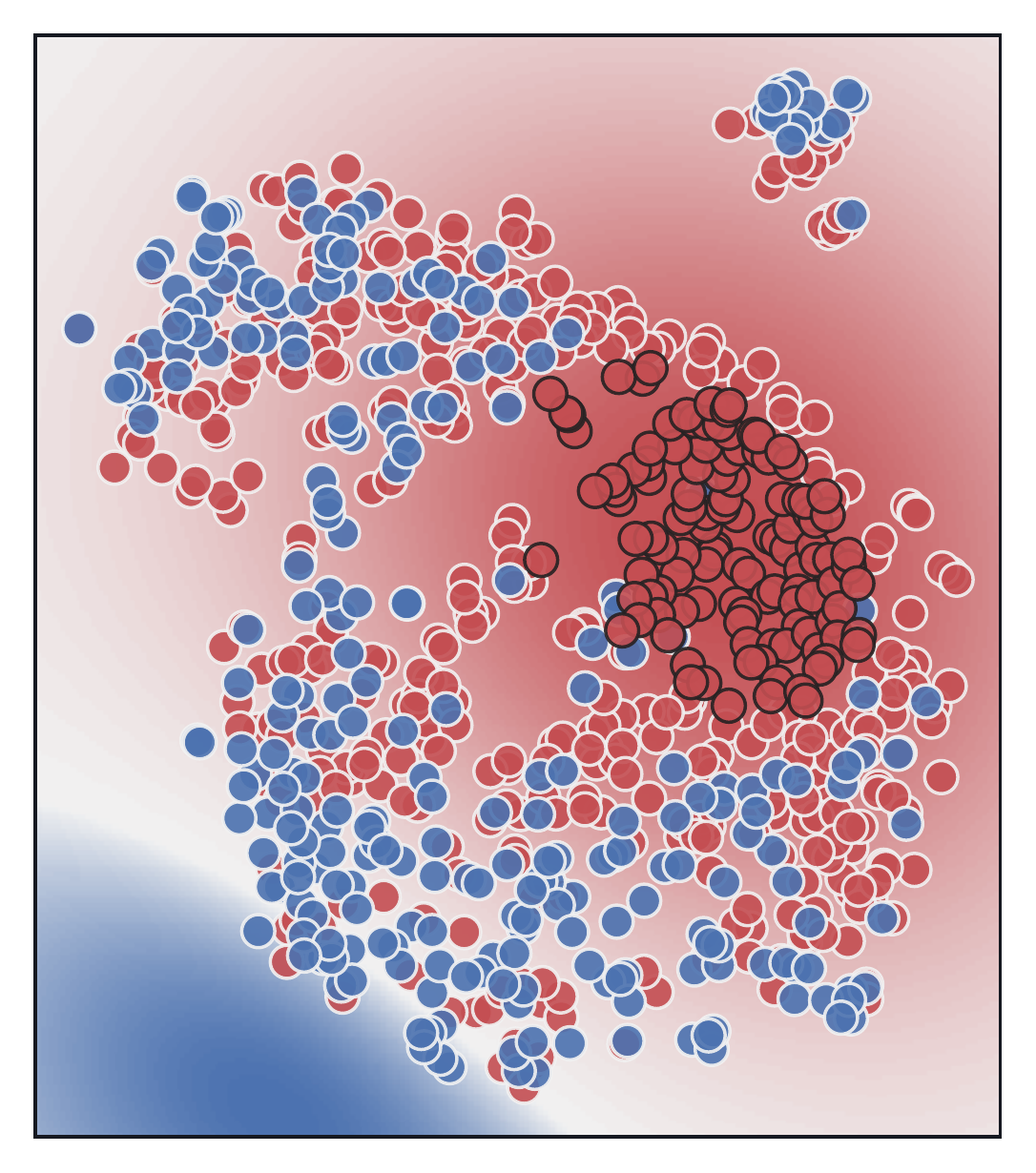}

\includegraphics[width=0.195\textwidth]{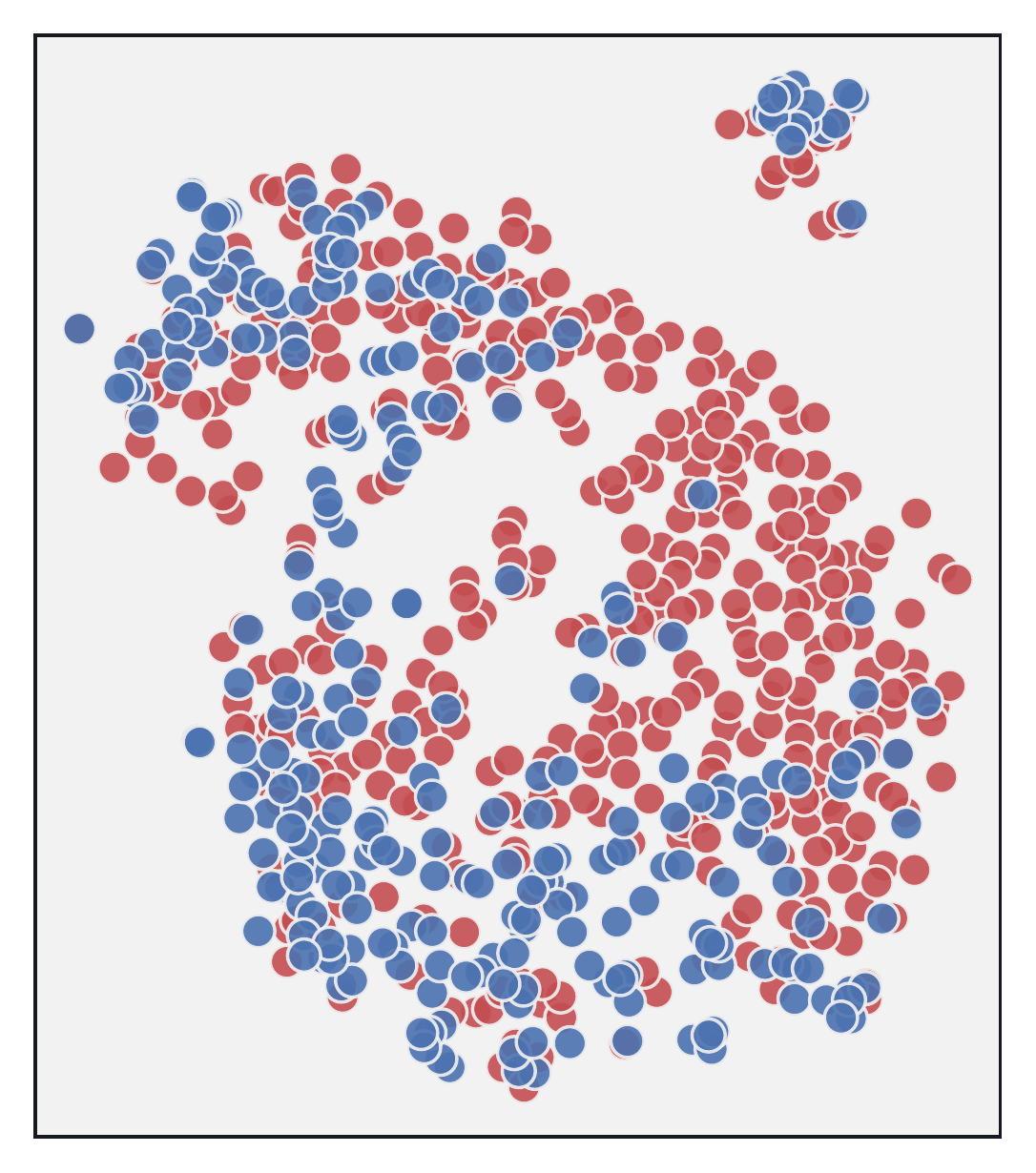}
\includegraphics[width=0.195\textwidth]{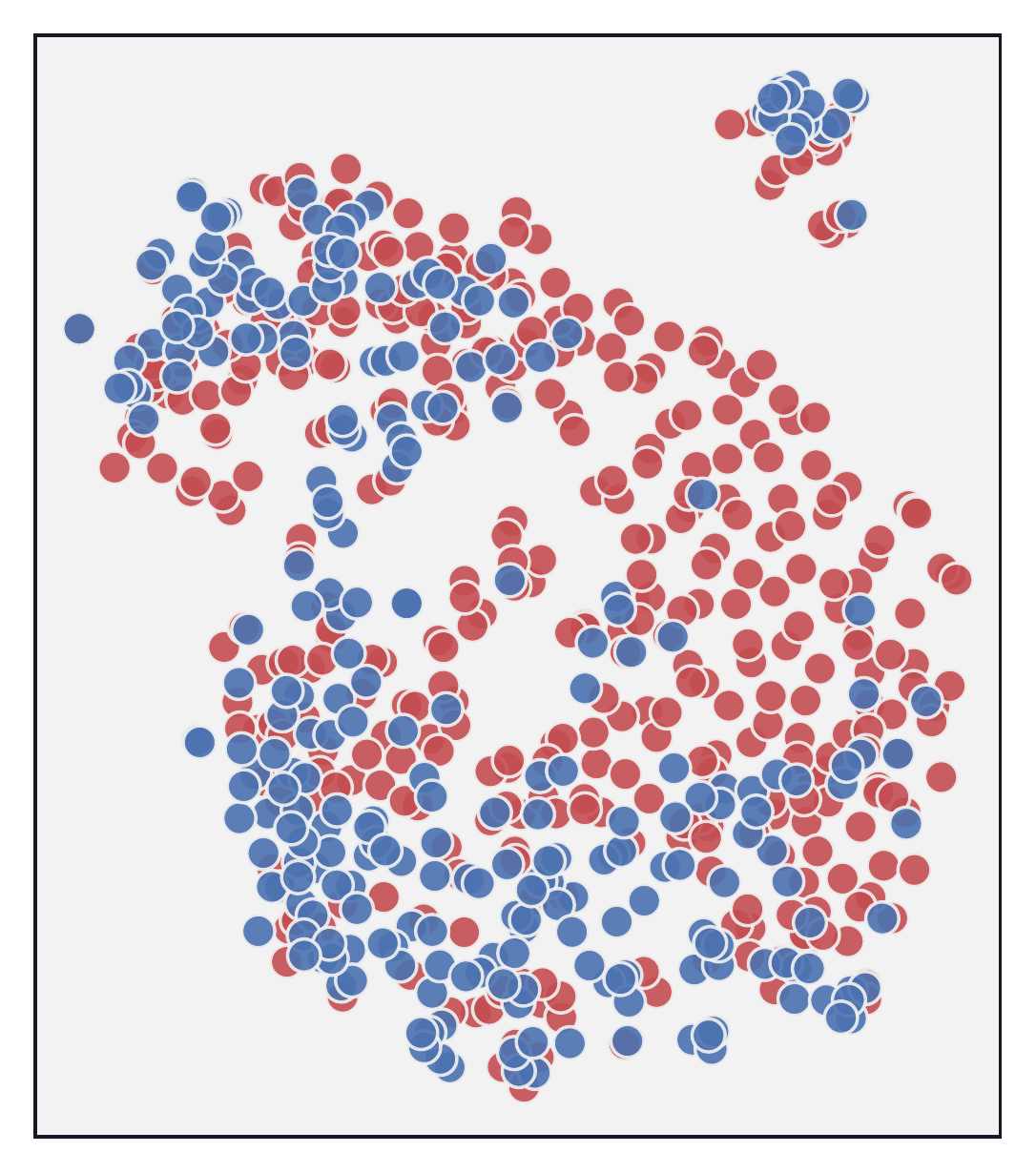}
\includegraphics[width=0.195\textwidth]{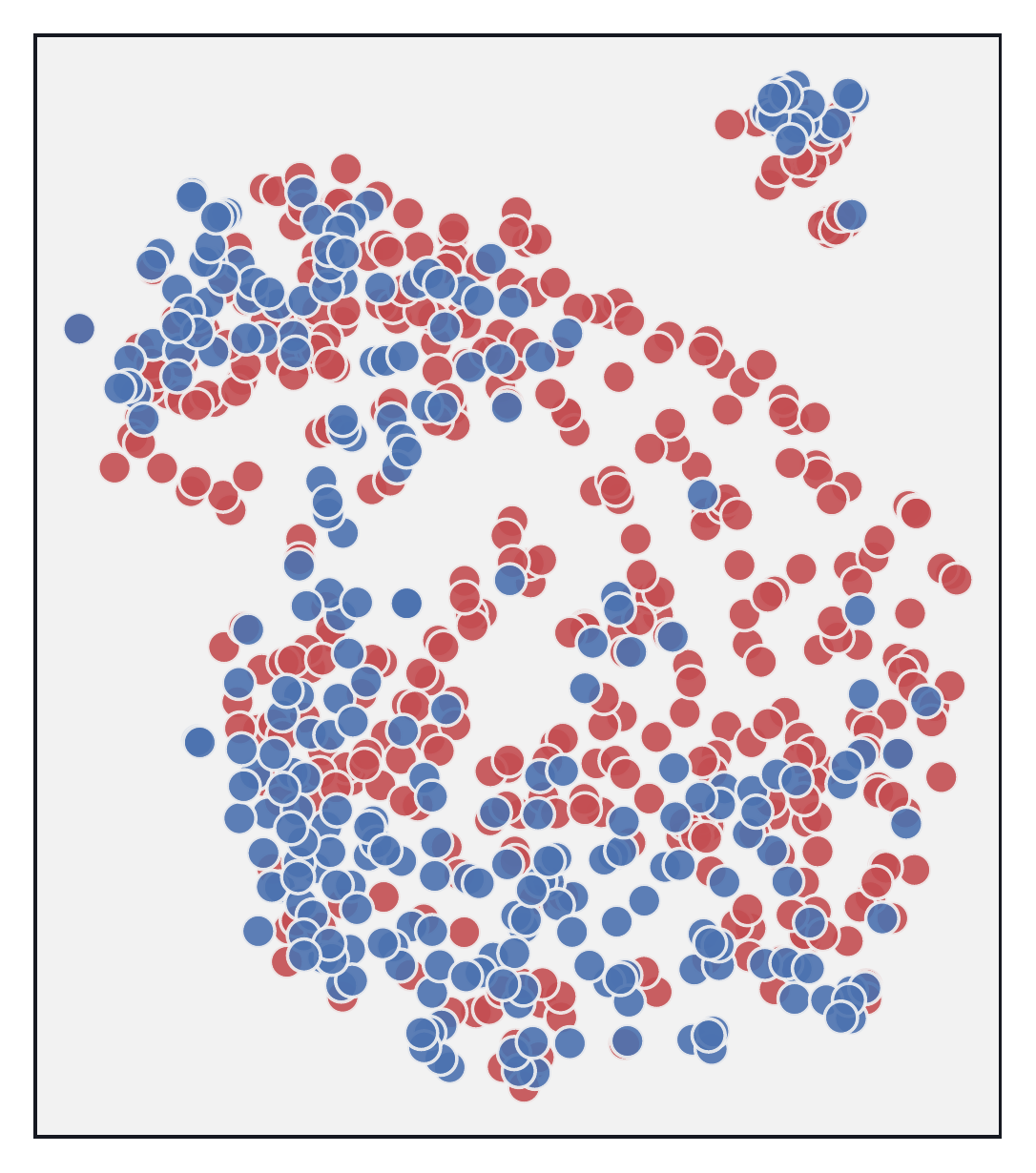}
\includegraphics[width=0.195\textwidth]{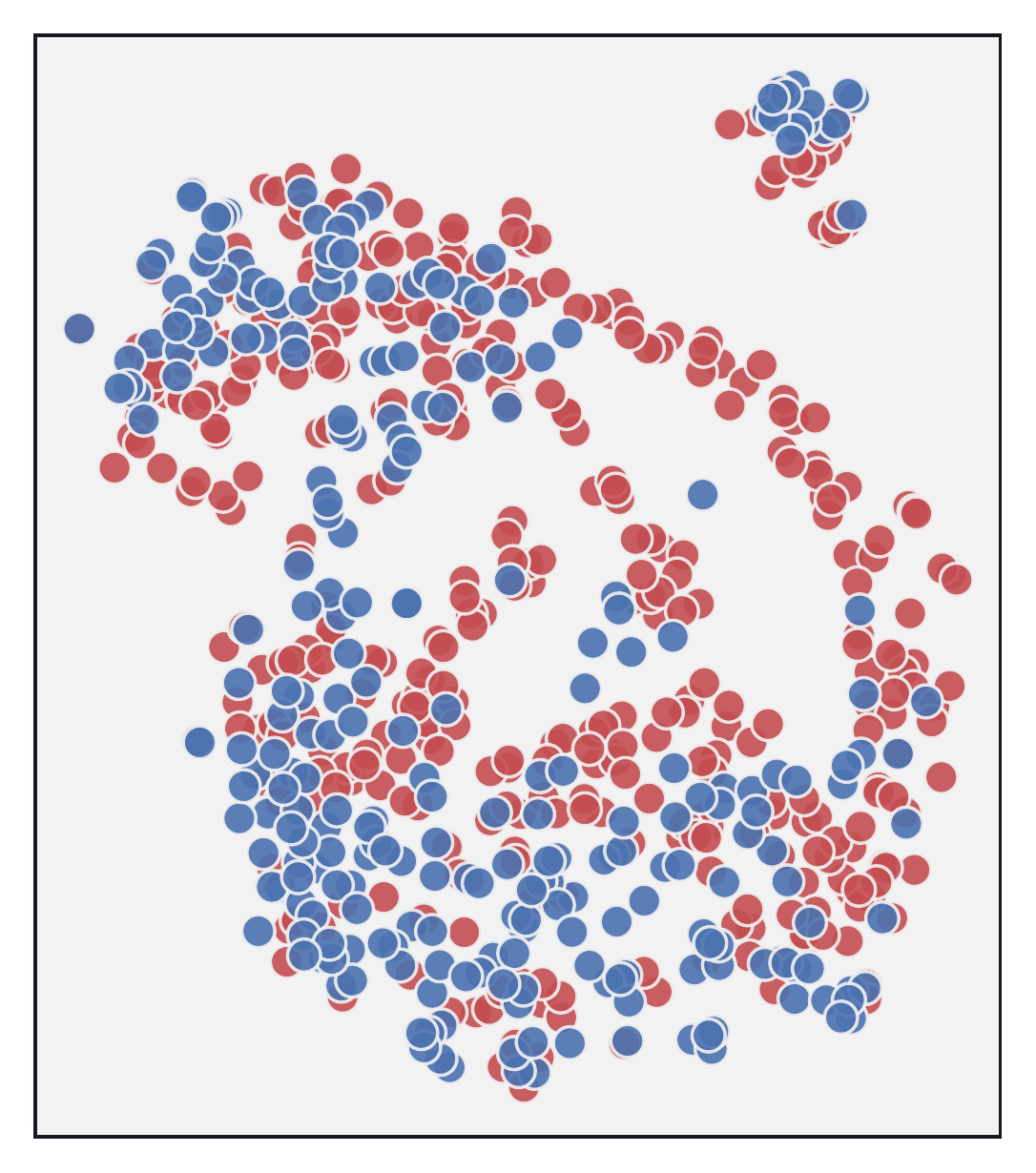}
\includegraphics[width=0.195\textwidth]{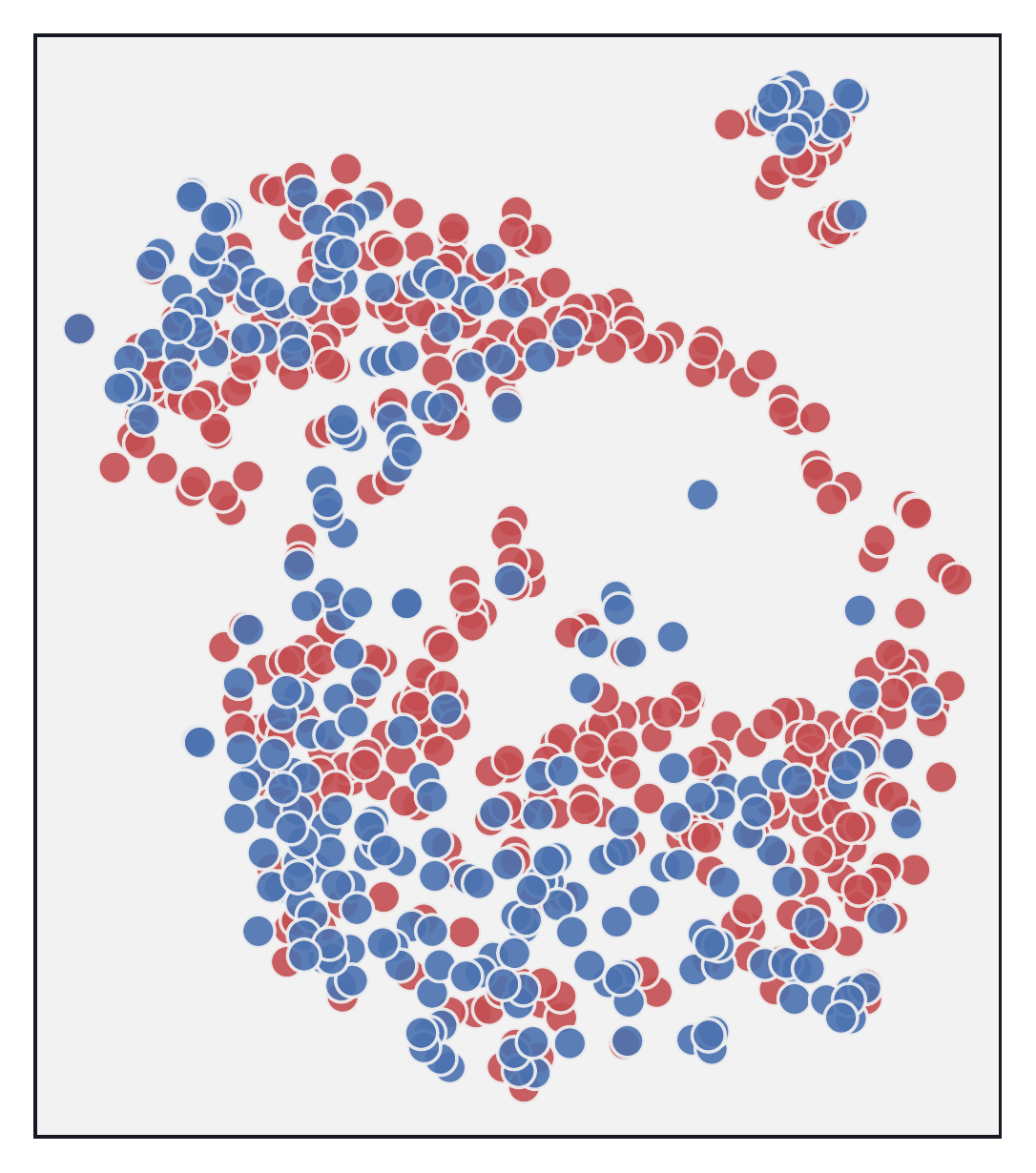}
\caption{Visualized impact of the $\gamma$ parameter of RBU on the shape of the mutual class potential and resulting undersampling regions. Top row: original dataset with highlighted majority objects selected for undersampling. Bottom row: dataset after undersampling. Values of $\gamma$, from the left: $1.0$, $2.5$, $5.0$, $10.0$, $25.0$.}
\label{fig:potential}
\end{figure*}

The pseudocode of the proposed method is presented in Algorithm~\ref{algorithm:rbu}. In addition to the collection of majority objects $K$ and the collection of minority objects $\kappa$, algorithm has two additional parameters: spread of the individual radial basis function $\gamma$, affecting the range of impact of the associated observation on the mutual class potential, and the undersampling ratio, with ratio equal to $1.0$ indicating that the majority objects are undersampled up to the point of achieving balanced class distribution. Furthermore, a visualization of the algorithms behavior for different values of $\gamma$ is presented in Figure~\ref{fig:potential}. As can be seen, the value of $\gamma$ parameter significantly impacts the shape of the resulting potential: using smaller values of $\gamma$ leads to a more complex potential field, affected in a given point in space only by the observations in its close proximity, whereas using larger values of $\gamma$ leads to a smooth potential. As a result, the choice of $\gamma$ affects the order of undersampling. For the smaller values of $\gamma$ removed observations are mostly a part of local clusters consisting of several majority and no minority observations, and these clusters are never completely removed. Furthermore, individual majority observations and majority observations located in a close proximity of minority observations tend to remain unaffected. On the other hand, for larger values of $\gamma$ a single cluster with a high concentration of majority objects is identified and the undersampling is performed solely within its bounds. When combined with a significant data imbalance this can lead to a potentially undesirable behavior of a very highly centralized undersampling, which may indicate the proclivity towards using lower values of $\gamma$.

\begin{algorithm}
\caption{Radial-Based Undersampling}
\textbf{Input:} collections of majority objects $K$ and minority objects $\kappa$ \\
\textbf{Parameters:} spread of radial basis function $\gamma$, undersampling $ratio$ \\
\textbf{Output:} undersampled collection of majority objects $K'$
\label{algorithm:rbu}	
\vspace{-0.5\baselineskip}

\hrulefill
\begin{algorithmic}[1]
\STATE \textbf{function} RBU($K$, $\kappa$, $\gamma$, $ratio$):
\STATE $K' \gets K$
\FOR{every object $K_i'$ in $K'$ and its associated potential $\Phi_i$}
\STATE $\Phi_i \gets \Phi(K_i', K, \kappa, \gamma)$
\ENDFOR
\WHILE{$\vert K \vert - \vert K' \vert < ratio \cdot (\vert K \vert - \vert \kappa \vert)$}
\STATE $x \gets $ object $K_i'$ from $K'$ with highest potential $\Phi_i$; in case of multiple selected objects break ties arbitrarily
\STATE discard $x$ from $K'$
\FOR{every object $K_i'$ in $K'$ and its associated potential $\Phi_i$}
\STATE $\Phi_i \gets \Phi_i - e^{-(\frac{\lVert K_i' - x \rVert_2}{\gamma})^{2}}$
\ENDFOR
\ENDWHILE
\STATE \textbf{return} $K'$
\end{algorithmic}
\end{algorithm}

\noindent\textbf{Main results.} The results of the experimental study described in \cite{koziarski2020radial} demonstrated the usefulness of RBU algorithm when compared to state-of-the-art over- and undersampling techniques. As demonstrated by the conducted computational complexity analysis, RBU offered a significant speed-up when compared to RBO, reducing the complexity from $\mathcal{O}(imn^2)$ to $\mathcal{O}(mn^2)$, with $n$ denoting the total number of observations, $m$ denoting the number of features, and $i$ denoting the number of iterations of RBO algorithm per synthetic observation, which was typically set between 1,000 and 10,000 in the conducted experiments. Interestingly, at the same time RBU achieved a statistically significantly better performance than RBO when combined with decision trees, and no statistically significant differences in other cases. Finally, an approach for evaluating the impact of dataset characteristics on the algorithms performance was introduced, which was used to demonstrate that RBU tends to outperform the reference methods on difficult datasets.

\section{RB-CCR: Radial-Based Combined Cleaning and Resampling}

The Radial-Based Combined Cleaning and Resampling (RB-CCR) \cite{rbccr} algorithm, as the name indicates, is an attempt at combining the design principles underlying both the CCR and RBO algorithms. Recall that CCR is an energy-based oversampling algorithm that relies on spherical regions, centered around the minority class observations, to designate areas in which synthetic minority observations should be generated. These spherical regions expand iteratively, with the rate of expansion inversely proportional to the number of neighboring observations belonging to the majority class. While computationally efficient and conceptually simple, using spherical regions to model the areas designed for oversampling has two limitations. First of all, it enforces constant rate of expansion of the sphere in every direction, irregardless of the exact position of the majority neighbors. Secondly, it does not utilize the information about the neighboring minority class observations. To address these issues RB-CCR proposes a novel sampling procedure, the aim of which is refining the original spherical regions via constraining the regions designed for oversampling based on the class potential. In the remainder of this section the proposed sampling procedure, as well as the way in which it can be integrated with the CCR algorithm, are described.

\noindent\textbf{Guided sampling procedure.} The proposed sampling procedure is based on the notion of class potential (Equation~\ref{eq:potential}). To reiterate, two previously identified drawbacks of the original CCR algorithm are: that the sphere expansion procedure progresses at a constant rate in every direction, irregardless of the exact position of the majority neighbors; and that it does not utilize the information about the position of neighboring minority class observations. Intuitively, neither of these is a desired behavior, since it can lead to a lower than expected expansion in the direction of minority observation clusters, and higher than expected expansion in the direction of majority observation clusters. While in theory an obvious modification that could address theses issues would be to exchange the spheres used by CCR to a more robust shapes, such as ellipsoids, and adjust the expansion step accordingly, in practice it is not clear how the later could be achieved. Alternatively, RB-CCR proposes to exploit the efficiency of first defining the sphere around the minority observation, and then partitioning it into sub-regions based on the class potential to more effectively guide sample generation.

\begin{figure*}[!htb]
\centering
\includegraphics[width=0.3\textwidth]{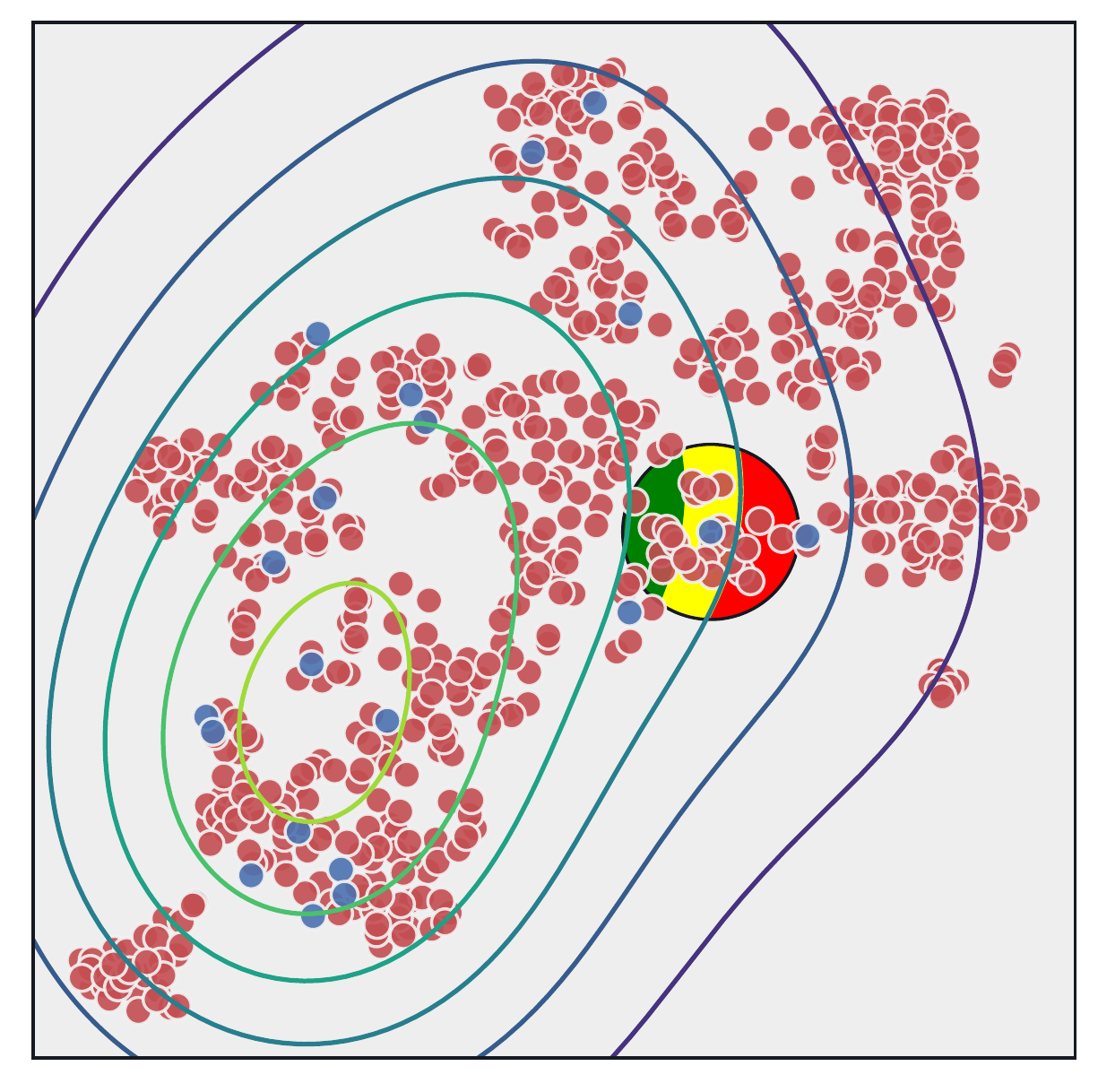}
\caption{An example of a sphere generated around a specific minority observation, partitioned into three regions: high potential (H), indicated with a green color, equal potential (E), indicated with a yellow color, and low potential (L), indicated with a red color. Note that the shape of the regions aligns with that of the produced potential field, indicated with a contour plot.}
\label{fig:example-regions-local}
\end{figure*}

The proposed strategy generates a number of candidate samples within the sphere to estimate the range of potential values, afterwards uses the minority class potential to divide a given sphere into three regions: low (L), equal (E) and high (H) potential regions, and finally selects a desired number of candidate observations exclusively from a chosen region, specified as a parameter of the algorithm. A more detailed formulation of the proposed strategy is presented in Algorithm~\ref{algorithm:sampling}, and an illustration of the sphere partitioning procedure is presented in Figure~\ref{fig:example-regions-local}.

\begin{algorithm}[!htb]
	\caption{Guided sampling procedure}
	\textbf{Input:} sampling seed $x$, sampling radius $r$, collection of minority observations $\mathcal{X}_{min}$ \\
	\textbf{Parameters:} radial basis function spread $\gamma$, sampling $region$ from which returned samples will be drawn, number of candidates $c$ used for potential range estimation, number of returned candidate samples $n$ \\
    \textbf{Output:} collection of synthetic minority observations $S$ located in the sampling $region$ around $x$
		
	\label{algorithm:sampling}	
	\vspace{-0.5\baselineskip}
	
	\hrulefill
	\begin{algorithmic}[1]
    	\STATE \textbf{function} sample($x$, $r$, $\mathcal{X}_{min}$, $\gamma$, $region$, $c$, $n$):
    	\STATE $C \gets \emptyset$ \COMMENT{collection of candidate samples}
    	\STATE $Z \gets \emptyset$ \COMMENT{collection of candidate potentials}
    	\FOR{\textit{i} \textbf{in} 1 \textbf{to} $c$}
		\STATE $C_i \gets$ random sample inside a $x$-centered sphere with radius $r$
		\STATE $Z_i \gets \Phi(C_i, \mathcal{X}_{min}, \gamma)$
        \ENDFOR
	    \STATE $bound_L \gets \Phi^0 - \frac{1}{3}(\Phi(x, \mathcal{X}_{min}, \gamma) - \min Z)$ \COMMENT{estimated low potential bound within the sphere}
	    \STATE $bound_H \gets \Phi^0 + \frac{1}{3}(\max Z - \Phi(x, \mathcal{X}_{min}, \gamma))$ \COMMENT{estimated high potential bound within the sphere}
	    \STATE $S \gets \{x\}$ \COMMENT{collection of suitable candidates}
	    \FOR{\textit{i} \textbf{in} 1 \textbf{to} $c$}
		\IF{$Z_i \leq bound_L$}
		\STATE $reg_i \gets L$ \COMMENT{$i$-th candidate in the low potential region}
		\ELSIF{$Z_i \geq bound_H$}
		\STATE $reg_i \gets H$ \COMMENT{$i$-th candidate in the high potential region}
		\ELSE
		\STATE $reg_i \gets E$ \COMMENT{$i$-th candidate in the equal potential region}
		\ENDIF
		\IF{$reg_i = region$}
		\STATE $S \gets S \cup \{C_i\}$
		\ENDIF
        \ENDFOR
	    \STATE $S \gets n$ samples randomly selected with replacement from $S$ 
	\STATE \textbf{return} $S$
	\end{algorithmic}
\end{algorithm}

The CCR algorithm generates samples with uniform probability from within entire sphere. Alternatively, Figure \ref{fig:example-regions-local} illustrates that RB-CCR divides the original sphere into three regions (L, E, H). The regions are defined according to the shape of the globally calculated minority class potential. Subsequent to the partitioning, sample generation can be restricted to a specific region. 
Intuitively, samples in the high potential regions can be regarded as having a higher probability of coming from the underlying minority class distribution than samples in the low potential regions. This, to some extents, parallels different variants of SMOTE, such as Borderline-SMOTE \cite{Han:2005} or Safe-Level-SMOTE \cite{Bunkhumpornpat:2009}, which focus on different types of observations to guide the sampling process. However, contrary to SMOTE variants, RB-CCR provides a flexibility to chose an appropriate sampling region for the target data within a single framework.

\noindent\textbf{Integrating guided sampling with the CCR algorithm.} The proposed sampling strategy can easily be integrated into the original CCR algorithm. Instead of the original sampling within the whole sphere, RB-CCR uses the guided sampling strategy described in the previous section. The rest of the steps, that is: sphere radius calculation, translation of majority observations, and calculation of the number of synthetic observations generated for each minority observations, remain unchanged. A pseudocode of the proposed RB-CCR algorithm is presented in Algorithm~\ref{algorithm:rb-ccr}.

\begin{algorithm}[!htb]
	\caption{Radial-Based Combined Cleaning and Resampling}
	\footnotesize
	\textbf{Input:} collections of majority observations $\mathcal{X}_{maj}$ and minority observations $\mathcal{X}_{min}$ \\
	\textbf{Parameters:} $energy$ budget for expansion of each sphere, radial basis function spread $\gamma$, sampling $region$ from which returned samples will be drawn, number of candidates $c$ used for potential range estimation \\
    \textbf{Output:} collections of translated majority observations $\mathcal{X}_{maj}'$ and synthetic minority observations $S$
		
	\label{algorithm:rb-ccr}	
	\vspace{-0.5\baselineskip}
	
	\hrulefill
	\begin{algorithmic}[1]
		\STATE \textbf{function} RB-CCR($\mathcal{X}_{maj}$, $\mathcal{X}_{min}$, $energy$, $\gamma$, $region$, $c$):
		\STATE $S \gets \emptyset$ \COMMENT{synthetic minority observations}
		\STATE $t \gets $ zero matrix of size $|\mathcal{X}_{maj}| \times m$, with $m$ denoting the number of features \COMMENT{translations of majority observations}
		\STATE $r \gets $ zero vector of size $|\mathcal{X}_{min}|$ \COMMENT{radii of spheres associated with the minority observations}
        \FORALL{minority observations $x_i$ in $\mathcal{X}_{min}$}
        \STATE $e$ $\gets$ $energy$ \COMMENT{remaining energy budget}
        \STATE $n_r \gets 0$ \COMMENT{number of majority observations inside the sphere generated around $x_i$}
        \FORALL{majority observations $x_j$ in $\mathcal{X}_{maj}$}
        \STATE $d_j \gets \lVert x_i - x_j \rVert_2$
        \ENDFOR
        \STATE sort $\mathcal{X}_{maj}$ with respect to $d$
        \FORALL{majority observations $x_j$ in $\mathcal{X}_{maj}$}
        \STATE $n_r \gets n_r + 1$
        \STATE $\Delta e \gets - (d_j - r_i) \cdot n_r$
        \IF{$e + \Delta e > 0$}
        \STATE $r_i \gets d_j$
        \STATE $e \gets e + \Delta e$
        \ELSE
        \STATE $r_i \gets r_i + \frac{e}{n_r}$
        \STATE \textbf{break}
        \ENDIF
        \ENDFOR
        \FORALL{majority observations $x_j$ in $\mathcal{X}_{maj}$}
        \IF{$d_j < r_i$}
        \STATE $t_j \gets t_j + \dfrac{r_i - d_j}{d_j} \cdot (x_j - x_i)$
        \ENDIF
        \ENDFOR
        \ENDFOR
        \STATE $\mathcal{X}_{maj}' \gets \mathcal{X}_{maj} + t$
        \FORALL{minority observations $x_i$ in $\mathcal{X}_{min}$}
        \STATE $g_i \gets \lfloor\dfrac{r_i^{-1}}{\sum_{k = 1}^{|\mathcal{X}_{min}|}{r_k^{-1}}} \cdot (|\mathcal{X}_{maj}| - |\mathcal{X}_{min}|)\rfloor$ \label{op:prop} 
		\STATE add sample($x_i$, $r_i$, $\mathcal{X}_{min}$, $\gamma$, $region$, $c$, $g_i$) to $S$
        \ENDFOR
		\STATE \textbf{return} $\mathcal{X}_{maj}'$, $S$
	\end{algorithmic}
\end{algorithm}

\begin{figure*}[!htb]
\centering
\begin{subfigure}[b]{0.23\textwidth}
  \includegraphics[width=\textwidth]{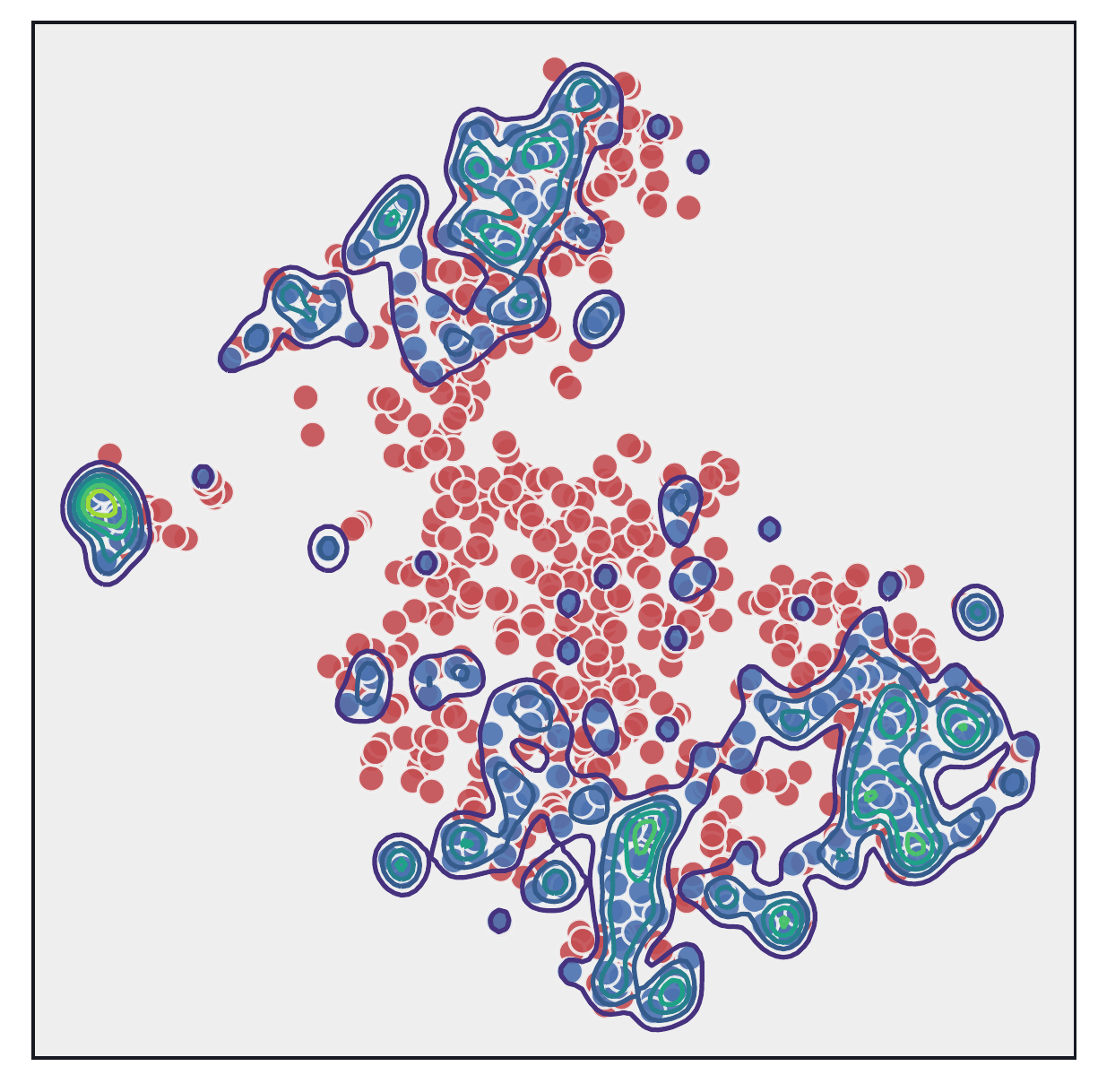}
  \caption{$\gamma = 0.1$}
\end{subfigure}
~
\begin{subfigure}[b]{0.23\textwidth}
  \includegraphics[width=\textwidth]{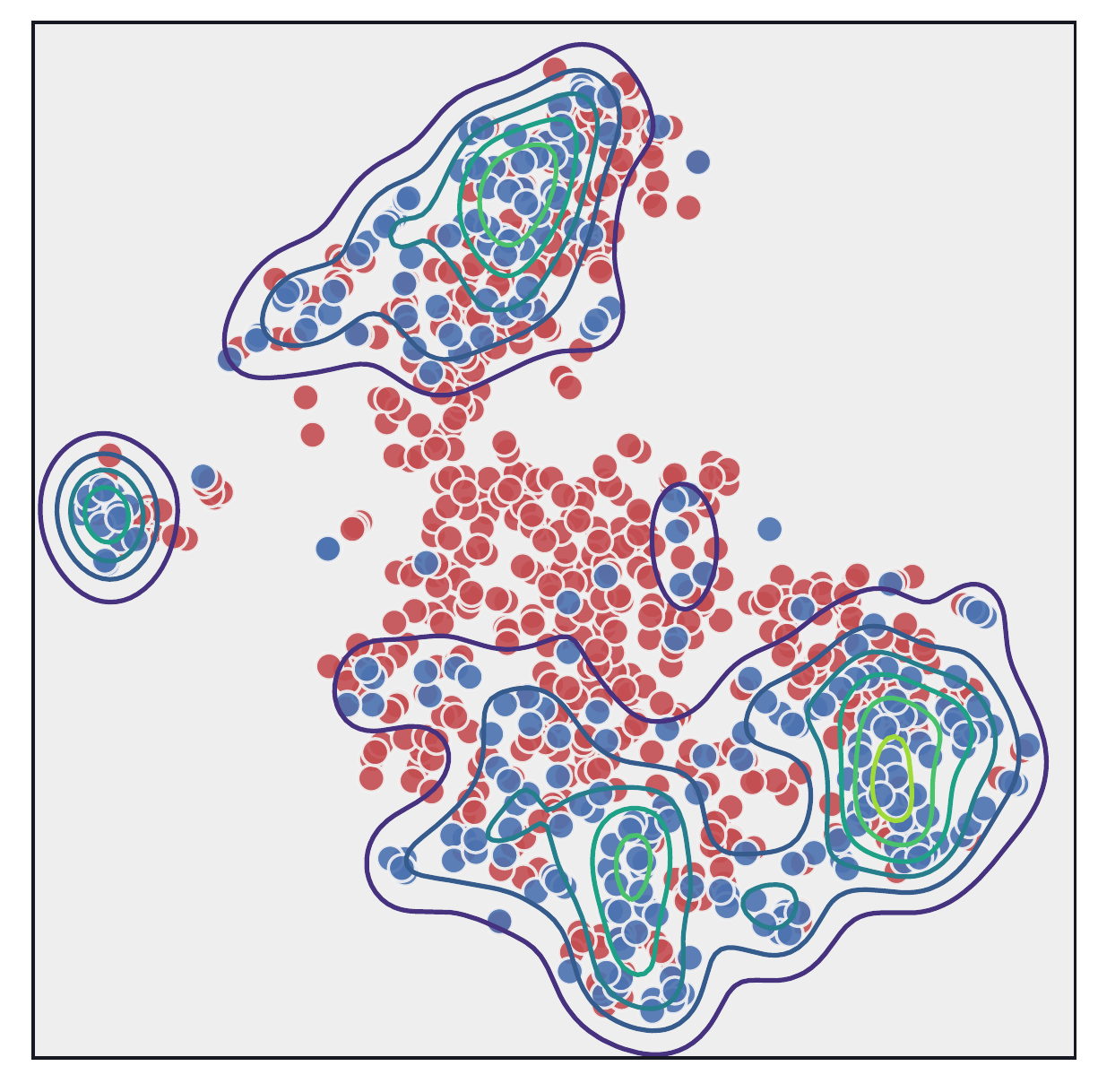}
  \caption{$\gamma = 0.25$}
\end{subfigure}
~
\begin{subfigure}[b]{0.23\textwidth}
  \includegraphics[width=\textwidth]{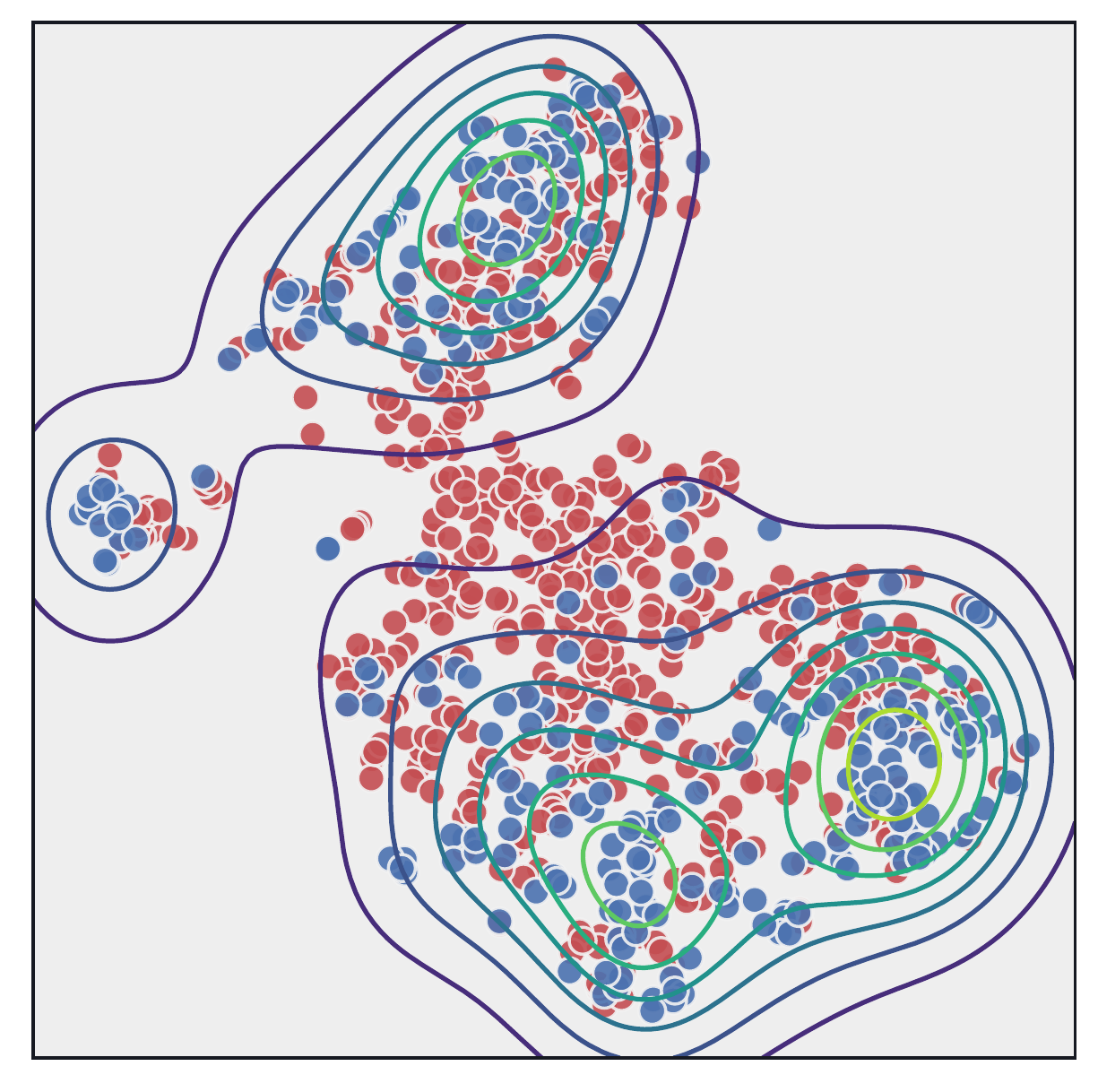}
  \caption{$\gamma = 0.5$}
\end{subfigure}
~
\begin{subfigure}[b]{0.23\textwidth}
  \includegraphics[width=\textwidth]{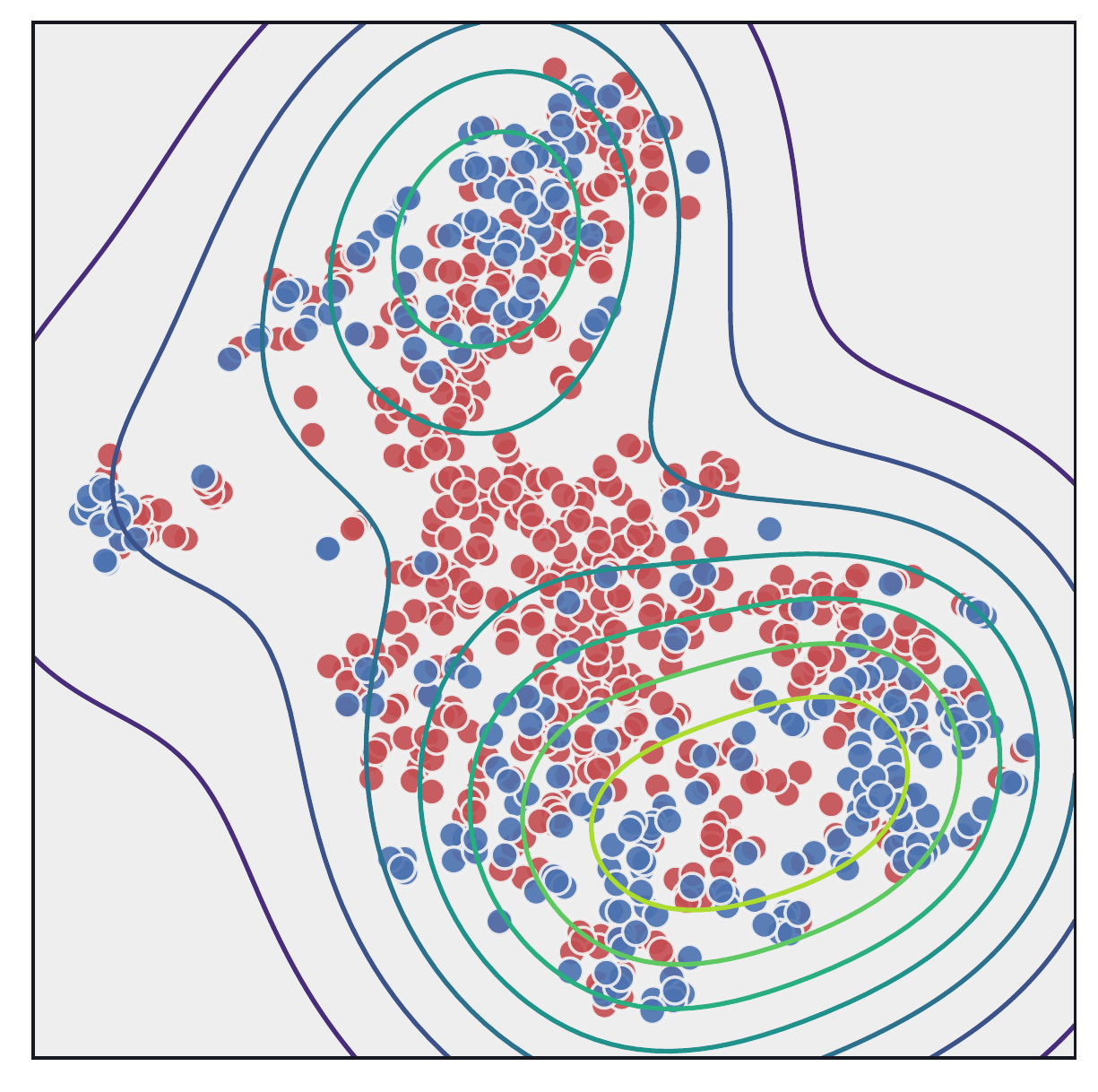}
  \caption{$\gamma = 1.0$}
\end{subfigure}
\caption{Visualization of the impact of $\gamma$ parameter on the shape of minority class potential.}
\label{fig:example-potential}
\end{figure*}

\begin{figure*}[!htb]
\centering
\begin{subfigure}[b]{0.23\textwidth}
  \includegraphics[width=\textwidth]{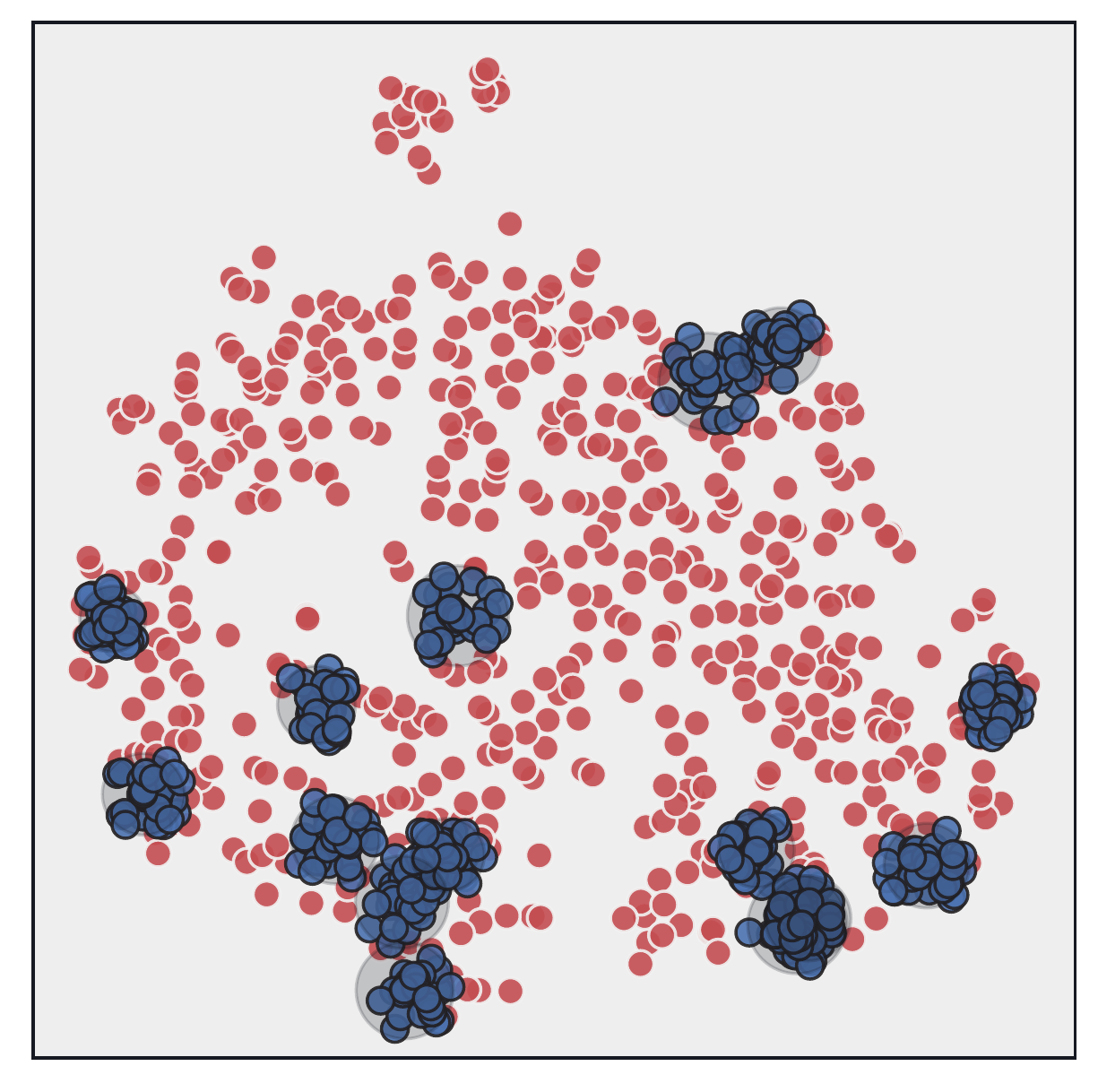}
  \caption{energy $= 0.1$}
\end{subfigure}
~
\begin{subfigure}[b]{0.23\textwidth}
  \includegraphics[width=\textwidth]{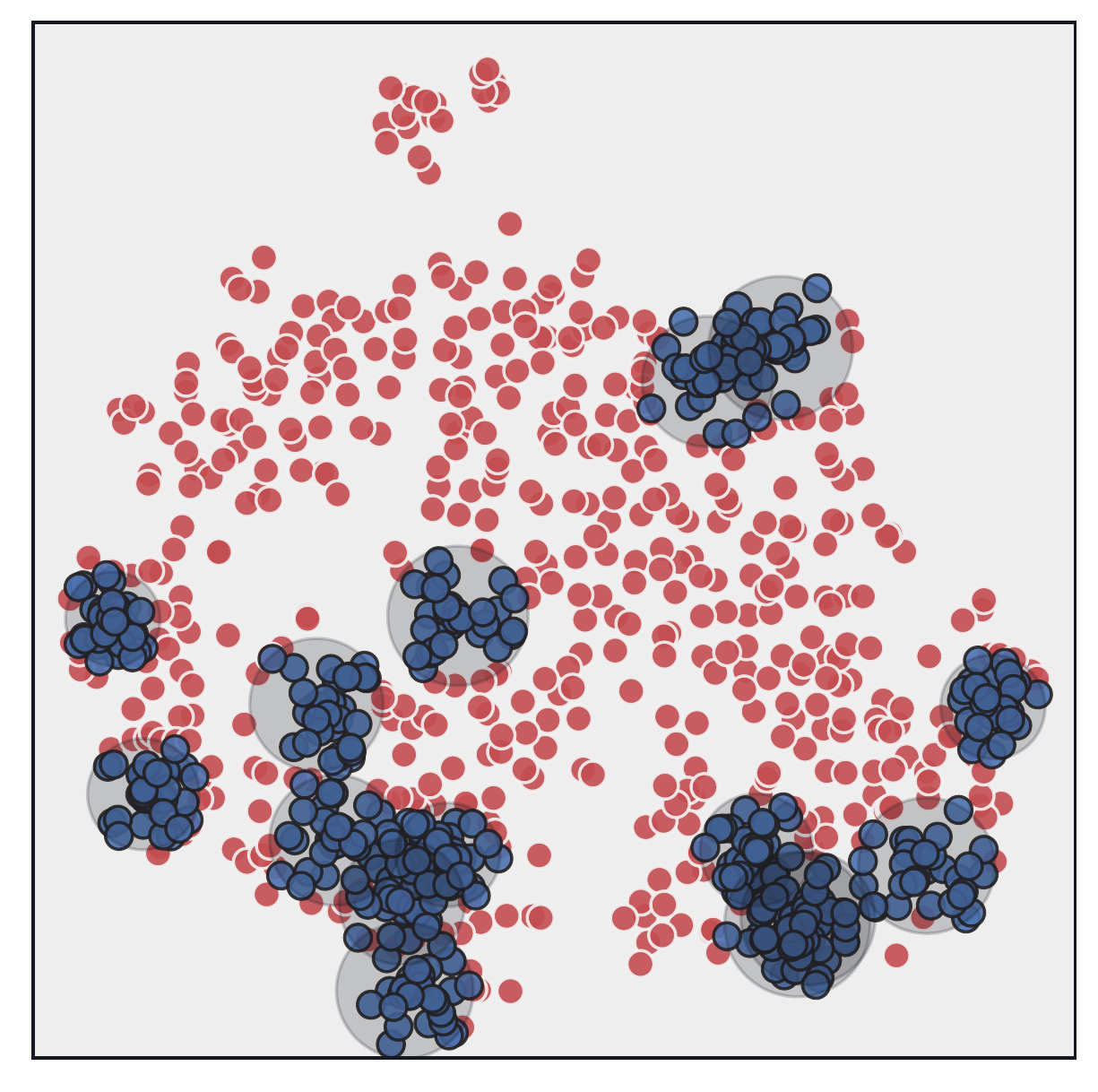}
  \caption{energy $= 0.25$}
\end{subfigure}
~
\begin{subfigure}[b]{0.23\textwidth}
  \includegraphics[width=\textwidth]{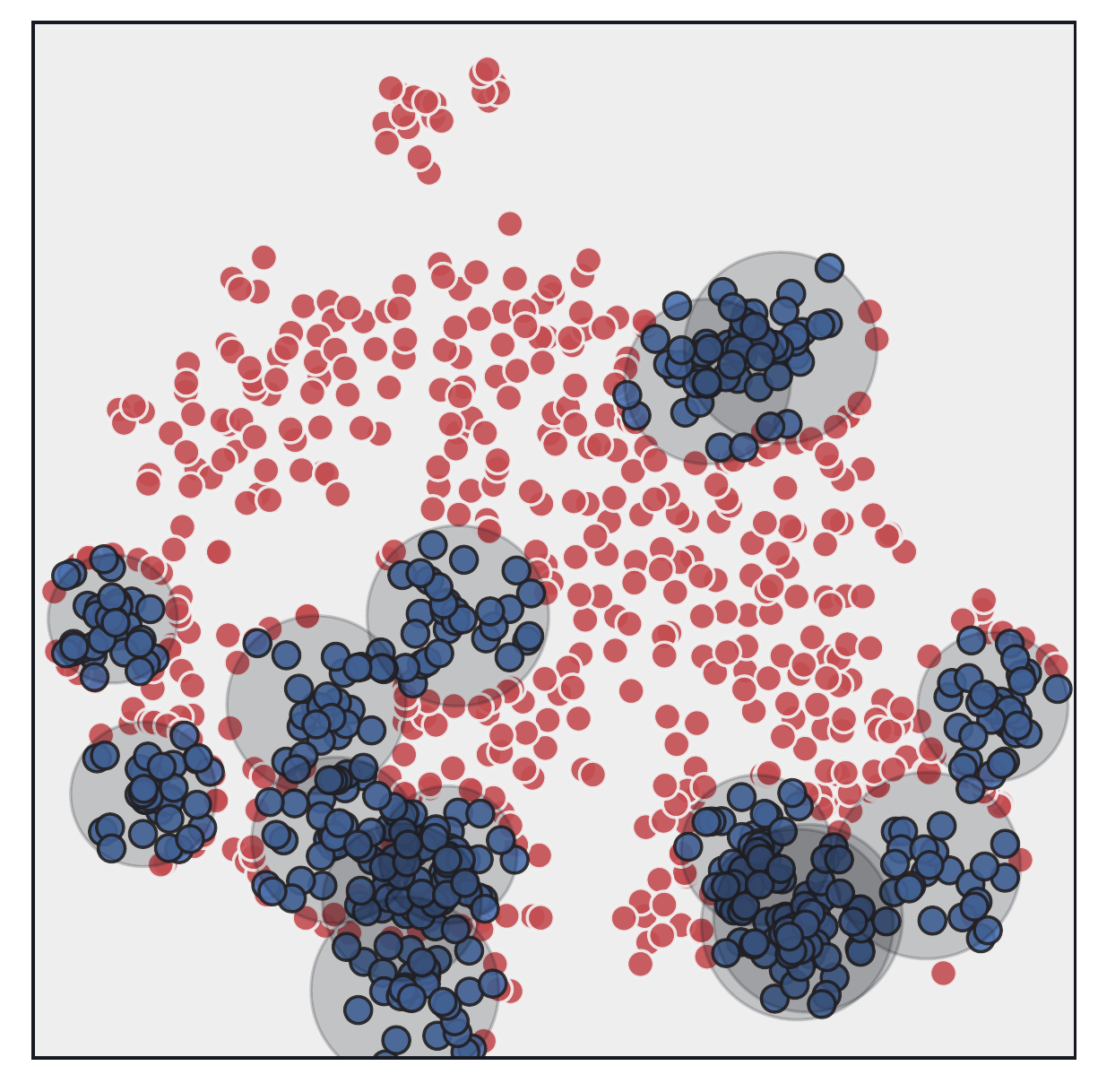}
  \caption{energy $= 0.5$}
\end{subfigure}
~
\begin{subfigure}[b]{0.23\textwidth}
  \includegraphics[width=\textwidth]{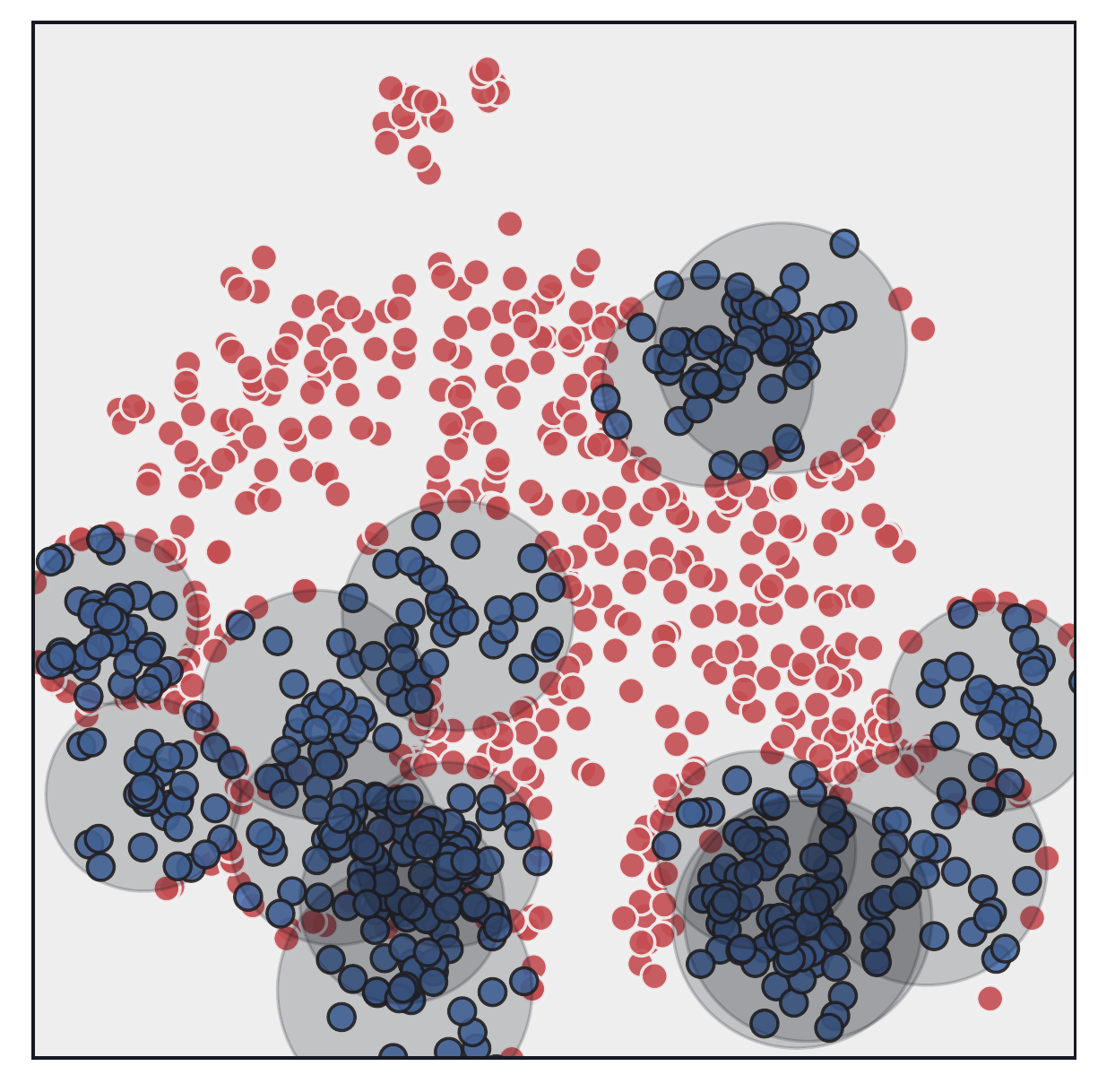}
  \caption{energy $= 1.0$}
\end{subfigure}
\caption{Visualization of the impact of $energy$ parameter on the sphere radius and corresponding region in which synthetic minority observations (indicated by dark outline) are being generated. Note that the majority observations within the sphere are being pushed outside during the cleaning step.}
\label{fig:example-sphere-radius}
\end{figure*}

\begin{figure*}[!htb]
\centering
\begin{subfigure}[b]{0.23\textwidth}
  \includegraphics[width=\textwidth]{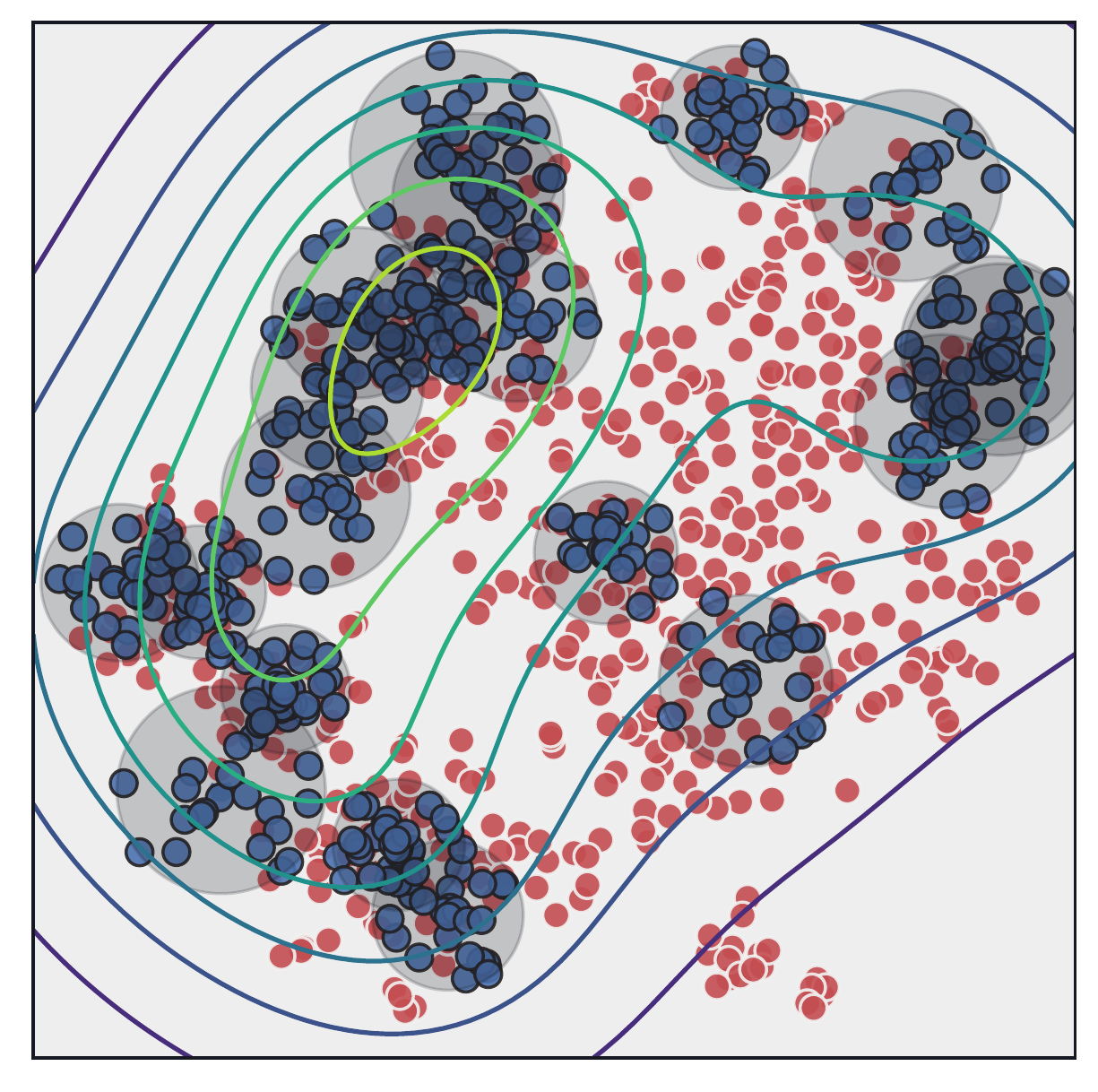}
  \caption{LEH}
\end{subfigure}
~
\begin{subfigure}[b]{0.23\textwidth}
  \includegraphics[width=\textwidth]{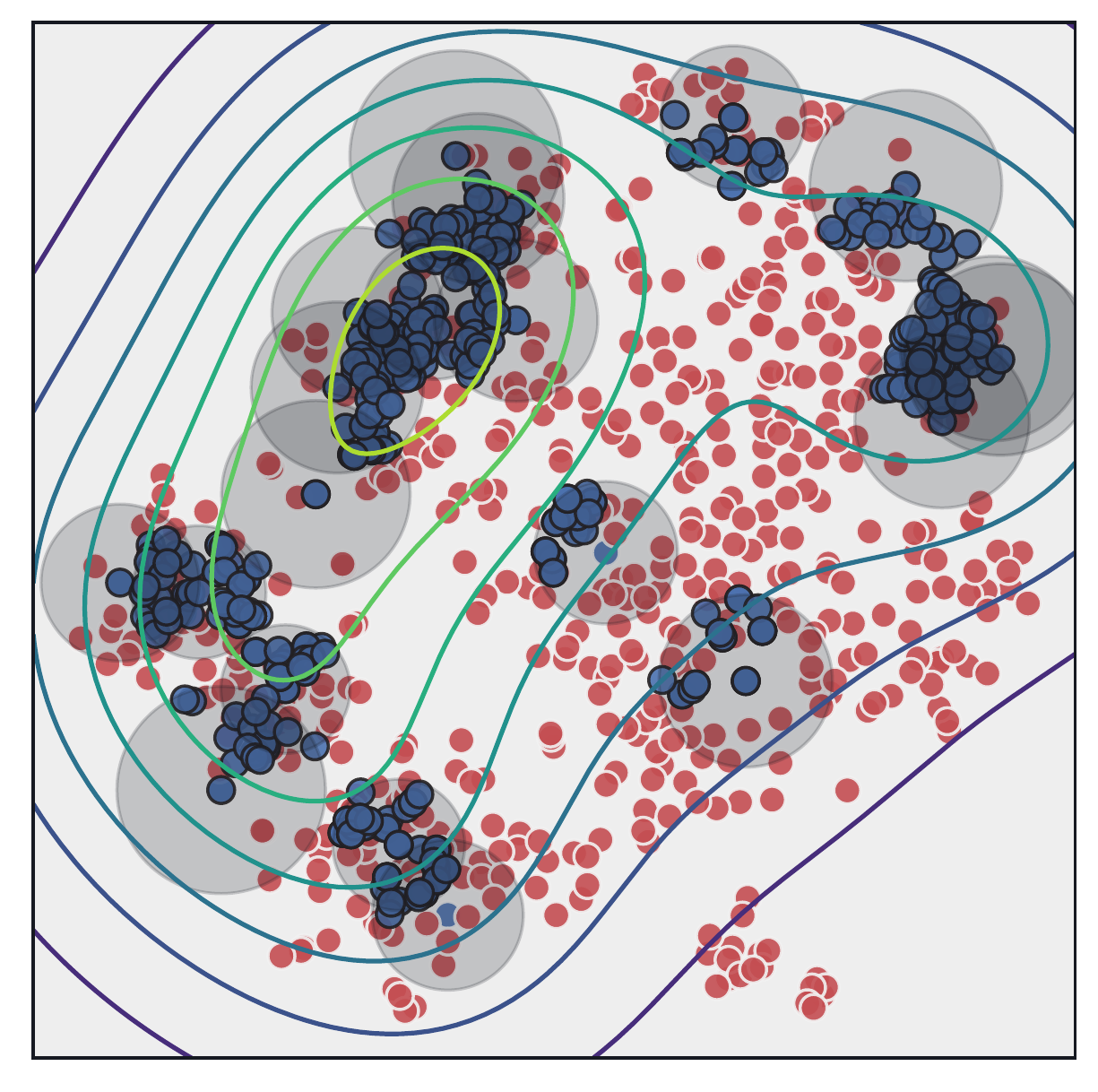}
  \caption{H}
\end{subfigure}
~
\begin{subfigure}[b]{0.23\textwidth}
  \includegraphics[width=\textwidth]{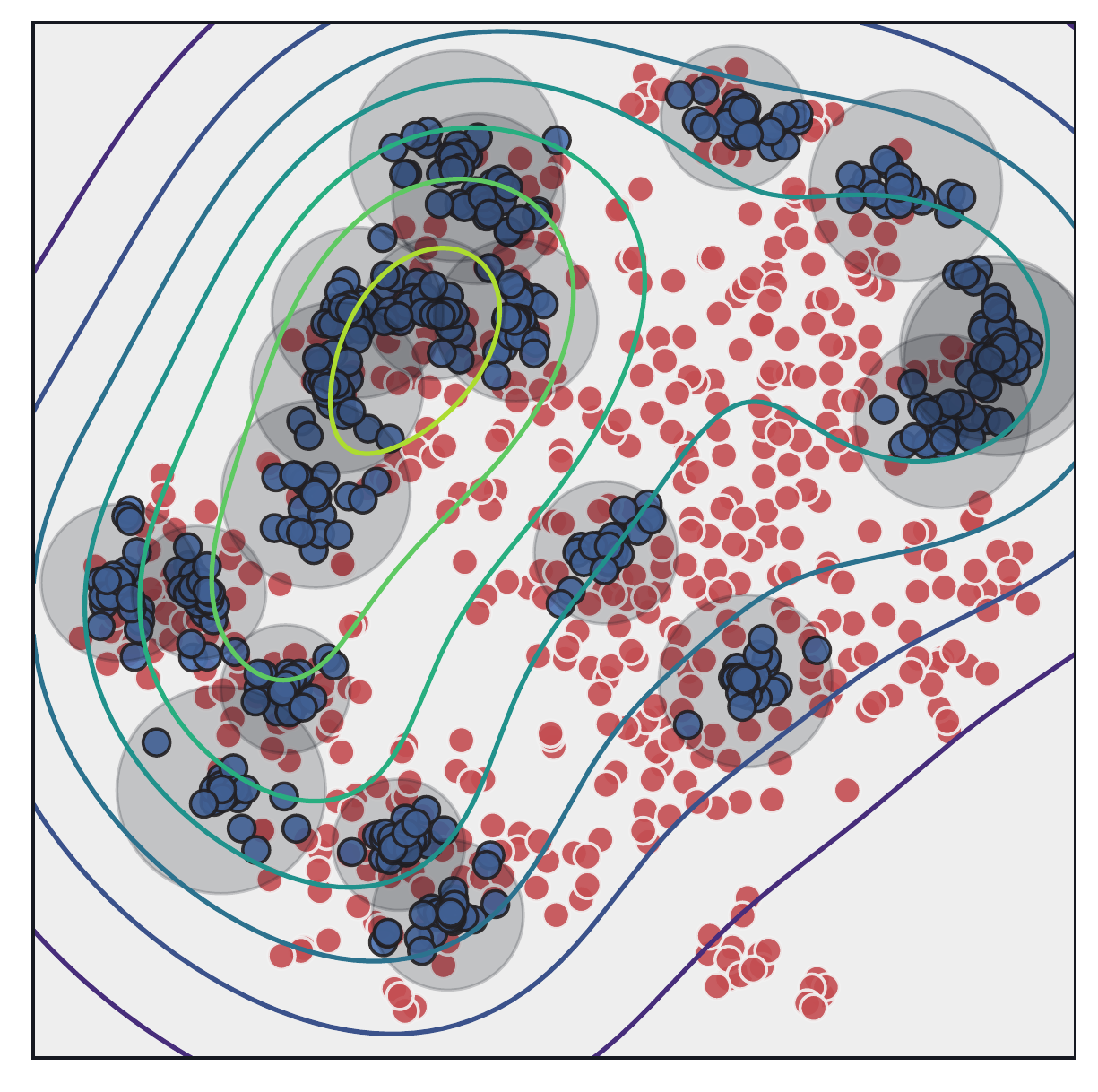}
  \caption{E}
\end{subfigure}
~
\begin{subfigure}[b]{0.23\textwidth}
  \includegraphics[width=\textwidth]{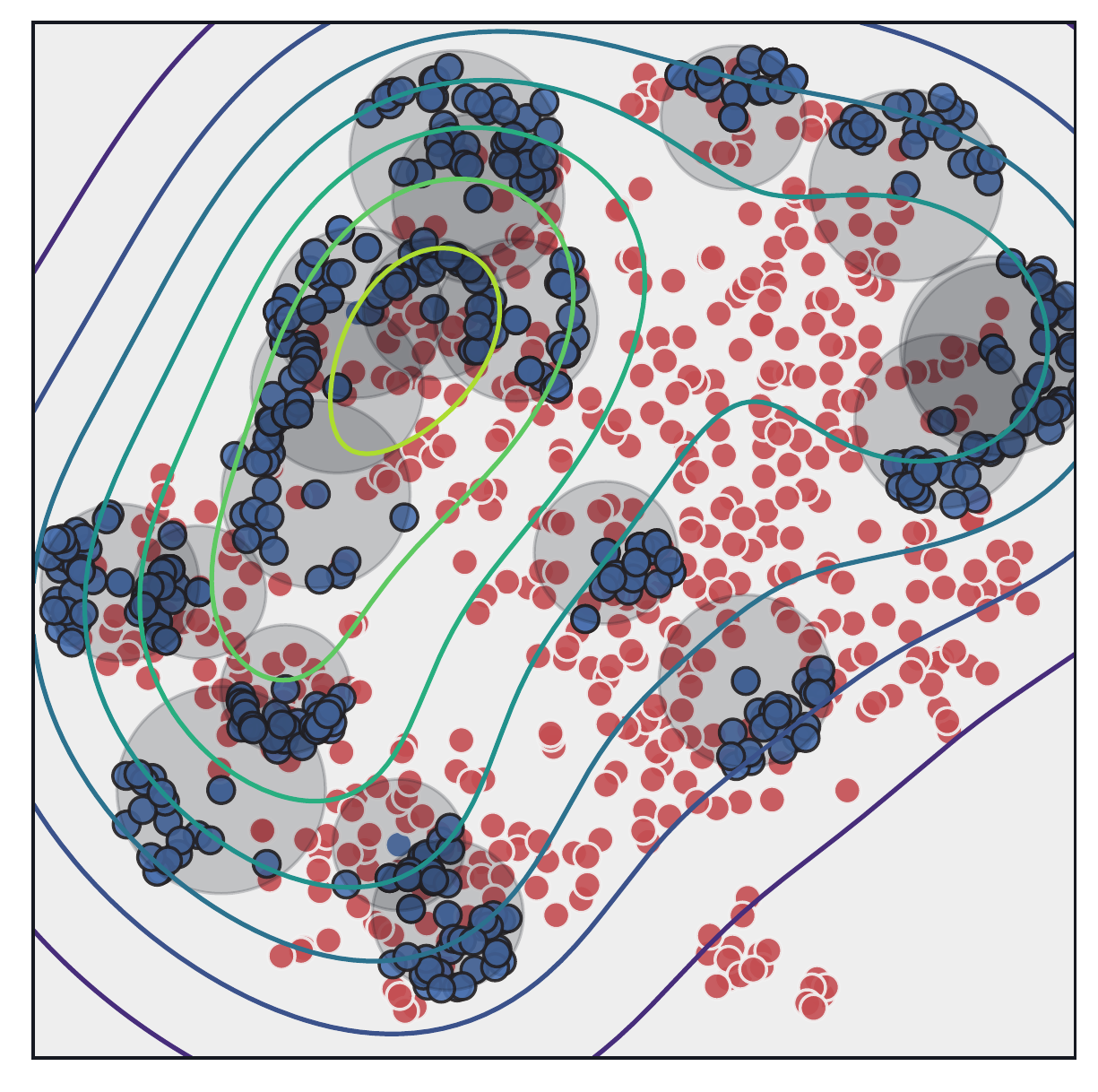}
  \caption{L}
\end{subfigure}
\caption{An example of the choice of sampling region on the distribution of generated minority observations. Baseline case, equivalent to sampling in all of the possible regions (LEH), was compared with sampling in the high (H), equal (E) and low (L) potential regions. Note that the distribution of generated
\label{fig:example-regions-global} observations aligns with the shape of the potential field.}
\end{figure*}

The behavior of the proposed algorithm changes depending on the choice of its three major hyperparameters: RBF spread $\gamma$, energy used for sphere expansion, and sampling region. The impact of $\gamma$ is illustrated in Figure~\ref{fig:example-potential}. As can be seen, $\gamma$ regulates the smoothness of the potential shape, with low values of $\gamma$ producing a less regular contour, conditioned mainly on the position of minority neighbors located in a close proximity. On the contrary, higher $\gamma$ values produce a smoother, less prone to overfitting potential, with a smaller number of distinguishable clusters. Secondly, the value of energy affects the radius of the produced spheres, which controls the size of sampling regions and the range of translations, as illustrated in Figure~\ref{fig:example-sphere-radius}. It is worth noting that as the energy approaches zero, the algorithm degenerates to random oversampling. The choice of the energy is also highly dependent on the dimensionality of the data, and has to be scaled to the number of features a given dataset contains, with higher dimensional datasets requiring higher energy to achieve a similar sphere expansion. Finally, the choice of the sampling region determines how the generated samples align with the minority class potential, which is demonstrated in Figure~\ref{fig:example-regions-global}. Sampling in all of the available regions (LEH), equivalent to the original CCR algorithm, completely ignores the potential and uses whole spheres as a region of interest. Sampling in the E region constraints samples in areas of potential approximately equal to the sphere center, that is the underlying, real minority observation. Sampling in the H region pushes the generated observations towards regions of the data space estimated to have a higher minority class probability, 
which can be interpreted as focusing the sampling process on generating samples that are safer, more resembling the original minority observations. The opposite is true for sampling in the L region.

\begin{figure*}[!htb]
\centering
\begin{subfigure}[b]{0.23\textwidth}
  \includegraphics[width=\textwidth]{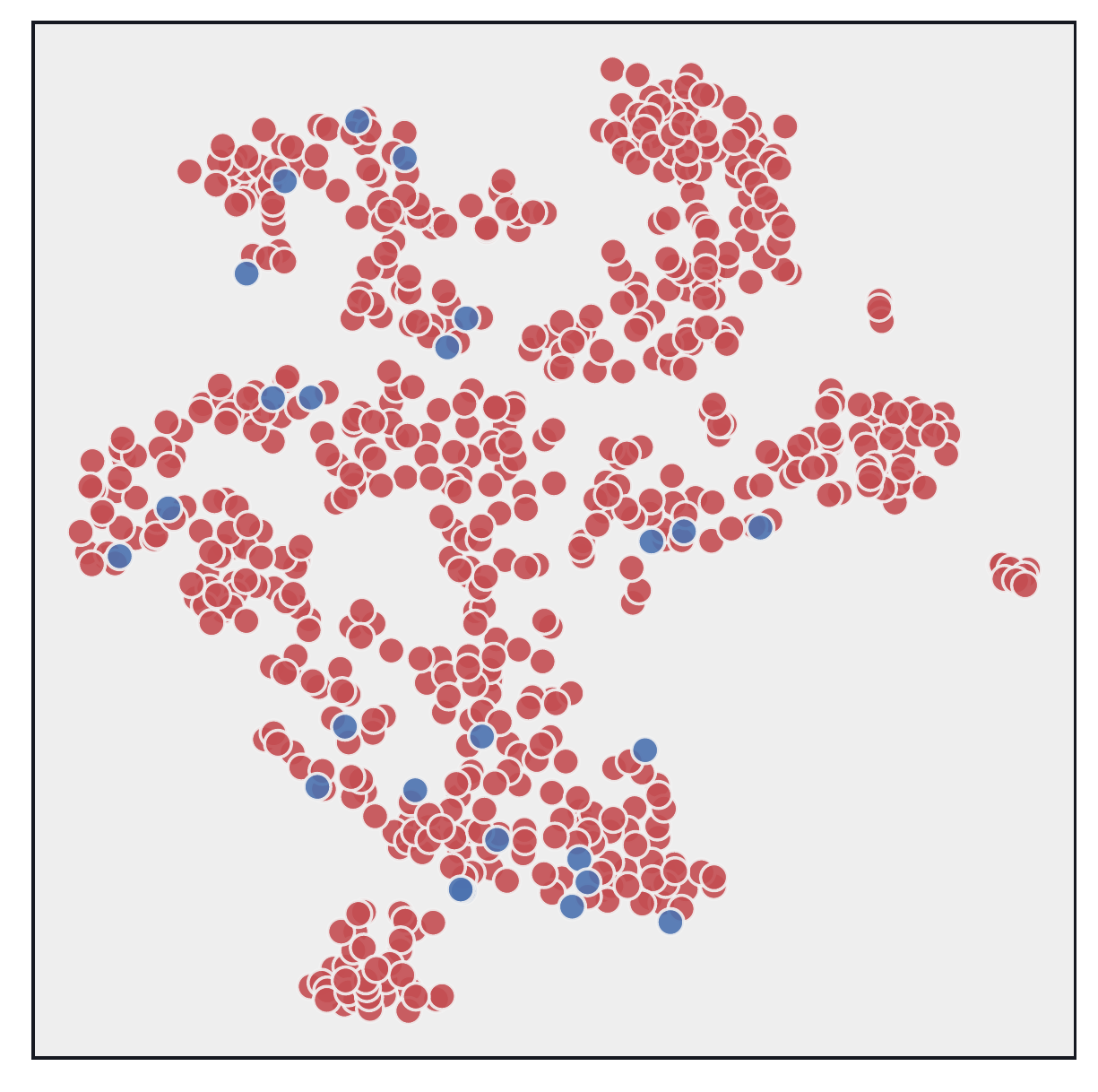}
  \caption{original data}
\end{subfigure}
~
\begin{subfigure}[b]{0.23\textwidth}
  \includegraphics[width=\textwidth]{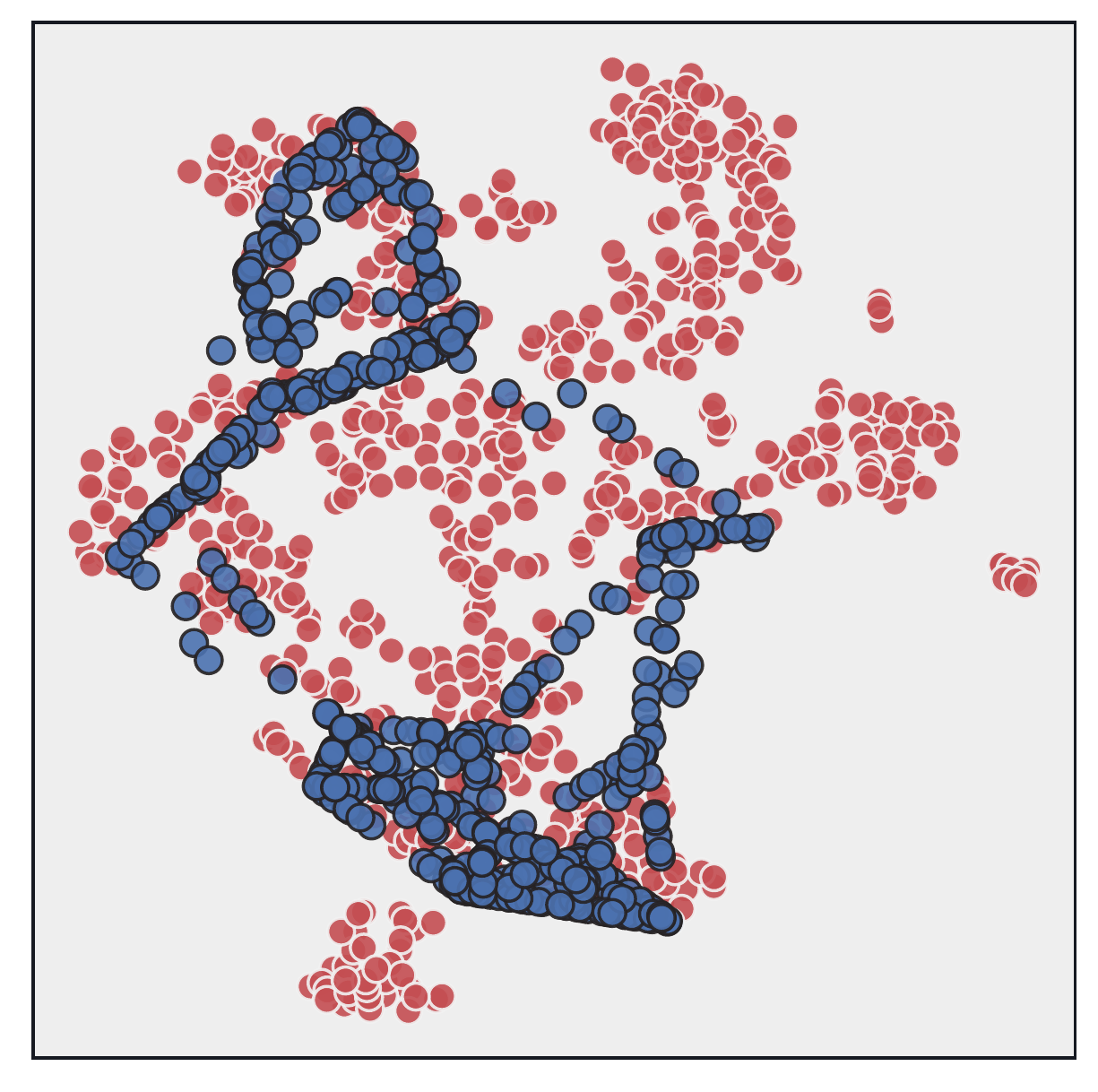}
  \caption{SMOTE}
\end{subfigure}
~
\begin{subfigure}[b]{0.23\textwidth}
  \includegraphics[width=\textwidth]{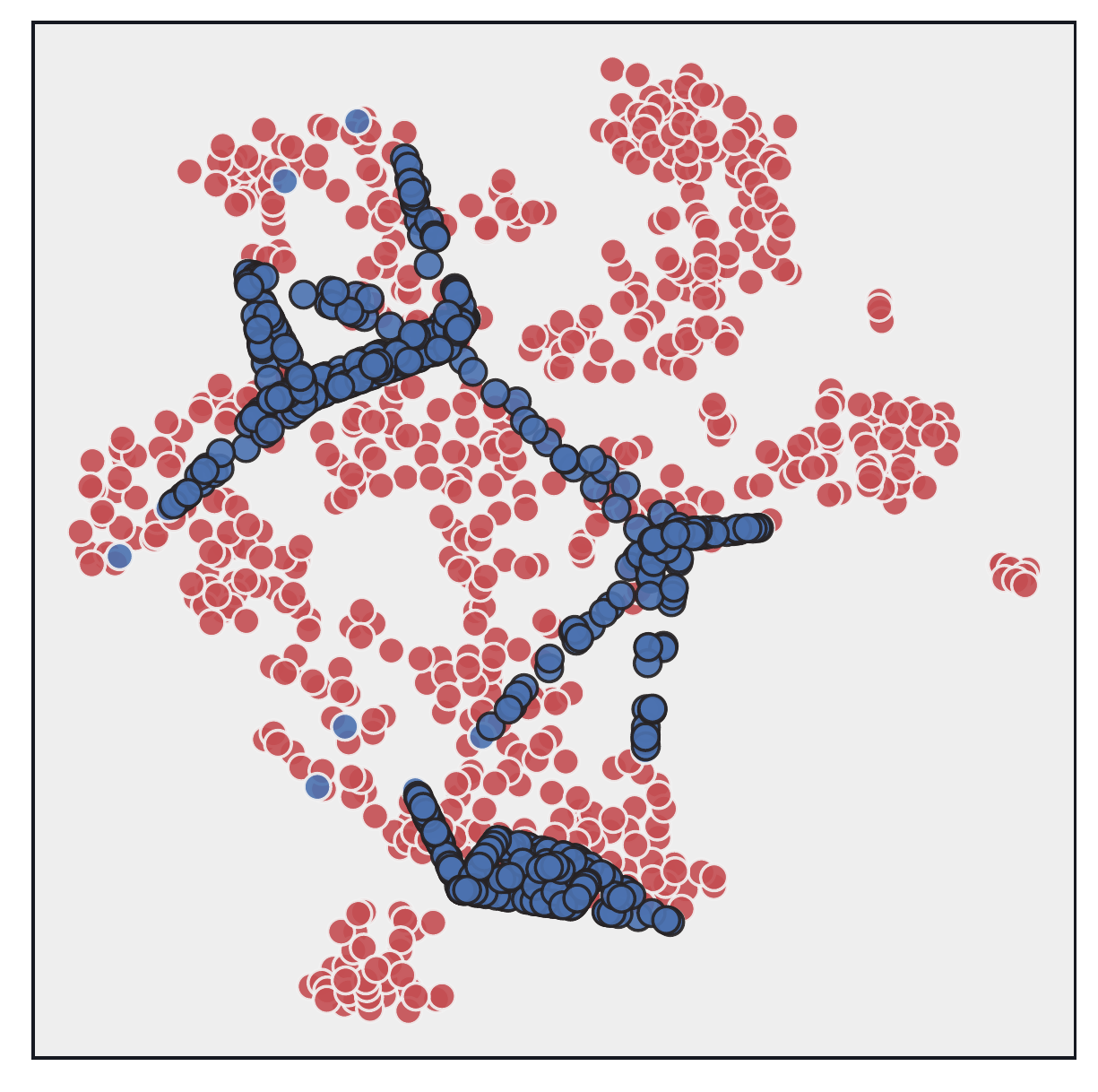}
  \caption{Bord}
\end{subfigure}

\begin{subfigure}[b]{0.23\textwidth}
  \includegraphics[width=\textwidth]{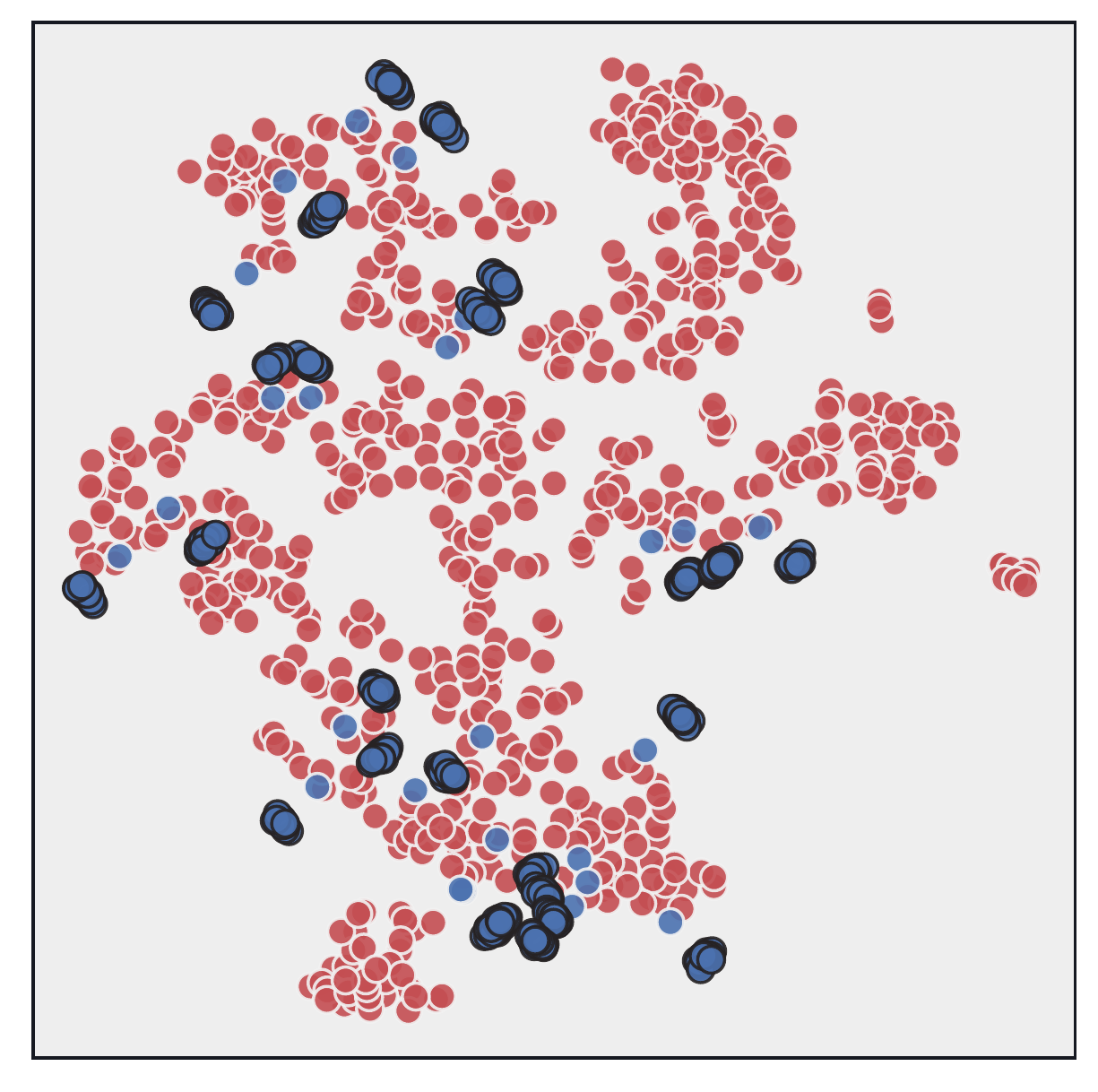}
  \caption{RBO}
\end{subfigure}
~
\begin{subfigure}[b]{0.23\textwidth}
  \includegraphics[width=\textwidth]{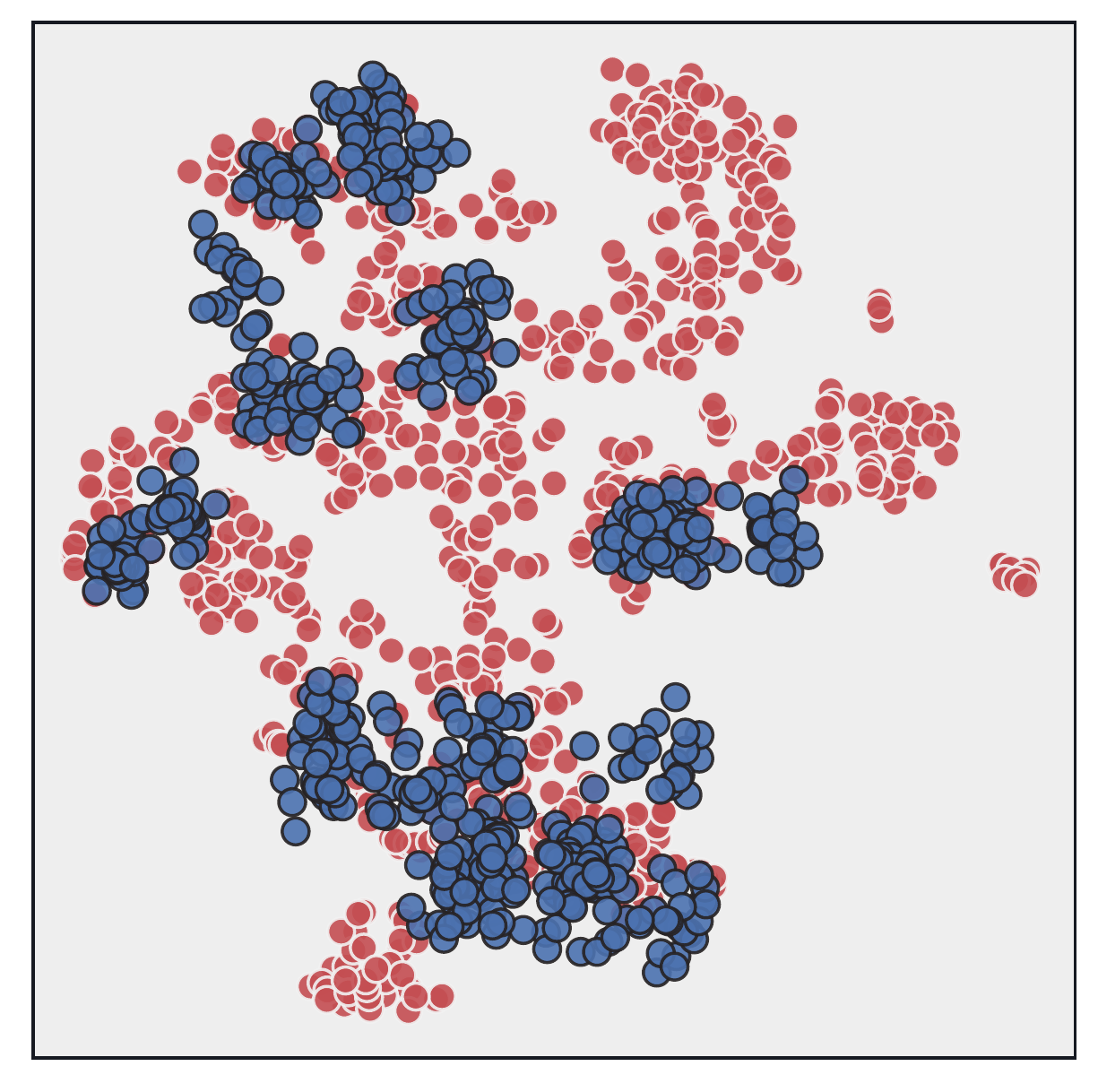}
  \caption{CCR}
\end{subfigure}
~
\begin{subfigure}[b]{0.23\textwidth}
  \includegraphics[width=\textwidth]{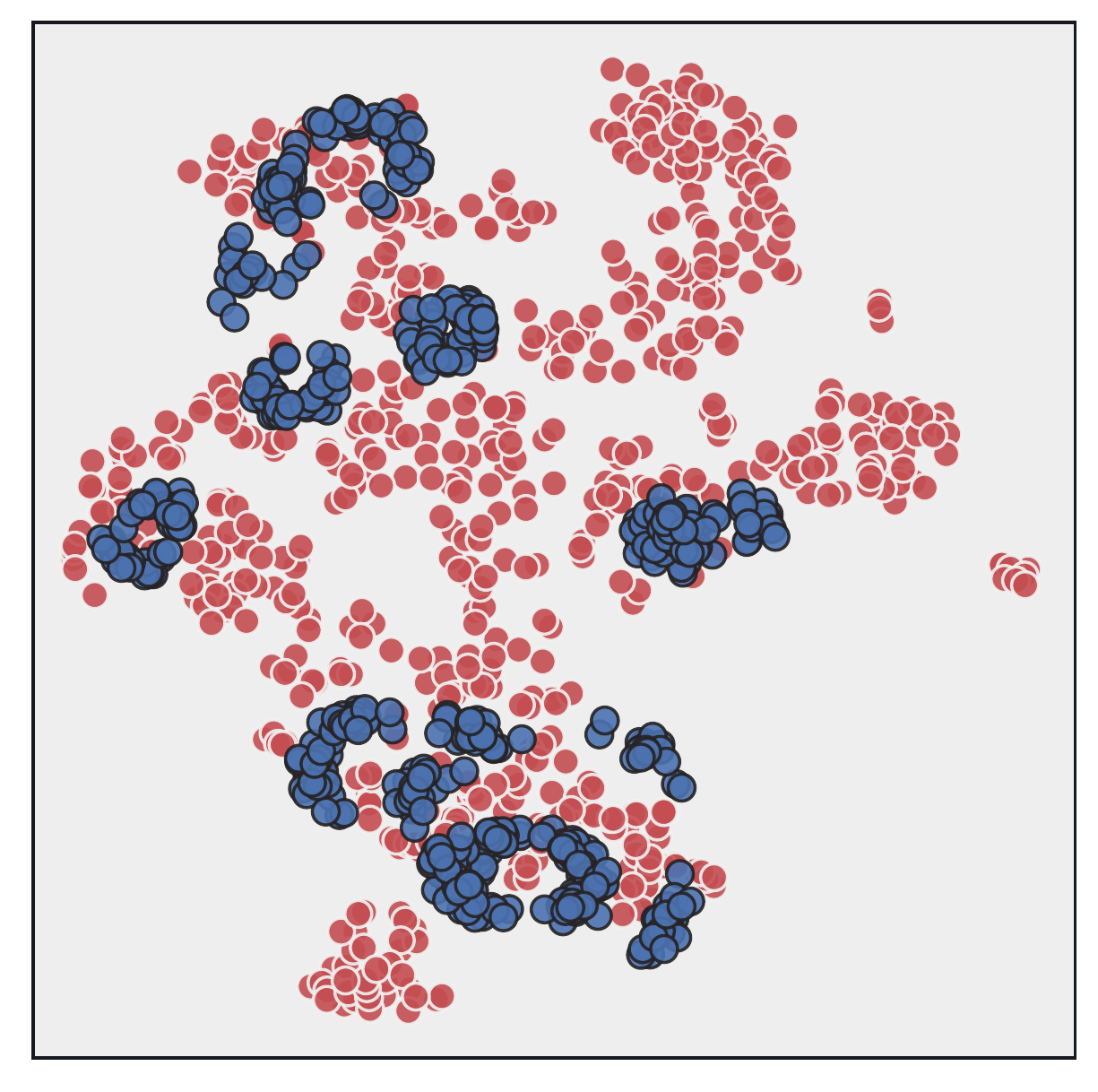}
  \caption{RB-CCR}
\end{subfigure}
\caption{A comparison of data distribution after oversampling with different algorithms on a highly imbalanced dataset with disjoint minority class distributions. SMOTE introduces a high degree of class overlap; Bord solves the problem only partially, still introducing some overlap, at the same time completely omitting oversampling around selected observations. RBO does not produce artificial overlap, but at the same time is very conservative during sampling, in particular within originally overlapping regions. CCR and RB-CCR produce a distribution that leads to a higher bias towards the minority class, both due to synthesizing observations around all of the instances, as well as the conducted translation of majority observations, at the same time minimizing class overlap. Compared to CCR, RB-CCR produces more constrained samples based on the underlying potential.}
\label{fig:example-method-comparison}
\end{figure*}

Finally, it is worth discussing how RB-CCR compares to the other oversampling algorithms. An illustration of differences between several popular methods is presented in Figure~\ref{fig:example-method-comparison}, with a highly imbalanced dataset characterized by a disjoint minority class distribution used as a benchmark. As can be seen, when compared to the SMOTE-based approaches, RB-CCR tends to introduce lower class overlap, which can occur for SMOTE when dealing with disjoint distributions, the presence of noise or outliers. RBO avoids the problem of sampling in the majority class regions, however, it produces very conservative and highly clustered samples. These can cause the classifier to overfit in a manner similar to random oversampling. RB-CCR avoids the risk of overfitting with larger regions of interest. Moreover, the larger regions enable a greater reduction in classifiers bias towards the majority class. The energy parameter facilitates the control of this behavior, with higher values of energy leading to less conservative sampling, and information provided by the class potential is used to fine-tune the shape of regions of interest within the sphere. This enables better control of the sampling.

\noindent\textbf{Main results.} During the experimental study described in \cite{rbccr} it was demonstrated that with a proper choice of sampling region RB-CCR outperforms CCR, and cross-validation can be reliably used to determine the proper choice of sampling region for a given dataset. The empirical assessment also shown that sampling the high potential region with RB-CCR generally produces significantly better precision and specificity, with less impact on recall than CCR. Thus, RB-CCR achieves a better balance in the precision-recall trade-off. Moreover, on average RB-CCR outperforms the considered reference methods with respect to recall, AUC and G-mean. Finally, during comparison of various classification and resampling algorithm combinations it was demonstrated that RB-CCR combined with a multi-layer perceptron tends to outperform all of the other considered algorithm combinations, making it a sensible default for a practical use.

\section{CSMOUTE: Combined Synthetic Oversampling and Undersampling}

The Combined Synthetic Oversampling and Undersampling (CSMOUTE) \cite{koziarski2020csmoute} algorithm, similar to RBU, attempts to extend the design technique previously utilized in the oversampling to the undersampling procedure. Specifically, it expands on the idea of using interpolation between nearby instances, previously introduced in the oversampling setting in the form of SMOTE \cite{chawla2002smote}. SMOTE became a cornerstone of numerous data-level strategies for handling data imbalance \cite{Fernandez:2018}. It was introduced as a direct response to the shortcomings of random oversampling, which was shown by Chawla et al. \cite{chawla2002smote} to cause overfitting in selected classification algorithms, such as decision trees. Instead of simply duplicating the minority instances, which was the case in random oversampling, SMOTE instead advocates for generating synthetic observations via data interpolation. It is worth noting that this approach does not remain without its on disadvantages, previously discussed in Section~\ref{sec:limitations}. Still, despite its flaws, SMOTE remains one of the most important approaches for handling data imbalance. 

\noindent\textbf{The algorithm.} At the hearth of the proposed approach lies the Synthetic Majority Undersampling Technique (SMUTE), a conceptually simple method that leverages the concept of interpolation of nearby instances, previously introduced in the oversampling setting in SMOTE. Specifically, the proposed algorithm focuses on the majority class observations, and at each iteration randomly selects one of them and one of their $k$ nearest neighbors, with $k$ being a parameter of the algorithm. Afterwards, a new instance is synthesized, similar to the SMOTE, added to the collection of majority observations, and at the same time two of the original instances are removed from the dataset. The pseudocode of the proposed approach is presented in Algorithm~\ref{algorithm:csmoute}. In contrast to the SMOTE, the proposed procedure is not motivated by the need to combat overfitting, which is not as pressing an issue in the case of undersampling, but instead the one to minimize the loss of information due to the instance removal. The idea is similar to various prototype selection approaches, in which the original collection of majority observations is reduced to a smaller collection of representative instances, e.g. by applying clustering. In fact, the proposed approach can be viewed as a special case of a local clustering, in which instead of using the whole set of majority observations only its subset is selected, in this case an extreme collection consisting only of two nearby observations, and a random noise is added to the resulting synthetic observation. However, in contrast with the traditional clustering approaches, likely to operate on either a whole collection of majority observations, or its much larger subset, proposed approach differs in two significant ways. First of all, by using only two observations during the reduction step, the capabilities of the algorithm to generate representative prototypes are reduced, an obvious drawback of the method. However, since the reduction step originates in the position of a particular existing observation, it allows us to guide the resampling procedure. Specifically, it enables us to select only the points of origin with particular characteristics, for instance only the borderline instances, or only the outliers, which is either not possible or not as easily achievable in the traditional clustering approaches. Similarly, it allows us to adjust the ratio of points of origin of a given type what will be used for reduction, or accommodate other local data characteristics. This is further facilitated by the second of the proposed approaches, the Combined Synthetic Oversampling and Undersampling Technique (CSMOUTE), which simply combines SMOTE oversampling with SMUTE undersampling at a given ratio of over- and undersampling.

\begin{algorithm}
\caption{Combined Synthetic Oversampling and Undersampling}
\textbf{Input:} collections of majority observations $\mathcal{X}_{maj}$ and minority observations $\mathcal{X}_{min}$ \\
\textbf{Parameters:} number of nearest neighbors used during oversampling $k_{SMOTE}$ and undersampling $k_{SMUTE}$, $ratio$ of data balancing done using oversampling \\
\textbf{Output:} resampled collections of majority observations $\mathcal{X}_{maj}'$ and minority observations $\mathcal{X}_{min}'$
\label{algorithm:csmoute}	
\vspace{-0.5\baselineskip}

\hrulefill
\begin{algorithmic}[1]
\STATE \textbf{function} SMOTE($k$, $n$):
\STATE $\mathcal{X}_{min}' \gets \mathcal{X}_{min}$
\WHILE{$\vert \mathcal{X}_{min}' \vert - \vert \mathcal{X}_{min} \vert < n$}
\STATE $x_1 \gets $ randomly selected observation from $\mathcal{X}_{min}$
\STATE $x_2 \gets $ randomly selected observation from $k$ nearest neighbors of $x_1$ belonging to $\mathcal{X}_{min}$
\STATE $r \gets $ random real number in $[0, 1]$
\STATE $x' \gets x_1 + r \cdot (x_2 - x_1)$
\STATE append $x'$ to $\mathcal{X}_{min}'$
\ENDWHILE
\STATE \textbf{return} $\mathcal{X}_{min}'$
\vspace{\baselineskip}
\STATE \textbf{function} SMUTE($k$, $n$):
\STATE $\mathcal{X}_{maj}' \gets \mathcal{X}_{maj}$
\WHILE{$\vert \mathcal{X}_{maj} \vert - \vert \mathcal{X}_{maj}' \vert < n$}
\STATE $x_1 \gets $ randomly selected observation from $\mathcal{X}_{maj}'$
\STATE $x_2 \gets $ randomly selected observation from $k$ nearest neighbors of $x_1$ belonging to $\mathcal{X}_{maj}'$
\STATE $r \gets $ random real number in $[0, 1]$
\STATE $x' \gets x_1 + r \cdot (x_2 - x_1)$
\STATE delete $x_1$ and $x_2$ from $\mathcal{X}_{maj}'$
\STATE append $x'$ to $\mathcal{X}_{maj}'$
\ENDWHILE
\STATE \textbf{return} $\mathcal{X}_{maj}'$
\vspace{\baselineskip}
\STATE \textbf{function} CSMOUTE($k_{SMOTE}$, $k_{SMUTE}$, $ratio$):
\STATE $n \gets \vert \mathcal{X}_{maj} \vert - \vert \mathcal{X}_{min} \vert$
\STATE $n_{SMOTE} \gets \textrm{round}(n \cdot ratio)$
\STATE $n_{SMUTE} \gets n - n_{SMOTE}$
\STATE $\mathcal{X}_{min}' \gets \textrm{SMOTE}(k_{SMOTE}, n_{SMOTE})$
\STATE $\mathcal{X}_{maj}' \gets \textrm{SMUTE}(k_{SMUTE}, n_{SMUTE})$
\STATE \textbf{return} $\mathcal{X}_{maj}'$, $\mathcal{X}_{min}'$
\end{algorithmic}
\end{algorithm}

\noindent\textbf{Main results.} In the study described in \cite{koziarski2020csmoute} it was demonstrated that SMUTE by itself is a reasonable alternative to other undersampling approaches, such as random undersampling, and that by combining SMUTE undersampling with SMOTE oversampling it is possible to further improve the algorithms performance compared to both stand-alone alternatives. Furthermore, it was demonstrated that CSMOUTE outperforms the considered reference methods. Finally, during the experimental analysis it was demonstrated that the performance of the algorithm shows a significant correlation with the characteristics of a dataset on which it is applied, with the performance improved on datasets consisting of a high proportion of outlier instances, and worsened on datasets with a high proportion of borderline instances. The fact that the algorithm displays a significantly different performance based on the global dataset characteristics suggests that it might be feasible to apply resampling selectively, based on the local characteristics of the data, to achieve a better performance. Specifically, a promising strategy might incorporate placing a particular focus on the outlier instances during the undersampling, at the same time treating borderline instances differently, for example by either excluding them completely or by applying oversampling instead.

\section{PA: Potential Anchoring}

The Potential Anchoring (PA) \cite{pa} algorithm also utilizes radial basis functions and the concept of class potential introduced with RBO, but contrary to the previously described radial-based approaches, the aim of PA is preserving the shape of the underlying class distributions. This is motivated by the fact that the majority of the existing oversampling strategies, such as Borderline-SMOTE \cite{Han:2005}, Safe-Level-SMOTE \cite{Bunkhumpornpat:2009}, ADASYN \cite{He:2008}, and MWMOTE \cite{barua2012mwmote}, modify the underlying class distribution in oversampling process, focusing the generation of synthetic observations in a specific regions. However, all of the aforementioned resampling strategies are at the same time based on different, often contradictory, ideas on where the resampling process should be focused. While all of them have their own niches of outperformance, out of necessity they are specialized, and it is often not clear which method, if any, is preferred in a general case.

Instead of using an ad-hoc strategy of boosting specific regions of data space, PA takes the approach of preserving the original shape of underlying class distribution. Specifically, this is achieved by treating the generated synthetic observations as optimization parameters, which are positioned to minimize the difference between the potential of the original and resampled observations, with a regularization constraint added to prevent overfitting during oversampling. The proposed framework is used in both over- and undersampling, presenting a unified framework for the two of them.

\noindent\textbf{Potential resemblance loss.} PA is based on the previously defined concept of potential function (Equation~\ref{eq:potential}). Recall that class potential can be viewed as a measure of density of observations from that class. The values of potential function are not, however, bound to any specific range, making it difficult to compare the relative shape of potential computed with respect to two different collections of observations. To mitigate this issue, a normalized potential function $\Psi$ is proposed. This function computes the potential for $k$ anchor points $\mathcal{A}$, and returns a vector of $k$ normalized potentials. More formally, the normalized potential function is defined as
\begin{equation}
    \Psi(\mathcal{X}, \mathcal{A}, \gamma) = \frac{1}{\sum_{i=1}^{k}{\Phi(\mathcal{A}_i, \mathcal{X}, \gamma)}} 
    \left[
    \begin{gathered}
    \Phi(\mathcal{A}_1, \mathcal{X}, \gamma) \\
    \Phi(\mathcal{A}_2, \mathcal{X}, \gamma) \\
    ... \\
    \Phi(\mathcal{A}_{k}, \mathcal{X}, \gamma)
    \end{gathered}
    \right].
    \label{eq:normalized-potential}
\end{equation}

Due to the non-negativity of $\Phi$, the values of $\Psi$ are also non-negative and range from 0 to 1. The normalized potential function describes the relative density of observations in any given anchor point $\mathcal{A}_i$. This property makes it possible to directly compare the outputs of $\Psi$ computed with respect to two different collections of observations, even if the collections differ in size, as will be the case during resampling.

Finally, based on the concept of normalized potential, the potential resemblance loss is defined. Given a collection of original observations $\mathcal{X}$, $k$ anchor points $\mathcal{A}$, collection of prototypes $\mathcal{P}$, that is generated observations the position of which we wish to optimize, their starting positions $\mathcal{P}^0$, radial basis function spread $\gamma$, and regularization coefficient $\lambda$, the potential resemblance loss is defined as
\begin{equation}
    \begin{aligned}
    \mathcal{L}(\mathcal{X}, \mathcal{A}, \mathcal{P}, \mathcal{P}^0, \gamma, \lambda) = & \sum_{i=1}^{k}{(\Psi(\mathcal{X}, \mathcal{A}, \gamma)_i - \Psi(\mathcal{P}, \mathcal{A}, \gamma)_i)^2} \\ 
    & + \lambda  \sum_{i=1}^{\mid \mathcal{P} \mid}{e^{-\left(\frac{\lVert \mathcal{P}_i - \mathcal{P}^0_i \rVert_2}{\gamma}\right)^{2}}}
    \end{aligned}.
    \label{eq:loss}
\end{equation}

The left-side term of the equation is a mean squared error between the normalized potential computed with respect to $\mathcal{X}$ and $\mathcal{P}$, and as such measures the difference in relative shape of the potential produced by these two collection of observations. The right-side term is a regularization term that measures the displacement of prototypes $\mathcal{P}$ from their starting positions. If $\mathcal{P}^0$ is created by a random sampling of $\mathcal{X}$, as is the case in the proposed approach, this term prevents the algorithm minimizing $\mathcal{L}$ from degenerating to random oversampling, since prototypes $\mathcal{P}$ will be displaced from their original positions. It is worth noting that even though in principle we would like to penalize the placement of $\mathcal{P}$ close to any of the observations from $\mathcal{X}$, in the conducted experiments considering only the starting positions $\mathcal{P}^0$ was sufficient to prevent the overfitting, at the same time being more computationally efficient.

\noindent\textbf{Algorithm.} Being equipped with a potential resemblance loss $\mathcal{L}$, the problem of imbalanced data resampling can be formulated as the optimization of prototype point positions $\mathcal{P}$ with respect to $\mathcal{L}$. While the proposed approach is motivated from the point of view of oversampling, it is also easily applicable to the undersampling, as the same principle of preserving the original class density can be applied in both cases. The main difference between the two is that while during the oversampling the original minority observations can be preserved, and a collection of additional prototypes $\mathcal{P}$ can be used as the synthetic observations, during the undersampling the original majority observations will be instead replaced with a smaller collection of prototypes preserving the original class potential.

The pseudocode of the proposed Potential Anchoring (PA) algorithm is presented in Algorithm~\ref{algorithm:pa}. The method combines over- and undersampling up to the point of achieving balanced class distribution, with the ratio of imbalance eliminated with either over- or undersampling treated as a parameter. First, $k$ anchor points, with respect to which normalized potential will be calculated, are generated via clustering of the collection of original observations $\mathcal{X}$. Second, the prototypes are initialized by randomly sampling with replacement from the collection of observations of a given class. Importantly, small random jitter is afterwards introduced to break the symmetry during the optimization. Finally, the prototypes are then optimized with respect to the potential resemblance loss function $\mathcal{L}$, separately for the majority and the minority class. This loss function is penalized with a regularization coefficient $\lambda$ in the case of oversampling. Throughout the conducted experiments $k$-means clustering was used to generate the anchor points. The differentiability of $\mathcal{L}$ was also leveraged, and the optimization was conducted using Adam optimizer \cite{kingma2014adam}.

\begin{algorithm*}[!htb]
	\caption{Potential Anchoring}
	\textbf{Input:} collection of original observations $\mathcal{X}$ divided into majority $\mathcal{X}_{maj}$ and minority $\mathcal{X}_{min}$ observations \\
	\textbf{Parameters:} $ratio$ of imbalance eliminated with oversampling, number of anchor points $k$, number of $iterations$, radial basis function spread $\gamma$, oversampling regularization coefficient $\lambda$, learning rate $\alpha$, random jitter used for initialization $\epsilon$ \\
    \textbf{Output:} collection of resampled observations $\mathcal{X}'$
		
	\label{algorithm:pa}	
	\vspace{-0.5\baselineskip}
	
	\hrulefill
	\begin{algorithmic}[1]
		\STATE \textbf{function} PA($\mathcal{X}_{maj}$, $\mathcal{X}_{min}$, $ratio$, $k$, $iterations$, $\gamma$, $\lambda$, $\alpha$, $\epsilon$):
		\STATE $\mathcal{A} \gets k$ anchor points obtained by clustering $\mathcal{X}$
		\STATE $n_{PAO} \gets ratio \cdot \left( \lvert \mathcal{X}_{maj} \rvert - \lvert \mathcal{X}_{min} \rvert \right)$
		\STATE $n_{PAU} \gets$ $\lvert \mathcal{X}_{maj} \rvert - \left(1 - ratio\right) \cdot \left( \lvert \mathcal{X}_{maj} \rvert - \lvert \mathcal{X}_{min} \rvert \right)$
		\STATE $\mathcal{P}_{min}^0 \gets n_{PAO}$ prototypes randomly selected with replacement from $\mathcal{X}_{min}$
		\STATE $\mathcal{P}_{maj}^0 \gets n_{PAU}$ prototypes randomly selected with replacement from $\mathcal{X}_{maj}$
		\STATE $\mathcal{P}_{min}, \mathcal{P}_{maj} \gets \mathcal{P}_{min}^0, \mathcal{P}_{maj}^0 $ with added random jitter $\epsilon$
		\FOR{i in 1..$iterations$}
		    \STATE perform optimization step on $\mathcal{P}_{min}$ w.r.t. $\mathcal{L}(\mathcal{X}_{min}, \mathcal{A}, \mathcal{P}_{min}, \mathcal{P}_{min}^0, \gamma, \lambda)$ using learning rate $\alpha$
		\ENDFOR
		\FOR{i in 1..$iterations$}
		    \STATE perform optimization step on $\mathcal{P}_{maj}$ w.r.t. $\mathcal{L}(\mathcal{X}_{maj}, \mathcal{A}, \mathcal{P}_{maj}, \mathcal{P}_{maj}^0, \gamma, 0)$ using learning rate $\alpha$
		\ENDFOR
		\STATE $\mathcal{X}' \gets \mathcal{X}_{min} \cup \mathcal{P}_{min} \cup \mathcal{P}_{maj}$
		\STATE \textbf{return} $\mathcal{X}'$
	\end{algorithmic}
\end{algorithm*}

\begin{figure}[!htb]
\centering
\begin{subfigure}[t]{0.22\textwidth}
  \includegraphics[width=\textwidth]{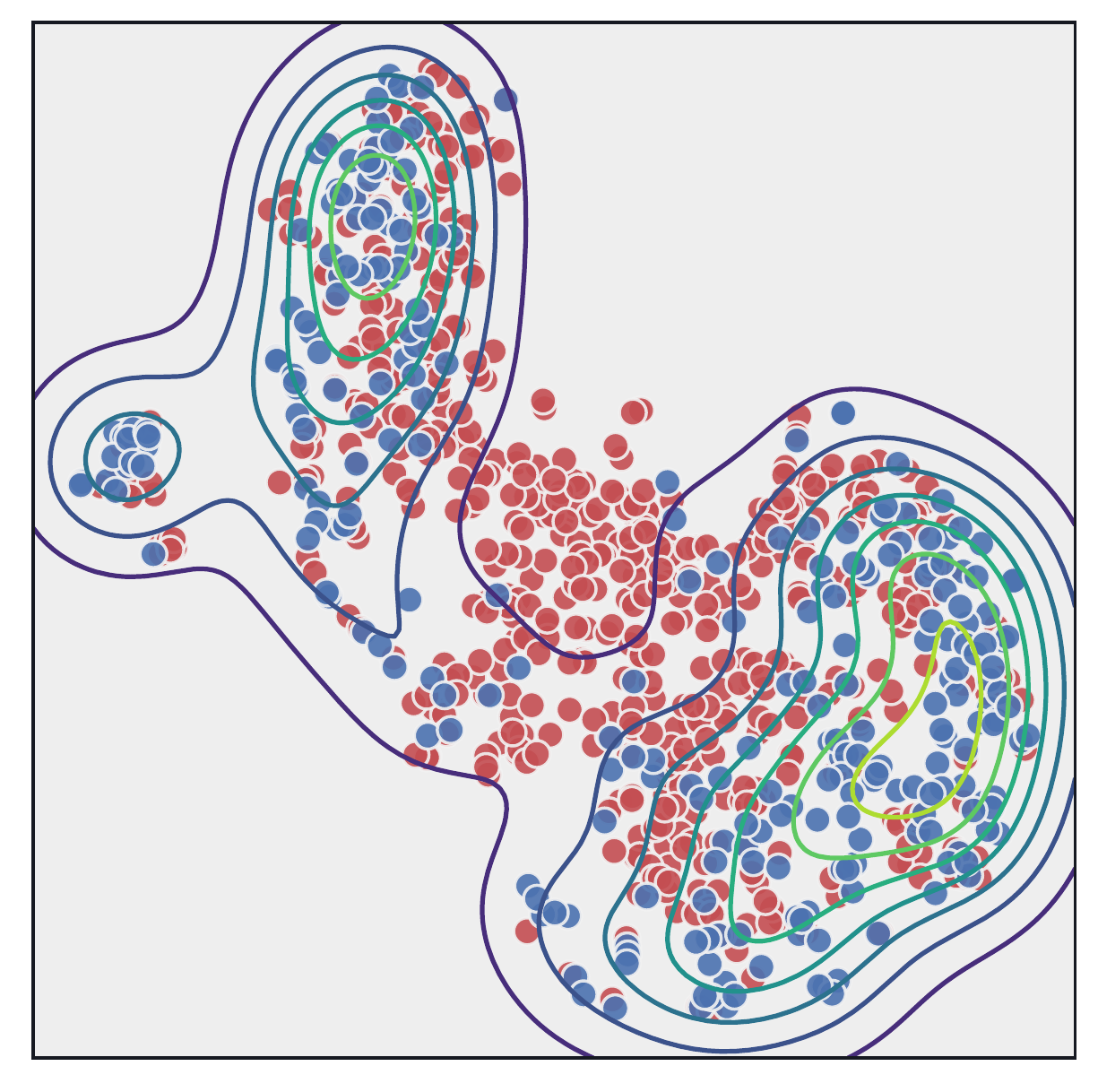}
  \caption{minority class potential}
\end{subfigure}
~
\begin{subfigure}[t]{0.22\textwidth}
  \includegraphics[width=\textwidth]{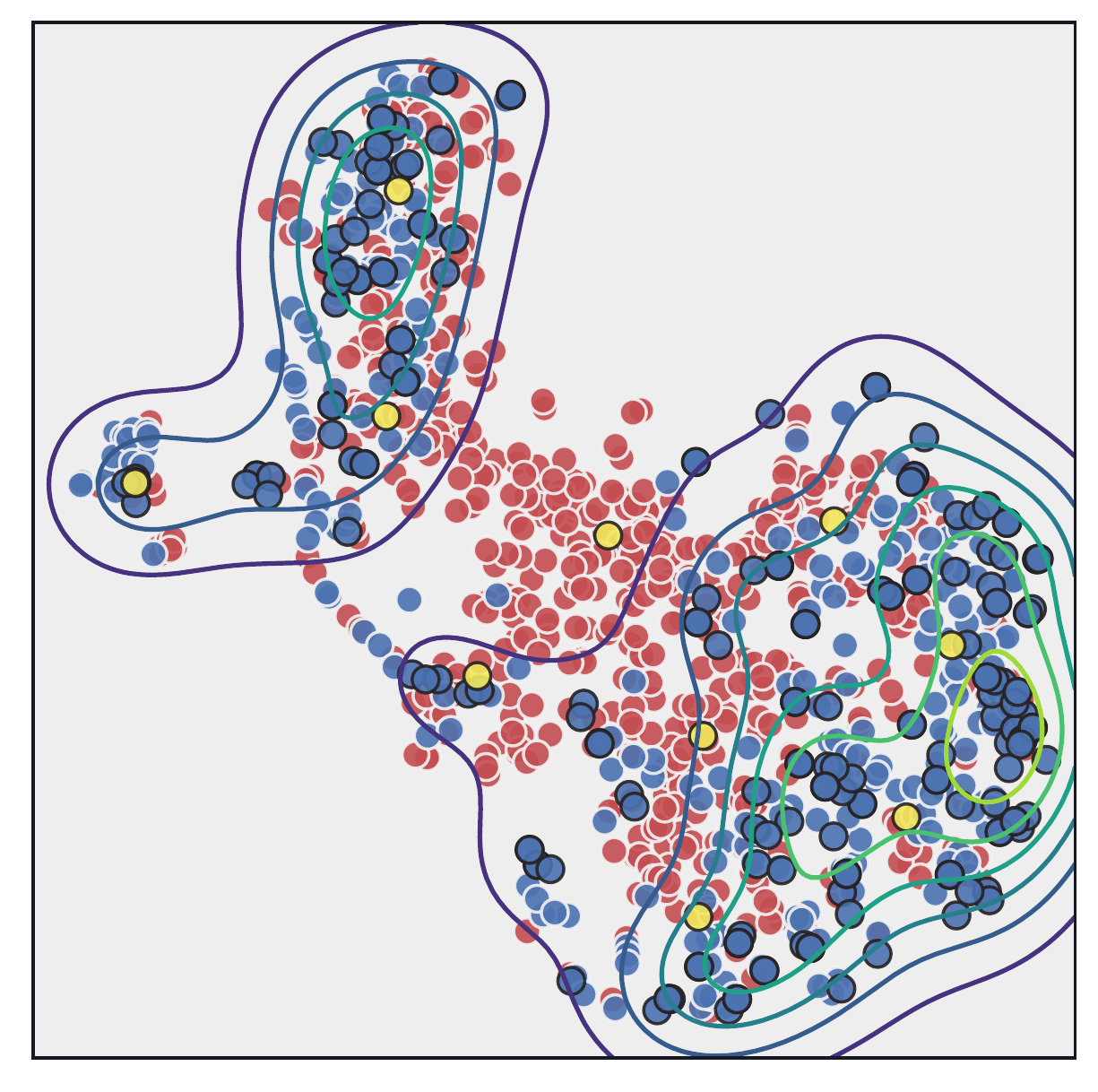}
  \caption{PAO oversampling}
\end{subfigure}

\begin{subfigure}[t]{0.22\textwidth}
  \includegraphics[width=\textwidth]{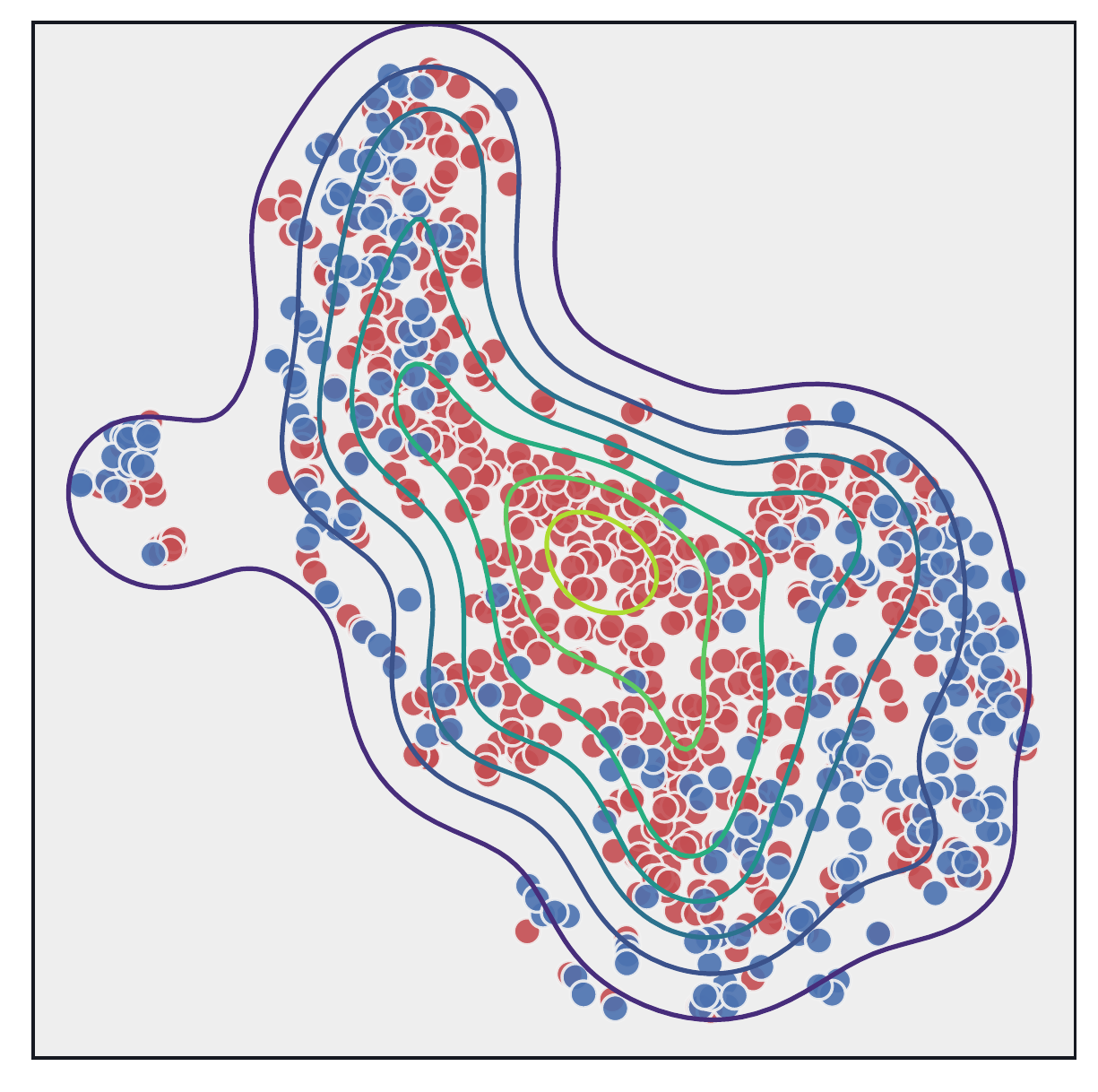}
  \caption{majority class potential}
\end{subfigure}
~
\begin{subfigure}[t]{0.22\textwidth}
  \includegraphics[width=\textwidth]{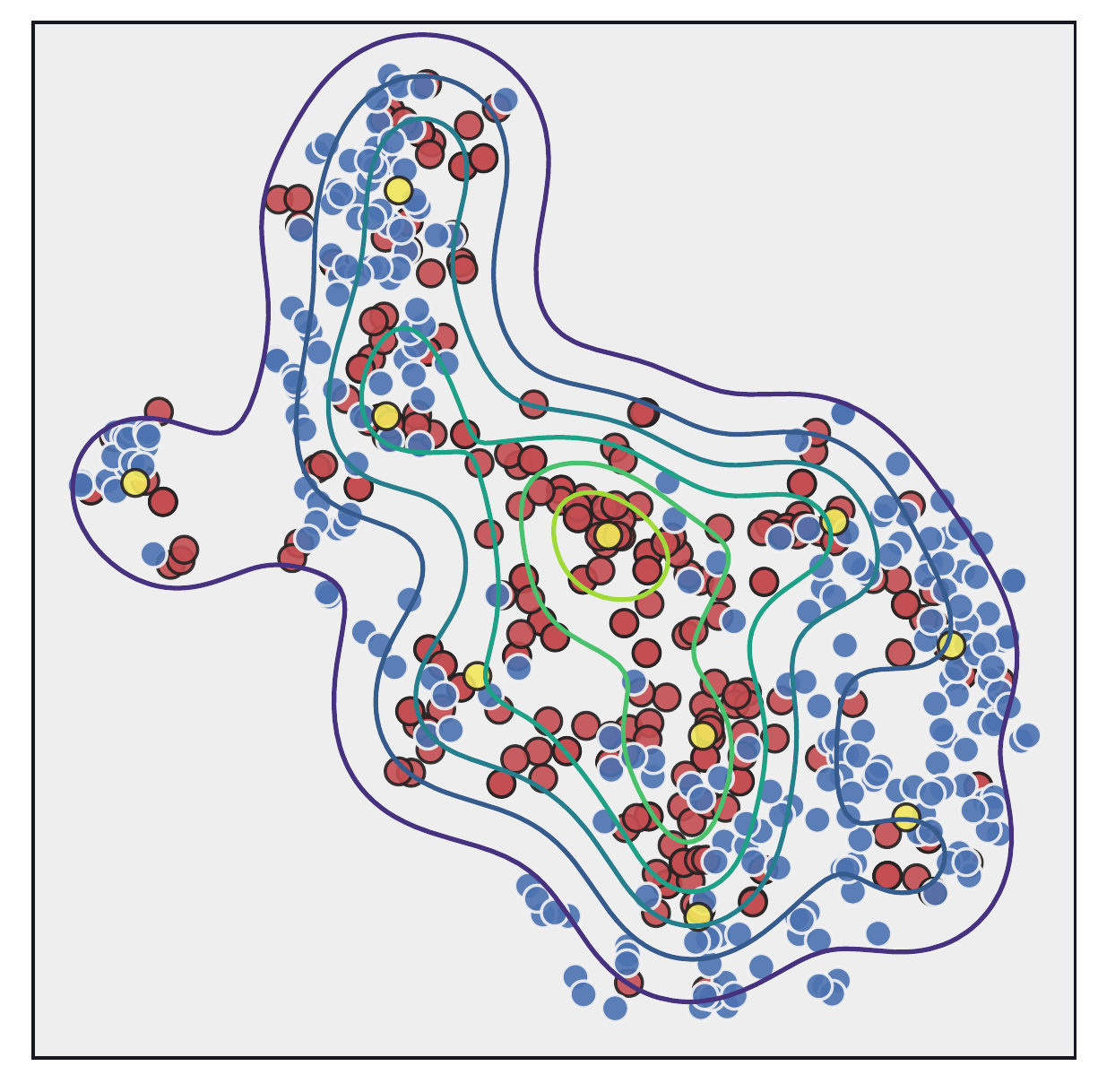}
  \caption{PAU undersampling}
\end{subfigure}
\caption{Example of the original minority and majority class potentials compared with the potential of synthetic observations generated by PAO and PAU. Anchor points denoted with a yellow color.}
\label{fig:example-pao-pau}
\end{figure}

\begin{figure*}[!htb]
\centering
\begin{subfigure}[b]{0.22\textwidth}
  \includegraphics[width=\textwidth]{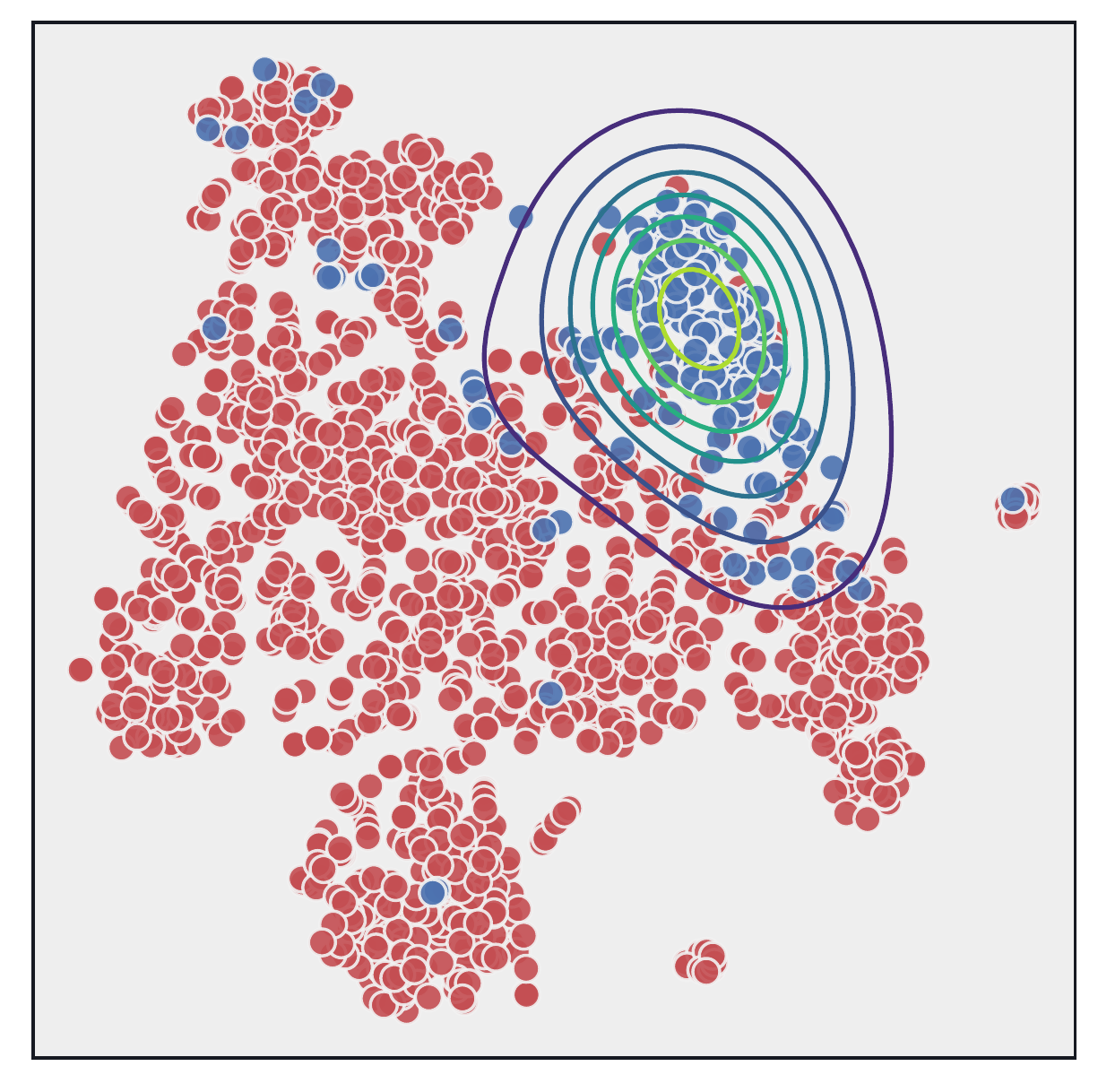}
  \caption{original dataset}
\end{subfigure}
~
\begin{subfigure}[b]{0.22\textwidth}
  \includegraphics[width=\textwidth]{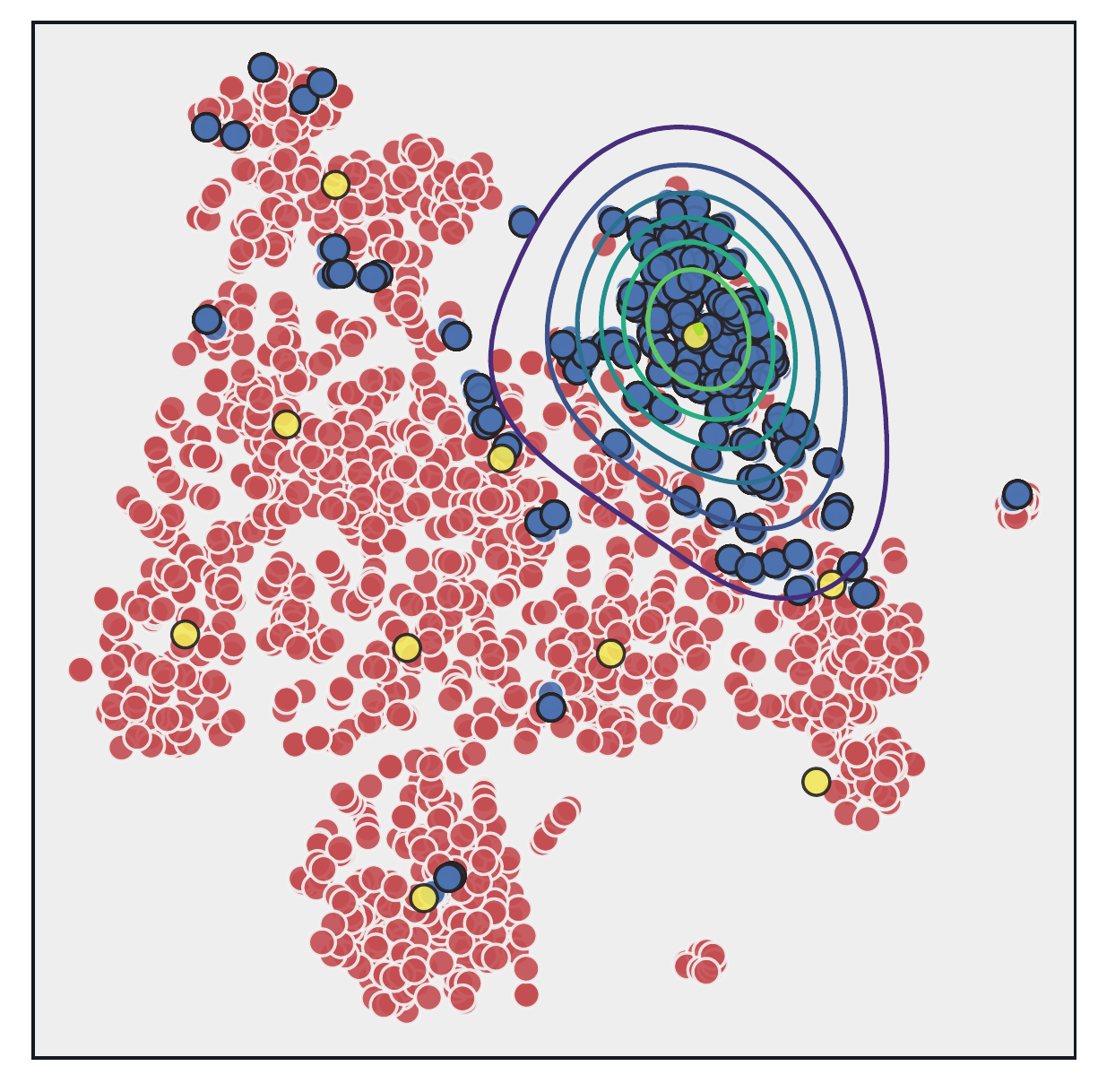}
  \caption{$\lambda = 0.0$}
\end{subfigure}
~
\begin{subfigure}[b]{0.22\textwidth}
  \includegraphics[width=\textwidth]{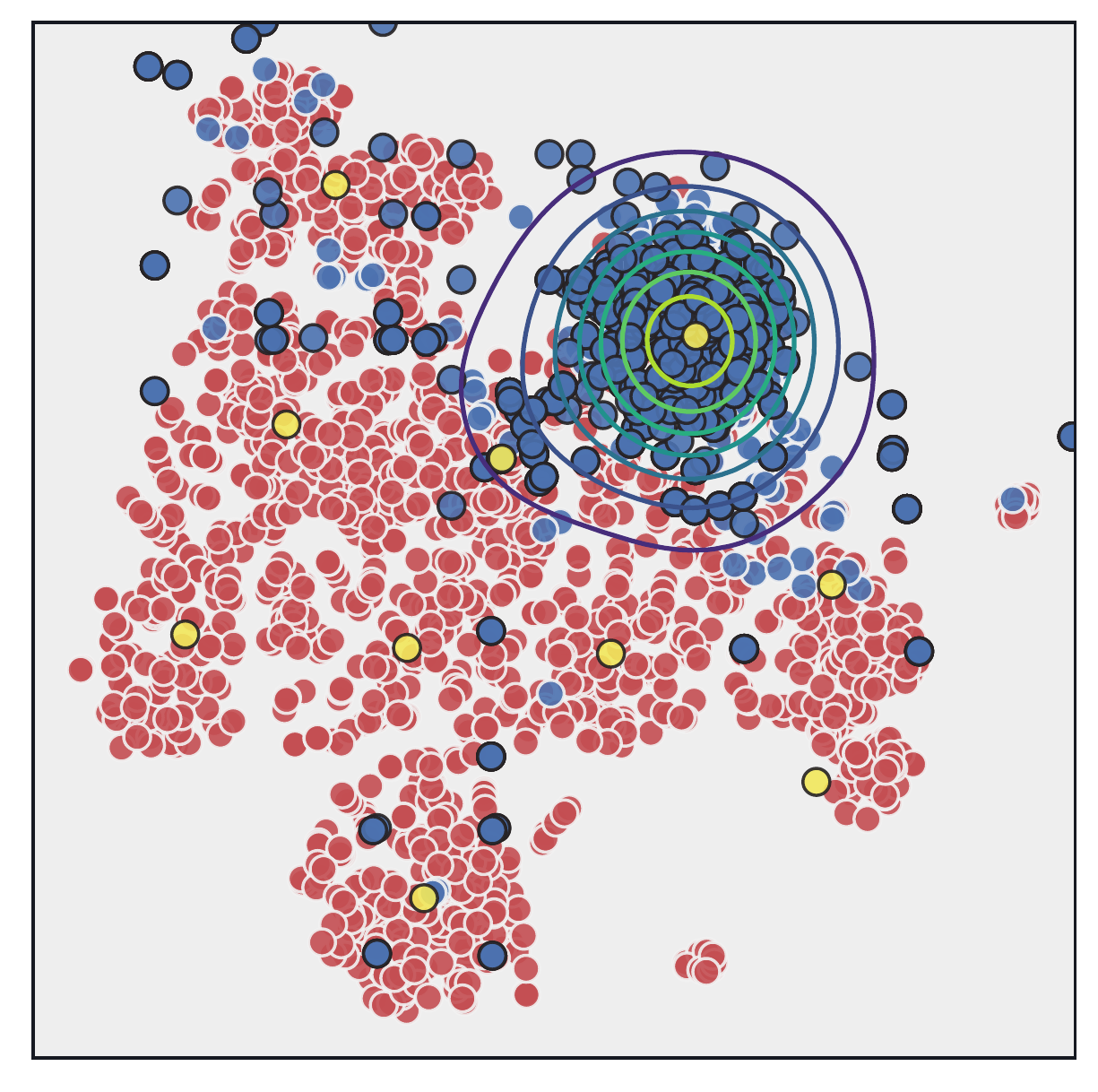}
  \caption{$\lambda = 0.001$}
\end{subfigure}
~
\begin{subfigure}[b]{0.22\textwidth}
  \includegraphics[width=\textwidth]{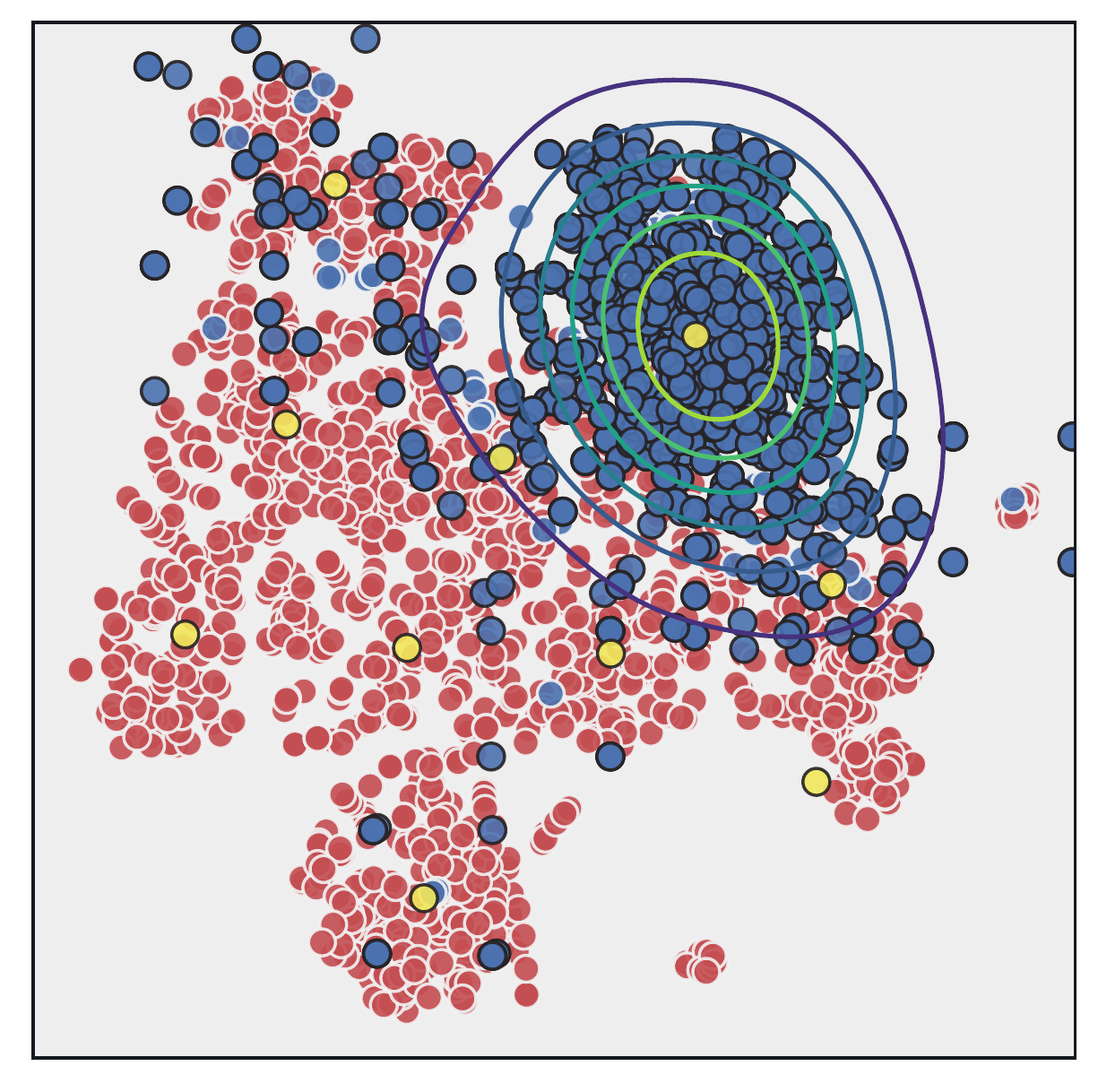}
  \caption{$\lambda = 10.0$}
\end{subfigure}
\caption{Visualization of the impact of $\lambda$ parameter on the behavior of PAO. Higher values of $\lambda$ lead to lower similarity between the original and generated potential shape, and higher spread of synthesized observations.}
\label{fig:example-lambda}
\end{figure*}

A particular case of PA involves eliminating the imbalance solely by either oversampling (PAO) or undersampling (PAU). The concept of class potential is illustrated in both cases in Figure~\ref{fig:example-pao-pau}. As can be seen, PA generates synthetic observations, the potential of which resembles the original despite the fact that it is being anchored in a small number of points. Secondly, the impact of regularization coefficient $\lambda$ on the behavior of PAO is illustrated in Figure~\ref{fig:example-lambda}. As can be seen, higher values of $\lambda$ lead to lower similarity between the original and generated potential shape, and higher spread of synthesized observations. Disabling regularization leads to minimal translation of the prototypes, and behavior closely resembling random oversampling.

\noindent\textbf{Main results.} The results of the experimental study described in \cite{pa} demonstrated that PAO and PAU individually outperform the previously proposed radial-based approaches, that is RBO and RBU, indicating the soundness of the strategy of preserving the original shape of class distributions. Furthermore, the strategy of combining over- and undersampling was also experimentally validated, with the best results achieved by a combination of small amount of oversampling and large amount of undersampling. Finally, the approach achieved a significantly better results than the reference methods, in particular in combination with SVM and MLP classifiers. The analysis based on the proposed data difficulty index indicated that the outperformance is particularly strong when PA is applied on a complex datasets, which are, however, not affected by the presence of noise, which significantly reduced the relative performance of PA.

%% file: Sources/Multiclass.tex
\chapter{Multiclass resampling strategies}
\label{chapter:multiclass}

\begin{center}
  \begin{minipage}{0.5\textwidth}
    \begin{small}
      In which the extension of a selected binary oversampling algorithms to the multiclass setting is presented.
    \end{small}
  \end{minipage}
  \vspace{0.5cm}
\end{center}

Up to this point the focus was put on the resampling algorithms that deal with a binary classification problems. However, in a vast number of real-life problems one is faced with multi-class datasets that are characterized by skewed distributions. While for binary problems relationships between the classes are easy to be defined, in multi-class problems the mutual relationships among classes are much more complex \cite{Wang:2012}. Popular methods dedicated to two-class problems cannot be directly applied to their multi-class counterparts, as they cannot model the relationships among classes and inherited learning difficulties, such as multiple classes overlapping or borderline instances being located among more than two decision regions. One may decompose a given problem into a set of binary sub-problems, yet this is an oversimplification that loses a valuable perspective on non-pairwise data characteristics \cite{Krawczyk:2016,Fernandez:2018}.

To mitigate the aforementioned difficulties a novel class decomposition strategy utilizing information coming from all of the classes was developed. Furthermore, two of the previously introduced binary oversampling algorithms, RBO and CCR, were extended to the multi-class domain by proposing a dedicated resampling strategies. In the remainder of this section both of the proposed algorithms are briefly introduced, and the main experimental results, described in more detail in the corresponding papers, are outlined.

\section{MC-RBO: Multiclass Radial-Based Oversampling}

The Multiclass Radial-Based Oversampling (MC-RBO) \cite{krawczyk2019radial} algorithm, as the name indicates, is an extension of the RBO algorithm to the multi-class setting. Specifically, it is based on an iterative procedure, during which classes are oversampled individually, one at the time. First of all, MC-RBO sorts the classes by the number of associated observations in the descending order. The majority class, that is the class with the highest number of observations, is initially picked. Afterwards, for each of the classes except the majority class, a collection of combined majority observations, consisting of a fraction of observations from each of already considered class, in constructed. Finally, the oversampling with RBO algorithm is performed using the observations from the currently considered class as a minority, and the combined majority observations as the majority class. The pseudocode of the proposed algorithm is presented in Algorithm~\ref{algorithm:mrbo}. Compared with an alternative strategy of adapting RBO to the multi-class setting, one-versus-all (OVA) class decomposition, the proposed method displays two advantages. First of all, it usually reduces the computational cost of the algorithm, since the collection of combined majority observations is often smaller than the collection of all observations. Secondly, it assigns equal weight to every class in the collection of combined majority observations, since each of them has the equal number of observations. This would not be the case in OVA decomposition, in which the classes with a higher number of observations could dominate the rest. Based on the empirical observations, using the proposed class decomposition strategy usually also leads to achieving a better performance during the classification.

\begin{algorithm}[!htb]
	\caption{Multi-Class Radial-Based Oversampling}
	\textbf{Input:} collection of observations $\mathcal{X}$, with $\mathcal{X}^{(c)}$ denoting a subcollection of observations belonging to class $c$ \\
	\textbf{Parameters:} spread of radial basis function $\gamma$, optimization $step$, number of $iterations$ per synthetic observation, $k$ nearest neighbors used for potential approximation \\
    \textbf{Output:} collection $\mathcal{S}$, with $\mathcal{S}_i$ denoting a collection of synthetic minority observations generated for $i$-th class
	\label{algorithm:mrbo}
	\vspace{-0.5\baselineskip}

	\hrulefill
	\begin{algorithmic}[1]
		\STATE \textbf{function} MC-RBO($\mathcal{X}$, $\gamma$, $step$, $iterations$, $k$):
		\STATE $\mathcal{S} \gets \emptyset$
		\STATE $C \gets $ collection of all classes, sorted by the number of associated observations in a descending order
		\FOR{$i \gets 1$ \textbf{to} $|C|$}
		\STATE $n_{classes} \gets $ number of classes with higher number of observations than $C_i$
		\IF{$n_{classes} = 0$}
		\STATE $\mathcal{S}_i \gets \emptyset$
		\ELSE
		\STATE $\mathcal{X}_{min} \gets \mathcal{X}^{(C_i)}$
		\STATE $\mathcal{X}_{maj} \gets \emptyset$
		\FOR{$j \gets 1$ \textbf{to} $n_{classes}$}
		\STATE add $\lfloor\frac{|\mathcal{X}^{(C_1)}|}{n_{classes}}\rfloor$ randomly chosen observations from $\mathcal{X}^{(j)} \cup \mathcal{S}_j$ to $\mathcal{X}_{maj}$
		\ENDFOR
		\STATE $\mathcal{S}_i \gets $ RBO($\mathcal{X}_{maj}$, $\mathcal{X}_{min}$, $\gamma$, $step$, $iterations$, $k$)
		\ENDIF
		\ENDFOR
		\STATE \textbf{return} $\mathcal{S}$
	\end{algorithmic}
\end{algorithm}

\begin{figure}
\centering
    \begin{subfigure}[t]{0.23\textwidth}
        \centering
        \includegraphics[width=1\textwidth]{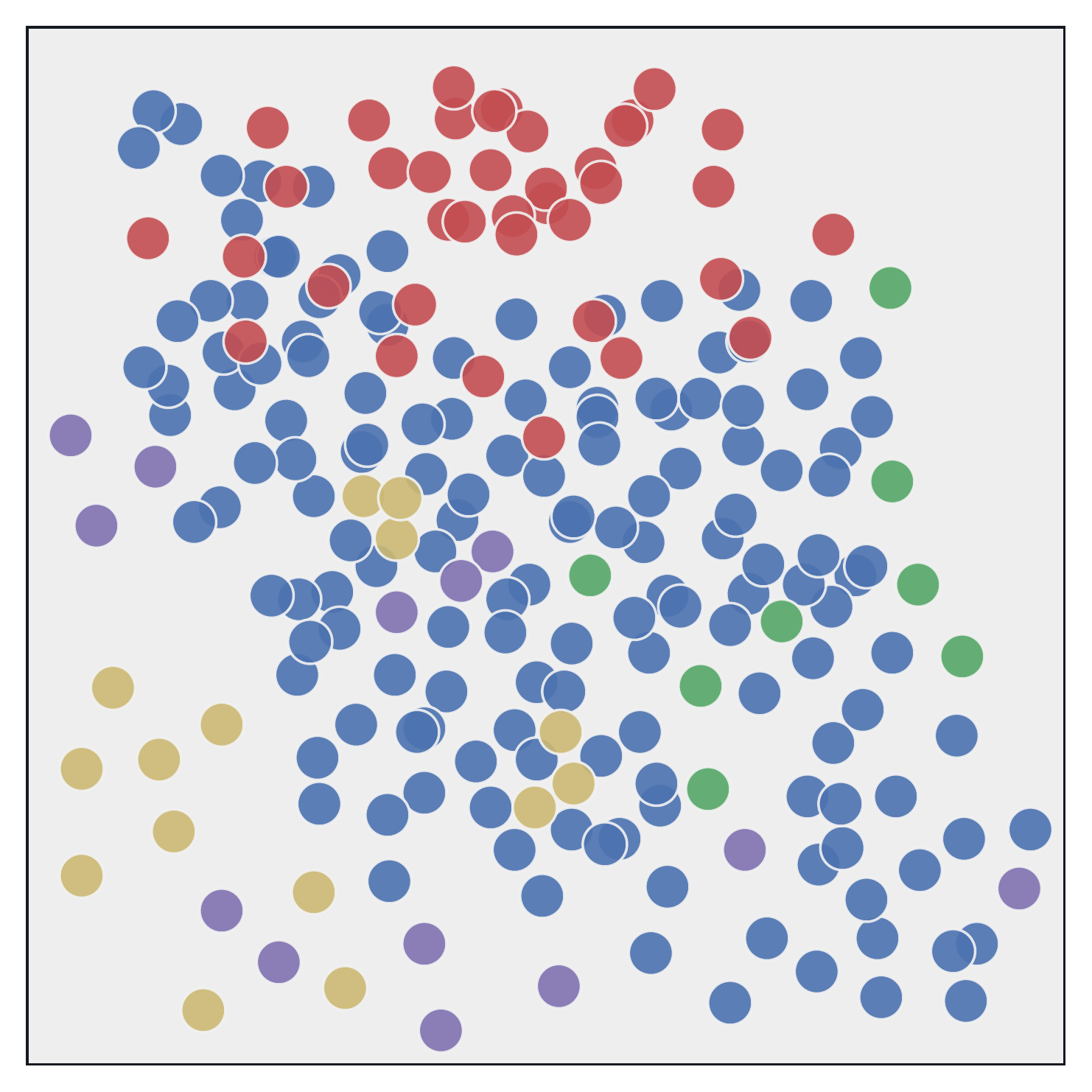}
        \caption{Initial dataset.}
    \end{subfigure}
    ~ 
    \begin{subfigure}[t]{0.23\textwidth}
        \centering
        \includegraphics[width=1\textwidth]{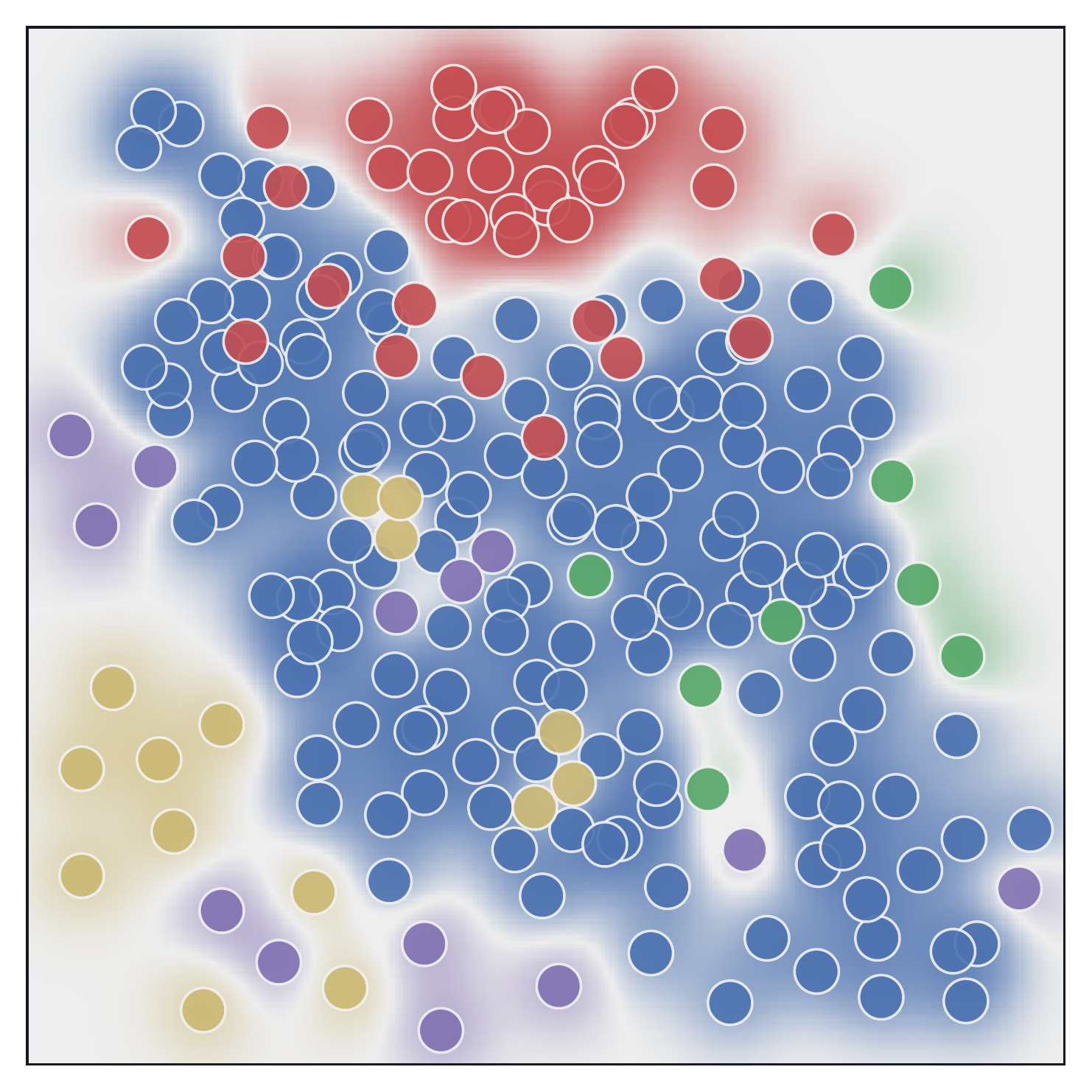}
        \caption{Class potentials.}
    \end{subfigure}
     ~ 
    \begin{subfigure}[t]{0.23\textwidth}
        \centering
        \includegraphics[width=1\textwidth]{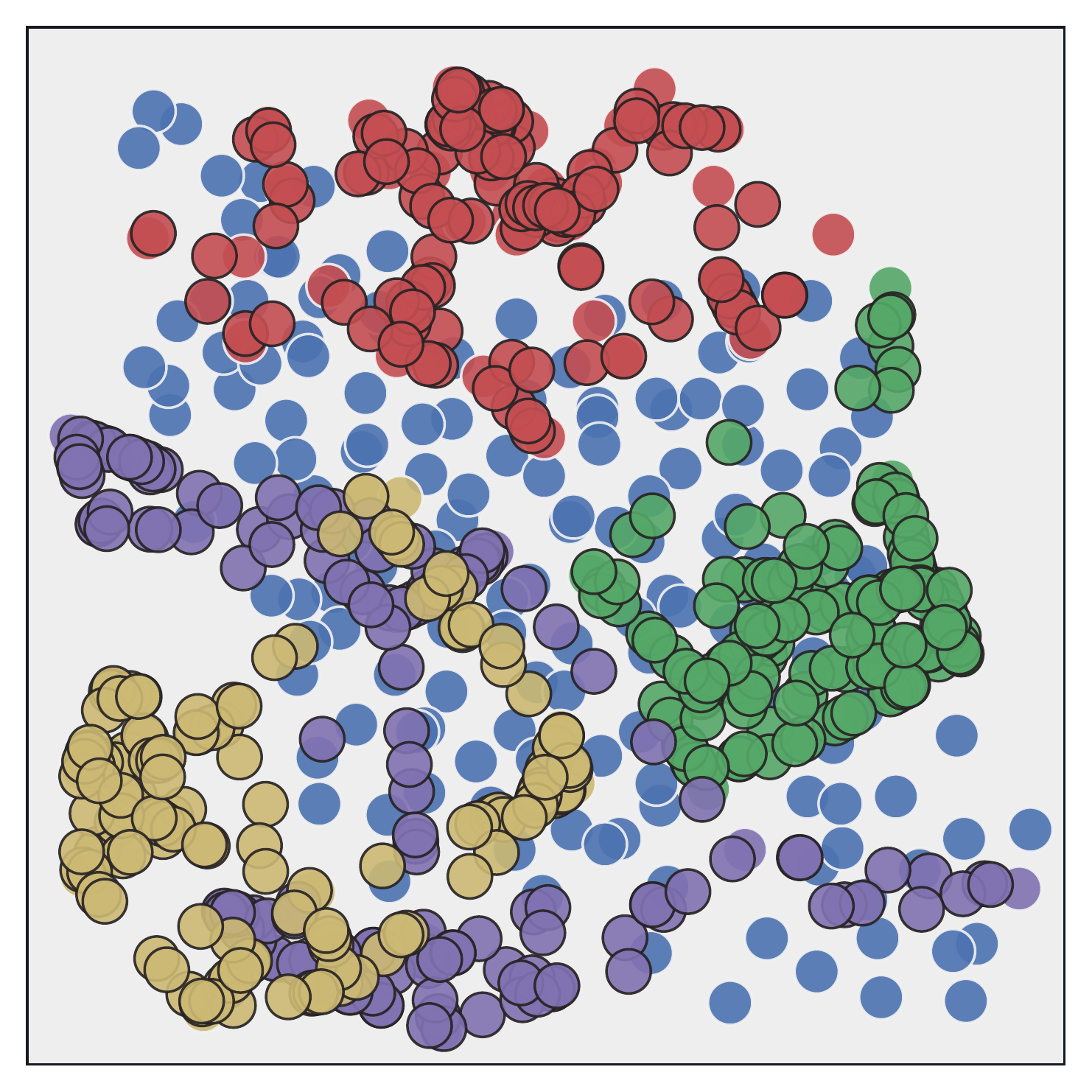}
        \caption{Dataset after balancing with STATIC-SMOTE.}
    \end{subfigure}
     ~ 
    \begin{subfigure}[t]{0.23\textwidth}
        \centering
        \includegraphics[width=1\textwidth]{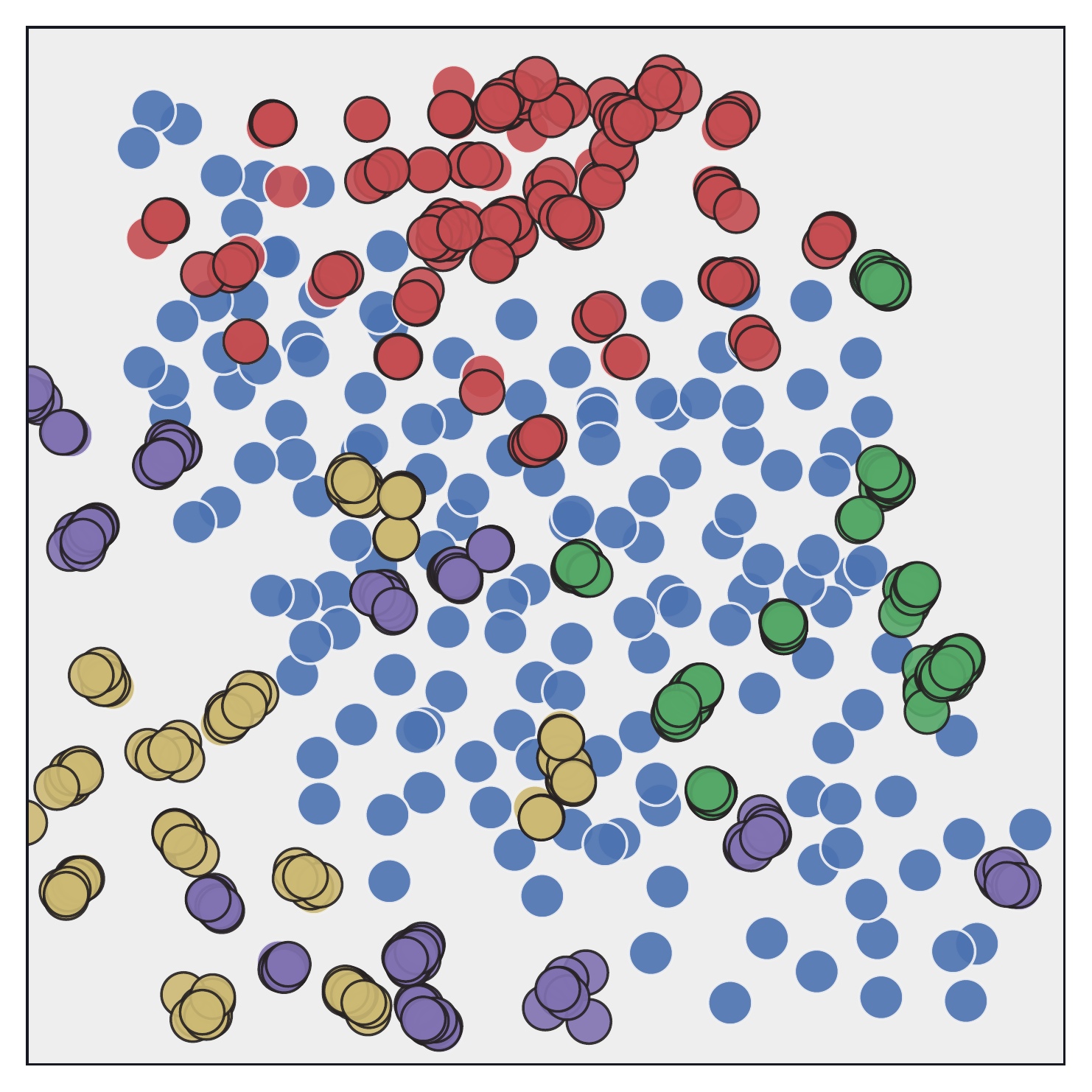}
        \caption{Dataset after balancing with MC-RBO.}
    \end{subfigure}
\caption{Example of an imbalanced dataset, calculated class potentials, and comparison of STATIC-SMOTE and MC-RBO oversampling. MC-RBO produces more conservative synthetic observations, which are much less likely to overlap the distributions of other classes.}
\label{fig:oversampling}
\end{figure}

An illustration of the behavior of MC-RBO compared with SMOTE is presented in Figure~\ref{fig:oversampling}. As can be seen, contrary to SMOTE, MC-RBO is resilient to the presence of outliers in a sense that generated synthetic minority observations do not overlap the clusters of other classes. Instead, in the case of outliers, they are located in a close proximity to the existing observations. On the other hand, when not affected by the distributions of other classes, they are located further away from the original observations. It should be noted that in the presented example, in either case, oversampled synthetic observations are located relatively close to the original observations. This spread can be adjusted by increasing the number of iterations.

\noindent\textbf{Main results.} During the experimental study described in \cite{krawczyk2019radial} it was first demonstrated that the proposed class decomposition strategy, when applied with RBO, outperforms both OVO and OVA class decomposition. Secondly, it was also demonstrated that MC-RBO achieves a statistically significantly better performance than a wide range of state-of-the-art methods for dealing with multiclass data imbalance, which included dedicated multiclass sampling approaches, inbuilt classification mechanisms, and a combination of binarization with various binary sampling techniques. In particular, it was discovered that MC-RBO offers a significantly better performance for small oversampling ratios, which led us to a conclusion that MC-RBO relies not on the number of created instances, but on the intelligent procedure that places them in important parts of the decision space.

\section{MC-CCR: Multiclass Combined Cleaning and Resampling}

\begin{figure*}
\centering
        \includegraphics[width=0.16\textwidth]{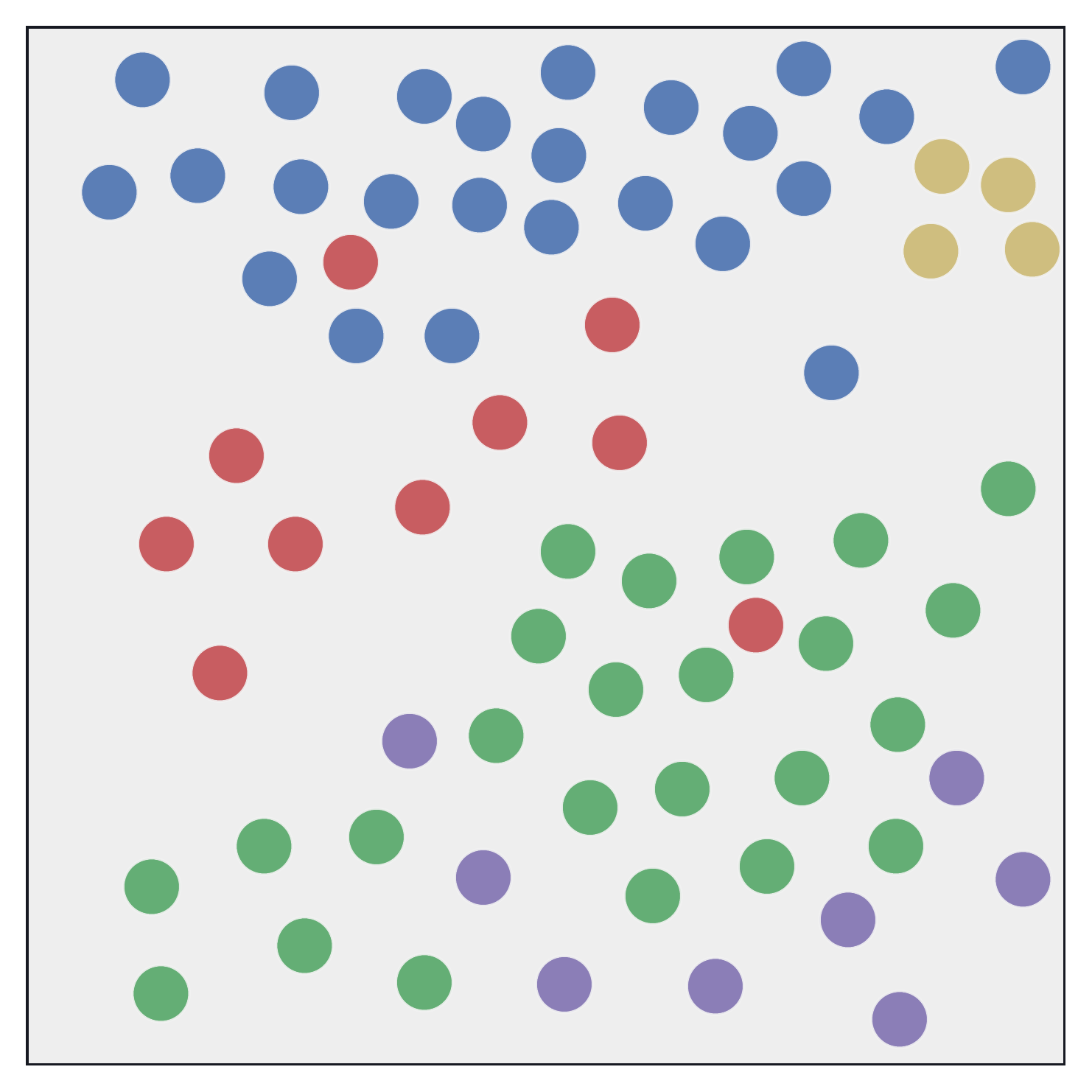}
        \includegraphics[width=0.16\textwidth]{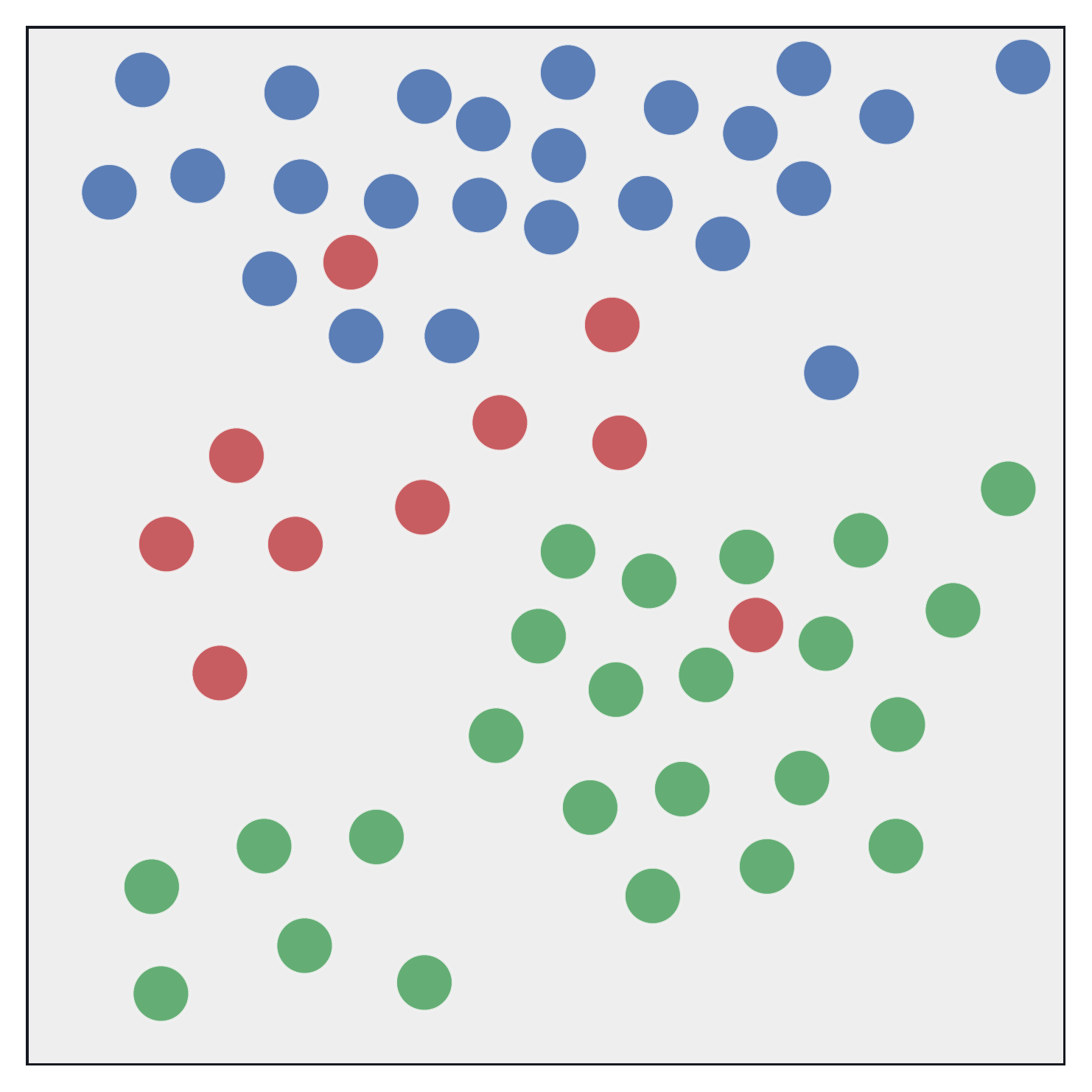}
        \includegraphics[width=0.16\textwidth]{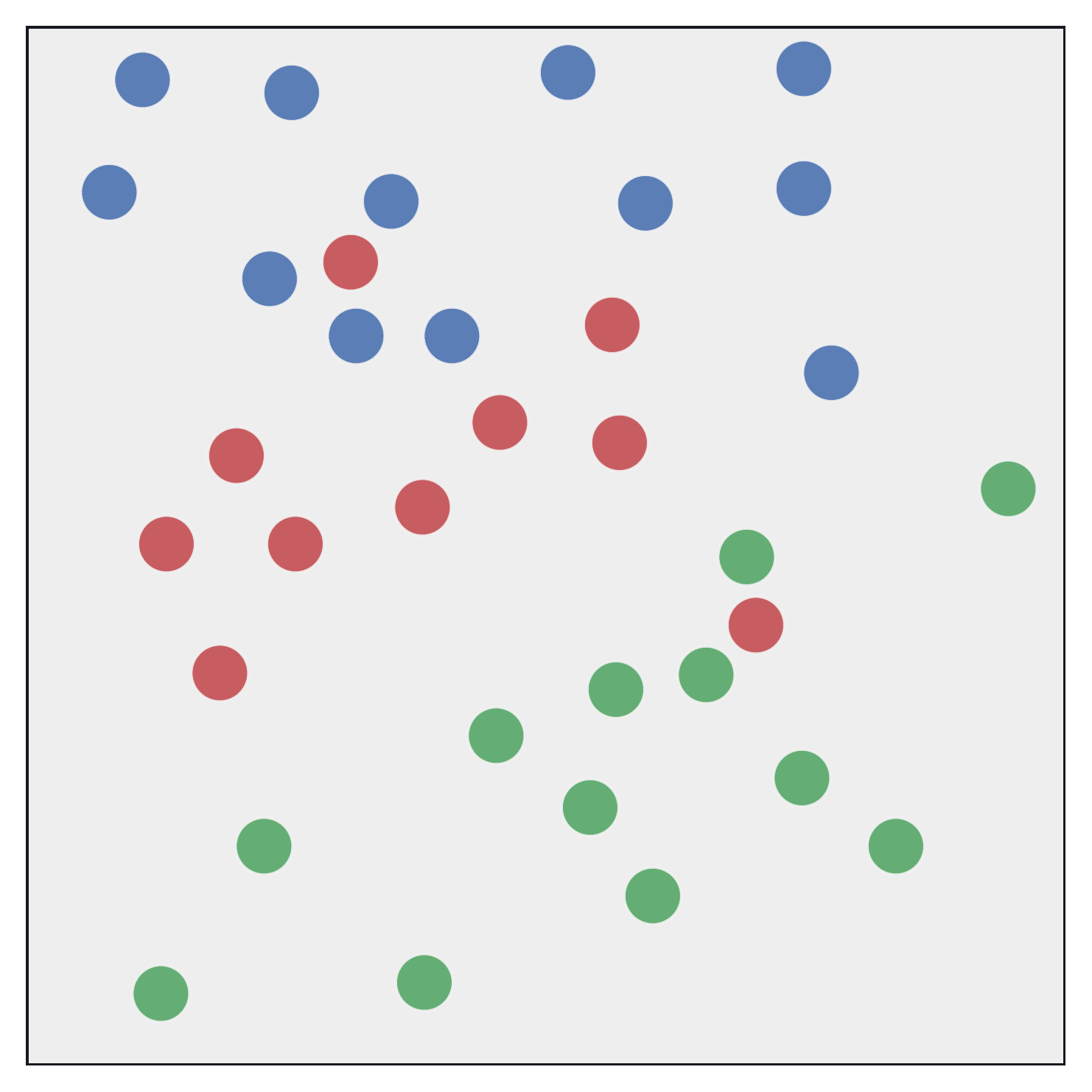}
        \includegraphics[width=0.16\textwidth]{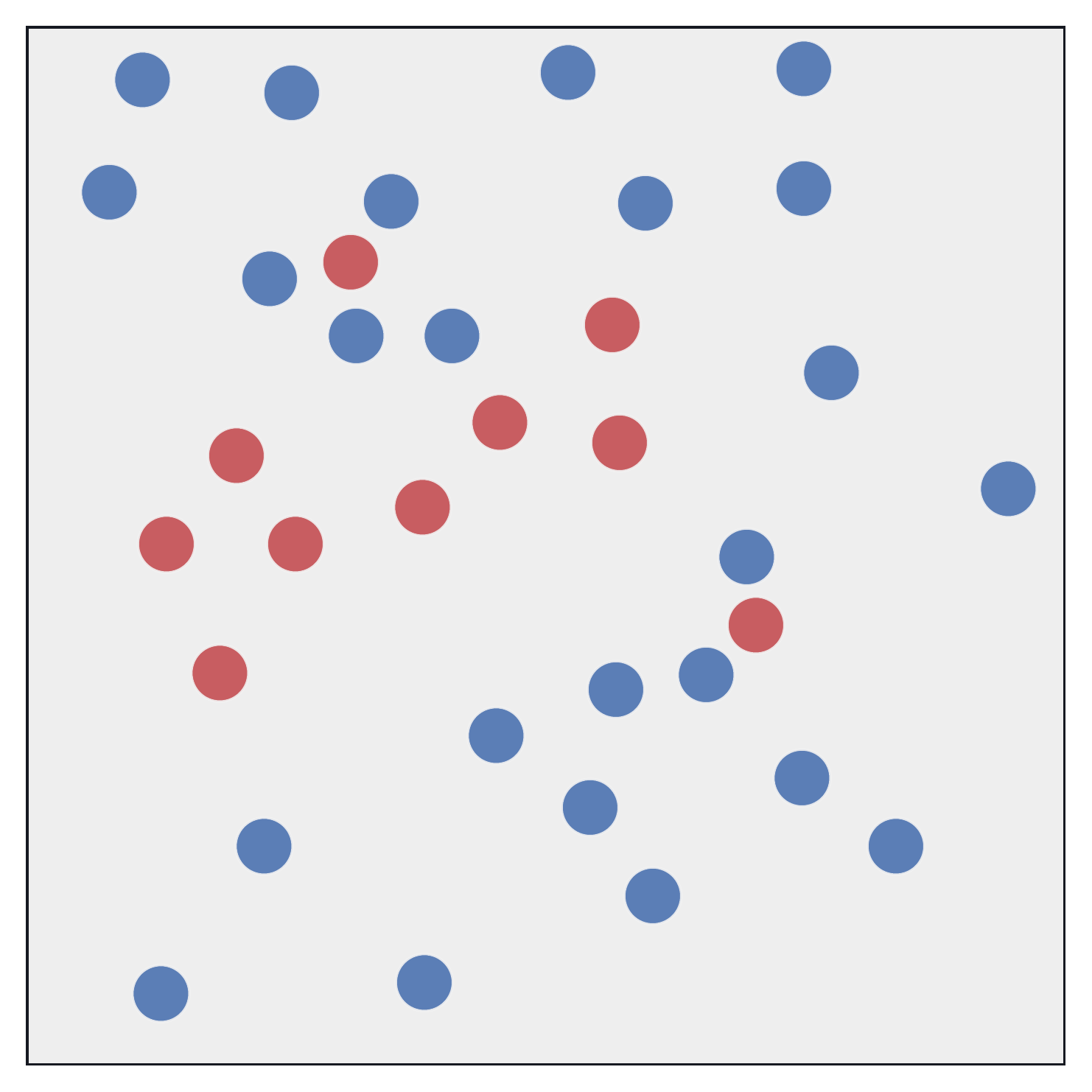}
        \includegraphics[width=0.16\textwidth]{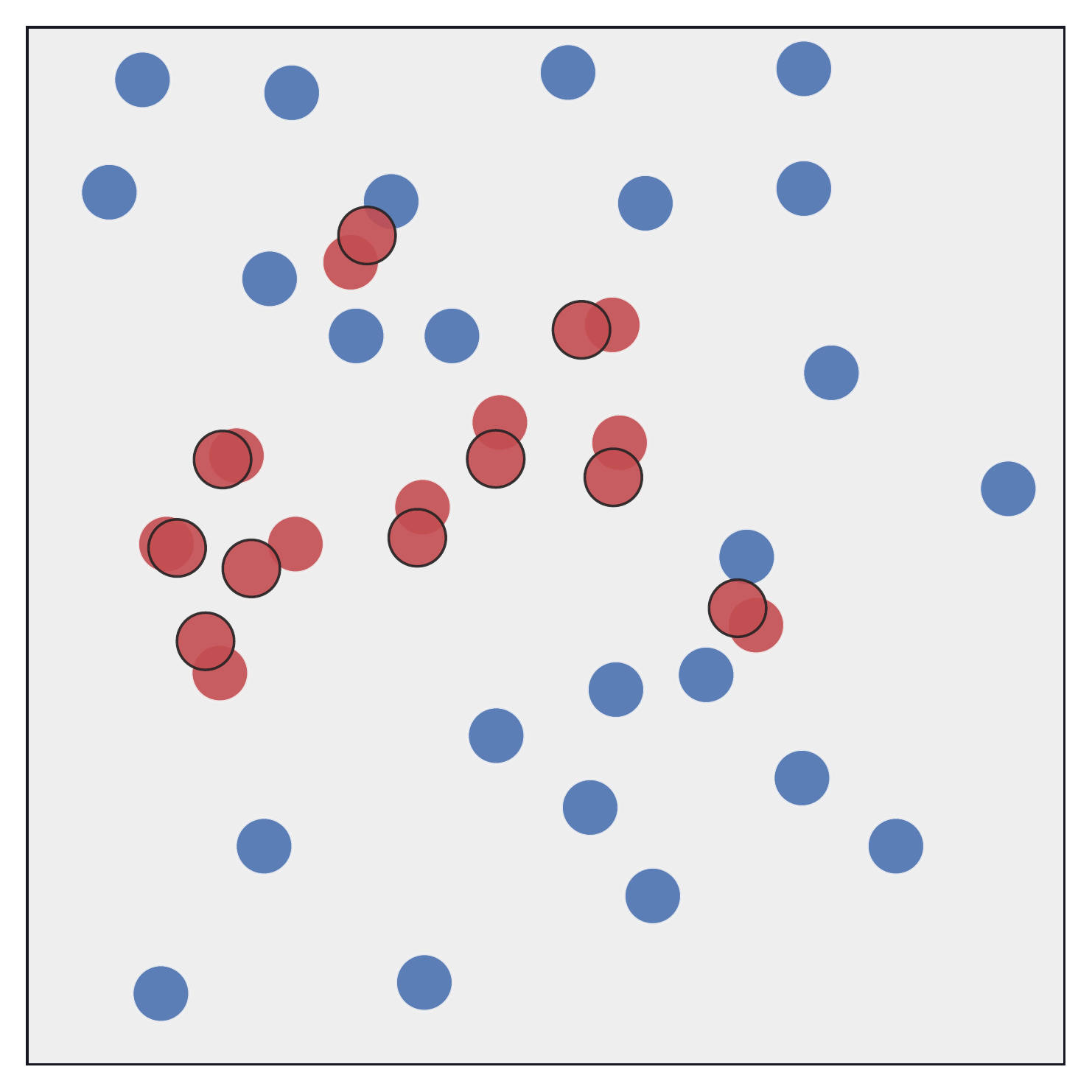}
        \includegraphics[width=0.16\textwidth]{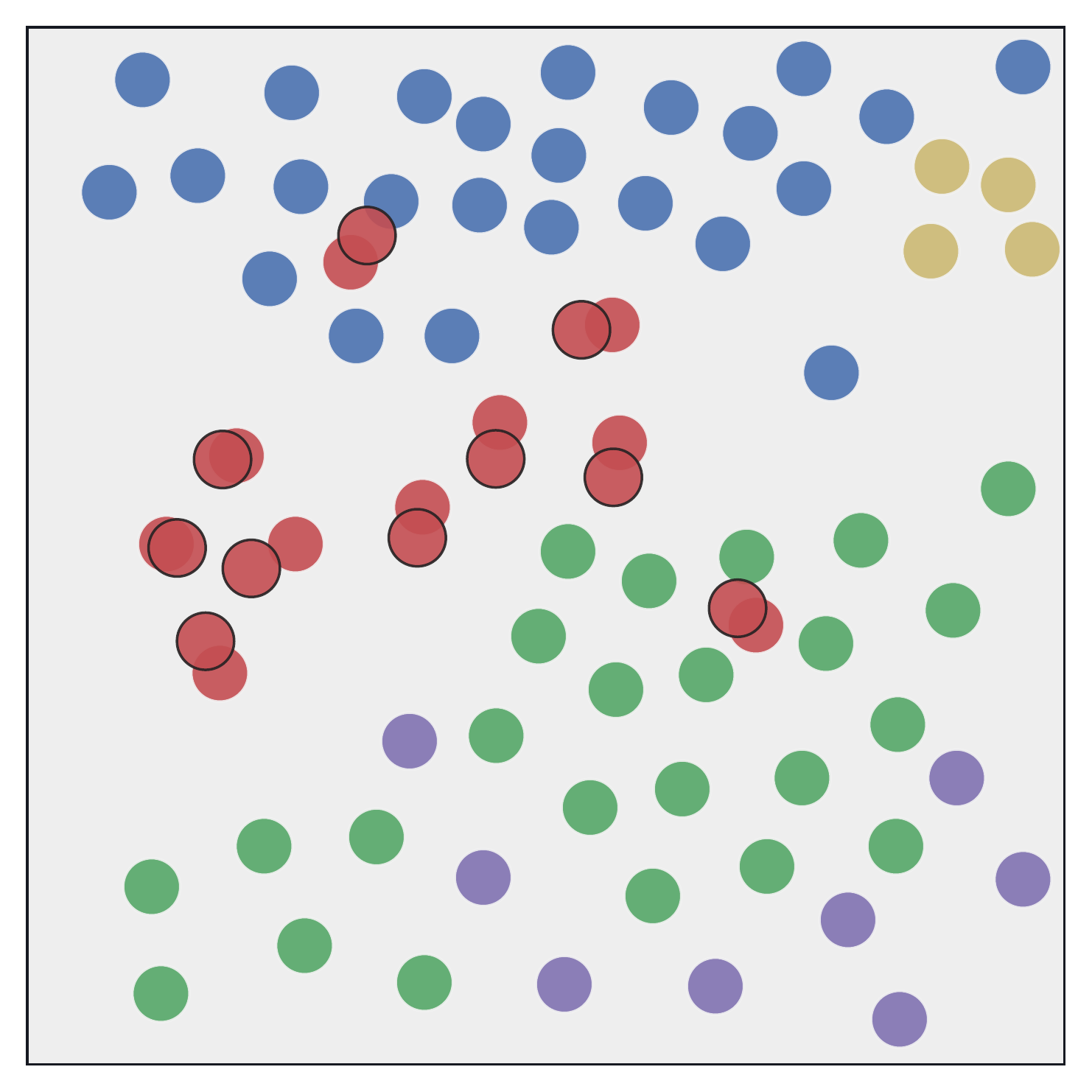}
\caption{An illustration of the multi-class problem decomposition used in MC-CCR. From the left: 1) original dataset, with one of the minority classes (in red) being currently oversampled, 2) classes with smaller number of observations are being temporarily excluded, 3) observations from the remaining majority classes are sampled in an equal proportion, 4) problem is converted to the binary setting by merging all of the majority observations into a single class, 5) cleaning and oversampling with binary CCR is applied, 6) generated synthetic observations are added to the original dataset, and the translations applied during the cleaning stage of binary CCR are preserved.}
\label{fig:mc}
\end{figure*}

The Multiclass Combined Cleaning and Resampling (MC-CCR) \cite{koziarski2020combined} algorithm combines the class decomposition approach introduced by MC-RBO with CCR translation and oversampling procedures. Just like MC-RBO, MC-CCR is an iterative approach, in which individual classes are being resampled one at a time using a subset of observations from already processed classes. MC-CCR utilized the same algorithmic backbone that was used in MC-RBO decomposition: first of all, the classes are sorted in the descending order by the number of associated observations. Secondly, for each of the minority classes, a collection of combined majority observations, consisting of a randomly sampled fraction of observations from each of the already considered classes, is constructed. Finally, a preprocessing with the CCR algorithm is performed using the observations from the currently considered class as a minority, and the combined majority observations as the majority class. Both the generated synthetic minority observations and the applied translations are incorporated into the original data, and the synthetic observations can be used to construct the collection of combined majority observations for later classes. A complete pseudocode of the proposed method is presented in Algorithm~\ref{algorithm:mc-ccr}. Furthermore, the behavior of the algorithm is illustrated in Figure~\ref{fig:mc}.

\begin{algorithm}[!htb]
	\caption{Multi-Class Combined Cleaning and Resampling}
	\textbf{Input:} collection of observations $\mathcal{X}$, with $\mathcal{X}^{(c)}$ denoting a subcollection of observations belonging to class $c$ \\
	\textbf{Parameters:} $energy$ budget for expansion of each sphere, $p$-norm used for distance calculation \\
    \textbf{Output:} collections of translated and oversampled observations $\mathcal{X}$
	\label{algorithm:mc-ccr}
	\vspace{-0.5\baselineskip}
    
	\hrulefill
	\begin{algorithmic}[1]
		\STATE \textbf{function} MC-CCR($\mathcal{X}$, $energy$, $p$):
		\STATE $C \gets $ collection of all classes, sorted by the number of associated observations in a descending order
		\FOR{$i \gets 1$ \textbf{to} $|C|$}
		\STATE $n_{classes} \gets $ number of classes with higher number of observations than $C_i$
		\IF{$n_{classes} > 0$}
		\STATE $\mathcal{X}_{min} \gets \mathcal{X}^{(C_i)}$
		\STATE $\mathcal{X}_{maj} \gets \emptyset$
		\FOR{$j \gets 1$ \textbf{to} $n_{classes}$}
		\STATE add $\lfloor\frac{|\mathcal{X}^{(C_1)}|}{n_{classes}}\rfloor$ randomly chosen observations from $\mathcal{X}^{(j)}$ to $\mathcal{X}_{maj}$
		\ENDFOR
		\STATE $\mathcal{X}_{maj}'$, $\mathcal{S} \gets $ CCR($\mathcal{X}_{maj}$, $\mathcal{X}_{min}$, $energy$, $p$)
		\STATE $\mathcal{X}^{(C_i)} \gets \mathcal{X}^{(C_i)} \cup \mathcal{S}$
		\STATE substitute observations used to construct $\mathcal{X}_{maj}$ with $\mathcal{X}_{maj}'$
		\ENDIF
		\ENDFOR
		\STATE \textbf{return} $\mathcal{X}$
	\end{algorithmic}
\end{algorithm}

Compared with an alternative strategy of adapting the CCR method to the multi-class task, one-versus-all (OVA) class decomposition, the proposed algorithm has two advantages over them. Firstly, it usually decreases the computational cost, since the collection of combined majority observations was often smaller than the set of all instances in our experiments. Secondly, it assigns equal weight to every class in the collection of combined majority observations since each of them has the same number of examples.
 This would not be the case in the OVA decomposition, in which the classes with a higher number of observations could dominate the rest.

It is also important to note that the proposed approach influences the behavior of underlying resampling with CCR. First of all, because only a subset of observations is used to construct the collection of combined majority observations, the cleaning stage applies translations only on that subset of observations: in other words, the impact of the cleaning step is limited. Secondly, it affects the order of the applied translations: it prioritizes the classes with a lower number of observations, for which the translations are more certain to be preserved, whereas the translations applied during the earlier stages while resampling more populated classes can be negated. While the impact of the former on the classification performance is unclear, it can be argued that at least the later is a beneficial behavior, since it further prioritizes least represented classes. Nevertheless, based on the empirical observations, using the proposed class decomposition strategy usually also leads to achieving a better performance during the classification than the ordinary OVA.

\begin{figure*}
\centering
    \begin{subfigure}[t]{0.18\textwidth}
    \captionsetup{font=scriptsize,labelfont=scriptsize}
        \centering        
        \includegraphics[width=\textwidth]{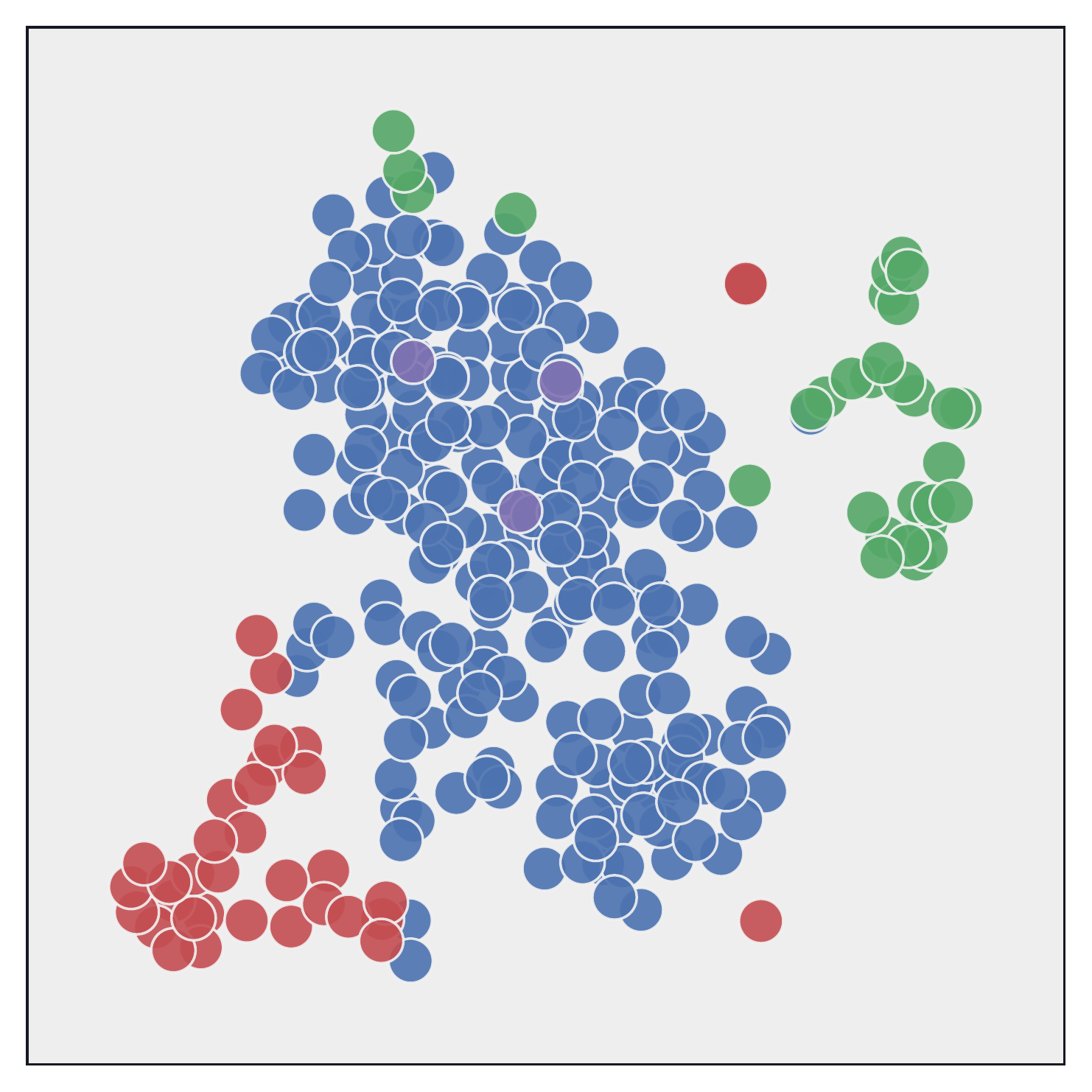}
        \caption{Original dataset}
    \end{subfigure}
    ~ 
    \begin{subfigure}[t]{0.18\textwidth}
    \captionsetup{font=scriptsize,labelfont=scriptsize}
        \centering        
        \includegraphics[width=\textwidth]{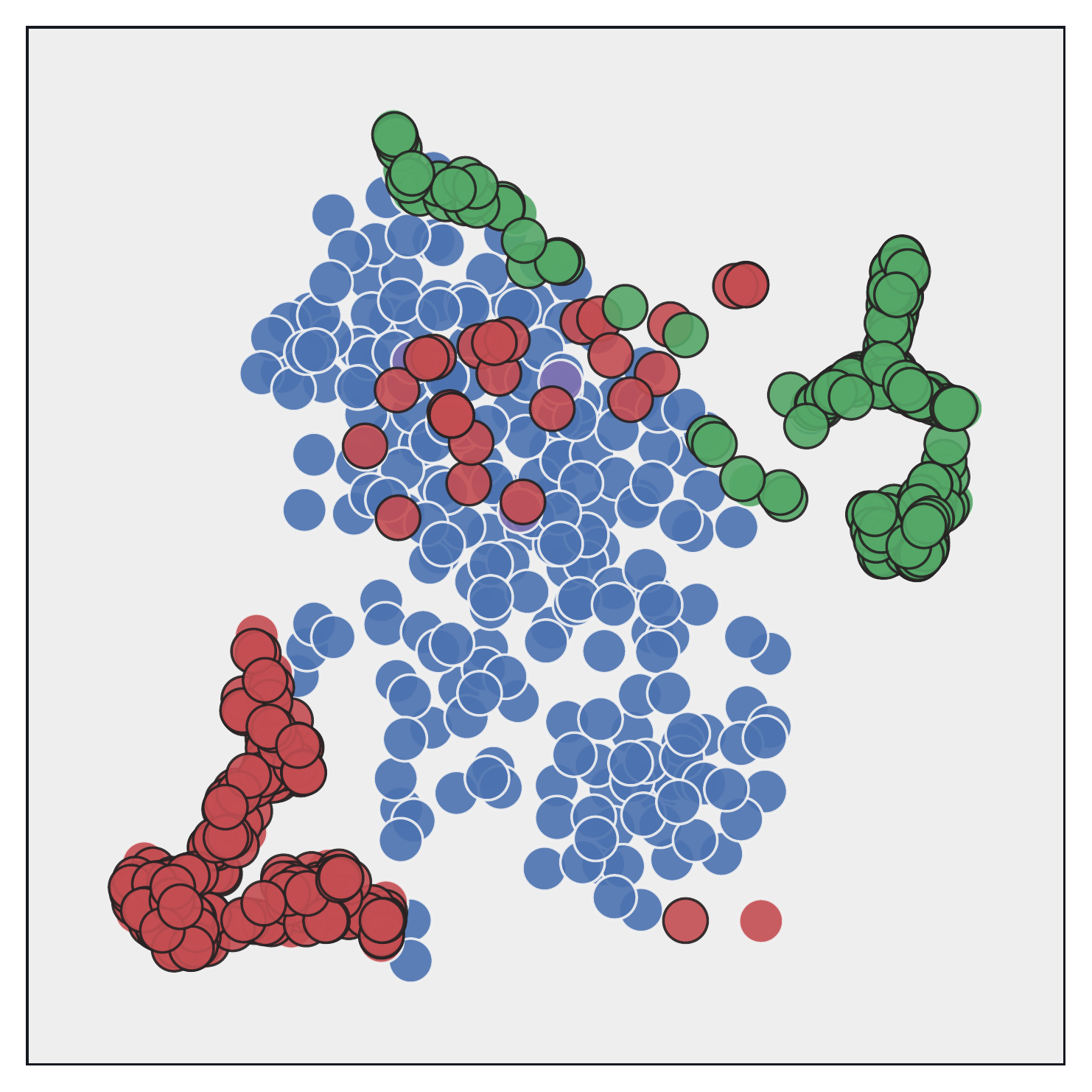}
        \caption{S-SMOTE}
    \end{subfigure}
    ~ 
    \begin{subfigure}[t]{0.18\textwidth}
    \captionsetup{font=scriptsize,labelfont=scriptsize}
        \centering        
        \includegraphics[width=\textwidth]{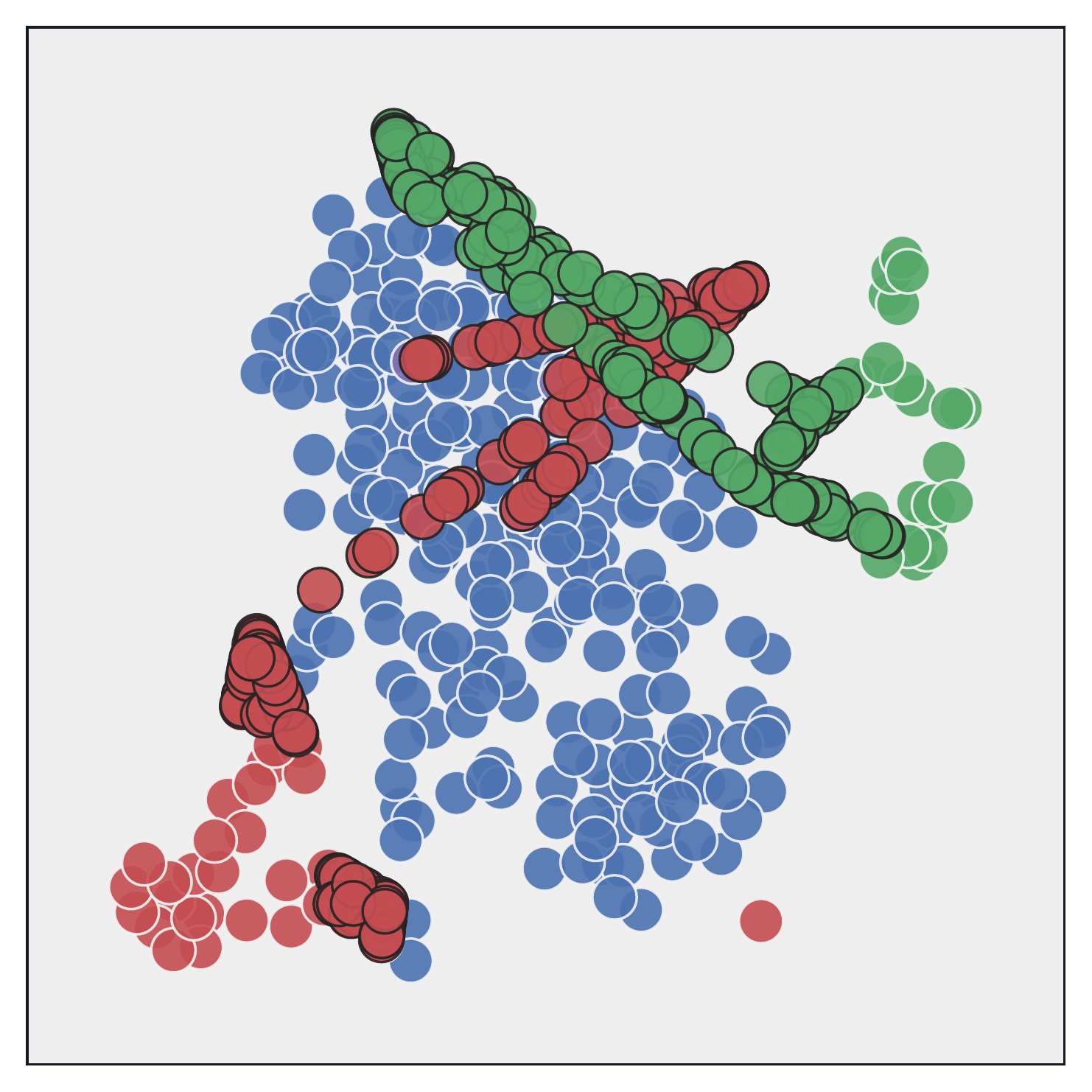}
        \caption{Borderline S-SMOTE}
    \end{subfigure}
    ~ 
    \begin{subfigure}[t]{0.18\textwidth}
    \captionsetup{font=scriptsize,labelfont=scriptsize}
        \centering        
        \includegraphics[width=\textwidth]{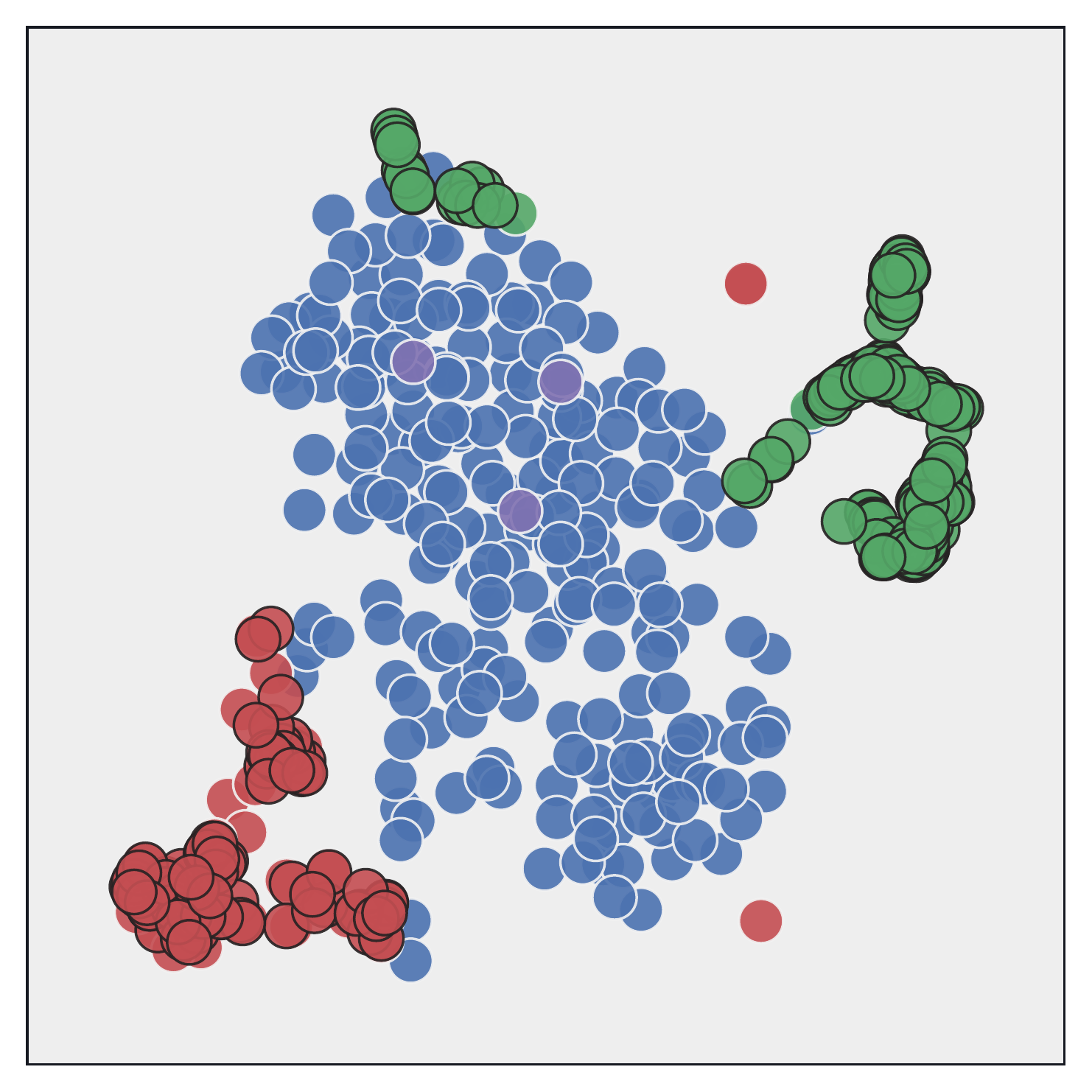}
        \caption{S-SMOTE + ENN}
    \end{subfigure}
    ~ 
    \begin{subfigure}[t]{0.18\textwidth}
    \captionsetup{font=scriptsize,labelfont=scriptsize}
        \centering        
        \includegraphics[width=\textwidth]{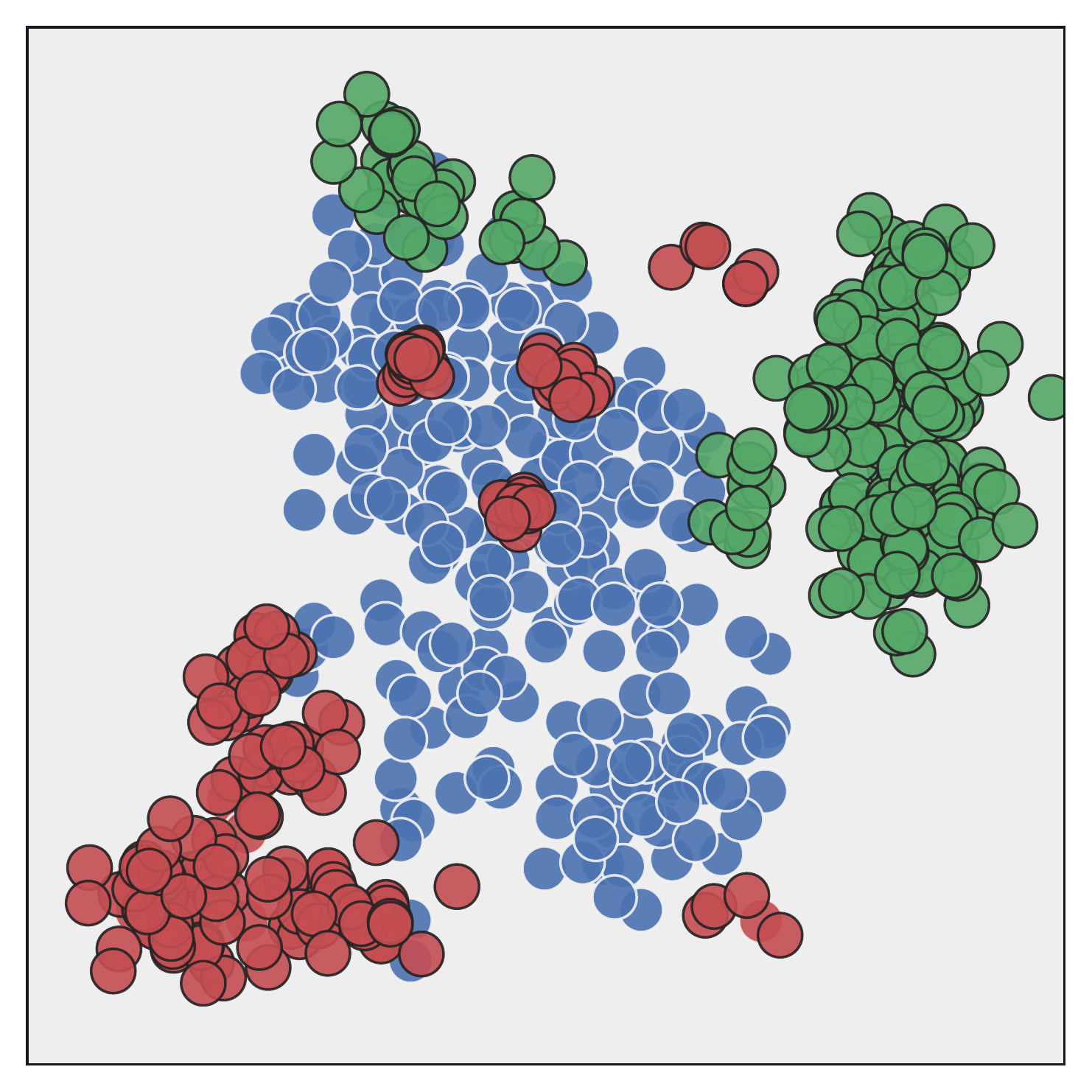}
        \caption{MC-CCR}
    \end{subfigure}
\caption{A comparison of different oversampling algorithms on a multi-class dataset with one majority class (in blue), two minority classes (in green and red), and several majority observations mislabeled as minority (in purple). Both S-SMOTE and Borderline S-SMOTE are susceptible to the presence of label noise and disjoint data distributions. Mechanisms of dealing with outliers, such as postprocessing with ENN, mitigate some of these issues, but at the same time completely exclude underrepresented regions. MC-CCR reduces the negative impact of mislabeled observations by constraining the oversampling regions around them, and at the same time does not ignore outliers not surrounded by majority observations.}
\label{fig:comp}
\end{figure*}

A comparison of the proposed MC-CCR algorithm with several SMOTE-based approaches is presented in Figure~\ref{fig:comp}. An example of a multi-class dataset with two minority classes, disjoint data distributions and label noise was used. As can be seen, S-SMOTE is susceptible to the presence of label noise and disjoint data distributions, producing synthetic minority observations overlapping the majority class distribution. Borderline S-SMOTE, while less sensitive to the presence of individual mislabeled observations, remains even more affected by the disjoint data distributions. Mechanisms of dealing with outliers, such as postprocessing with ENN, mitigate both of these issues, but at the same time exclude entirely underrepresented regions, likely to occur in the case of high data imbalance or small total number of observations. MC-CCR reduces the negative impact of mislabeled observations by constraining the oversampling regions around them, and at the same time, does not ignore outliers not surrounded by majority observations.

\noindent\textbf{Main results.} In the experimental study described in \cite{koziarski2020combined} the usefulness of the proposed resampling approach was empirically evaluated. The experiment began with an ablation study, the goal of which was validation of the design choices behind MC-CCR algorithm. While intuitively motivated, the individual components of MC-CCR are heuristic in nature, and it is not clear whether they actually lead to a better results. The algorithms three main components: the cleaning strategy (the way MC-CCR handles the majority instances located inside the generated spheres), the selection strategy (the approach of assigning greater probability of generating new instances around the minority observations with small associated sphere radius), and the multi-class decomposition strategy, were each considered as a parameter of the algorithm to determine their best configuration. The results of this part of the study served as a confirmation of the design choices, with each of the components enabled leading to the best average performance. In particular, the choice of the cleaning strategy proved to be vital to achieving a satisfactory performance, and conducting no cleaning produced significantly worse results. 

With all of its components enabled, MC-CCR achieved a statistically significantly better results than the considered reference methods, especially in combination with classifiers using the concept of decision tree induction (C5.0) and minimal distance classifiers ($k$-NN). Furthermore, MC-CCR turned out to be very robust to the label noise and it was marked by the smallest decrease in predictive performance depending on the label noise level or the number of classes affected by the noise. Based on the conducted experiments, one can observe that the proposed method works very well for both noisy and no-noise data.

%% file: Sources/Applications.tex
\chapter{Applications in histopathology}
\label{chapter:applications}

\begin{center}
  \begin{minipage}{0.5\textwidth}
    \begin{small}
      In which the main results achieved with the proposed resampling algorithms in the histopathological image classification domain are discussed.
    \end{small}
  \end{minipage}
  \vspace{0.5cm}
\end{center}

One of the domains particularly affected by the negative impact of data imbalance is the medical data classification. The reason for this is twofold. First, the imbalance in the medical data is a widespread issue, in particular when dealing with rare conditions, which is further exacerbated by the usually costly data acquisition and labeling process. Second, the cost of incorrect prediction in general, and false negatives in particular, can be especially high in the medical domain, since it can have a real impact on human health and life.

From the data resampling standpoint, particularly difficult subset of medical problems are the imbalanced medical image recognition tasks. This is due to the inherent characteristics of image data \cite{cyganek2013object}, which is typically high-dimensional, must be provided in high volumes to ensure a satisfactory performance of convolutional neural networks, and due to its spacial properties can be ill-suited for some of the standard mechanisms utilized by the data-level algorithms, such as data interpolation used by SMOTE-based methods.

In this chapter the summary of results achieved in the course of three different studies is presented. The first one \cite{koziarski2018convolutional} was focused on evaluating the impact of data imbalance on the performance of convolutional neural networks, and the possibility of reducing the negative impact of data imbalance by applying methods operating in the image space, including CCR and RBO resampling algorithms. The second one \cite{koziarski2019radial} extended the data resampling approach to conduct the resampling in the high-level feature space extracted from a trained convolutional neural network, with a particular focus put on the RBU algorithm. Finally, in the last study \cite{diagset} a novel large-scale dataset for prostate cancer multi-class classification was proposed. As a part of this study various factors affecting the performance of convolutional neural networks were experimentally evaluated, which included data imbalance. In the course of this study MC-CCR algorithm was used to reduce the negative impact of data imbalance.

\section{Assessing the impact of data imbalance}

\begin{figure}[!htb]
\centering
\begin{subfigure}[t]{0.32\textwidth}
  \includegraphics[width=\textwidth]{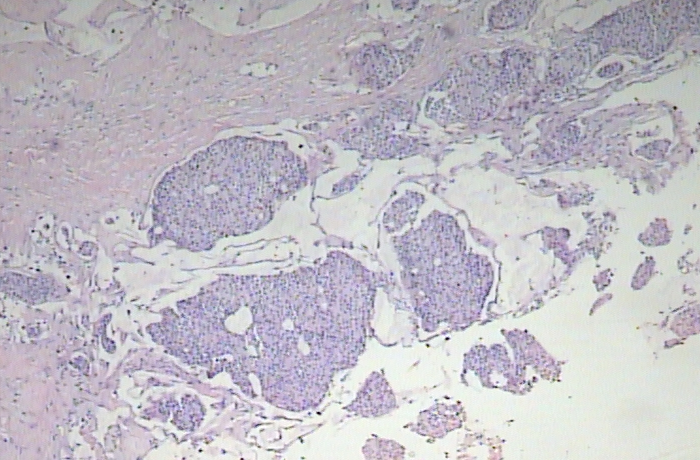}
  \caption{$40\times$}
\end{subfigure}
~
\begin{subfigure}[t]{0.32\textwidth}
  \includegraphics[width=\textwidth]{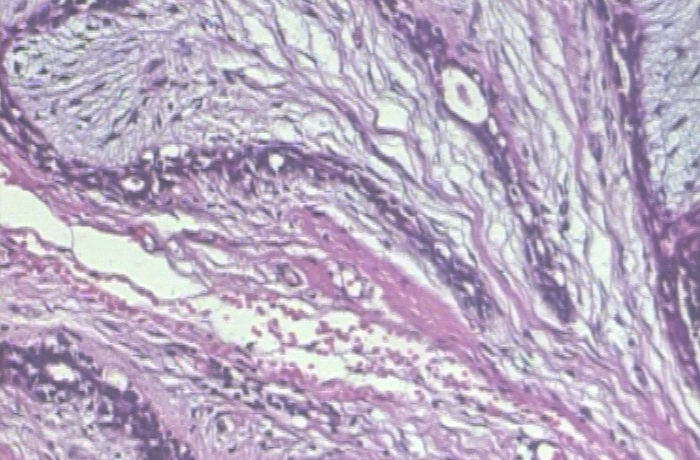}
  \caption{$100\times$}
\end{subfigure}

\begin{subfigure}[t]{0.32\textwidth}
  \includegraphics[width=\textwidth]{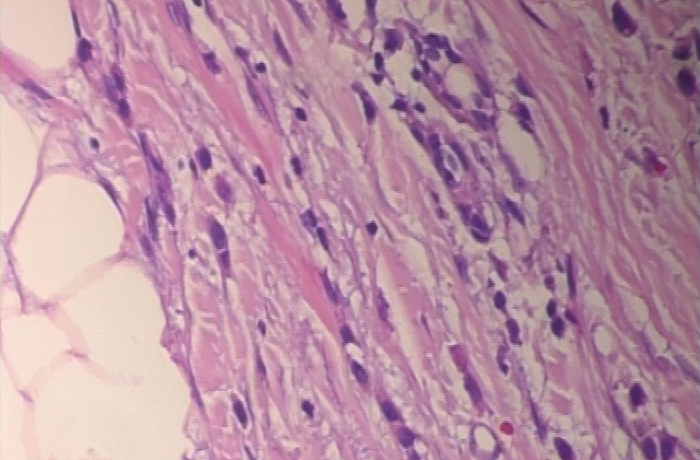}
  \caption{$200\times$}
\end{subfigure}
~
\begin{subfigure}[t]{0.32\textwidth}
  \includegraphics[width=\textwidth]{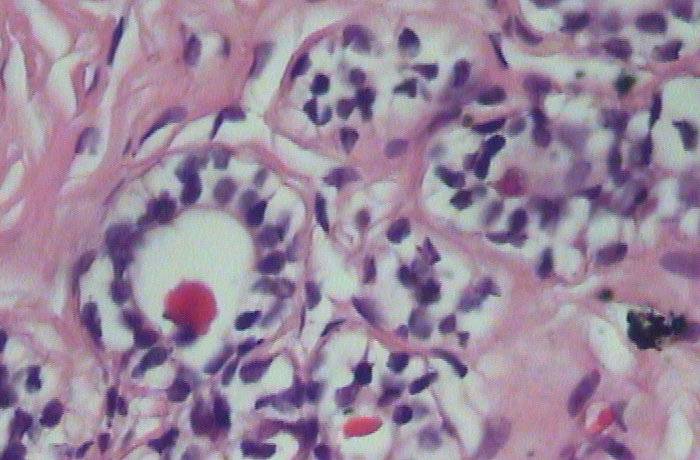}
  \caption{$400\times$}
\end{subfigure}
\caption{Sample images from BreakHis dataset at different magnifications.}
\label{fig:breakhis-images}
\end{figure}

In the first of the conducted studies \cite{koziarski2018convolutional} the impact of data imbalance on the performance of convolutional neural networks trained in the histopathological image classification task was experimentally evaluated. The experiments were based on the Breast Cancer Histopathological Database (BreakHis) \cite{spanhol2016dataset}, which contained 7909 microscopic images of breast tumor tissue, extracted using magnification factors $40\times$, $100\times$, $200\times$ and $400\times$, with approximately 2000 images per magnification factor. Each image had the dimensionality of $700\times460$ pixels and an associated binary label, indicating whether the sample was benign or malignant. An example of BreakHis images is presented in Figure~\ref{fig:breakhis-images}.

By default, BreakHis dataset displays the imbalance of approximately 2.0, with the malignant samples belonging to the majority class. During the conducted experiments an undersampling of the data up to the point of achieving the desired imbalance ratio (IR) was performed. The values of IRs $\in \{1.0, 2.0, ..., 10.0\}$ were considered. Importantly, for each IR the same total number of samples, that is 676 training and 336 test images, was used. It was the maximum amount of data making it possible to produce every considered IR. The same total number of samples for each IR was kept, as opposed to decreasing the number of samples from the minority class and keeping the size of the majority class constant. It allowed us to avoid the issue of decreasing amount of data, which could be another factor affecting the classification performance.

\begin{figure}[!t]
\centering
\subfloat[accuracy]{\includegraphics[width=0.33\linewidth]{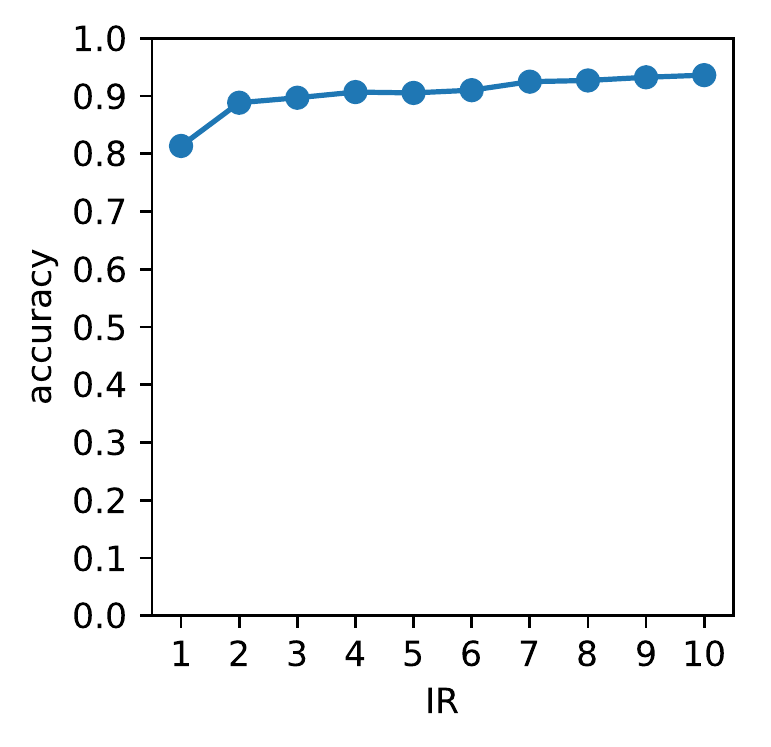}%
\label{fig:accuracy}}
\hfil
\subfloat[precision]{\includegraphics[width=0.33\linewidth]{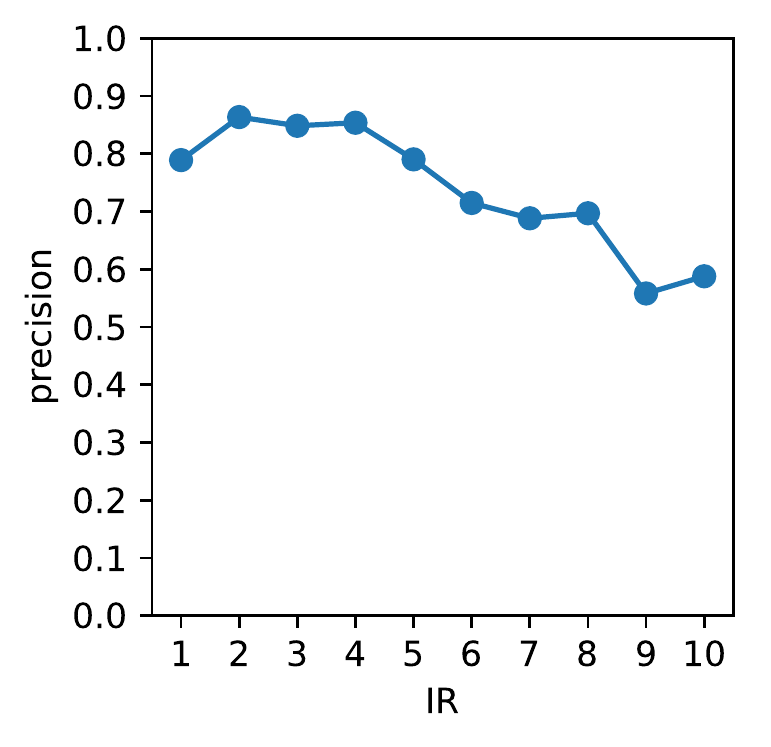}%
\label{fig:precision}}
\hfil
\subfloat[recall]{\includegraphics[width=0.33\linewidth]{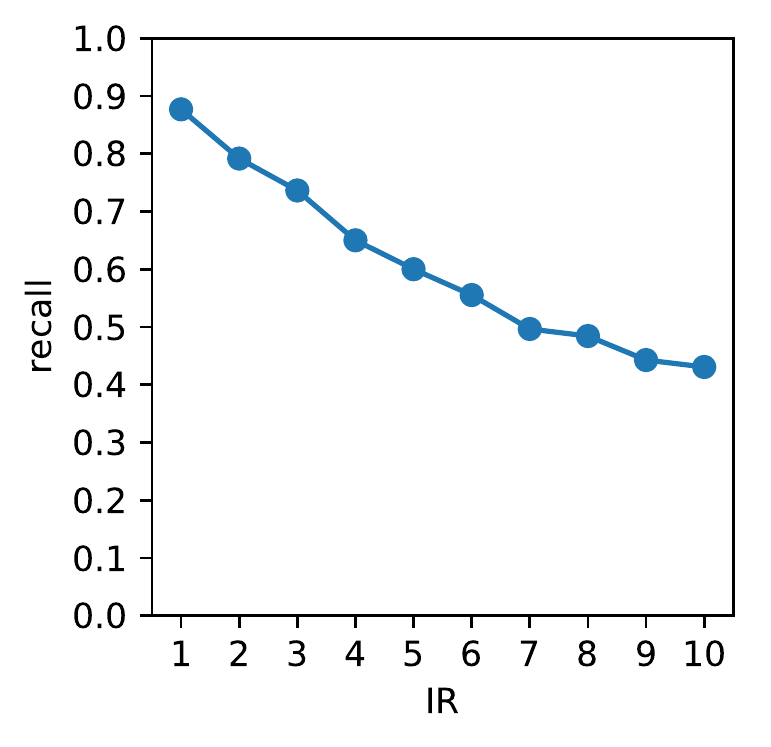}%
\label{fig:recall}}

\subfloat[F-measure]{\includegraphics[width=0.33\linewidth]{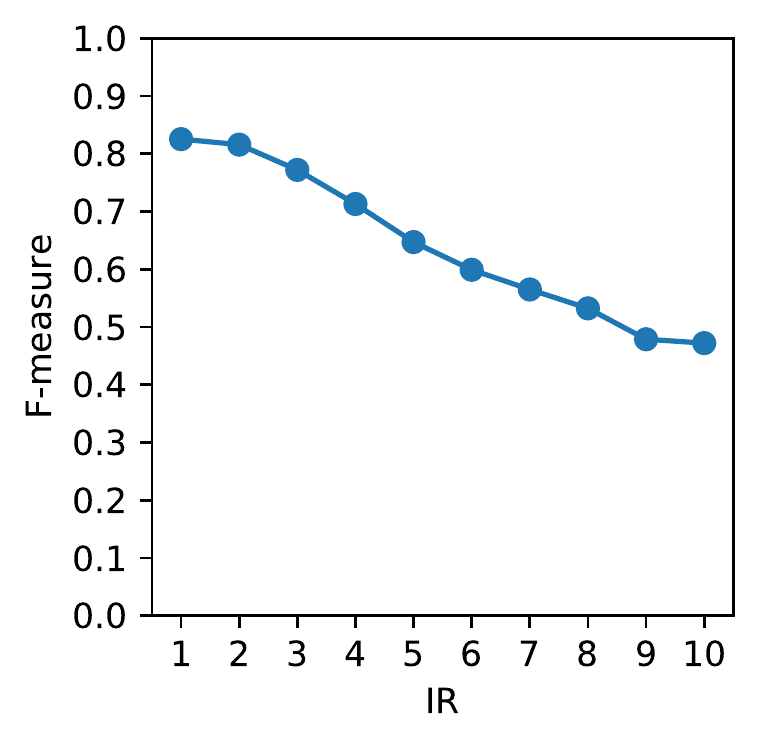}%
\label{fig:fmeasure}}
\hfil
\subfloat[AUC]{\includegraphics[width=0.33\linewidth]{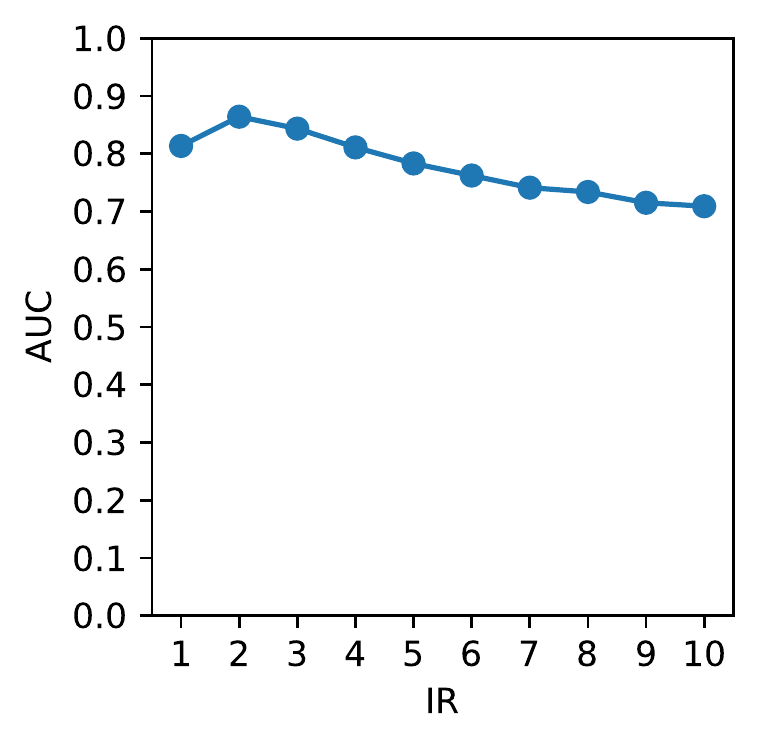}%
\label{fig:auc}}
\hfil
\subfloat[G-mean]{\includegraphics[width=0.33\linewidth]{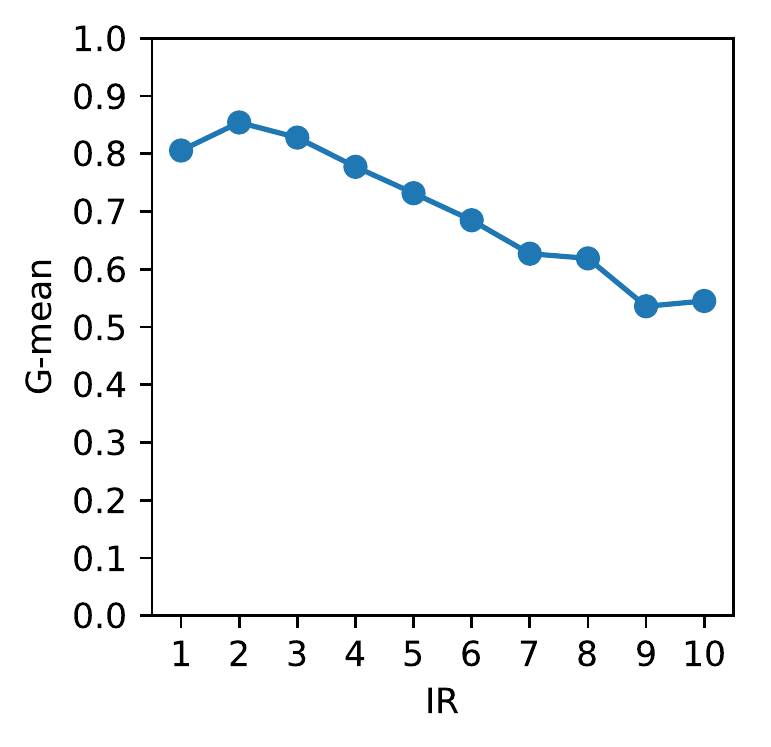}%
\label{fig:gmean}}
\caption{The impact of data imbalance ratio (IR) on the average values of various performance metrics.}
\label{fig:imba}
\end{figure}

\noindent\textbf{The impact of data imbalance on the classification performance.} The goal of the first experiment was evaluating to what extent data imbalance affects the classification performance. To this end the original BreakHis dataset was undersampled up to the point of achieving the desired imbalance ratio (IR), at the same time keeping the total number of observations from both classes constant. The values of IR $\in \{1.0, 2.0, ..., 10.0\}$ were used. Results of this part of the experimental study, averaged over all folds and magnification factors, are presented in Figure~\ref{fig:imba}. As can be seen, the accuracy is not an appropriate performance metric in the imbalanced data setting: it increases steadily with IR, despite the accompanying decrease in both precision and recall. On the other hand, all of the remaining measures indicate a significant drop in performance, especially for higher values of IR. For instance, for the balanced distributions an average value of F-measure above 0.8 was observed, whereas for the IR = 10.0 it drops below 0.5, despite the total number of observations being the same. This indicates that data imbalance has a significant impact on the classifiers behavior and a noticeable decrease in performance can be expected for higher IR. It should be noted that for low values of IR, that is 2 and 3, a better precision, AUC and G-mean were actually observed than for the balanced data distribution. This behavior may suggest that depending on the optimization criterion, slight data imbalance can actually be beneficial for the performance of the model. In the case of the histopathological data, especially if the majority class consists of the images of malignant tissue.

\noindent\textbf{Evaluation of different strategies for dealing with data imbalance.} The goal of the second experiment was comparing various strategies of dealing with data imbalance and assessing which, and under what conditions, lead to the best performance. In this experiment the values of IR $\in \{2.0, 3.0, ..., 10.0\}$ were considered, and the imbalance was grouped into four categories: low ($2.0 - 4.0$), medium ($5.0 - 7.0$), high ($8.0 - 10.0$) and any ($2.0 - 10.0$). For each category the results were averaged over the corresponding values of IR. In total, 8 different strategies of dealing with data imbalance were evaluated: weighted loss (W. Loss) \cite{dong2018imbalanced}, batch balancing (B. Balance), random oversampling (ROS), SMOTE \cite{chawla2002smote}, CCR \cite{koziarski2017ccr}, RBO \cite{koziarski2019neuro}, random undersampling (RUS), and the Neighborhood Cleaning Rule (NCL) \cite{laurikkala2001improving}, with all of the resampling algorithms applied in the image space. To assess the statistical significance of the results a Friedman ranking test with a Shaffer post-hoc analysis were performed at the significance level $\alpha = 0.05$. The results are presented in Table~\ref{table:rankings}. As can be seen, there was no single method that achieved best performance on all levels of imbalance and for all of the performance measures. In general, CCR, RBO, RUS and NCL methods achieved the highest rank in at least one of the settings. For low imbalance levels NCL achieved the best performance for all three combined metrics: F-measure, AUC and G-mean. However, in none of the cases did it achieve a statistically significantly better results than the baseline. For higher levels of imbalance RBO achieved the best rank in most cases, with statistically significant differences. While most of the approaches led to an improvement in performance compared to the baseline at least in some settings, two methods, weighted loss and SMOTE, achieved a noticeably worse performance than the other strategies.

\begin{sidewaystable*}
\footnotesize
\caption{Average ranks achieved by various techniques of dealing with data imbalance for the specified imbalance ratio (IR). Best performance observed for a given ratio was denoted in bold. The number of times a method achieved statistically significantly better or worse performance than one of the other methods was denoted in subscript with, respectively, a plus or a minus sign.}
\label{table:rankings}
\centering
\begin{tabularx}{\textwidth}{lcY|YY|YYYY|YY}
\toprule
& & & \multicolumn{2}{c|}{\textbf{Inbuilt mechanisms}} & \multicolumn{4}{c|}{\textbf{Oversampling strategies}} & \multicolumn{2}{c}{\textbf{Undersampling strategies}} \\
\midrule
& IR & Baseline & W. Loss & B. Balance & ROS & SMOTE & CCR & RBO & RUS & NCL \\
\midrule
\parbox[t]{2mm}{\multirow{4}{*}{\rotatebox[origin=c]{90}{Precision}}} & 2-4 & \textbf{1.17}\textsubscript{+6, -0} & 2.83\textsubscript{+4, -0} & 7.33\textsubscript{+0, -3} & 6.42\textsubscript{+0, -2} & 4.92\textsubscript{+0, -1} & 5.08\textsubscript{+0, -1} & 6.42\textsubscript{+0, -2} & 7.50\textsubscript{+0, -3} & 3.33\textsubscript{+2, -0} \\
 & 5-7 & \textbf{2.33}\textsubscript{+4, -0} & 4.00\textsubscript{+0, -0} & 6.75\textsubscript{+0, -2} & 6.83\textsubscript{+0, -2} & 4.67\textsubscript{+0, -0} & 3.92\textsubscript{+0, -0} & 6.50\textsubscript{+0, -2} & 7.25\textsubscript{+0, -2} & 2.75\textsubscript{+4, -0} \\
 & 8-10 & \textbf{3.00}\textsubscript{+1, -0} & 3.67\textsubscript{+1, -0} & 6.00\textsubscript{+0, -0} & 5.58\textsubscript{+0, -0} & 5.00\textsubscript{+0, -0} & 4.58\textsubscript{+0, -0} & 5.83\textsubscript{+0, -0} & 7.25\textsubscript{+0, -2} & 4.08\textsubscript{+0, -0} \\
\cmidrule(r){2-11}
 & 2-10 & \textbf{2.17}\textsubscript{+6, -0} & 3.50\textsubscript{+4, -0} & 6.69\textsubscript{+0, -4} & 6.28\textsubscript{+0, -3} & 4.86\textsubscript{+1, -1} & 4.53\textsubscript{+2, -1} & 6.25\textsubscript{+0, -3} & 7.33\textsubscript{+0, -5} & 3.39\textsubscript{+4, -0} \\
\midrule
\parbox[t]{2mm}{\multirow{4}{*}{\rotatebox[origin=c]{90}{Recall}}} & 2-4 & 8.92\textsubscript{+0, -6} & 7.42\textsubscript{+0, -5} & 3.83\textsubscript{+2, -0} & 3.50\textsubscript{+2, -0} & 6.67\textsubscript{+0, -1} & 4.00\textsubscript{+2, -0} & 3.54\textsubscript{+2, -0} & \textbf{2.62}\textsubscript{+3, -0} & 4.50\textsubscript{+1, -0} \\
 & 5-7 & 8.75\textsubscript{+0, -5} & 7.33\textsubscript{+0, -4} & 3.54\textsubscript{+3, -0} & 2.62\textsubscript{+3, -0} & 5.75\textsubscript{+0, -1} & 4.38\textsubscript{+1, -0} & \textbf{2.33}\textsubscript{+4, -0} & 3.17\textsubscript{+3, -0} & 7.12\textsubscript{+0, -4} \\
 & 8-10 & 8.83\textsubscript{+0, -5} & 7.25\textsubscript{+0, -4} & 2.88\textsubscript{+3, -0} & 3.54\textsubscript{+3, -0} & 6.00\textsubscript{+0, -2} & 4.83\textsubscript{+1, -0} & 2.21\textsubscript{+4, -0} & \textbf{2.04}\textsubscript{+4, -0} & 7.42\textsubscript{+0, -4} \\
\cmidrule(r){2-11}
 & 2-10 & 8.83\textsubscript{+0, -7} & 7.33\textsubscript{+0, -5} & 3.42\textsubscript{+4, -0} & 3.22\textsubscript{+4, -0} & 6.14\textsubscript{+1, -4} & 4.40\textsubscript{+3, -0} & 2.69\textsubscript{+4, -0} & \textbf{2.61}\textsubscript{+4, -0} & 6.35\textsubscript{+1, -5} \\
\midrule
\parbox[t]{2mm}{\multirow{4}{*}{\rotatebox[origin=c]{90}{F-measure}}} & 2-4 & 4.25\textsubscript{+0, -0} & 3.67\textsubscript{+0, -0} & 6.58\textsubscript{+0, -1} & 5.50\textsubscript{+0, -0} & 5.67\textsubscript{+0, -0} & 4.25\textsubscript{+0, -0} & 5.33\textsubscript{+0, -0} & 7.00\textsubscript{+0, -1} & \textbf{2.75}\textsubscript{+2, -0} \\
 & 5-7 & 7.17\textsubscript{+0, -1} & 6.42\textsubscript{+0, -0} & 5.08\textsubscript{+0, -0} & 4.25\textsubscript{+0, -0} & 4.92\textsubscript{+0, -0} & \textbf{3.17}\textsubscript{+1, -0} & 4.25\textsubscript{+0, -0} & 5.08\textsubscript{+0, -0} & 4.67\textsubscript{+0, -0} \\
 & 8-10 & 7.33\textsubscript{+0, -4} & 5.83\textsubscript{+0, -0} & 3.83\textsubscript{+1, -0} & 3.67\textsubscript{+1, -0} & 5.75\textsubscript{+0, -0} & 3.75\textsubscript{+1, -0} & \textbf{3.42}\textsubscript{+1, -0} & 5.08\textsubscript{+0, -0} & 6.33\textsubscript{+0, -0} \\
\cmidrule(r){2-11}
 & 2-10 & 6.25\textsubscript{+0, -1} & 5.31\textsubscript{+0, -0} & 5.17\textsubscript{+0, -0} & 4.47\textsubscript{+0, -0} & 5.44\textsubscript{+0, -0} & \textbf{3.72}\textsubscript{+1, -0} & 4.33\textsubscript{+0, -0} & 5.72\textsubscript{+0, -0} & 4.58\textsubscript{+0, -0} \\
\midrule
\parbox[t]{2mm}{\multirow{4}{*}{\rotatebox[origin=c]{90}{AUC}}} & 2-4 & 6.50\textsubscript{+0, -0} & 4.58\textsubscript{+0, -0} & 5.96\textsubscript{+0, -0} & 4.58\textsubscript{+0, -0} & 5.50\textsubscript{+0, -0} & 4.29\textsubscript{+0, -0} & 4.67\textsubscript{+0, -0} & 5.17\textsubscript{+0, -0} & \textbf{3.75}\textsubscript{+0, -0} \\
 & 5-7 & 8.75\textsubscript{+0, -6} & 7.42\textsubscript{+0, -4} & 4.17\textsubscript{+1, -0} & 3.17\textsubscript{+2, -0} & 5.08\textsubscript{+1, -0} & 3.50\textsubscript{+2, -0} & \textbf{3.12}\textsubscript{+2, -0} & 3.33\textsubscript{+2, -0} & 6.46\textsubscript{+0, -0} \\
 & 8-10 & 8.58\textsubscript{+0, -5} & 7.17\textsubscript{+0, -4} & 2.75\textsubscript{+4, -0} & 3.42\textsubscript{+3, -0} & 6.25\textsubscript{+0, -3} & 4.50\textsubscript{+1, -0} & \textbf{2.33}\textsubscript{+4, -0} & 2.67\textsubscript{+4, -0} & 7.33\textsubscript{+0, -4} \\
\cmidrule(r){2-11}
 & 2-10 & 7.94\textsubscript{+0, -7} & 6.39\textsubscript{+0, -5} & 4.29\textsubscript{+2, -0} & 3.72\textsubscript{+3, -0} & 5.61\textsubscript{+1, -1} & 4.10\textsubscript{+2, -0} & \textbf{3.38}\textsubscript{+4, -0} & 3.72\textsubscript{+3, -0} & 5.85\textsubscript{+1, -3} \\
\midrule
\parbox[t]{2mm}{\multirow{4}{*}{\rotatebox[origin=c]{90}{G-mean}}} & 2-4 & 7.00\textsubscript{+0, -0} & 5.42\textsubscript{+0, -0} & 5.08\textsubscript{+0, -0} & 4.58\textsubscript{+0, -0} & 5.92\textsubscript{+0, -0} & 4.08\textsubscript{+0, -0} & 4.33\textsubscript{+0, -0} & 4.67\textsubscript{+0, -0} & \textbf{3.92}\textsubscript{+0, -0} \\
 & 5-7 & 8.75\textsubscript{+0, -6} & 7.58\textsubscript{+0, -4} & 4.33\textsubscript{+1, -0} & 3.08\textsubscript{+2, -0} & 5.33\textsubscript{+1, -0} & 3.67\textsubscript{+2, -0} & 3.00\textsubscript{+2, -0} & \textbf{2.92}\textsubscript{+3, -0} & 6.33\textsubscript{+0, -1} \\
 & 8-10 & 8.67\textsubscript{+0, -5} & 7.33\textsubscript{+0, -4} & 2.67\textsubscript{+4, -0} & 3.50\textsubscript{+3, -0} & 6.25\textsubscript{+0, -3} & 4.58\textsubscript{+1, -0} & \textbf{2.25}\textsubscript{+4, -0} & 2.58\textsubscript{+4, -0} & 7.17\textsubscript{+0, -4} \\
\cmidrule(r){2-11}
 & 2-10 & 8.14\textsubscript{+0, -7} & 6.78\textsubscript{+0, -5} & 4.03\textsubscript{+2, -0} & 3.72\textsubscript{+4, -0} & 5.83\textsubscript{+1, -3} & 4.11\textsubscript{+2, -0} & \textbf{3.19}\textsubscript{+4, -0} & 3.39\textsubscript{+4, -0} & 5.81\textsubscript{+1, -3} \\
\bottomrule
\end{tabularx}
\end{sidewaystable*}

\begin{figure}[!t]
\centering
\subfloat[F-measure, baseline]{\includegraphics[width=0.33\linewidth]{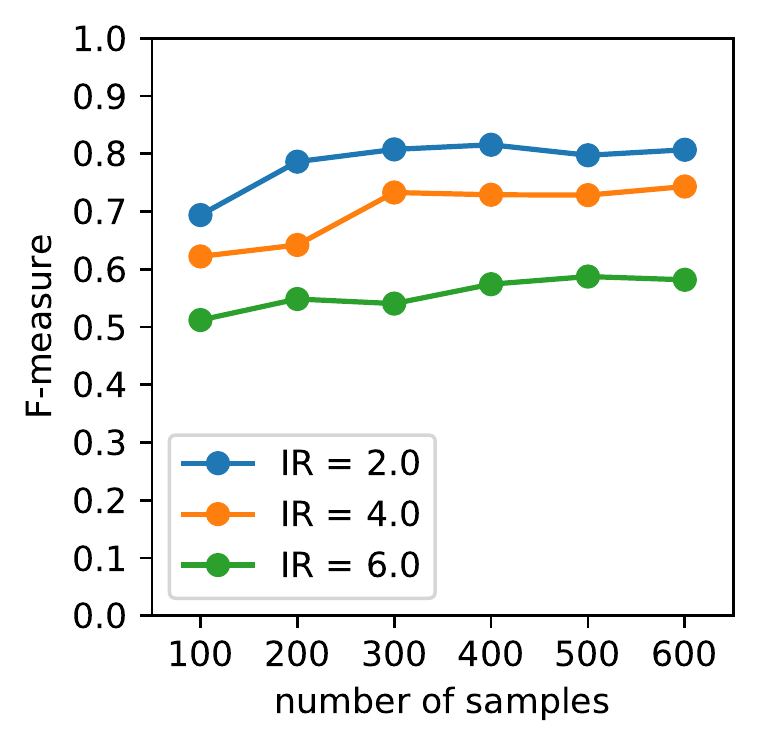}}
\hfil
\subfloat[AUC, baseline]{\includegraphics[width=0.33\linewidth]{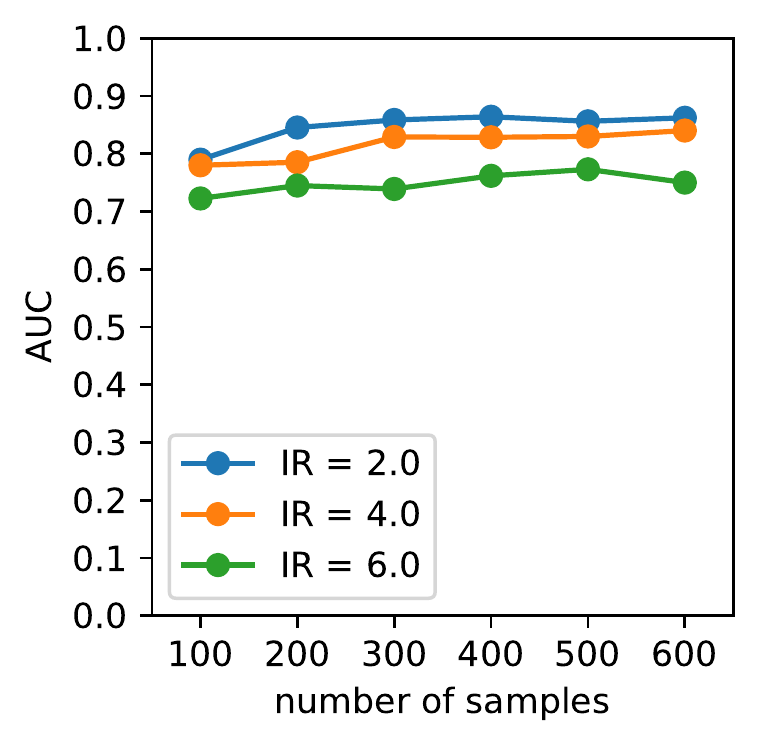}}
\hfil
\subfloat[G-mean, baseline]{\includegraphics[width=0.33\linewidth]{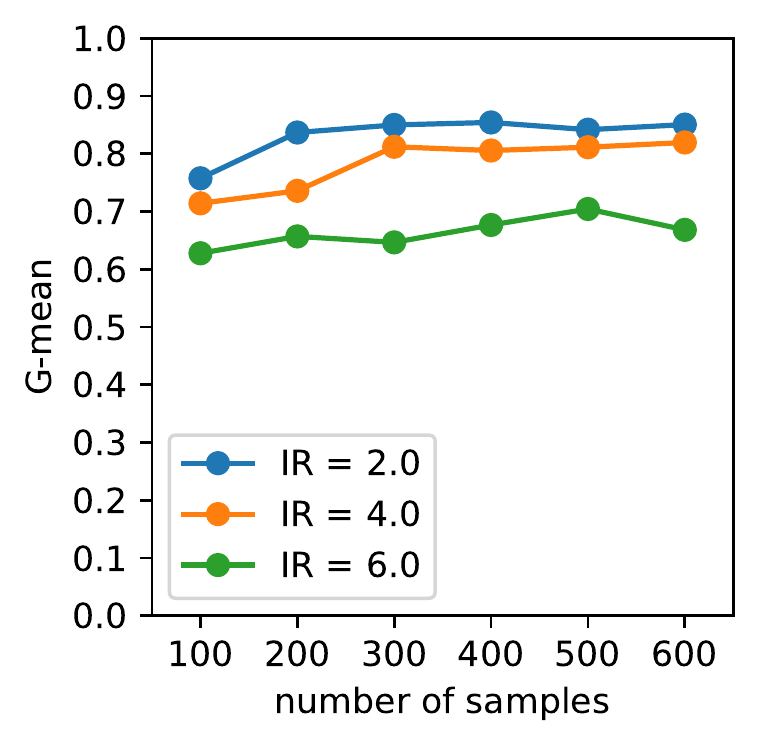}}

\subfloat[F-measure, NCL]{\includegraphics[width=0.33\linewidth]{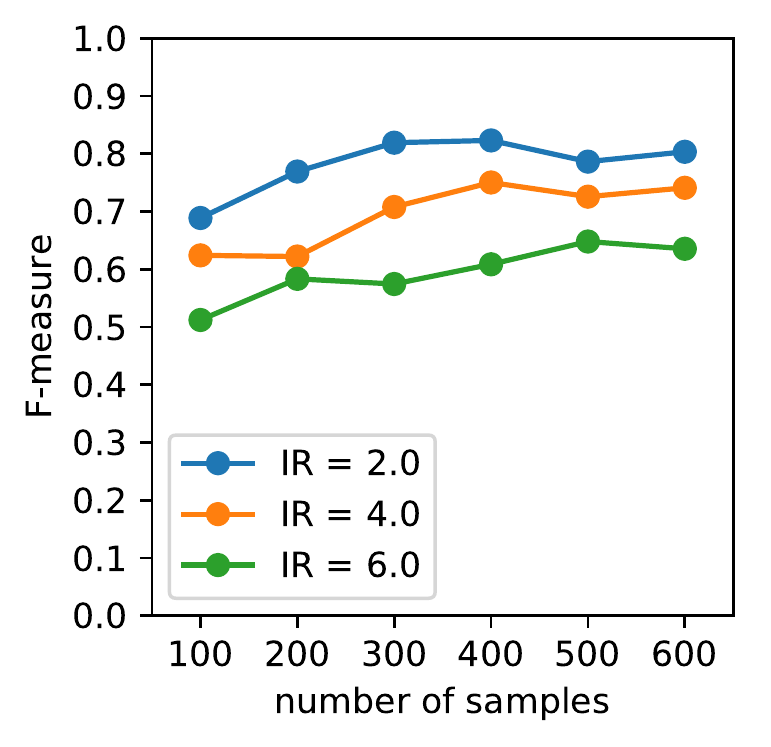}}
\hfil
\subfloat[AUC, NCL]{\includegraphics[width=0.33\linewidth]{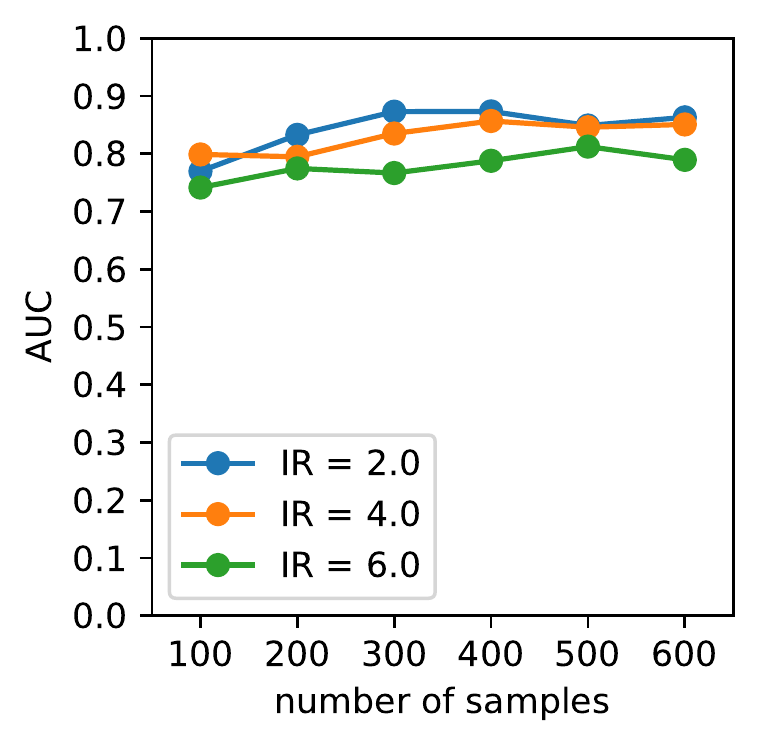}}
\hfil
\subfloat[G-mean, NCL]{\includegraphics[width=0.33\linewidth]{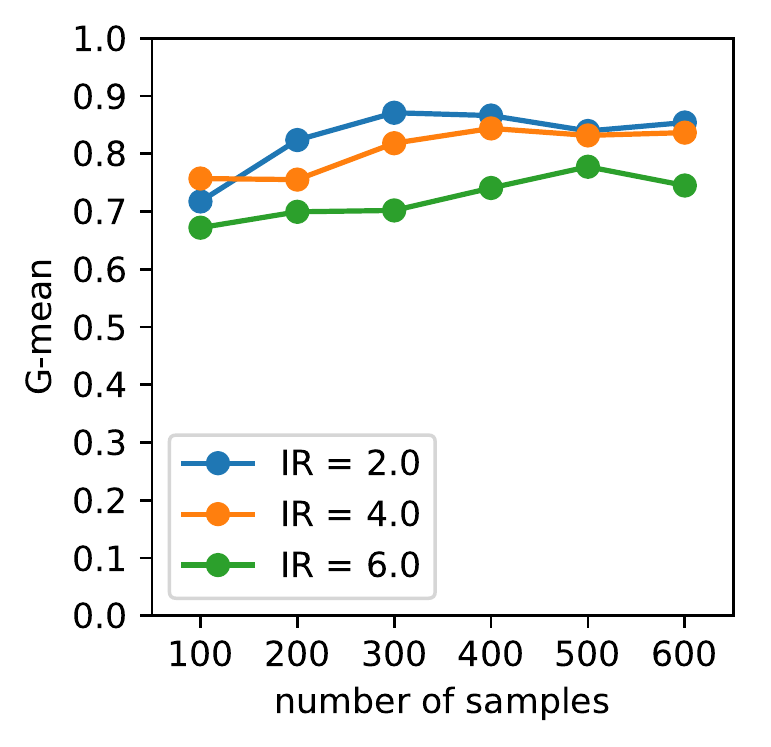}}

\subfloat[F-measure, RBO]{\includegraphics[width=0.33\linewidth]{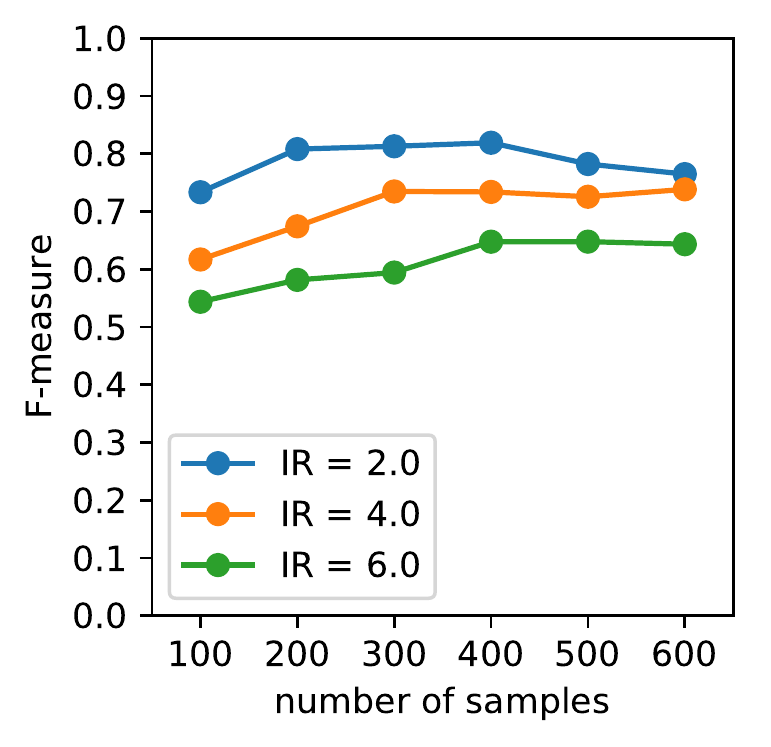}}
\hfil
\subfloat[AUC, RBO]{\includegraphics[width=0.33\linewidth]{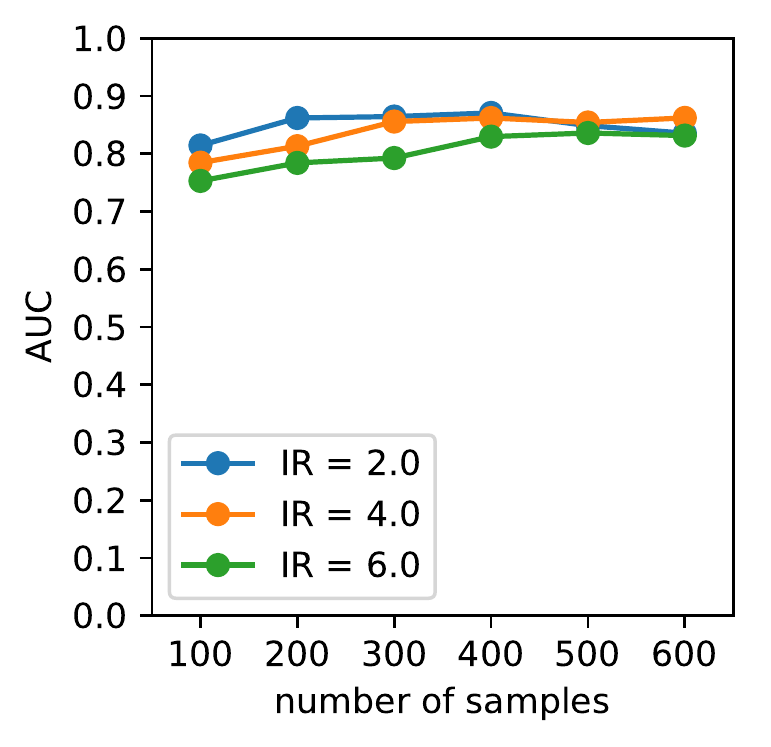}}
\hfil
\subfloat[G-mean, RBO]{\includegraphics[width=0.33\linewidth]{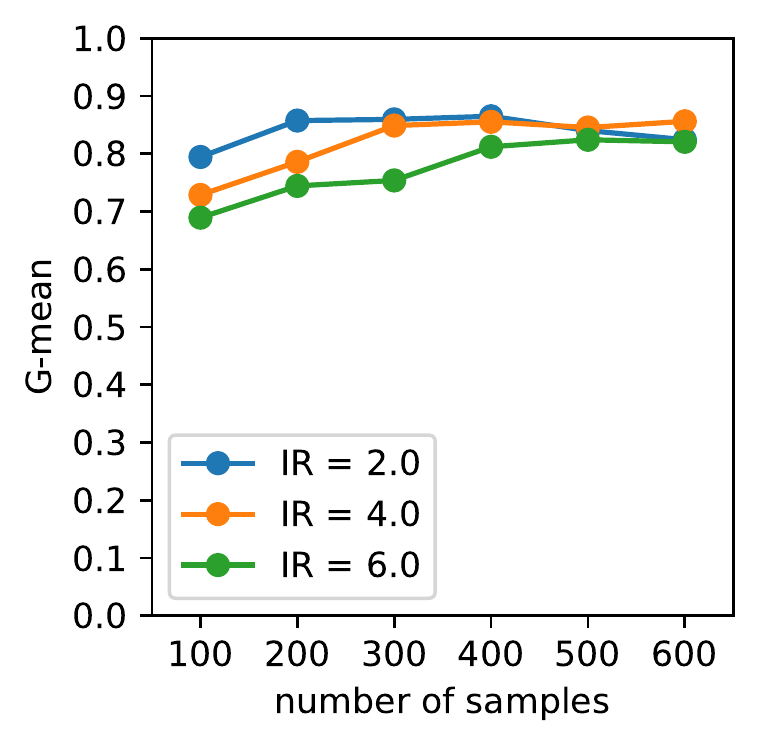}}
\caption{The impact of the number of training observations on average values of various performance metrics, either on the original data (top row), after undersampling with NCL (middle row) or oversampling with RBO (bottom row).}
\label{fig:n_samples}
\end{figure}

\noindent\textbf{The value of new data in the presence of data imbalance.} The goal of the third experiment was evaluating to what extent increasing the amount of training data improves the performance for various levels of imbalance. A the total number of training observations $\in \{100, 200, ..., 600\}$, and IR $\in \{2.0, 4.0, 6.0\}$, were considered. In addition to the baseline case, in which no strategy of dealing with imbalance was employed, two best-performing resampling techniques were used: NCL and RBO. The average values of the combined performance measures are presented in Figure~\ref{fig:n_samples}. As can be seen, in the baseline case data imbalance decreases the value of new observations. For the case of IR = 6.0, even after increasing the number of training samples six times, the same performance as the one observed for IR = 4.0 was not achieved for any of the considered metrics. In other words, even when more training data from both minority and majority distributions was used, due to the inherent data imbalance a worse performance was achieved. To a smaller extent this trend is visible also between IR = 2.0 and IR = 4.0, especially when F-measure is considered. Using one of the resampling techniques prior to classification partially reduced this trend: in this case, after increasing the number of samples it was possible to outperform the case with 100 training samples.

\begin{figure}[!t]
\centering
\subfloat[F-measure, IR 2-4]{\includegraphics[width=0.33\linewidth]{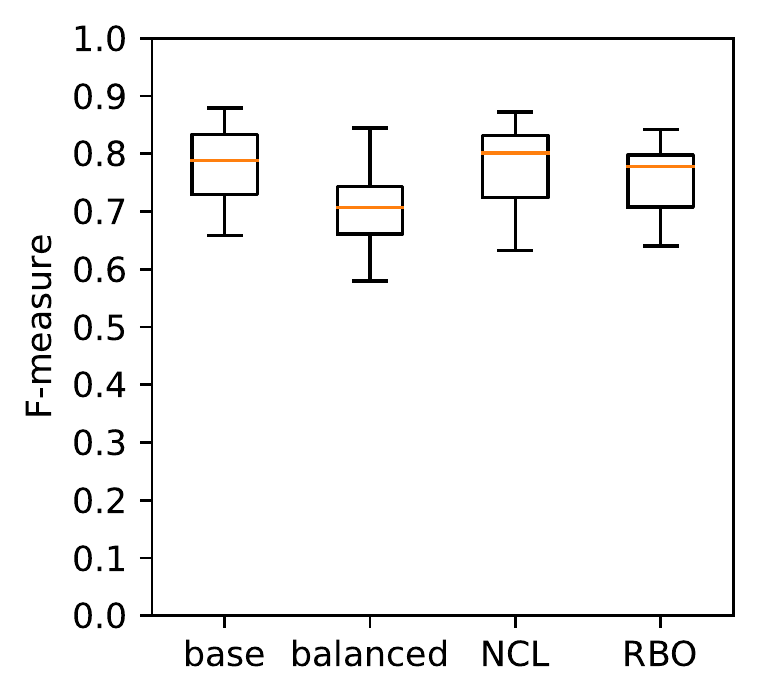}}
\hfil
\subfloat[AUC, IR 2-4]{\includegraphics[width=0.33\linewidth]{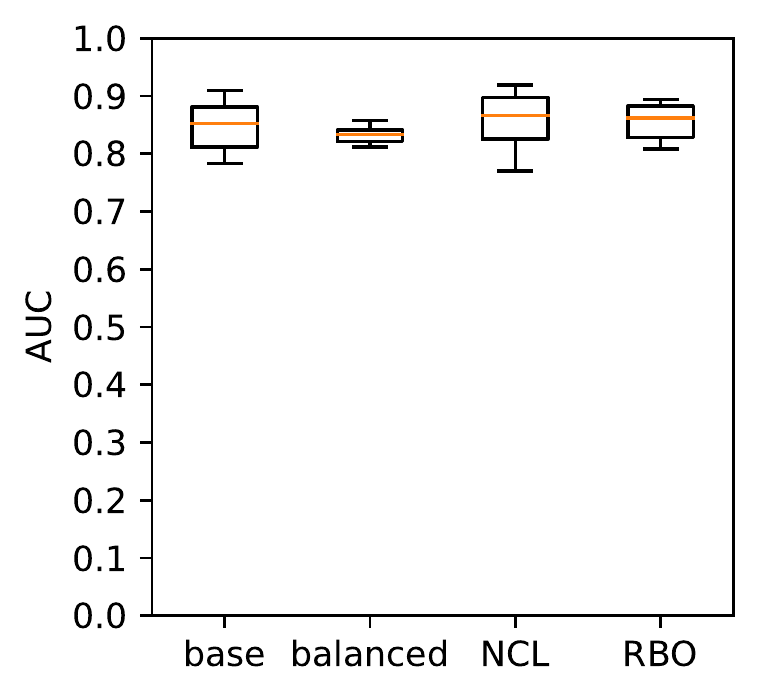}}
\hfil
\subfloat[G-mean, IR 2-4]{\includegraphics[width=0.33\linewidth]{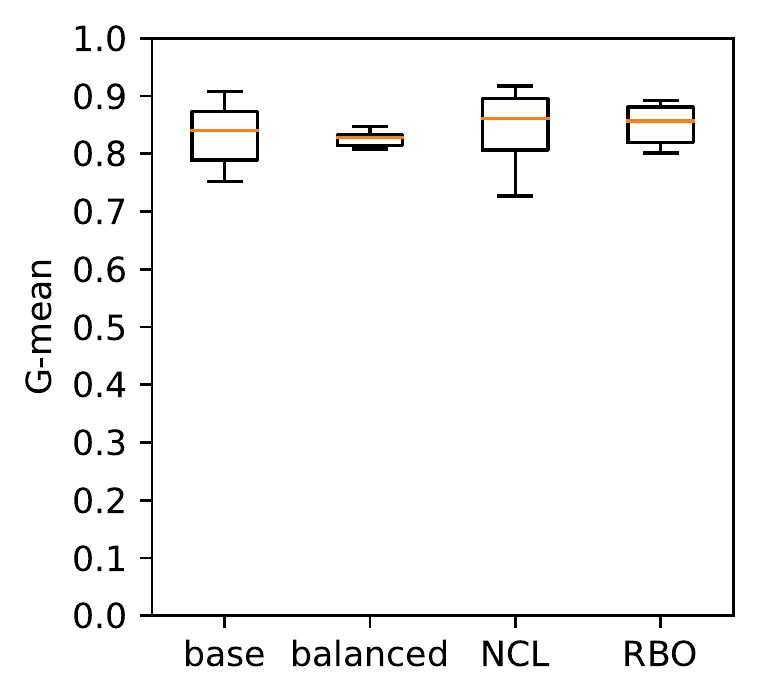}}

\subfloat[F-measure, IR 5-7]{\includegraphics[width=0.33\linewidth]{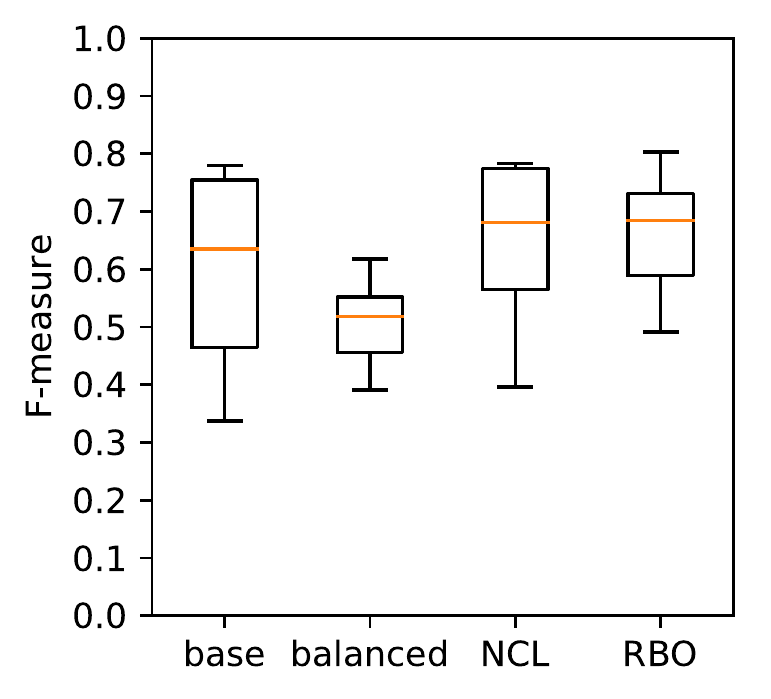}}
\hfil
\subfloat[AUC, IR 5-7]{\includegraphics[width=0.33\linewidth]{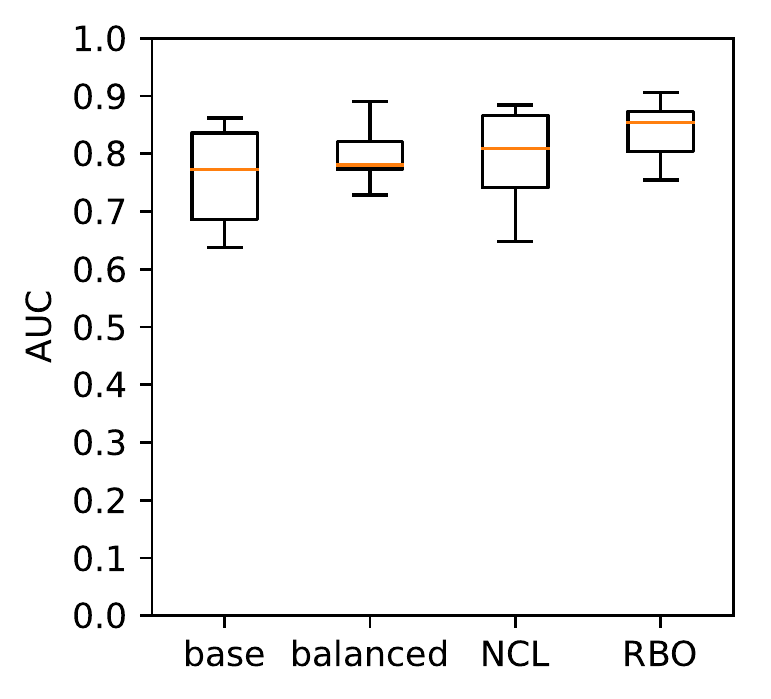}}
\hfil
\subfloat[G-mean, IR 5-7]{\includegraphics[width=0.33\linewidth]{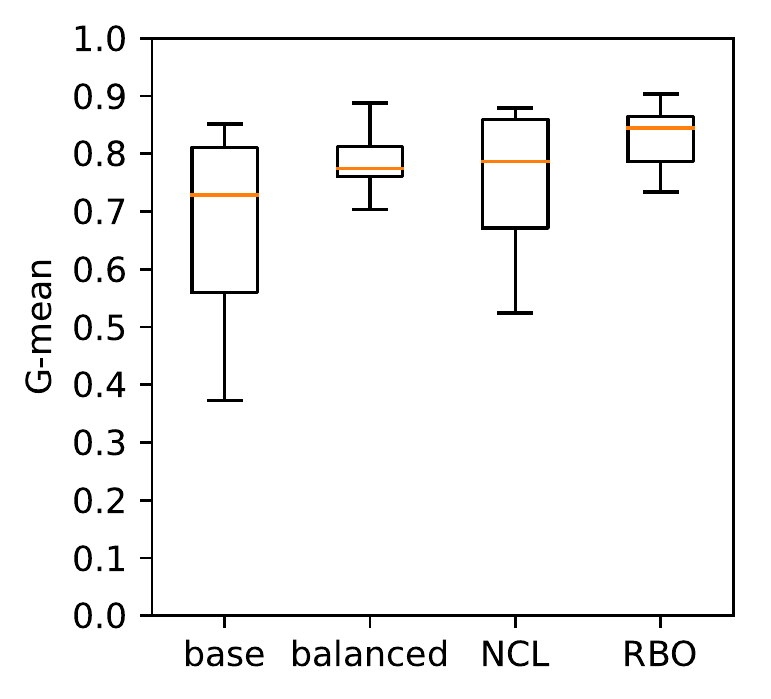}}

\subfloat[F-measure, IR 8-10]{\includegraphics[width=0.33\linewidth]{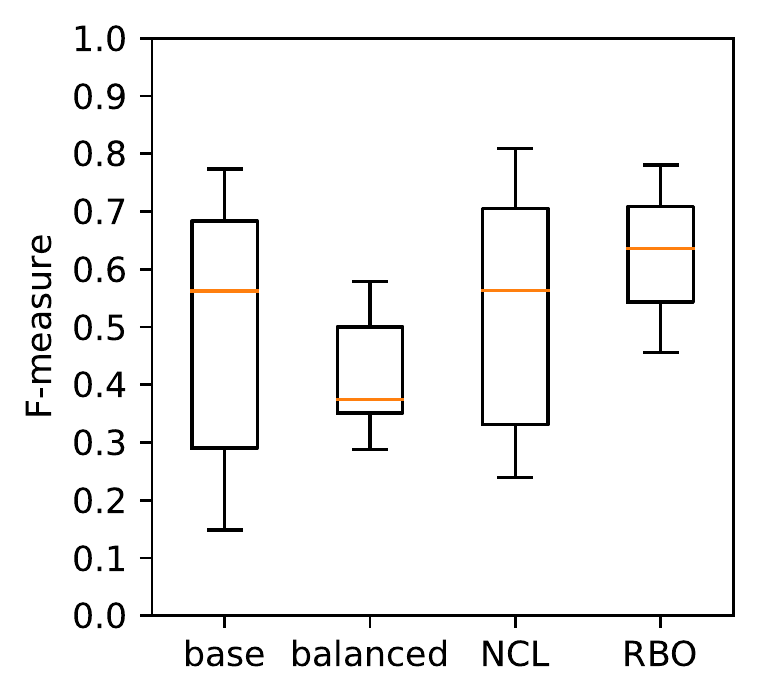}}
\hfil
\subfloat[AUC, IR 8-10]{\includegraphics[width=0.33\linewidth]{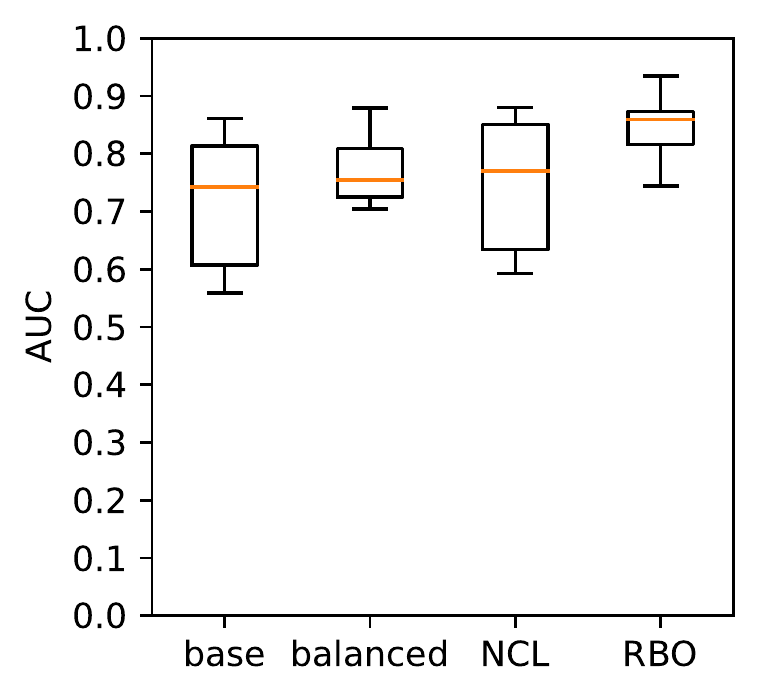}}
\hfil
\subfloat[G-mean, IR 8-10]{\includegraphics[width=0.33\linewidth]{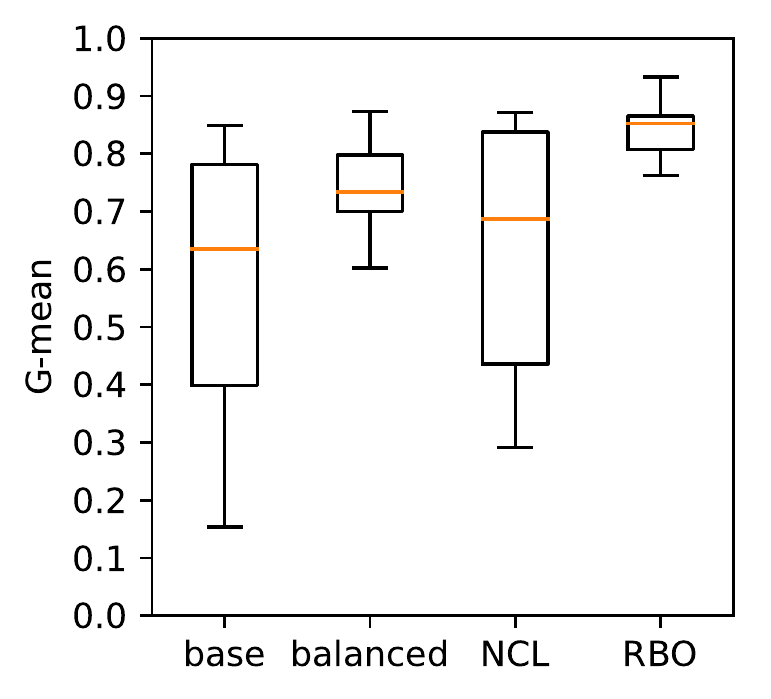}}
\caption{Average values of various performance metrics. Baseline case, in which both training and test data was imbalanced, was compared with the case in which only test data was imbalanced. Performance for NCL and RBO was also included for reference.}
\label{fig:test_only_imba}
\end{figure}

\noindent\textbf{The strategy of balancing training distribution during data acquisition.} In the previous experiments both training and test data distributions were modified while adjusting the imbalance ratio. However, when dealing with real data one does not have an option of adjusting test distribution. Still, in some cases it is possible to influence the imbalance of training data: for instance, in the case of histopathological images we can have at our disposal a larger quantity of unannotated images, and the main cost is associated with the annotation process. It is, therefore, possible to select the images designed for annotation so that their distribution is balanced. The goal of the final experiment was evaluating whether such data acquisition strategy is beneficial for the classification performance. To this end two variants were evaluated: the baseline case, in which both training and test data distribution were imbalanced with IR $\in \{2.0, 3.0, ..., 10.0\}$, and the balanced case, in which only test distribution was imbalanced and training data consisted of an equal number of randomly selected samples from both classes. The results of this experiment are presented in Figure~\ref{fig:test_only_imba}. For reference, the performance observed on data balanced with NCL and RBO was also included. As can be seen, for low values of IR a worse performance can actually be observed after balancing the training data according to all of the combined performance metrics. This trend is most noticeable for F-measure. Furthermore, the observed F-measure was also higher in the baseline case for higher IR. On the other hand, balancing training data improved the AUC and G-mean for medium and high levels of imbalance. In all of the cases, using the original, imbalanced training data distribution and balancing it with one of the considered resampling strategies led to an improvement in performance.

\noindent\textbf{Main results.} The conclusions made based on the observed results can be summarized as follows:
\begin{itemize}
\item Medium and high data imbalance levels have a significant negative impact on the classification performance, irregardless of the chosen performance measure. However, for some of the considered measures, at low level of imbalance an improved performance was observed, which may suggest that small data imbalance can actually be beneficial in a specific settings. Especially the latter finding should be further confirmed on additional benchmark datasets.
\item Some of the popular strategies of dealing with data imbalance, namely using weighted loss and oversampling data with SMOTE, significantly underperformed in the conducted experiments. Techniques that achieved the best results were NCL and RBO resampling algorithms. This leads to a conclusion that developing a novel strategies of handling data imbalance, designed specifically for dealing with images, might be necessary to achieve a satisfactory performance in the histopathological image recognition task.
\item Data imbalance negatively impacts the value of additional training data. Even when more data from both minority and majority class was used, data imbalance made it impossible to achieve a performance observed for lower imbalance ratios. This can be partially mitigated by using an appropriate strategy of handling data imbalance.
\item Depending on data imbalance ratio and the metric used to measure classification performance, balancing training data during acquisition can have a negative impact on the performance when compared to sampling training data with the same imbalance ratio as test data. In all of the considered cases, applying resampling on imbalanced data was preferable approach to balancing data during acquisition.
\end{itemize}

\section{Small-scale binary breast cancer image classification with RBU}

In the second of the conducted studies \cite{koziarski2019radial} the possibility of performing data resampling in the space of high-level features extracted from a previously trained convolutional neural network was examined. The investigation was focused on the RBU algorithm due to the fact that undersampled data produced with such approach can be afterwards fed back to the neural network: while the undersampling is performed in the feature space, every feature representation can be easily reverted to the associated original image representation. On the contrary, when using oversampling algorithms that create synthetic observations, once the observations are generated, they cannot be easily transformed back to the image representation.

\begin{figure}
\centering
\includegraphics[width=0.7\linewidth]{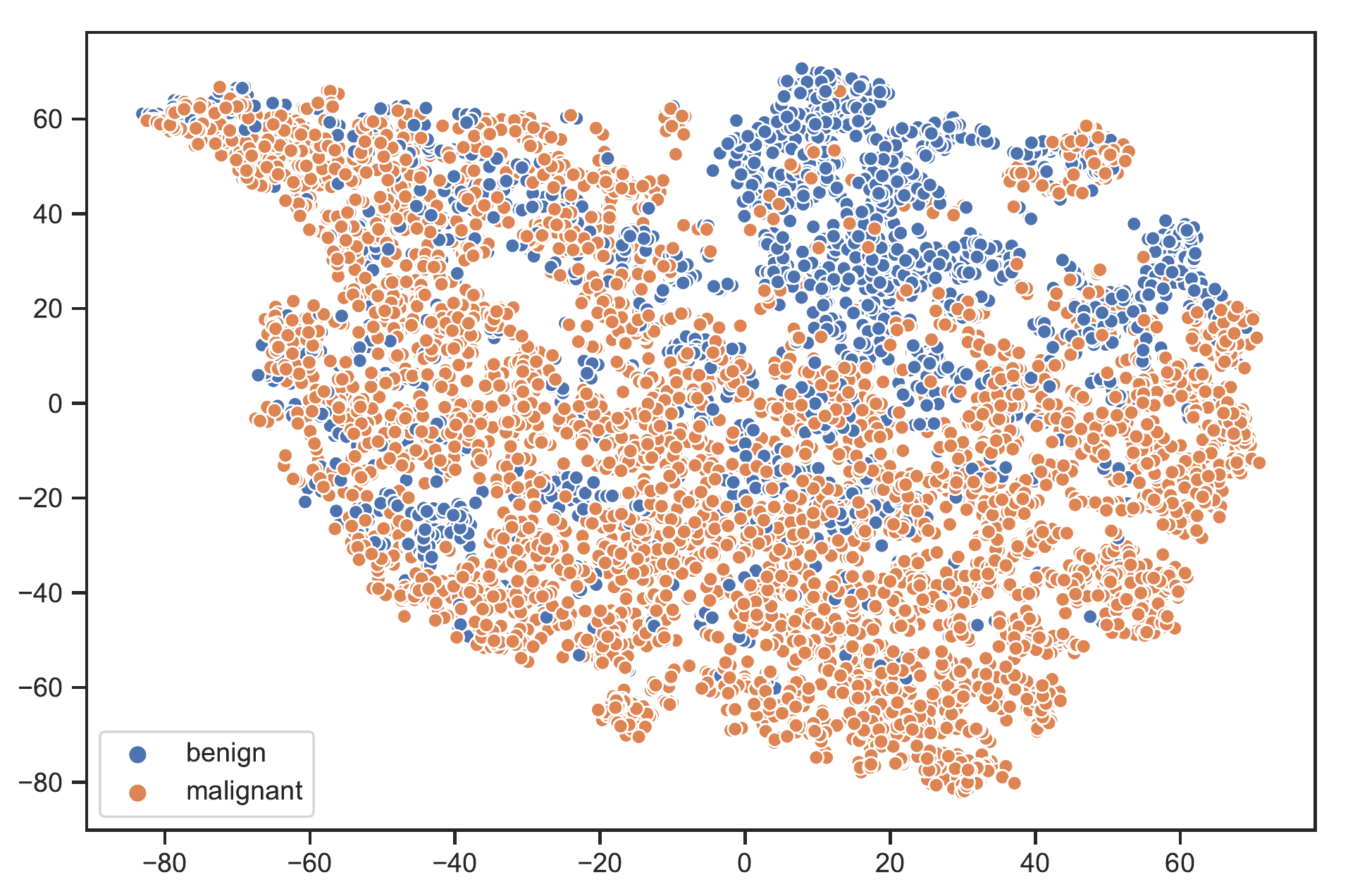}
\caption{Visualization of high-level feature representations extracted using VGG16 convolutional neural network from BreakHis dataset, projected into the two-dimensional space with PCA and t-SNE.}
\label{fig:isbu_cnn}
\end{figure}

Majority of the existing undersampling algorithms, including RBU, is based on the possibility of object comparison using the euclidean distance metric. However, image data, such as histopathological tissue, is ill-suited for such comparison. To extract more suitable feature representation that could be employed during undersampling, high-level feature representations extracted from the last layer of a trained VGG16 network \cite{simonyan2014very} were used. Resulting 4096-dimensional features were later used to perform undersampling in the applicable cases. Visualization of features extracted from the original BreakHis dataset, projected into the two-dimensional space with PCA and t-SNE, is presented in Figure~\ref{fig:isbu_cnn}. As can be seen, the resulting features represent a typical case of an imbalanced data classification problem, with a clearly distinguishable minority object cluster and several additional, smaller minority clusters overlapping the majority distribution.

\noindent\textbf{Impact of undersampling ratio.} In the first stage of the conducted experimental study the question of the impact of undersampling ratio on the algorithms performance was considered. To this end two algorithms were compared: random undersampling (RUS) and RBU. The values of ratio $\in \{0.0, 0.1, 0.2, ..., 1.0\}$ were considered, with ratio equal to 0 indicating that not a single majority object was undersampled, and ratio equal to 1 indicating that majority objects were undersampled up to the point of achieving a balanced class distribution.

\begin{figure*}
\centering
\includegraphics[width=0.85\textwidth]{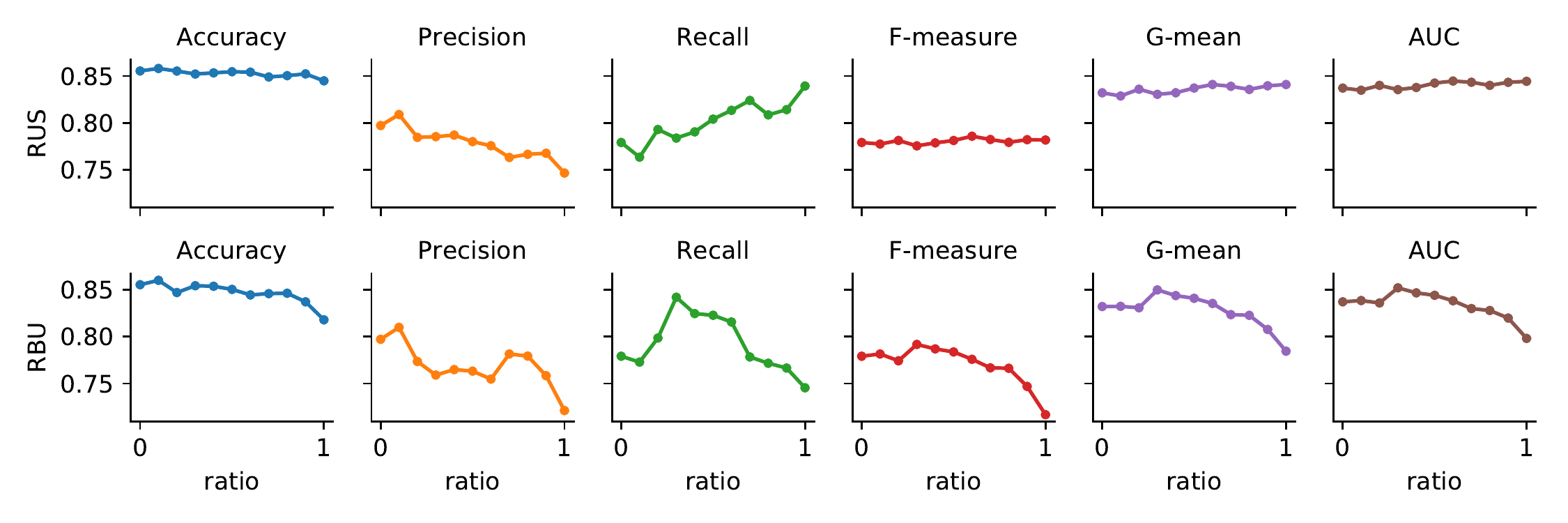}
\caption{Comparison of the impact of undersampling ratio on classification performance for RUS and RBU algorithms.}
\label{fig:ratio}
\end{figure*}

The results of this experiment are presented in Figure~\ref{fig:ratio}. As can be seen, for the random undersampling a peak in performance with respect to the combined metrics, that is F-measure, G-mean and AUC, was observed for an undersampling ratio equal to 0.6; however, the change was relatively insignificant, with a small improvement in performance compared to the baseline case with no undersampling applied, irregardless of the chosen undersampling ratio. Furthermore, the impact on precision and recall was fairly typical for the undersampling, with recall increasing and precision decreasing as more objects were discarded. On the other hand, for RBU a more significant increase in performance was observed at the ratio equal to 0.3, followed by a rapid decrease as the undersampling ratio increased further. Interestingly, contrary to the random undersampling, applying RBU at higher undersampling ratios was actually noticeably detrimental to the classification performance. Furthermore, as opposed to the random undersampling, discarding a large quantity of objects actually negatively impacted the classifiers recall: this could be caused by the fact that RUS discards objects proportionally to their density, preserving the general shape of the majority class distribution, while RBU alters the distribution shape, which seems to be detrimental to the classifier as a larger quantity of objects is discarded.

\noindent\textbf{Comparison with reference methods.} In the second stage of the conducted experimental study the performance of RBU was compared with that of selected reference resampling strategies. Namely, in addition to the random oversampling (ROS) and undersampling (RUS), SMOTE oversampling algorithm \cite{chawla2002smote}, Neighborhood Cleaning Rule (NCL) \cite{laurikkala2001improving}, Instance Hardness Threshold method (IHT) \cite{smith2014instance} and Tomek links undersampling (TL) \cite{tomek1976two} were considered. Finally, the baseline case in which no resampling was applied to the data was also included. For the reference undersampling strategies on which the type of feature representation could have an impact on the performance, both original image representation as well as the high-level features extracted using convolutional neural network (denoted with subscript \txtsub{CNN}) were considered.

The observed results are presented in Table~\ref{tab:final}. As can be seen, the proposed RBU algorithm achieved the best results with respect to all of the combined performance metrics, that is F-measure, G-mean and AUC. Despite being outperformed by individual reference methods with respect to either precision or recall, RBU achieved the best balance between the two out of the considered resampling algorithms. This indicates its usefulness in the task of imbalanced image classification, particularly in the recognition of histopathological tissue. It is also worth noting that using the high-level feature representations, extracted from a convolutional neural network, was not always beneficial for the reference undersampling algorithms: while a notable improvement in performance was observed for the IHT algorithm, a slightly worse results were observed for the NCL, and no significant change was observed for TL.

\begin{table*}[htb]
\scriptsize
\caption{Final comparison of different resampling strategies on BreakHis classification performance.}
\label{tab:final}
\centering
\begin{tabularx}{\textwidth}{lYYYYYYYYYYY}
\hline\noalign{\smallskip}
Metric & Base & RUS & ROS & SMOTE & NCL & NCL\txtsub{CNN} & IHT & IHT\txtsub{CNN} & TL & TL\txtsub{CNN} & RBU\txtsub{CNN} \\
\noalign{\smallskip}
\hline
\noalign{\smallskip}
Precision & .797 & .767 & .797 & .789 & .780 & .793 & .638 & .667 & \textbf{.800} & .798 & .759 \\
Recall & .779 & .818 & .801 & .793 & .821 & .799 & \textbf{.931} & .905 & .780 & .777 & .842 \\
F-measure & .779 & .784 & \textbf{.792} & .783 & .791 & .789 & .751 & .762 & .779 & .778 & \textbf{.792} \\
G-mean & .832 & .840 & .843 & .837 & .845 & .842 & .818 & .831 & .831 & .831 & \textbf{.850} \\
AUC & .837 & .844 & .847 & .841 & .849 & .846 & .829 & .837 & .837 & .837 & \textbf{.852} \\
\noalign{\smallskip}
\hline
\end{tabularx}
\end{table*}

\noindent\textbf{Main results.} The results of the conducted experimental analysis indicate the suitability of the approach of undersampling in the high-level convolutional neural network feature space, in particular in combination with RBU, which outperformed reference methods with respect to all of the considered combined performance metrics. Furthermore, the conducted analysis of the impact of the undersampling ratio on the classification performance shed some light on the algorithms behavior, namely the observed decrease in the performance for higher numbers of undersampled objects, which is likely caused by the fact that contrary to the random approaches, RBU alters the shape of the majority class distribution.

\section{Large-scale multi-class prostate cancer classification with MC-CCR}

In the third of the conducted studies \cite{diagset} the problem of large-scale multi-class prostate cancer classification was considered. While the two previous studies focused explicitly on the problem of data imbalance, in the last one a general methodology for the patch-level recognition and slide-level diagnosis was developed, and data imbalance was only one of the recognized data difficulty factors that were considered. Furthermore, in contrast to the two previously described studies that were based on a relatively small benchmark datasets, during the last study a novel large-scale dataset annotated by expert histopathologists was developed. As a result, the last study describes a whole pipeline that was developed with the aim of supporting the process of the medical diagnosis in a real-life setting.

Due to the fact that this study has a scope exceeding the area of interest of this thesis, in the remainder of this section only the excerpts relevant to the issue of data imbalance were included, that is description of the developed dataset and the results of experiments identifying data difficulty factors which were present in a patch-level recognition task. Other contributions, including whole classification pipeline based on multi-scale ensembles of convolutional neural networks, and a methodology of converting patch-level prediction maps to the scan-level diagnosis, were described in more detail in the paper itself \cite{diagset}.

\begin{figure*}
\centering
\includegraphics[width=\textwidth]{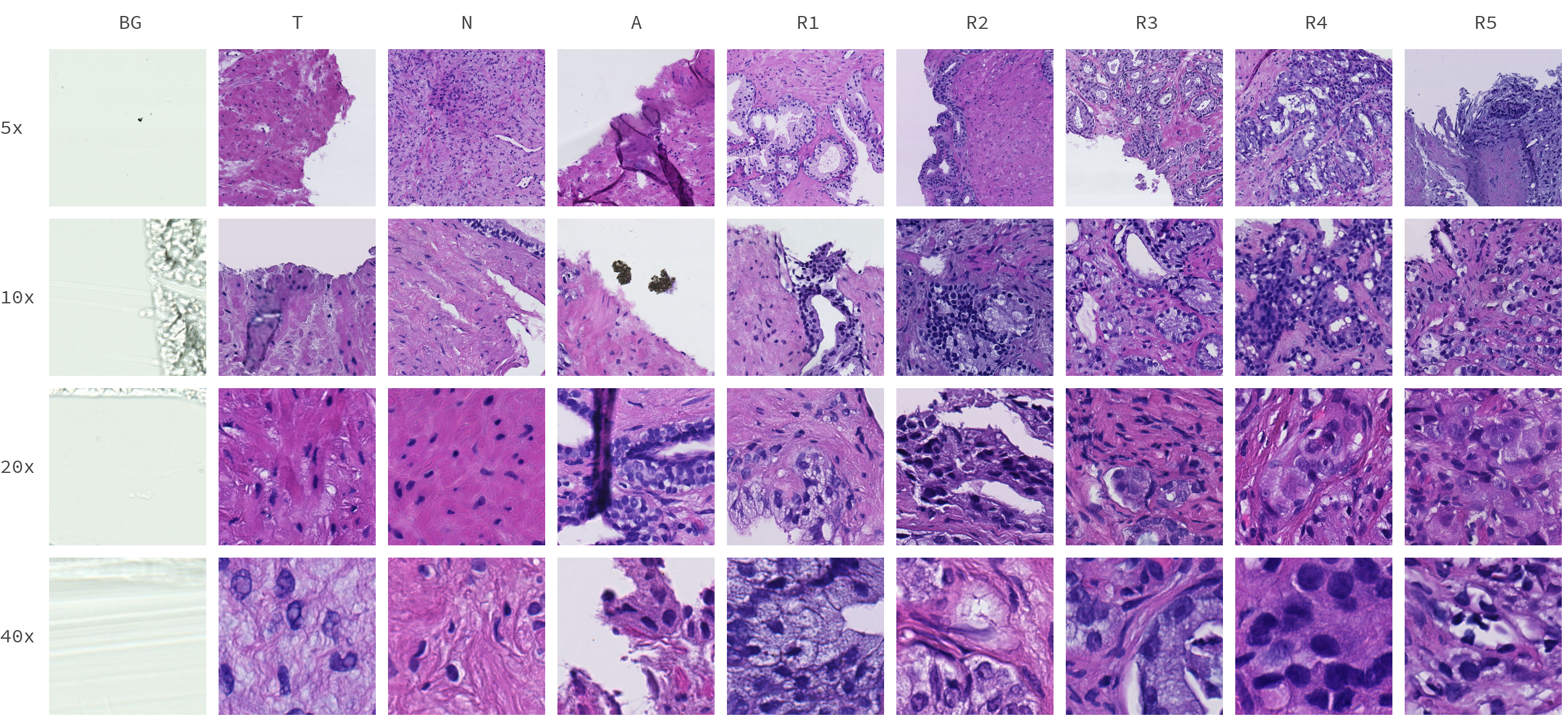}
\caption{Randomly selected samples of image patches from DiagSet-A, extracted at different magnifications (rows) and containing different classes of tissue (columns).}
\label{fig:diagset-samples}
\end{figure*}

\noindent\textbf{Dataset.} During the data preparation stage two separate datasets were constructed, one containing a patch-level data with detailed multi-class annotations, and the second containing whole histopathological scans with a binary diagnosis indicating the presence of cancerous tissue on the scan. The remainder of this section focuses on the first of the datasets, DiagSet-A, as it was the basis of experiments focused on data imbalance. DiagSet-A consists of small image patches extracted from the underlying 
whole slide imaging (WSI) scans, with labels assigned based on the annotation made by human histopathologists. Patches with a size of $256\times256$ were extracted from the scans with a stride of 128, at 4 different magnification levels: $40\times$, $20\times$, $10\times$ and $5\times$. Each patch was assigned a single label out of 9 possible classes: scan background (BG), tissue background (T), normal, healthy tissue (N), acquisition artifact (A), or one of the 1-5 Gleason \cite{Albelda1993,Konig2004,Gleason1966} grades (R1-R5). Samples for each class and magnification level are presented in Figure~\ref{fig:diagset-samples}.

During the labeling process a histopathologist annotated larger WSI regions as belonging to one of the defined classes. Due to the nature of the labeling process some patches can be covered by annotations only partially, or contain multiple overlapping annotations. To translate these annotations to labels on the patch level the following procedure was used: on the highest magnification level, that is $40\times$, a label was assigned if and only if only a single class annotation with overlap ratio equal to or higher than 0.75 was present. In this case that annotation label was assigned as a class associated with a given patch. If either none of the classes overlapped the patch at a specified ratio, or multiple contradictory labels were present, the patch was not assigned any class. Secondly, on lower magnification levels, that is $20\times$, $10\times$, or $5\times$, a patch was first divided into smaller $40\times$ patches (4 in case of $20\times$ magnification, 16 in case of $10\times$, and 64 in case of $5\times$). Each $40\times$-level patch was assigned a label according to the previously described procedure. Finally, the most severe of the $40\times$-level labels were assigned as a final label for the lower magnification patch. For instance, if given $20\times$-level patch could be divided into one $40\times$-level patch with a label N, two $40\times$-level patches with R3 label, and one $40\times$-level patch with R4 label, the R4 label would be assigned to the $20\times$-level patch.

Due to the length of the annotation process, as well as the time required to train described machine learning models on a large quantities of data, and for the sake of the experimental study described in later part of this section, DiagSet-A was divided into two parts. First, DiagSet-A.1, consists of 238 WSI scans annotated by the histopathologists. DiagSet-A.1 was used in the majority of the preliminary experiments described in the remainder of this section. Secondly, DiagSet-A.2, which compared to the DiagSet-A.1 consists of 190 additional training scans, as well as a single additional validation and test scan. Importantly, DiagSet-A.2 also introduced additional class, BG, which was not originally present in the dataset. DiagSet-A.2 was used during the final evaluation of the proposed approach, and can be treated as a final version of the dataset. Detailed number of scans and patches extracted for both version of the dataset are presented in Table~\ref{table:diagset-size}, whereas their class distribution is presented in Table~\ref{table:diagset-distrib}.

\begin{table*}
\caption{Class distribution of the DiagSet-A.1 (initial dataset, used in the preliminary experiments) and DiagSet-A.2 (final dataset).}
\label{table:diagset-distrib}
\centering
\begin{tabular}{llllllllll}
\toprule
\textbf{Version} & \textbf{BG} & \textbf{T} & \textbf{N} & \textbf{A} & \textbf{R1} & \textbf{R2} & \textbf{R3} & \textbf{R4} & \textbf{R5} \\
\midrule
DiagSet-A.1 & 0.0\% & 24.8\% & 44.6\% & 2.0\% & 0.6\% & 0.8\% & 6.6\% & 17.1\% & 3.4\% \\
DiagSet-A.2 & 6.6\% & 23.8\% & 35.3\% & 8.3\% & 0.7\% & 0.7\% & 6.1\% & 15.4\% & 3.0\% \\
\bottomrule
\end{tabular}
\end{table*}

\begin{table*}
\caption{Detailed number of scans and extracted patches in the specific data partitions.}
\label{table:diagset-size}
\centering
\begin{tabular}{llllll}
\toprule
\textbf{Version} & \textbf{Partition} & \textbf{\# of scans} & \textbf{Magnification} & \textbf{\# of patches} \\
\midrule
DiagSet-A.1 & train & 156 & $40\times$ & 850,860 \\
 &  &  & $20\times$ & 273,321 \\
 &  &  & $10\times$ & 90,746 \\
 &  &  & $5\times$ & 32,981 \\
\cmidrule{2-5}
 & validation & 41 & $40\times$ & 215,071 \\
 &  &  & $20\times$ & 70,013 \\
 &  &  & $10\times$ & 23,189 \\
 &  &  & $5\times$ & 8,374 \\
\cmidrule{2-5}
 & test & 41 & $40\times$ & 267,691 \\
 &  &  & $20\times$ & 86,888 \\
 &  &  & $10\times$ & 28,945 \\
 &  &  & $5\times$ & 10,507 \\
\cmidrule{3-5}
 &  & 238 &  & 1,958,586 \\
\midrule
DiagSet-A.2 & train & 346 & $40\times$ & 1,250,661 \\
 &  &  & $20\times$ & 398,201 \\
 &  &  & $10\times$ & 132,882 \\
 &  &  & $5\times$ & 48,782 \\
\cmidrule{2-5}
 & validation & 42 & $40\times$ & 245,441 \\
 &  &  & $20\times$ & 77,889 \\
 &  &  & $10\times$ & 25,294 \\
 &  &  & $5\times$ & 8,977 \\
\cmidrule{2-5}
 & test & 42 & $40\times$ & 284,032 \\
 &  &  & $20\times$ & 91,125 \\
 &  &  & $10\times$ & 30,086 \\
 &  &  & $5\times$ & 10,836 \\
\cmidrule{3-5}
 &  & 430 &  & 2,604,206 \\
\bottomrule
\end{tabular}
\end{table*}

\noindent\textbf{Experimental study.} As a part of the conducted experimental study some of the factors that might have limited the performance achieved by convolutional neural networks in the patch recognition task were analysed. Due to the computational constraints the experiments focused on two architectures: VGG19, the model that achieved the best performance in the binary setting in the previous stage of experiments, and ResNet50, as an representative of a more recent architecture that underperformed in the conducted tests. Three factors that might have affected the performance of a specific neural networks were distinguished: availability of data, presence of label noise, and data imbalance.

\noindent\textbf{Availability of data.} A known difficulty associated with using machine learning algorithms in the histopathology is high cost associated with data acquisition, which requires a tedious and time-consuming labeling by an experienced medical practitioner. As a result, the availability of the labeled scans is limited. Despite the fact that due to their high resolution the number of image patches extracted from a single scan can be relatively high, an argument can be made that the patches extracted from a single scan are likely to be roughly similar. At the same time, there is a great variation in the appearance of cancer tissue, which makes the process of generalizing from a limited amount of data difficult.

To experimentally evaluate the impact of the amount of available data on the performance of a network trained in the patch recognition task, the number of training observations was artificially decreased and the model was retrained from scratch. Afterwards, the performance of the model was evaluated on a fixed validation set in the binary class setting. Specifically, the number of scans in $\{6, 21, 36, 51, ..., 156\}$ was considered, with 15 training scans incrementally added to the pool at each evaluation stage. Both VGG19 and ResNet50 architectures were examined, and the data was extracted at two different magnifications: $40\times$ and $10\times$. It is worth noting that despite the same pool of scans used for both magnifications, due to the used patch extraction procedure, the total number of patches was approximately 10 times higher at the $40\times$ magnification.

\begin{figure}
\centering
\includegraphics[width=0.7\linewidth]{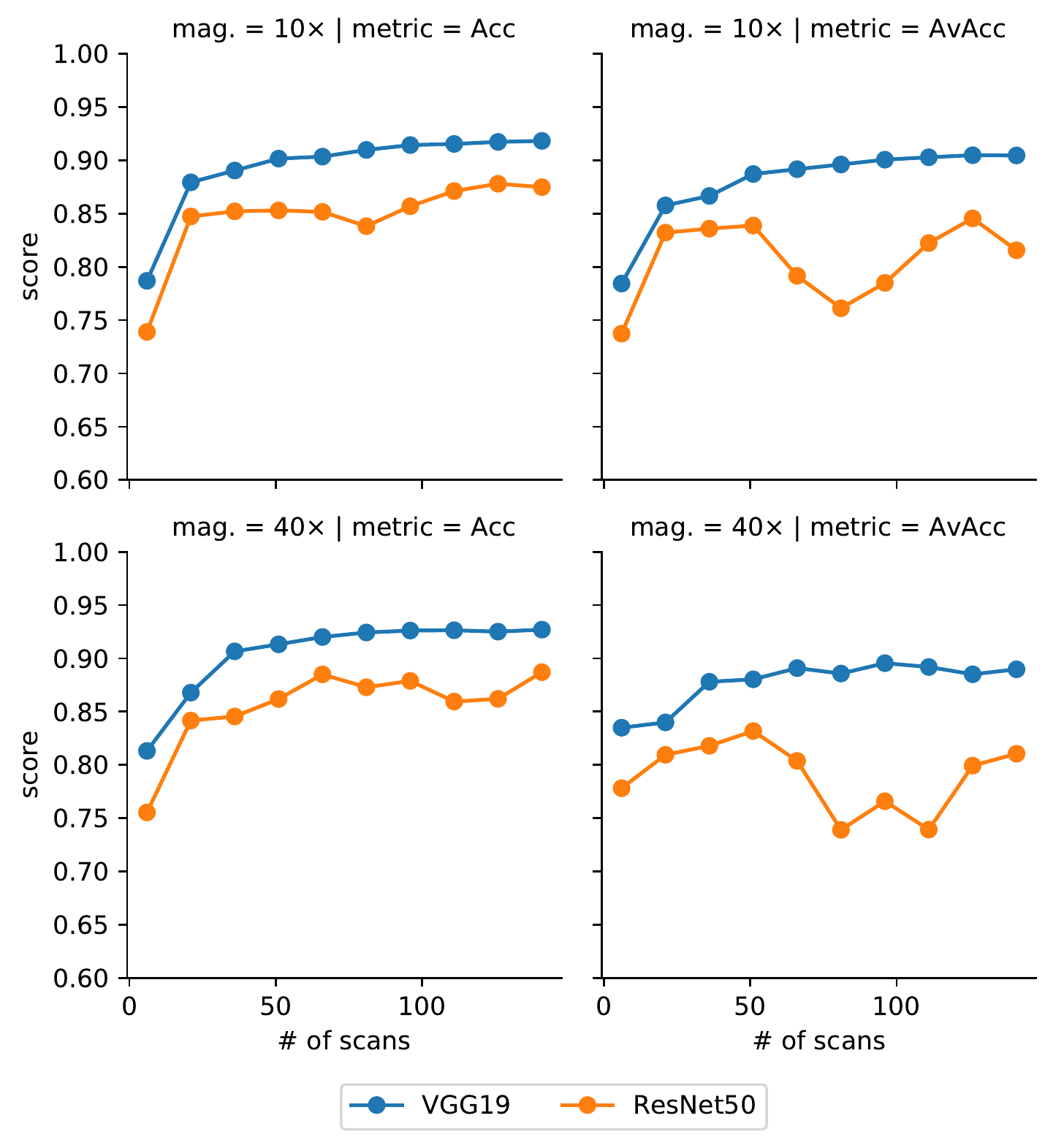}
\caption{Visualization of the impact of increasing amount of training data (measured as the number of scans from which the patches were extracted) on the classification performance.}
\label{fig:n_scans}
\end{figure}

The results of this experiment are presented in Figure~\ref{fig:n_scans}. In addition to the standard classification accuracy (Acc), which was included for reference, average accuracy (AvAcc) \cite{Branco:2017} was also used as a metric appropriate for evaluation of performance in multi-class imbalanced setting: even though some of the experiments were conducted on a binary variant of dataset, AvAcc was used throughout the experiments for consistency. Several observations can be made based on the obtained results. First of all, VGG19 network outperformed ResNet50 architecture regardless of the amount of supplied data, scan magnification or chosen performance metric, which is consistent with the results from the previous experiments. Secondly, while the performance of VGG19 network increased monotonically (with the exception of AvAcc at $40\times$ magnification, for which a slight oscillations were present), a much higher variance was observed for ResNet50: this suggests a higher susceptibility of this model to not only quantity but also the quality of data, causing the training to be less stable. This is pronounced even further when considering the AvAcc metric, indicating that in particular the performance on the minority classes is affected. Finally, while the performance of VGG19 to a large extent saturates at the higher number of training scans, in particular at the $40\times$ magnification, for which more patches are available, the performance of ResNet50 kept improving, albeit with large oscillations. It is unclear whether further increase in the number of training scans would decrease the gap in performance between the two architectures.

\begin{figure}
\centering
\includegraphics[width=\linewidth]{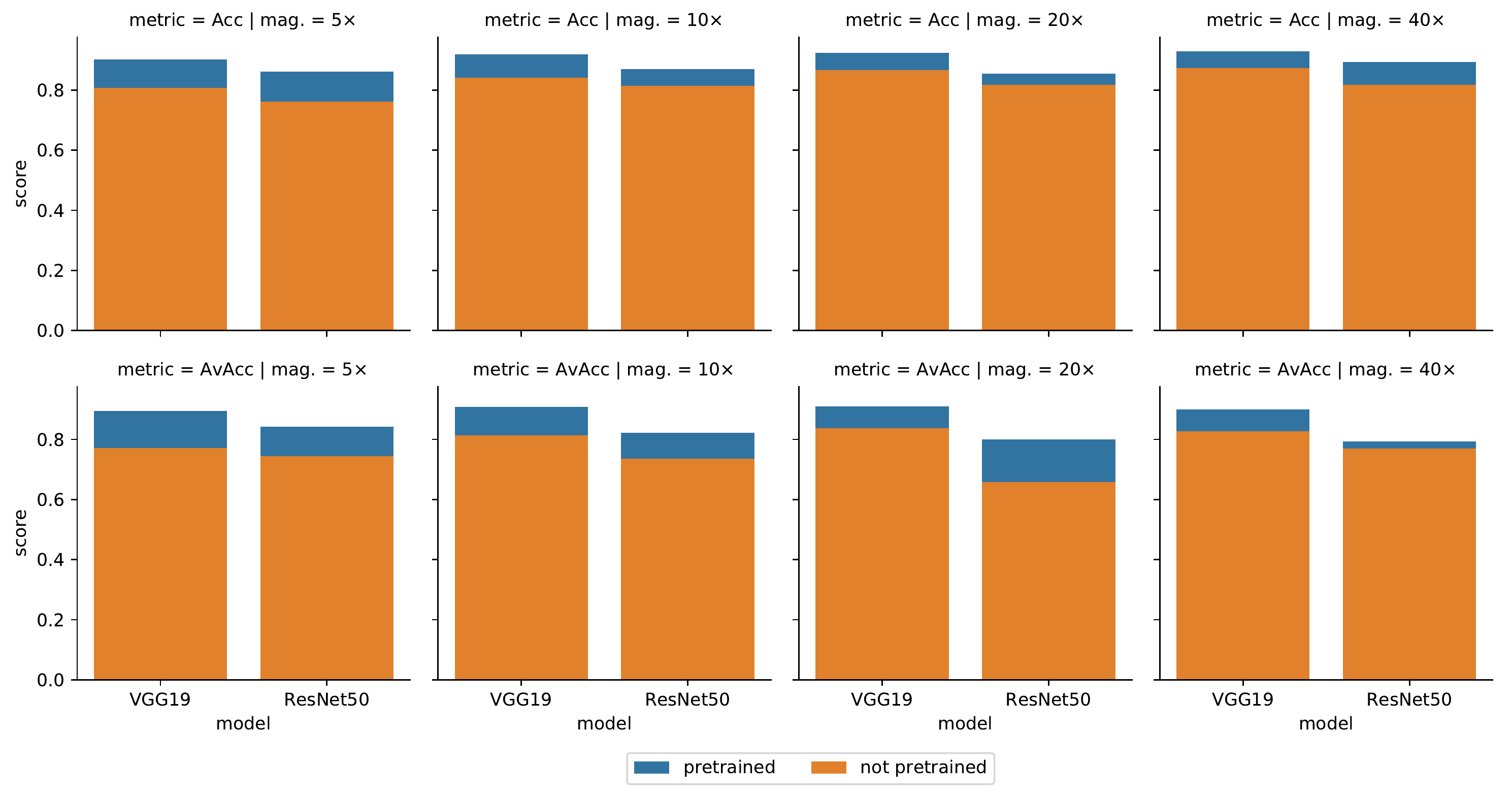}
\caption{Visualization of the impact of applying weight transfer, with the weights extracted from a model trained on the ImageNet dataset.}
\label{fig:pretraining}
\end{figure}

A common strategy for reducing the negative impact of low quantity of the available data is using transfer learning, in particular weight transfer from a model trained on another task. However, it is not clear to what extent weight transfer is suitable with a significantly different problem domains of the source and target models, which is the case for an original model trained on a natural image dataset, such as ImageNet, later transferred to the histopathological image recognition task. To experimentally evaluate the usefulness of weight transfer in such scenario the final performance of models either trained from scratch, or initialized with the weights transferred from a model trained on the ImageNet dataset, were compared. All possible magnifications were examined, with VGG19 and ResNet50 once again used as the underlying models in the binary class setting. The observed results are presented in Figure~\ref{fig:pretraining}. As can be seen, in every single case model trained using weight transfer achieved better performance than the same model trained from scratch, both with respect to accuracy and average accuracy of the predictions. This indicates the suitability of transfer learning, even when the original model was trained on a vastly different problem domain.

\noindent\textbf{Presence of label noise.} Another data difficulty factor likely to affect the performance of models in the histopathological image recognition task is presence of label noise. Gleason grading can be a highly subjective task, even at the level of the whole slide. But even more difficulty is associated with the labeling of a specific regions: a number of cases in which medical practitioners labeled different regions as containing cancerous tissue, despite achieving the same diagnosis on the level of the whole scan, was observed during the data acquisition process. Yet another factor contributing to the presence of label noise is the difficulty associated with the exact labeling of intertwining cancerous and non-cancerous tissue: medical practitioners tend to operate on lower magnifications levels than those considered by the proposed patch recognition system, which leads to labeling cancerous regions as a whole, without exclusion of small tissue patches that by itself might not be categorized as cancerous. 

\begin{figure}
\centering
\includegraphics[width=0.7\linewidth]{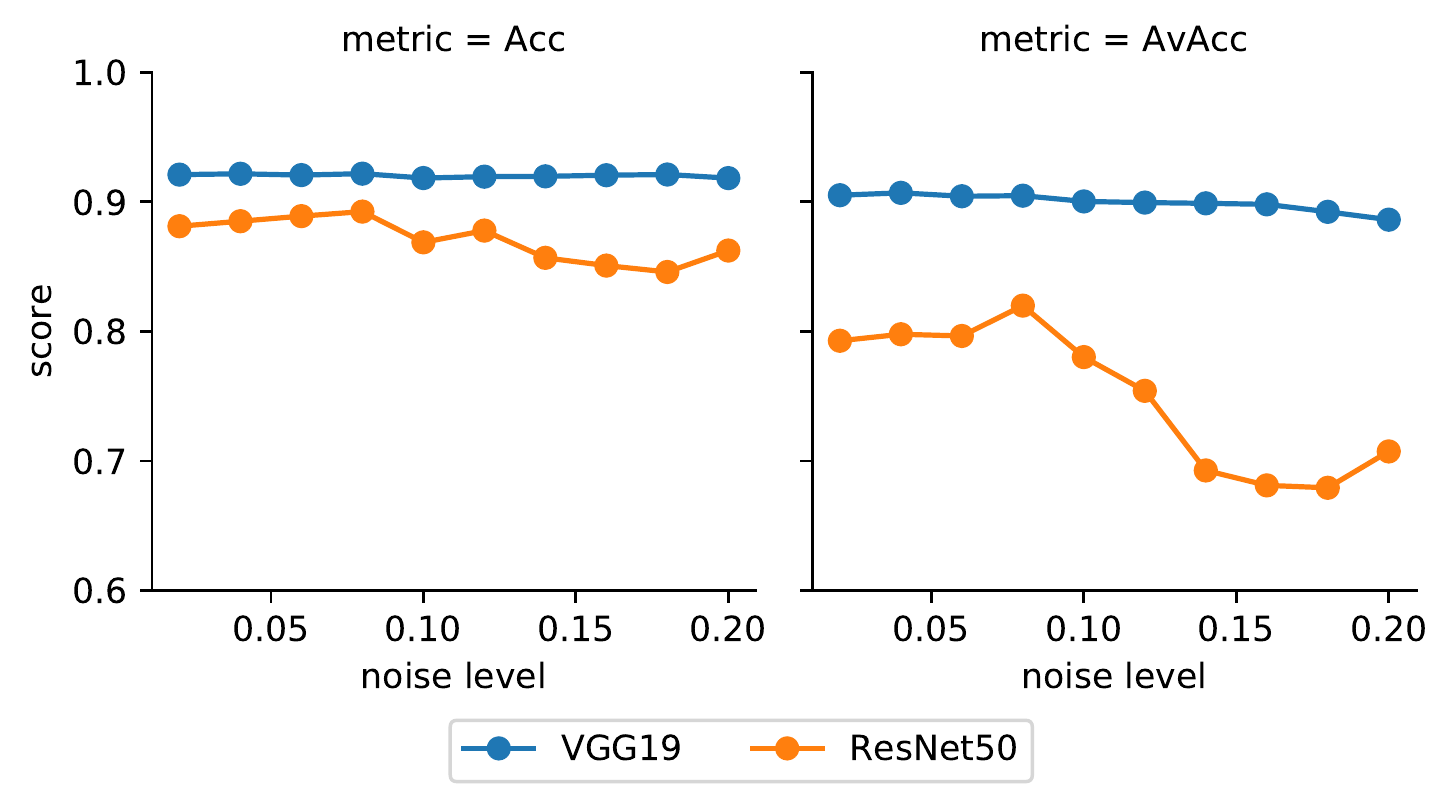}
\caption{Visualization of the impact of label noise level on the classification performance.}
\label{fig:noise_level}
\end{figure}

As a result of the aforementioned factors, presence of label noise is very likely in the considered data, but both its degree and its impact on the models' performance remain unknown. However, in an attempt at evaluating said impact an experiment was conducted, in which the label noise was artificially introduced to the original data. Specifically, the binary classification setting was considered at several noise levels, with the label noise introduced by randomly changing the label of a given observation to another class with a probability equal to noise level $\in \{0.02, 0.04, ..., 0.20\}$. It is worth noting that due to the inherent data imbalance, with more than twice patches being originally labeled as non-cancerous, described procedure actually decreased the degree of imbalance proportionally to the noise level. The observed results are presented in Figure~\ref{fig:noise_level}. As can be seen, presence of label noise has a significantly higher impact on the performance for ResNet50 architecture than it does for VGG19. Furthermore, for both of the considered models the negative impact was higher with respect to the AvAcc than the standard accuracy, indicating that even though the proportion of majority observations in the dataset decreased, model became more heavily biased towards the majority class.

\begin{figure}
\centering
\includegraphics[width=0.49\linewidth]{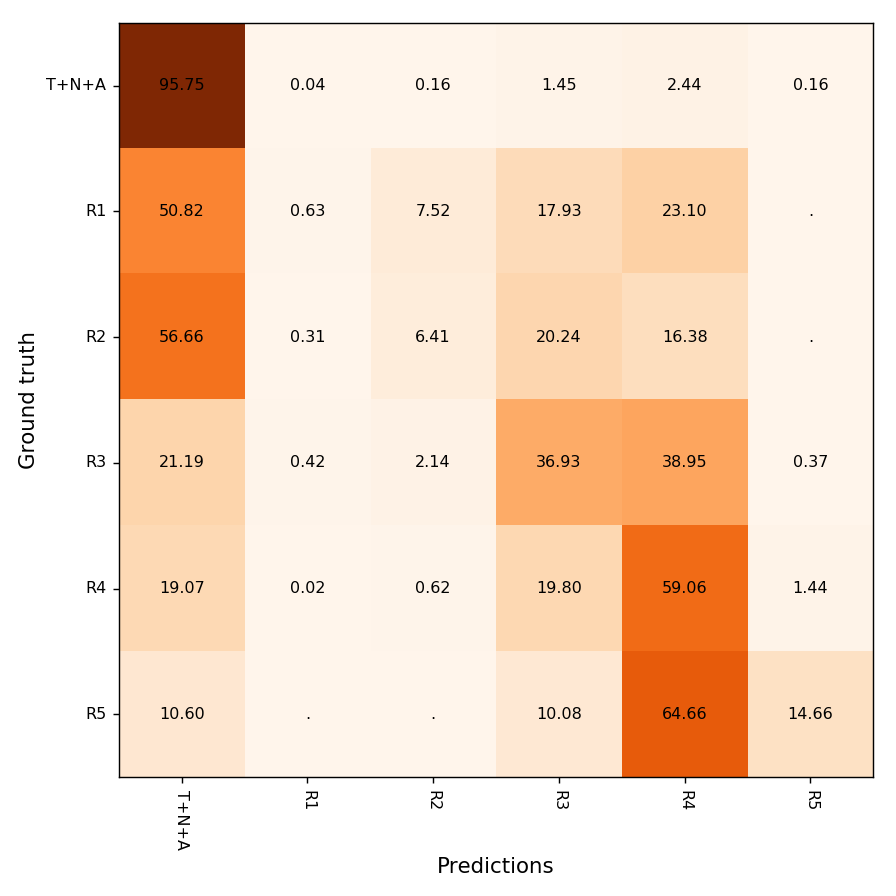}
\includegraphics[width=0.49\linewidth]{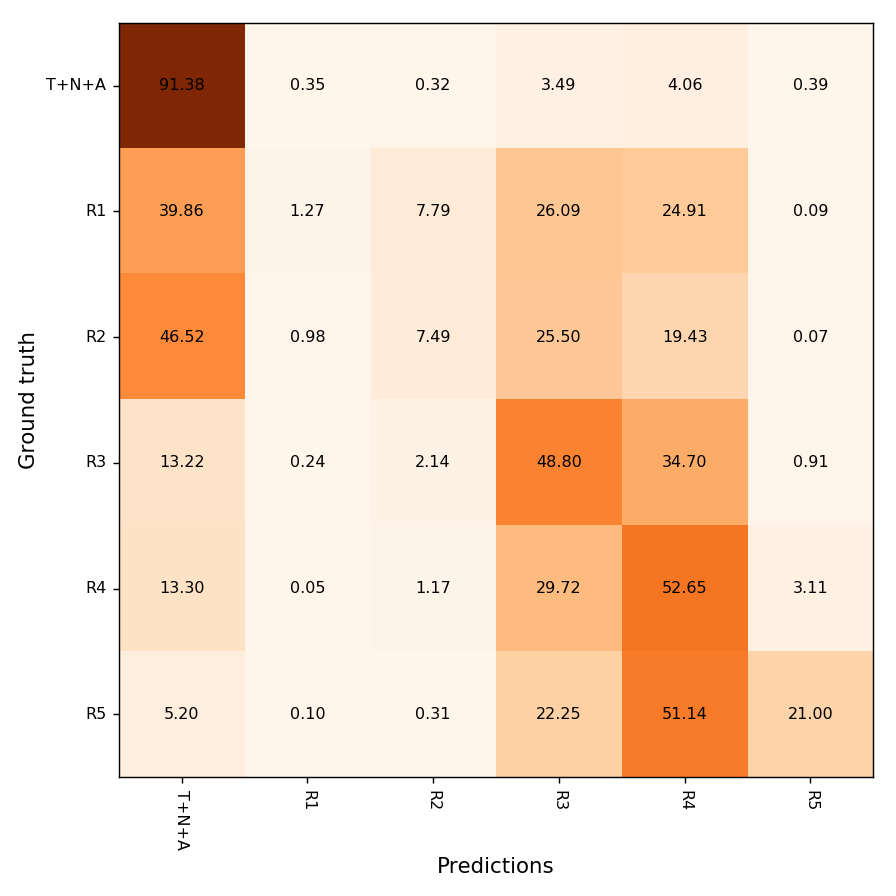}
\caption{Confusion matrices of predictions made by a network trained in a default training mode (left) and a network trained using batch balancing strategy (right).}
\label{fig:cm-diagset-a1-batch-balancing}
\end{figure}

\noindent\textbf{Data imbalance.} Finally, yet another data difficulty factor affecting the presented dataset is presence of data imbalance: a disproportion between the number of observations belonging to individual classes. Data imbalance is known to negatively affect the performance of various classification algorithms, including convolutional neural networks \cite{buda2018systematic}. Several studies \cite{koziarski2018convolutional,kwolek2019breast,koziarski2019radial} considered the subject of data imbalance in small-scale histopathological image recognition, with smaller benchmark datasets used as a basis for evaluation. It is however worth noting that data imbalance poses a particular challenge for smaller datasets, in which the number of observations from the minority class might be insufficient to enable successful learning. To evaluate to what extent data imbalance affects the performance of models trained on the DiagSet dataset, and in particular whether the traditional methods for dealing with data imbalance are suitable for reducing its negative impact, two experiments were conducted. First of all, a traditional training strategy of convolutional neural network was compared with batch balancing, an approach in which at each training iteration batch of images is sampled randomly, with artificially introduced even class distribution. Batch balancing is conceptually very similar to random oversampling (ROS) in the mini-batch training mode, with the only difference between the two being the fact that ROS freezes the minority observations that will be over-represented during the initial oversampling, whereas batch balancing over-represents each observation at an equal rate. Both methods were applied in the multi-class Gleason setting, in which every Gleason grade was considered separately, and the remaining classes (T, N and A) were merged into a single non-cancerous class. The results of this experiment are presented in Table~\ref{table:batch-balancing}. As can be seen, using the batch balancing strategy made it possible to improve the performance of a network with respect to AvAcc at 3 out of 4 magnifications. However, in every case applying batch balancing also led to a worst classification accuracy, indicating that the performance in recognition of minority classes was increased at the expense of majority classes. The improvement of performance for minority classes was also not substantial, which is further illustrated in Figure~\ref{fig:cm-diagset-a1-batch-balancing} with a comparison of confusion matrices of predictions of the network trained in both modes at $40\times$ magnification.

\begin{table}
\caption{Comparison of the performance of a VGG19 network trained with an imbalanced data distribution with a network trained using batch balancing strategy.}
\label{table:batch-balancing}
\centering
\begin{tabular}{llll}
\toprule
\textbf{Metric} & \textbf{Magnification} & \textbf{Imbalanced} & \textbf{Balanced} \\
\midrule
Acc & $40\times$ & \textbf{84.56} & 81.78 \\
 & $20\times$ & \textbf{83.97} & 80.31 \\
 & $10\times$ & \textbf{82.73} & 77.93 \\
 & $5\times$ & \textbf{79.80} & 75.45 \\
\midrule
AvAcc & $40\times$ & 35.57 & \textbf{37.10} \\
 & $20\times$ & 38.89 & \textbf{39.26} \\
 & $10\times$ & \textbf{39.76} & 39.54 \\
 & $5\times$ & 38.93 & \textbf{39.80} \\
\bottomrule
\end{tabular}
\end{table}

Secondly, various data-level approaches for handling data imbalance were evaluated. Specifically, the considered case was one in which high-level image features were extracted from the last fully-connected layer of a previously trained VGG19 network, training data was resampled in the feature space, and finally the oversampled, high-level features were classified using a separate SVM classifier with the RBF kernel. High-level features were used instead of the original image representation due to the fact that image data is ill-suited for traditional interpolation-based techniques, such as SMOTE \cite{chawla2002smote}. In this stage of experiments several oversampling algorithms were considered: random oversampling (ROS), previously mentioned SMOTE, SMOTE combined with Tomek links (SMOTE+TL) \cite{tomek1976two}, Borderline SMOTE (Bord) \cite{Han:2005} and MC-CCR \cite{koziarski2020combined}. As a reference, the case in which no oversampling was performed (None) was also included. In every case, oversampling was performed on a 4096-dimensional feature vector extracted from a previously trained convolutional neural network. Due do the computational limitations, in this stage of the experiments only the $10\times$ and $5\times$ magnifications were considered. At this stage a complete multi-class setting was used, that is every class was considered separately. Results are presented in Table~\ref{table:resampling}. As can be seen, using this approach led to a greater improvement in performance with respect to AvAcc than the one observed when using batch balancing strategy, with MC-CCR achieving the best results. However, this improvement was accompanied by a larger decrease in standard classification accuracy, which in general was greater for the methods that achieved better AvAcc. While this is an expected behavior, this illustrates the point that despite the fact that improvement in performance on minority classes is possible, it is usually accompanied by a decrease in performance on majority classes. This is important in particular in the problem domain of histopathological image classification, where it is not clear what cost is associated with incorrect classification of minority classes, and in particular how the cost changes depending on whether we are interested only in the binary diagnosis, or we need a specific Gleason discrimination.

\begin{table*}
\small
\caption{Comparison of different data oversampling strategies applied prior to SVM-based classification.}
\label{table:resampling}
\centering
\begin{tabular}{llllllll}
\toprule
\textbf{Metric} & \textbf{Magnification} & \textbf{None} & \textbf{ROS} & \textbf{SMOTE} & \textbf{SMOTE+TL} & \textbf{Bord} & \textbf{MC-CCR} \\
\midrule
Acc & $10\times$ & \textbf{66.93} & 64.05 & 64.19 & 63.93 & 61.59 & 58.38 \\
 & $5\times$ & \textbf{65.18} & 61.19 & 61.12 & 60.72 & 58.72 & 56.40 \\
\midrule
AvAcc & $10\times$ & 42.69 & 45.71 & 44.78 & 44.81 & 43.61 & \textbf{48.73} \\
 & $5\times$ & 40.42 & 41.85 & 41.29 & 41.23 & 40.72 & \textbf{43.10} \\
\bottomrule
\end{tabular}
\end{table*}

\noindent\textbf{Main results.} Based on the results of the conducted experiments it can be concluded that lack of data, presence of label noise and data imbalance are all factors that influence the performance of convolutional neural networks in the patch recognition task. Furthermore, some of those factors tended to have a higher negative impact on ResNet50, used as an example of deeper, more recent architecture, than it did on VGG19. This suggests that the resilience of VGG19 to said factors can make it more suitable for this specific problem domain, and highlights the necessity of development of novel architectures and training strategies, less prone to the negative impact of data difficulty factors.

From the point of view of data imbalance the observed results demonstrate that in a practical setting the imbalance itself is not the sole difficulty factor affecting the behavior of classification algorithms, but one of the factors that have to be considered holistically to achieve a satisfactory performance. Based on the observed results it can be concluded that applying the MC-CCR algorithm in a high-level feature space enabled us to achieve a better performance than the reference methods, possibly due to a higher resilience to said difficult factors, in particular the presence of noise, which was demonstrated in previous studies \cite{koziarski2020combined}.

%% file: Sources/Summary.tex
\chapter{Summary}
\label{chapter:summary}

\begin{center}
  \begin{minipage}{0.5\textwidth}
    \begin{small}
      In which the thesis is concluded and the possible future research directions are discussed.
    \end{small}
  \end{minipage}
  \vspace{0.5cm}
\end{center}

In this thesis the problem of data imbalance, one of the most prevalent challenges affecting the contemporary machine learning, was considered. A range of resampling algorithms aimed at reducing the negative impact of data imbalance was designed, a methodology of extending the chosen strategies to the multi-class setting was considered, and the usefulness of the proposed methods was empirically evaluated, both on the standard benchmark datasets as well as in the histopathological image recognition task.

\section{Main contributions}

The bulk of the work focused on the development of binary resampling algorithms. Two main approaches for utilizing local data characteristics in the resampling process were considered. First of all, using radial basis functions to model local class density. This approach was first used in the Radial-Based Oversampling algorithm to designate the regions of interest in which oversampling should be conducted. The concept of class potential was later translated to the undersampling context in the Radial-Based Undersampling algorithm, in which it was used to determine the order of observation removal. Finally, it culminated in the form of Potential Anchoring algorithm, which used class potential not to boost the specific regions, but instead preserve the original shape of class distributions in a unified over- and undersampling framework.

The second was an energy-based approach introduced in the Combined Cleaning and Resampling algorithm, which was used to regulate the size of oversampling regions based on the local majority class neighborhood. This methodology was later extended in the form of Radial-Based Combined Cleaning and Resampling algorithm, which first used the energy-based methodology to form general regions of interest for oversampling, and afterwards fine-tuned their shape based on the radial-based approach. 

Afterwards, the Synthetic Majority Undersampling Technique was also introduced, which extended the neighborhood-based interpolation of SMOTE to the undersampling context. This approach was combined with the original SMOTE algorithm, which was empirically shown in a preliminary study to produce a promising results compared to methods based solely on either over- or undersampling, a notion that was also later utilized in the Potential Anchoring algorithm.

A class decomposition strategy was developed that was used to translate chosen binary oversampling algorithms to the multi-class setting in the form of Multiclass Radial-Based Oversampling and Multiclass Combined Cleaning and Resampling algorithms. The approach was in both cases empirically demonstrated to display a favorable performance when compared to the existing strategies for handling multi-class imbalance.

Finally, the developed methods were applied in the histopathological image recognition context. The impact of data imbalance on the performance of convolutional neural networks was investigated, and various data difficulty factors that should be considered when dealing with data imbalance in the image recognition task were identified. Selected methods were used to resample the data in the space of high-level features extracted from a convolutional neural networks, showing promising results when compared to the traditional data-level strategies for handling data imbalance. 

\section{Future research directions}

Several promising future research directions have been identified in the outlined papers. PA, despite its good performance on the examined datasets, was particularly susceptible to the presence of noise in data. Possible strategy for mitigation of this issue is a development of preprocessing techniques that would reduce the number of outliers, or modification of the cost function itself to reduce their impact. On the other hand, one of the limitations of the CCR algorithm was the regular shape of spherical regions of interest generated by that method, which expand in every direction equally, regardless of the position of majority class observations. RB-CCR partially solved that issue, but other strategies of fine-tuning could also be considered. RB-CCR could also benefit from a better strategy for choosing the resampling regions, possibly on a per-observation basis. Finally, CSMOUTE in its current form is a preliminary work, aimed at showing that combining over- and undersampling within a single dataset can be beneficial. A next step for extending the method would be introducing a mechanism for determining the specific regions in which either over- or undersampling should be performed, instead of performing both globally as it is currently done. A multi-class extensions of all of the remaining methods could also be considered, with a decomposition strategy adjusted specifically to the undersampling setting, as well as separate extensions for handling data streams and highly-dimensional data, both of which are factors recognized in the literature as limiting the performance of traditional data-level algorithms. Finally, from the standpoint of applying the proposed data-level algorithms in the image classification task, it seems worthwhile to consider alternative data representations other than features extracted from a classification networks. In particular features extracted from autoencoders and generative adversarial networks could be considered, as they were already used as a basis for classification in the literature.